\documentclass{article}

\usepackage[preprint]{neurips_2025}

\usepackage[utf8]{inputenc} % allow utf-8 input
\usepackage[T1]{fontenc}    % use 8-bit T1 fonts
\usepackage{hyperref}       % hyperlinks
\usepackage{url}            % simple URL typesetting
\usepackage{booktabs}       % professional-quality tables
\usepackage{amsfonts}       % blackboard math symbols
\usepackage{nicefrac}       % compact symbols for 1/2, etc.
\usepackage{microtype}      % microtypography     % colors
\usepackage{amsmath}
\usepackage{framed}
\usepackage{amsthm}
\usepackage{mathrsfs}
\usepackage{amssymb}
\usepackage{enumitem}
\newtheorem{lemma}{Lemma}
\newtheorem{theorem}{Theorem}
\newtheorem{conjecture}{Conjecture}
\usepackage{graphicx}
\usepackage{caption}
\usepackage{float}
\usepackage{tabularx}
\usepackage{booktabs}
\usepackage{makecell}
\usepackage{geometry}
\usepackage{graphicx}
\usepackage[perpage,para]{footmisc}
\usepackage[svgnames, dvipsnames, table]{xcolor}
\usepackage{tikz} 
\definecolor{HeaderBlue}{RGB}{31,74,125}
\definecolor{RowGreen}{RGB}{217,240,211}
\definecolor{RowYellow}{RGB}{254,242,204}
\definecolor{RowOrange}{RGB}{253,225,152}

\newcolumntype{C}{>{\centering\arraybackslash}X}

\title{AI Mathematician: Towards Fully Automated Frontier Mathematical Research}

\author{%
    Yuanhang Liu$^{4}$\thanks{Indicates equal contribution.}$^{\phantom{*}}$ \textbf{,}
    Yanxing Huang$^{3*}$\textbf{,}
    Yanqiao Wang$^{4*}$\textbf{,}
    Peng Li$^{2}$\thanks{\;Correspondence to Peng Li <lipeng@air.tsinghua.edu.cn>, Yang Liu <liuyang2011@tsinghua.edu.cn>.},\;
    Yang Liu$^{1,2}$\footnotemark[2] \\
$^{1}$ Dept. of Comp. Sci. \& Tech., Institute for AI, Tsinghua University, Beijing, China\\
$^{2}$ Institute for AI Industry Research (AIR), Tsinghua University, Beijing, China\\
$^{3}$ Department of Mathematical Sciences, Tsinghua University \\
$^{4}$ Qiuzhen College, Tsinghua University
}
\begin{document}

\maketitle

\begin{abstract}
  Large Reasoning Models (LRMs) have made significant progress in mathematical capabilities in recent times. However, these successes have been primarily confined to competition-level problems. In this work, we propose AI Mathematician (AIM) framework, which harnesses the reasoning strength of LRMs to support frontier mathematical research. We have identified two critical challenges of mathematical research compared to competition, {\it the intrinsic complexity of research problems} and {\it the requirement of procedural rigor}. To address these challenges, AIM incorporates two core strategies: an exploration mechanism to foster longer solution paths, and the pessimistic reasonable verification method to ensure reliability.
  This early version of AIM already exhibits strong capability in tackling research-level tasks. We conducted extensive experiments across several real-world mathematical topics and obtained promising results. AIM is able to autonomously construct substantial portions of proofs and uncover non-trivial insights within each research area. These findings highlight the potential of LRMs in mathematical discovery and suggest that LRM-based agent systems could significantly accelerate mathematical research in the future.
\end{abstract}

\begin{table}[htbp]
\centering
\label{tab:research}
\setlength{\tabcolsep}{4.5pt}
\begin{tabularx}{0.9\textwidth}{*{4}{C}}
\toprule[0.9pt]
\multicolumn{4}{>{\columncolor{HeaderBlue}}c}{\textcolor{white}{\textbf{\large Mathematical Research Problems}}} \\
\cmidrule[0.4pt]{1-4}
\rowcolor{HeaderBlue!15}
\makecell{\textbf{Quantum} \\ \textbf{Algorithm} \\ \textbf{Problem}} &
\makecell{\textbf{Absorbing} \\ \textbf{Boundary} \\ \textbf{Condition}} & 
\makecell{\textbf{High Contrast} \\ \textbf{Limit}} & 
\makecell{\textbf{Homogenization} \\ \textbf{Problem}} \\
\midrule[0.4pt]
\cellcolor{RowGreen}\makecell{AIM effectively\\ completes the\\ problem  with \\ a   detailed\\ solution  process.} & 
\cellcolor{RowGreen}\makecell{AIM provides\\ a substantially \\complete proof \\for this problem.} & 
\cellcolor{RowYellow}\makecell{AIM completes \\ main proof of \\ the conclusion \\ and produces \\other \\correct results.} & 
\cellcolor{RowOrange}\makecell{AIM provides \\ partially correct \\conclusions and\\reasoning, \\ offering \\instructive \\guidance for\\ this problem.} \\
\bottomrule[0.8pt]
\end{tabularx}
\vspace{1em}
\caption{Summary of mathematical problems addressed by our proposed AIM.}
\end{table}

\section{Introduction}

Mathematics has long been considered as the pinnacle of human intellect. After the emergency of large language models (LLMs), the capability border of AI systems has been expanded to an unprecedented level. Among these advancements, the progress in mathematical reasoning stands out as one of the most surprising and closely watched developments. It also brings to mind the earliest insights we had about artificial intelligence~\citep{newell1957logic}.

Only four years ago, in 2021, the most advanced LLM GPT-3 175B can not solve more than 35\% of grade school math problems even after thousands of training steps~\citep{cobbe2021trainingverifierssolvemath}. Then in early 2024, state-of-the-art models can already solve more than 50\% of high-school level math problems, while their accuracy on the former dataset already reaches around 90\%~\citep{shao2024deepseekmathpushinglimitsmathematical}. And within the last year 2024, we have witnessed the breakthrough of Large Reasoning Models (LRMs), which brings a significant boost in math reasoning~\citep{openai2024openaio1card, deepseekai2025deepseekr1incentivizingreasoningcapability, kimiteam2025kimik15scalingreinforcement}. Recently the cutting-edge model OpenAI o3 and o4-mini even achieved nearly 100\% accuracy on the most challenging competition level AIME 2024 and AIME 2025 datasets, surpassing almost all that humans can do~\citep{openai2025o3}.

The rapidly advancing mathematical capabilities of LRMs seem to make it possible for them to undertake frontier mathematical research. In this work, we take initial steps in this promising and underexplored direction, reporting preliminary yet encouraging results to inspire future exploration.

We propose an LRM-based mathematical research agent framework named AIM (AI Mathematician), with a special focus on frontier mathematical research. We have identified two prime challenges in transitioning from mathematical competitions to mathematical research:

\begin{itemize}
    \item \textbf{Complexity}. Frontier math papers typically require tens of pages of proofs of lemmas or intermediate steps towards a theorem, while competitions require attendants to accomplish technically sophisticated short proofs with hours. The complexity of mathematical research is also reflected in the breadth of domain knowledge it involves.
    \item \textbf{Procedural Rigor}. Frontier mathematical research mainly works with proof problems without an explicit answer. Their correctness relies on the procedural rigor and vast manual reviews. Currently, we still lack widely accepted and effective methods to assess a proof in natural language.
\end{itemize}

We proposed two approaches to fill the gap between competition level problem solving and frontier mathematical research. They appear to be simple techniques that already effectively enabled LRMs to possess to some extent the ability of frontier mathematical research. To conclude, our contributions can be divided into three parts:

\begin{itemize}
    \item \textbf{Exploration and Memory Mechanism} that tackles a research topic into multiple steps. The AIM agent can explore around the statements of the given problem and obtain several lemmas. It will finally try to complete the solution when time is right.
    \item \textbf{Pessimistic Reasonable Verification} (PRV) that requires an LRM to generate multiple different reviews to the same proof, and selects the worst review as the final assessment of it. In our experiments this method can notably improve the output quality.
    \item \textbf{Preliminary Experiments} on some frontier math problems. We have done some comprehensive experiments and manual reviews on several open problems in math. Our AIM can already find the proper approach to solve these problems, and made some impressive findings. Solutions generated by AIM still contain some flaws, but most of them could be resolved with the help of a professor, and result in a concrete discovery in mathematics.
\end{itemize}

\section{Method}

\subsection{Overview}

\begin{figure}[htbp]
    \centering
    \includegraphics[width=\textwidth]{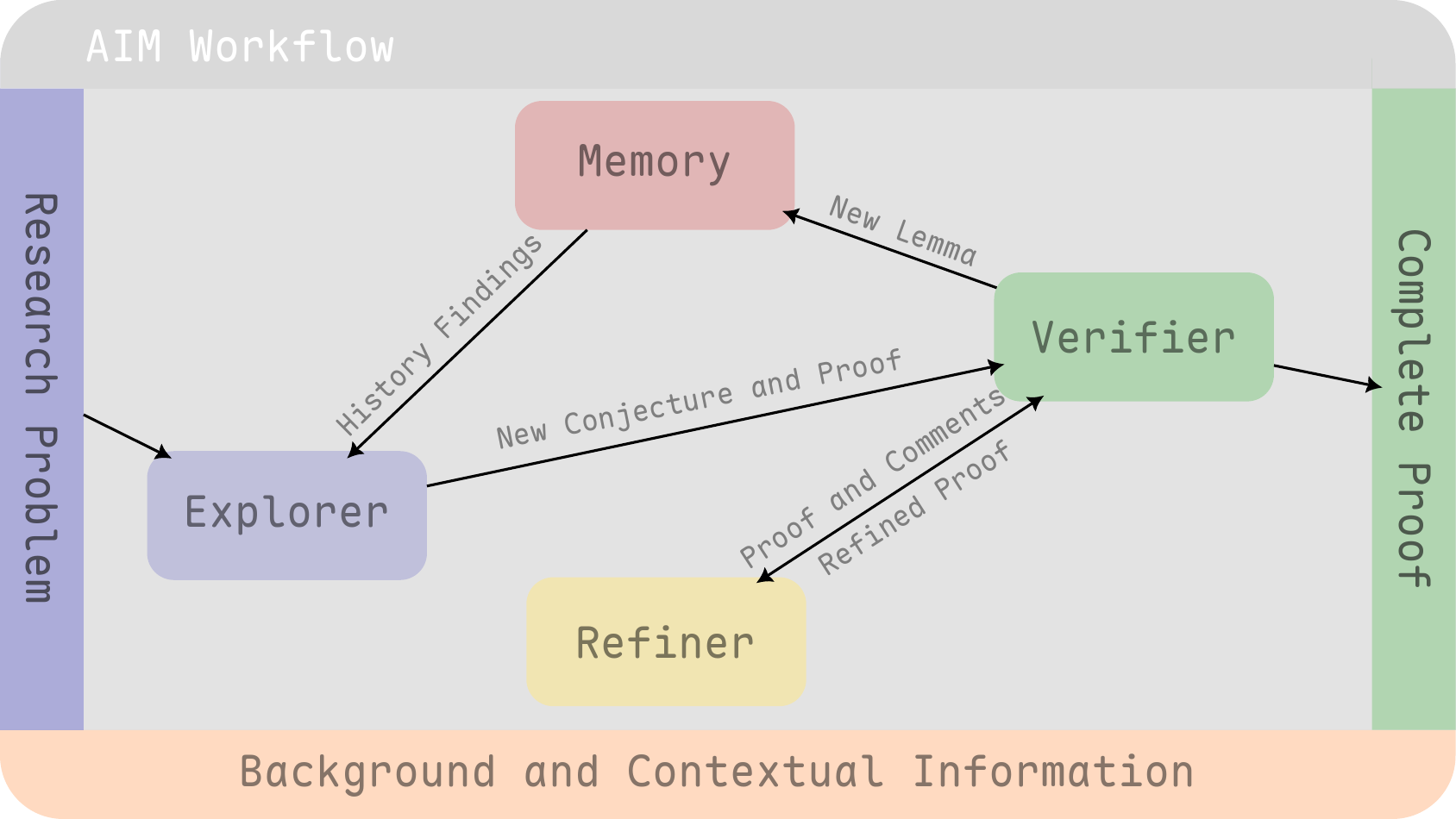}
    \caption{The illustration of AIM workflow. AIM adopts a direct workflow to explore through the given research topic. AIM used an exploration and memory mechanism to tackle down a research problem into multi-step exploration, and verification and refining mechanism to guarantee the correctness and quality of proofs.}
    \label{fig:aim_workflow}
\end{figure}

The overview of our proposed AIM framework is shown in Figure~\ref{fig:aim_workflow}. It consists of three LRM-based agents in this early version of AIM, which are {\it explorer}, {\it verifier}, and {\it refiner}. These agents will interact with each other and form two action loops, an exploration loop that enables longer reasoning paths, and a refine loop that guarantees the output quality. These action loops will be further discussed in the following subsections.

To leverage AIM for mathematical research, a human expert should firstly provide some background and contextual information of this research topic. This could include the definitions of terminologies, and some preliminary conclusions required for this problem. These contents will then be formatted and treated as a system prompt, which is visible to all three agents.

After this, we can directly pass the research problem to the agent, which could be a certain intermediate problem, whether minor or significant, or the research topic as a whole. AIM will then try to solve this problem step by step, and report all its findings to the user, which could be especially useful for human mathematicians.

\subsection{Exploration and Memory Mechanism}

When faced with complex research problems, even state-of-the-art LRMs are unable to produce correct and rigorous solutions in a single attempt. In our experiments, providing an LRM with the necessary background and directly prompting it to solve a research-level problem often leads to incorrect conclusions and unsound proofs lacking the required rigor.

This is where our exploration mechanism comes into play. Rather than attempting to solve the problem outright, the explorer first engages in open-ended reasoning and exploration around the topic. It formulates a set of intermediate observations and hypotheses, presenting them as conjectures, each accompanied by a detailed proof.

Importantly, the explorer refrains from asserting final conclusions unless it has sufficient confidence in the result. This entire mechanism is implemented purely through prompt engineering. In practice, the explorer demonstrates the ability to distinguish uncertain situations and typically only presents final proofs after a comprehensive exploration process.

The conjectures proposed during exploration are extracted and stored in memory. Following a verification step, valid conjectures are promoted to lemmas. The exploration process is then invoked iteratively, either until the problem is solved or a predefined exploration limit is reached. In each iteration, previously validated lemmas are provided to guide further reasoning and facilitate deeper exploration.

\subsection{Verification and Refine Mechanism}

LRMs are prone to errors when proving lemmas and often produce proofs that lack the rigor required in mathematical reasoning. While we currently lack a formal method for verifying natural language proofs, we implement the Pessimistic Reasonable Verification (PRV) process to address this challenge.

For each generated proof, we employ a separate, high-capacity LRM as a verifier to critically evaluate the reasoning and identify potential flaws. The verifier performs multiple independent reviews in parallel, and the proof is rejected if any one of these reviews deems it incorrect. This aligns with the standard in mathematical practice, where a valid proof must convincingly satisfy all reviewers. PRV thus serves as a practical simulation of such rigorous peer review in an automated setting.

Moreover, each LRM-based review is required to include a detailed, constructive feedback explaining the reasoning behind any objections. These rationales are then used to improve the proof. A dedicated refiner agent receives the original conjecture, its proof, and the comments from the verifier, and attempts to revise the proof to address the identified issues. If necessary, the refiner is also allowed to make minor adjustments to the conjecture itself. Additionally, correction suggestions can be provided to ensure that the refiner module consistently incorporates certain fixed essential requirements when refining proofs.

The refined proof is then returned to the verifier for re-evaluation. This creates an iterative refinement-verification loop that continuously enhances proof quality. Once a proof passes verification, it is accepted as a lemma and reintroduced into the exploration loop. If it fails verification after several iterations, it is eventually discarded.

\section{Background and Related Work}

\subsection{Large Reasoning Model}

Large Reasoning Models (LRMs) represent a new class of large language models characterized by their strong reasoning capabilities and the ability to generate exceptionally long chains of thought~\citep{openai2024openaio1card}. The concept was first introduced by OpenAI~\citep{openai2024openaio1card}, while DeepSeek-R1 independently developed and disclosed an effective training methodology for such models~\citep{deepseekai2025deepseekr1incentivizingreasoningcapability}.
LRMs are commonly trained using reinforcement learning with verifiable reward signals, particularly in domains such as mathematics and programming. More recent research has also explored alternative training strategies that significantly reduce the reliance on annotated data~\citep{zuo2025ttrltesttimereinforcementlearning, sun2025zerosearchincentivizesearchcapability, wang2025reinforcementlearningreasoninglarge}. Most existing LRMs have been trained primarily on high school level or competition level mathematical problems, and we do not have a quantitative metric on their performance on theorem proving. Nevertheless, LRMs have already shown promising capabilities in tackling advanced mathematical proofs. With a carefully designed reasoning workflow, they hold the potential to address significantly more complex, research-level problems while mitigate certain limitations.

\subsection{Agents}

Current agentic AI systems are driven by LLMs, which often incorporate with prompt engineering, dynamic task decomposition, persistent memory, and predefined workflow~\citep{sapkota2025aiagentsvsagentic}. Agent systems can endow large language models with greater autonomy, enabling them to perform multi-step tasks, observe their environment, and invoke external tools, among other capabilities.
LLM-based agent systems have already demonstrated strong capabilities in various domains~\citep{Wang_2024}. And with the advancement of LLMs and LRMs, recently some researchers have attempted to leverage agent systems for frontier research. AI Scientist~\citep{lu2024aiscientistfullyautomated} and AIGS~\citep{liu2024aigsgeneratingscienceaipowered} primarily attempted to automate research tasks in the field of AI. AlphaEvolve~\citep{deepmind_alphaevolve_2025} adopted a coding agent for mathematical research, and made remarkable progress on several long-standing math problems. These works are primarily centered around coding tasks or tasks that can be transformed into coding problems. To the best of our knowledge, we are the first to explore automated general mathematical research with LLM agent, and already obtained certain achievements.

\section{Experiments}

We employed AIM to tackle four mathematically challenging research problems, three proved theorems, and one open problem. Directly applying a single LRM to these problems yields incorrect results and does not constitute a valid proof process in mathematical theory. 

We conducted a systematic analysis and interpretation of preliminary experimental results, categorizing the data into three distinct groups with color-coded annotations: \textcolor{Green}{\textbf{correct proofs (``[Correct]'')}}, 
\textcolor{Blue}{\textbf{plausible inferences with missing details (``[Vague]'')}},  and \textcolor{Red}{\textbf{erroneous conclusions (``[Error]'')}}. Our experiments mainly used two distinct models, DeepSeek-R1~\citep{deepseekai2025deepseekr1incentivizingreasoningcapability} and OpenAI o4-mini~\citep{openai2025o3}. The output syntax of OpenAI o4-mini is encoded in Unicode, which we subsequently transcribed into standard \LaTeX with DeepSeek-V3~\citep{deepseek-ai_deepseek-v3_2025} for better readability.

The following provides a brief overview of the targeted mathematical research problems and the current progress of the proofs produced by AIM.
\begin{framed}
\begin{itemize}[left=0pt]
    \item \textbf{Quantum Algorithm Problem}. Linear Combination of Hamiltonian Simulation (LCHS,~\citep{PhysRevLett.131.150603}) method is an efficient method in scientific computation. Black-Scholes-Merton model is a fundamental mathematical framework in finance for pricing European-style options. This problem is to use LCHS to simulate the BSM model and design a corresponding quantum algorithm. AIM can basically solve the problem with a relatively detailed proof.

    \item \textbf{Absorbing Boundary Condition}.
This is a classic theoretical problem in analytical mathematics involving equations, where precise approximate solutions are derived through boundary construction. 
Given the method, AIM can basically complete the proof of this problem.
    \item \textbf{High Contrast Limit}. This is an error analysis problem concerning the parametric limits of the Lame-Stokes system, representing a complex theoretical issue in mathematical analysis and an important special case in practical applications. AIM gives the main process and proof of the core conclusion, although there are some irregularities in the derivation. In addition, the agent draws other correct conclusions.
 
    \item \textbf{Homogenization Problem}. Homogenization theory is a central field of study in mathematical analysis and applied mathematics, focusing on deriving the equivalent macroscopic properties of materials or systems with multiscale structures, such as periodic or stochastic features. In the process of exploring this problem, AIM gives some correct conclusions and ideas, which is of guiding significance for mathematicians to complete this research.

\end{itemize}
\end{framed}

\subsection{Quantum Algorithm Problem}
\subsubsection{Problem Description}
Our problem is applying Linear Combination of Hamiltonian Simulation lemma ~\citep{PhysRevLett.131.150603} to deal with Black-Scholes-Merton partial differential equation. Our input of the problem description is as follow:

\begin{framed}
Lemma (Linear combination of Hamiltonian simulation, LCHS): For \( t \in [0, T] \), let \( A(t) \in \mathbb{C}^{N \times N} \) be decomposed into Hermitian and anti-Hermitian parts such that \( A(t) = L(t) + iH(t) \), where \( L(t) = \frac{1}{2}[A(t) + A^{\dagger}(t)] \) and \( H(t) = \frac{1}{2i}[A(t) - A^{\dagger}(t)] \). Assume that \( L(t) \) is positive semi-definite for all \( t \in [0, T] \). Denoting the time ordering operator by \( \mathcal{T} \), we have \( \mathcal{T}\exp\left\{-\int_{0}^{t}A(s)ds\right\} = \int_{\mathbb{R}}\eta(k)u(t, k)dk \), where \( u(t, k) \) is the propagator for a time-dependent Hamiltonian simulation problem such that \( u(t, k) = \mathcal{T}\exp\left\{-i\int_{0}^{t}[H(s)+kL(s)]ds\right\} \), and \( \eta(k) = \frac{1}{\pi(1+k^2)} \) is the kernel function with respect to \( k \). 

BSM model: The PDE in the BSM model is given by 
\[
\frac{\partial V(S, t)}{\partial t} + \frac{1}{2}\sigma^2S^2\frac{\partial^2 V}{\partial S^2} + rS\frac{\partial V}{\partial S} - rV = 0.
\]
Boundary conditions are characterized by:
\( V(0,t) = 0 \) for all \( 0 < t \leq T \); 
\( V(S,t) \rightarrow S - Ke^{-r(T-t)} \) for \( S \rightarrow \infty \); 
\( V(S,T) = \max\{S - K, 0\} \). 

Your tasks are to: Use the idea of LCHS to deal with the BSM model. Design a quantum algorithm to solve the equation. And compute the complexity (query complexity to state preparation oracle and operator oracles, gate complexity, ancilla qubits, (parameters: norm of boundary and initial, T, approximation error \(\epsilon\), etc)).
\end{framed}

\subsubsection{Summary of Proof Results}
We use AIM based on DeepSeek-R1 to explore this problem.
    
The solution of our framework AIM addresses the Black-Scholes-Merton (BSM) model by transforming the PDE into a heat equation and leveraging quantum algorithms. The core steps are:
\begin{itemize}
    \item  {\bf PDE Transformation and Spatial Discretization}: Use variable substitution to convert the BSM PDE into a standard heat equation. Then discretize the spatial variable (Lemma 1).
    \item  {\bf Operator Decomposition and Integral Discretization}: Apply the Linear Combination of Hamiltonian Simulation (LCHS) lemma to decompose the heat equation’s solution operator into a continuous linear combination of unitary propagators. Then approximate the continuous integral via dscretization (Lemmas 2-3).
    \item  {\bf Quantum Implementation and Complexity Analysis}: Use the Linear Combination of Unitaries (LCU) framework and Hamiltonian simulation techniques for quantum implementation, and analyze complexity (Lemmas 3-7).
\end{itemize}

The overall approach is correct. Transforming the BSM partial differential equation into the heat equation and disctretizing the spatial variable first, then using LCHS lemma to transform equations and discretizing integrals are all correct and detailed. However, the process of complexity calculation is vague and insufficiently detailed, with some errors.
    \subsubsection{Detailed Analysis of Selected Proofs}
\begin{framed}
  \small
1.The Black-Scholes-Merton (BSM) partial differential equation (PDE) can be transformed into the standard heat equation via a logarithmic substitution. Specifically, under the transformation \( x = \ln(S/K) + (r - \sigma^2/2)(T - t) \), \( \tau = \sigma^2(T - t)/2 \), and \( V(S,t) = e^{-r(T-t)}U(x,\tau) \), the BSM PDE reduces to:

\[
\frac{\partial U}{\partial \tau} = \frac{\partial^2 U}{\partial x^2}.
\]

Discretizing this PDE spatially yields a matrix equation \( \partial_\tau |U\rangle = -B|U\rangle \), where \( B = -\partial_x^2 \). The operator \( B \) is positive semi-definite and can be directly interpreted as \( B = L \), where \( L = -\partial_x^2 \). While the original Hermitian decomposition \( B = L + iH \) (with \( H = 0 \)) trivializes the LCHS lemma, the dissipative nature of \( B \) permits alternative quantum simulation methods tailored to parabolic equations.
\begin{proof}

**Step 1: Transform BSM to the Heat Equation**

Start with the BSM PDE:
\[
\frac{\partial V}{\partial t} + \frac{1}{2}\sigma^2S^2\frac{\partial^2 V}{\partial S^2} + rS\frac{\partial V}{\partial S} - rV = 0.
\]
Apply the substitution \( V(S,t) = e^{-r(T-t)}U(x,\tau) \), \( x = \ln(S/K) + (r - \sigma^2/2)(T - t) \), and \( \tau = \sigma^2(T - t)/2 \). Compute derivatives rigorously:
- \( \partial_t V = e^{-r(T-t)}\left(rU - \frac{\sigma^2}{2}\partial_\tau U - (r - \sigma^2/2)\partial_x U\right) \),
- \( \partial_S V = e^{-r(T-t)} \partial_x U \cdot \frac{1}{S} \),
- \( \partial_S^2 V = e^{-r(T-t)} \left(\partial_x^2 U - \partial_x U\right) \cdot \frac{1}{S^2} \).

Substitute into the BSM PDE:
\[
e^{-r(T-t)}\left[ rU - \frac{\sigma^2}{2}\partial_\tau U - (r - \sigma^2/2)\partial_x U + \frac{1}{2}\sigma^2(\partial_x^2 U - \partial_x U) + r\partial_x U - rU \right] = 0.
\]
Multiply through by \( e^{r(T-t)} \), cancel terms, and simplify:
\[
-\frac{\sigma^2}{2}\partial_\tau U + \frac{\sigma^2}{2}\partial_x^2 U = 0 \implies \partial_\tau U = \partial_x^2 U.
\]

**Step 2: Spatial Discretization**

Discretize \( x \) on a grid with spacing \( \Delta x \). Approximate \( \partial_x^2 U \) via finite differences:
\[
\partial_x^2 U \approx \frac{U_{j+1} - 2U_j + U_{j-1}}{\Delta x^2}.
\]
Let \( D_2 \) be the discretized second derivative matrix. The operator becomes:
\[
B = -D_2.
\]

**Step 3: Positive Semi-Definite Operator**

The matrix \( B = -D_2 \) is positive semi-definite because \( D_2 \) (a centered second derivative discretization) is negative semi-definite. This aligns with the dissipative nature of the heat equation.

**Step 4: Implications for Quantum Simulation**

While the original decomposition \( B = L + iH \) (with \( H = -ic\partial_x \)) is invalid due to the absence of a first-order term, the operator \( B \) itself is suitable for quantum simulation methods designed for parabolic equations. Specifically, the time evolution \( \partial_\tau |U\rangle = -B|U\rangle \) can be simulated using Trotterization or other dissipative quantum algorithms, bypassing the need for the LCHS lemma.
\end{proof}

\textcolor{Green}{$\textbf{[Correct]}$ \small Variable substitution and equation transformation are correct. Discretization of the spatial variable is correct. And AIM finds the reduced PDE trivializes the LCHS lemma. It's right!}
\end{framed}
Firstly, AIM transforms the BSM equation to a heat equation. This step simplifies the problem. Then AIM discretizes the spatial variable. This makes it satisfy the conditions of the theorem.
\begin{framed}
\small
2.The integral representation in the LCHS lemma for the BSM model can be approximated with precision \(\epsilon\) using \(O(1/\epsilon^2)\) terms through an adaptive discretization of the \(k\)-integral, leveraging the rapid decay of the Cauchy kernel \(\eta(k) = \frac{1}{\pi(1 + k^2)}\).

\begin{proof}

**Truncation Error Analysis**:  
The integral \(I = \int_{-\infty}^{\infty} \eta(k) u(t, k) dk\) is truncated to \([-K, K]\). The tail error is bounded by:
\[
\int_{|k| > K} \eta(k) dk = \frac{2}{\pi} \int_{K}^{\infty} \frac{1}{1 + k^2} dk = \frac{2}{\pi} \left(\frac{\pi}{2} - \arctan(K)\right) \approx \frac{1}{\pi K} \quad \text{for } K \gg 1.
\]
Setting \(\frac{1}{\pi K} \leq \epsilon/2\) gives \(K \geq \frac{2}{\pi \epsilon}\). Thus, \(K = O(1/\epsilon)\).

**Adaptive Discretization**:  
1. **Central Interval \([-K, K]\)**: The entire truncated domain \([-K, K]\) must be discretized. To bound the discretization error, use a uniform grid with spacing \(\Delta k = \Theta(\epsilon)\). The number of points is \(N = O(K / \Delta k) = O(1/\epsilon^2)\). The discretization error using the trapezoidal rule scales as \(O(K \Delta k^2 \cdot \max |\partial_k^2 (\eta(k) u(t, k))|)\). Since \(\eta(k)\) and \(u(t, k)\) are smooth, \(\max |\partial_k^2 (\eta(k) u(t, k))|\) is bounded over \([-K, K]\). The total error becomes \(O(K \Delta k^2) = O\left(\frac{1}{\epsilon} \cdot \epsilon^2\right) = O(\epsilon)\).

**Outer Interval Contribution**:  
The truncated integral already includes all significant contributions, as the tail beyond \(K = O(1/\epsilon)\) contributes at most \(\epsilon/2\). The central interval \([-K, K]\) is approximated with \(O(1/\epsilon^2)\) points, ensuring the total discretization error is \(O(\epsilon)\). Additional adaptive methods in outer intervals (e.g., geometrically spaced grids) would only reduce logarithmic factors, which are asymptotically dominated by \(O(1/\epsilon^2)\).

**Total Terms**:  
The uniform grid over \([-K, K]\) requires \(O(1/\epsilon^2)\) terms. While adaptive grids in outer regions could marginally reduce constants, the dominant term remains \(O(1/\epsilon^2)\).

**Final Error**:  
Truncation error \(O(\epsilon/2)\) + discretization error \(O(\epsilon/2) = O(\epsilon)\). Thus, \(O(1/\epsilon^2)\) terms suffice for \(\epsilon\)-precision approximation.
\end{proof}

\textcolor{Green}{$\textbf{[Correct]}$ \small Truncate and discretize the integral correctly.}
\end{framed}

AIM analyzes the truncation error and determines the discretization spacing. This lays the groundwork for the construction of quantum algorithms.

\begin{framed}
\small
3.The solution operator for the heat equation derived from the BSM model, \( U(\tau) = e^{\tau B} \), can be approximated via the LCHS lemma as a linear combination of Hamiltonian simulations. Specifically, the integral representation \(\int_{\mathbb{R}} \eta(k) e^{-i k L \tau} dk\) (where \( L = -B \)) can be discretized into \( M = O(1/\epsilon^2) \) terms with quadrature weights \( \eta(k_j)\Delta k_j \). Each term \( e^{-i k_j L \tau} \) is a Hamiltonian simulation of \( L \) scaled by \( k_j \tau \). The total query complexity to the \( L \)-oracle is \( \tilde{O}\left( \|L\| T / \epsilon^3 \right) \), where \( T = \frac{2\tau}{\sigma^2} \), and the gate complexity scales polynomially with \( \|L\| \), \( T \), \( 1/\epsilon \), and logarithmic factors.

Proof:

**Step 1: Integral Representation via LCHS**
From Lemma-0, the solution operator for the heat equation is \( U(\tau) = e^{\tau B} \). Let \( L = -B \), a positive semi-definite operator. By the LCHS lemma with \( H = 0 \), we express:
\[
e^{\tau B} = \int_{\mathbb{R}} \eta(k) e^{-i k L \tau} dk,\quad \eta(k) = \frac{1}{\pi(1 + k^2)}.
\]

**Step 2: Quadrature Discretization**
Lemma-1 guarantees the integral can be discretized into \( M = O(1/\epsilon^2) \) terms with weights \( \eta(k_j)\Delta k \), yielding:
\[
e^{\tau B} \approx \sum_{j=1}^M \eta(k_j)\Delta k \cdot e^{-i k_j L \tau}.
\]
The truncation and discretization error is \( O(\epsilon) \) due to the \( O(1/k^2) \) decay of \( \eta(k) \).

\textcolor{Green}{$\textbf{[Correct]}$ \small Properly utilizing the LCHS lemma and memory lemmas for problem transformation.}

\end{framed}

\begin{framed}
\small
**Step 3: Hamiltonian Simulation Costs**
Each term \( e^{-i k_j L \tau} \) requires simulating \( L \) for time \( t_j = k_j \tau \). Using optimal Hamiltonian simulation algorithms (e.g., quantum signal processing or Taylor series) with query complexity \( O\left( \|L\| t_j + \log(1/\epsilon_{\text{sim}}) \right) \), set \( \epsilon_{\text{sim}} = \epsilon/M \) to bound the total simulation error by \( \epsilon \). Since \( k_j = O(1/\epsilon) \) dominates the quadrature, \( t_j = O(\tau / \epsilon) \). The per-term complexity becomes:
\[
O\left( \|L\| \tau / \epsilon + \log(M/\epsilon) \right) = \tilde{O}\left( \|L\| \tau / \epsilon \right).
\]

**Step 4: Total Query Complexity**
Summing over \( M = O(1/\epsilon^2) \) terms, the total queries to the \( L \)-oracle are:
\[
\tilde{O}\left( M \cdot \frac{\|L\| \tau}{\epsilon} \right) = \tilde{O}\left( \frac{\|L\| \tau}{\epsilon^3} \right).
\]
Expressing in terms of \( T = \frac{2\tau}{\sigma^2} \) and absorbing constants:
\[
\tilde{O}\left( \frac{\|L\| T}{\epsilon^3} \right).
\]

**Step 5: Ancilla and Gate Complexity**
The LCU framework requires \( \log M = O(\log(1/\epsilon)) \) ancilla qubits. Gate complexity inherits \( \tilde{O}(M) \)-scaling from the superposition state, augmented by \( O(\text{poly}(\log(\|L\| T / \epsilon))) \) factors from simulation subroutines. The dominant terms remain polynomial in \( \|L\| \), \( T \), and \( 1/\epsilon \), with polylogarithmic corrections.
\\ \textcolor{red}{$\textbf{[Error]}$ \small There are some mistakes about complexity computing. And the calculation process lacks detail.}
\end{framed}

Based on the previous memory lemmas, AIM constructs a quantum algorithm with Hamiltonian simulation and Linear Combination of Unitaries (LCU). Then computes the complexity. Although there are some mistakes and lacks some detail, the core methods are correct.

\subsection{Absorbing Boundary Condition}

\subsubsection{Problem Description}
Absorbing boundary conditions, formulated through modifications of conventional boundary formulations, significantly enhance solution clarity and computability at domain boundaries while preserving mathematical rigor. This specialized boundary treatment achieves mathematically precise approximations by establishing quantitative error control mechanisms between approximate and exact solutions, thereby maintaining theoretical soundness while improving computational tractability. Our study employs  to demonstrate systematic verification of solution well-posedness and uniqueness for the heat conduction equation under specifically designed absorbing boundary constraints ~\citep{abs}.

Let \( u \) be a function defined on \( (t,x) \in [0,\infty] \times \mathbb{R}^n \) and satisfy the following equations:
\[
\begin{cases}
\mathcal{L}u = \partial_t u - \Delta u + \rho(x)u = 0, & t > 0,\ x \in \mathbb{R}^n \\
u = u_0, & t = 0
\end{cases}
\]
where the operator \( \mathcal{L} = \partial_t - \Delta + \rho(x) \), \( \Delta \) is the Laplace operator, and \( \rho(x) \) is a non-negative function of \( x \).

We focus on a bounded open set \( \Omega \) in \( \mathbb{R}^n \), and denote \( \partial\Omega = \mathcal{S} \). By restricting the function \( u \) to the bounded open set \( \Omega \), we investigate the behavior of the solution within this bounded open set. The absorbing boundary conditions are as follows:

\begin{align}
&\begin{cases}
\mathcal{L}u = 0, & [0,\infty) \times \Omega \\
\frac{\partial u}{\partial \nu} = -\beta u - \sum_{k=1}^{m} \alpha_k (\partial_t - \Delta_{\mathcal{S}})\varphi_k = \mathcal{F}_1 u, & [0,\infty) \times \mathcal{S} \\
u = u_0, & t = 0
\end{cases}
&& 
\end{align}

\[
\begin{cases}
(\partial_t - \Delta_{\mathcal{S}} + d_k)\varphi_k = u, & t \geq 0 \\
\varphi_k = 0, & t = 0
\end{cases}
\]

The problem needed proved is that given any \( u_0 \in L^2(\Omega) \), there exists a unique 
\[
u \in C([0,T];L^2(\Omega)) \cap H^1([0,T];H^{-1}(\Omega)) \cap L^2([0,T];H^1(\Omega)),
\]
\[
\varphi_k \in C([0,T];L^2(\mathcal{S})) \cap H^1([0,T];H^{-1}(\mathcal{S})) \cap L^2([0,T];H^1(\mathcal{S}))
\]
satisfy the equations above.

\subsubsection{Summary of Proof Results}
In our experimental framework, we established a systematic approach for constructing and validating approximated solutions to the governing equations. The methodology comprises three key phases: 1) Formulation of approximate solutions through the Galerkin method; 2) Convergence analysis via energy estimation techniques; 3) Regularity verification and final convergence proof.

The experiment demonstrates that AIM first proves a key conclusion of energy estimation through rigorous mathematical derivation and inequality techniques. The obtained energy estimate is then used to prove the uniqueness of the solution. After considering these, AIM constructed the space of the approximated solutions, proving the existence and convergence of the approximated solution. Finally, the above lemma conclusions are integrated to prove the core theorem of our problem.

The analysis of AIM's theorem demonstration reveals both methodological strengths and areas requiring refinement. As evidenced by the proof trajectory, the system establishes a well-structured logical framework  and appropriately applies mathematical induction with rigorous bounding techniques . Crucially, AIM consistently identifies valid mathematical propositions and preserves the integrity of critical intermediate steps throughout the derivation process.At the same time, there are some shortcomings and problems in AIM. On the one hand, the details of the proof are not clear enough. On the other hand, there are also errors in the intermediate proof process.

\subsubsection{Detailed Analysis of Selected Proofs}
We explain some specific results of AIM. 
\begin{framed}
    \small 
**A priori energy estimate for the coupled system**  
Let \( u \) and \( \varphi_k \) (\( k = 1, \dots, m \)) satisfy the system (2.10) with \( u_0 \in L^2(\Omega) \) and \( \varphi_k(0) = 0 \). Then, there exists a constant \( C > 0 \) depending on \( T, \beta, \alpha_k, d_k \), but independent of \( u_0 \), such that:  
\[
\begin{aligned}
&\sup_{t \in [0,T]} \|u(t)\|_{L^2(\Omega)}^2 + \int_0^T \|\nabla u(t)\|_{L^2(\Omega)}^2 dt + \int_0^T \|u(t)\|_{L^2(\mathcal{S})}^2 dt \\
&+ \sum_{k=1}^m \left( \sup_{t \in [0,T]} \|\varphi_k(t)\|_{L^2(\mathcal{S})}^2 + \int_0^T \|\nabla_{\mathcal{S}} \varphi_k(t)\|_{L^2(\mathcal{S})}^2 dt \right) \leq C \|u_0\|_{L^2(\Omega)}^2.
\end{aligned}
\]
\end{framed}
At the beginning of the proof, AIM gives a critical consistent energy estimation inequality. This is a very important intermediate conclusion and a key step in solving this problem. 

\begin{framed}
    \small**Revised Proof of the a priori estimate:**  
Define the total energy functional:  
\[
E(t) = \frac{1}{2}\|u(t)\|_{L^2(\Omega)}^2 + \sum_{k=1}^m \frac{\alpha_k^2 d_k}{2\beta'}\|\varphi_k(t)\|_{L^2(\mathcal{S})}^2, \quad \beta' = \beta + \sum_{k=1}^m \alpha_k.
\]

\textcolor{Green}{\textbf{[Correct]} \small The agent consider the energy method, and show the correct energy functional form.}

\end{framed}
\begin{framed}
    \small
    
**Energy estimate for \( u \):** Multiply the \( u \)-equation by \( u \) and integrate over \( \Omega \):  
\[
\frac{1}{2}\frac{d}{dt}\|u\|_{L^2(\Omega)}^2 + \|\nabla u\|_{L^2(\Omega)}^2 + \beta' \|u\|_{L^2(\mathcal{S})}^2 = \sum_{k=1}^m \alpha_k d_k \int_{\mathcal{S}} u \varphi_k \, dS.
\]
Apply Young's inequality to the right-hand side:  
\[
\sum_{k=1}^m \alpha_k d_k \int_{\mathcal{S}} u \varphi_k \, dS \leq \frac{\beta'}{2}\|u\|_{L^2(\mathcal{S})}^2 + \sum_{k=1}^m \frac{\alpha_k^2 d_k^2}{2\beta'}\|\varphi_k\|_{L^2(\mathcal{S})}^2.
\]
Subtract the Young's bound and rearrange:  
\[
\frac{1}{2}\frac{d}{dt}\|u\|_{L^2(\Omega)}^2 + \|\nabla u\|_{L^2(\Omega)}^2 + \frac{\beta'}{2}\|u\|_{L^2(\mathcal{S})}^2 \leq \sum_{k=1}^m \frac{\alpha_k^2 d_k^2}{2\beta'}\|\varphi_k\|_{L^2(\mathcal{S})}^2.
\]

\textcolor{Green}{\textbf{[Correct]} \small Here, the agent gets the accurate result with the energy method and the application of inequality techniques. }

\end{framed}
\begin{framed}

\small
**Energy estimate for \( \varphi_k \):** Multiply the \( \varphi_k \)-equation by \( \frac{\alpha_k^2 d_k}{\beta'} \varphi_k \) and integrate:  
\[
\frac{\alpha_k^2 d_k}{2\beta'} \frac{d}{dt}\|\varphi_k\|_{L^2(\mathcal{S})}^2 + \frac{\alpha_k^2 d_k}{\beta'} \|\nabla_{\mathcal{S}} \varphi_k\|_{L^2(\mathcal{S})}^2 + \frac{\alpha_k^2 d_k^2}{2\beta'} \|\varphi_k\|_{L^2(\mathcal{S})}^2 \leq \frac{\alpha_k^2}{2\beta'} \|u\|_{L^2(\mathcal{S})}^2.
\]
Summing over \( k \), the right-hand side becomes:  
\[
\sum_{k=1}^m \frac{\alpha_k^2}{2\beta'} \|u\|_{L^2(\mathcal{S})}^2 \leq \sum_{k=1}^m \frac{\alpha_k}{2} \|u\|_{L^2(\mathcal{S})}^2 \leq \frac{\beta' - \beta}{2} \|u\|_{L^2(\mathcal{S})}^2.
\]

\textcolor{Green}{\textbf{[Correct]} \small Similarly, the same approach is applied for the assistant function.}
\end{framed}
Here is a partial proof process for the  consistent energy estimation.
AIM uses energy estimation method and combines the techniques of inequality to gradually deduce the estimation of the 
 solution and the auxiliary functions.
The entire derivation process is correct and rigorous, and the steps are sufficient. 
This is a very clear proof process, which meets our requirements and expectations for mathematical theoretical proof.

\begin{framed}
    \small **Step 1: Construct \( H^1 \)-Conforming Subspaces**  
Let \( \{w_j\} \subset H^1(\Omega) \) and \( \{\theta_{k,l}\} \subset H^1(\mathcal{S}) \) be eigenfunctions of the Laplace-Beltrami operators \( -\Delta_\Omega \) and \( -\Delta_{\mathcal{S}} \) with Neumann/Dirichlet boundary conditions (as appropriate). These form orthonormal bases in \( L^2(\Omega) \) and \( L^2(\mathcal{S}) \), respectively, and are dense in \( H^1(\Omega) \) and \( H^1(\mathcal{S}) \). Define \( V_N = \text{span}\{w_1, \dots, w_N\} \) and \( W_{k,N} = \text{span}\{\theta_{k,1}, \dots, \theta_{k,N}\} \).

\textcolor{Green}{\textbf{[Correct]} \small The agent applies the Galerkin method and construct the function spaces.}
\end{framed}
\begin{framed}
\small
**Step 2: Formulate the Projected Equations**  
For test functions \( v \in V_N \) and \( \psi \in W_{k,N} \), the Galerkin approximations satisfy the projected weak form:
\[
\begin{aligned}
&\int_\Omega \partial_t u^N v \, dx + \int_\Omega \nabla u^N \cdot \nabla v \, dx + \int_\Omega \rho u^N v \, dx \\
&\quad + \beta \int_{\mathcal{S}} u^N v \, ds + \sum_{k=1}^m \alpha_k \int_{\mathcal{S}} (\partial_t \varphi_k^N - \Delta_{\mathcal{S}} \varphi_k^N) v \, ds = 0, \\
\\
&\int_{\mathcal{S}} \partial_t \varphi_k^N \psi \, ds + \int_{\mathcal{S}} \nabla_{\mathcal{S}} \varphi_k^N \cdot \nabla_{\mathcal{S}} \psi \, ds + d_k \int_{\mathcal{S}} \varphi_k^N \psi \, ds = \int_{\mathcal{S}} u^N \psi \, ds.
\end{aligned}
\]

**Step 3: Solve the ODE System**  
Substituting \( u^N = \sum_{j=1}^N a_j(t) w_j \) and \( \varphi_k^N = \sum_{l=1}^N b_{k,l}(t) \theta_{k,l} \) into the projected equations yields a linear ODE system for \( a_j(t) \) and \( b_{k,l}(t) \). The coefficients are smooth (as eigenfunctions are \( C^\infty \)), ensuring continuity. By the Cauchy-Lipschitz theorem, unique solutions exist on \( [0,T] \).

\textcolor{Green}{\textbf{[Correct]} \small The agent gets the ODE satisfied with the approximated solution.}

\textcolor{Blue}{\textbf{[Vague]} \small It is better to have more detailed process.}
\end{framed}
Here is the proof step of the existence and uniqueness of the solution to the equation. The agent considers the basis functions in the functional space of the overall space and boundary space respectively, and constructs the existence space of approximated solutions.

After that, the agent constructs the form of the approximated solutions and the weak form of the satisfying equations. It also expresses the ODE system of the approximated solutions , and uses mathematical theorem to complete the proof of the existence and uniqueness of approximated solutions. This is a clear and accurate proof idea and mathematical process.
\begin{framed}
    \small **Step 6: Passage to the Limit**  
For fixed \( v \in V_{N_0} \) and \( \psi \in W_{k,N_0} \), the projected equations hold for all \( N \geq N_0 \). As \( N \to \infty \):
- Weak convergence handles volume integrals (e.g., \( \int \nabla u^N \cdot \nabla v \, dx \to \int \nabla u \cdot \nabla v \, dx \)).
- Strong trace convergence ensures \( \int_{\mathcal{S}} u^N v \, ds \to \int_{\mathcal{S}} u v \, ds \) and similar boundary terms.
- The density of \( \bigcup_{N} V_N \) in \( H^1(\Omega) \) and \( \bigcup_{N} W_{k,N} \) in \( H^1(\mathcal{S}) \) extends the equations to all \( v \in H^1(\Omega) \), \( \psi \in H^1(\mathcal{S}) \).

Thus, \( (u, \{\varphi_k\}) \) satisfies (2.10) in the weak sense. Uniqueness (by lemma) confirms the full sequence converges.

\textcolor{Blue}{$\textbf{[Correct]}$ \small Here the regularity should have more deprivation process to make the proof sufficient accurate.}

\textcolor{Red}{  $\textbf{[Error]}$    \small  The uniqueness should be prove by the difference function rather than the sequence convergence.}
\end{framed}

The proof of convergence is not completely correct. The weak convergence properties proposed by the agent are correct, but it is necessary to prove later that this limited function is the solution to the original equation. And further analysis and verification of regularity are also required.

Although there are still some proof errors, the proof results given by the agent are already relatively complete and rigorous. The agent gave the correct proving steps and the accurate derivation process, and used the correct mathematical method to obtain the conclusion of this mathematical theory.

\subsection{High Contrast Limit}
\subsubsection{Problem Description}
Given some specific setting of the function space with boundary condition, we consider the following transmission problem.
    \begin{equation}
    \begin{cases} 
        \mathcal{L}_{\lambda, \mu} \mathbf{u_{\epsilon}} = 0 & \text{in } \Omega \setminus \overline{D_{\epsilon}}, \\ 
        \mathcal{L}_{\widetilde{\lambda}, \widetilde{\mu}} \mathbf{u_{\epsilon}} = 0 & \text{in } D_{\epsilon}, \\
        \mathbf{u_{\epsilon}}|_{-} = \mathbf{u}|_{+} \text{ and } \left.\frac{\partial \mathbf{u_{\epsilon}}}{\partial \nu_{(\widetilde{\lambda}, \widetilde{\mu})}}\right|_{-} = \left.\frac{\partial \mathbf{u_{\epsilon}}}{\partial \nu_{(\lambda, \mu)}}\right|_{+} & \text{on } \partial D_{\epsilon},  \\
        \left.\frac{\partial \mathbf{u_{\epsilon}}}{\partial \nu_{(\lambda, \mu)}}\right|_{\partial \Omega} = g \in H_{\mathbb{R}}^{-\frac{1}{2}}(\partial \Omega) \quad \text{and} \quad \mathbf{u_{\epsilon}}|_{\partial \Omega} \in H_{\mathbb{R}}^{\frac{1}{2}}(\partial \Omega).
    \end{cases}
    \end{equation}

    This problem is equivalent to
    \begin{equation}
    \begin{cases}
    \mathcal{L}_{\lambda(x),\mu(x)} \mathbf{u_{\epsilon}} = \nabla \cdot \left[\lambda(x)(\nabla \cdot \mathbf{u_{\epsilon}})\mathbb{I}_d + 2\mu(x)\mathbb{D}(\mathbf{u_{\epsilon}})\right] = 0 \quad \text{in } \Omega,\\

    \left.\frac{\partial \mathbf{u_{\epsilon}}}{\partial \nu_{\lambda,\mu}}\right|_{\partial \Omega} = g \in H_{\mathbb{R}}^{-\frac{1}{2}}(\partial \Omega), \quad \mathbf{u_{\epsilon}}|_{\partial \Omega} \in H_{\mathbb{R}}^{\frac{1}{2}}(\partial \Omega),
    \end{cases}
    \end{equation}
    where $\lambda(x)=\lambda I_{\Omega \setminus D_{\epsilon}} + \widetilde{\lambda} {I_D}_{\epsilon}$, $\mu(x)=\mu I_{\Omega \setminus D_{\epsilon}} + \widetilde{\mu} {I_D}_{\epsilon}$. The existence and uniqueness of this problem comes from elliptic condition with the help of Lax-Milgram Theorem.
    
 Let $\widetilde{\lambda}\to\infty$ and $\widetilde{{\mu}}$ fixed, then we arrive at the equations about the \textbf{incompressible inclusion limit}. In this case, the transmission problem is a coupled \textbf{Lame-Stokes system}:
    \begin{equation}
    \begin{cases} 
    \mathcal{L}_{\lambda, \mu} \mathbf{u_{\epsilon}} = 0 & \text{in } \Omega \setminus \overline{D_{\epsilon}}, \\ 
    \mathcal{L}_{\widetilde{\mu}}(\mathbf{u_{\epsilon}}, p_{\epsilon}) = 0 \:\text{and} \operatorname{div} \:\mathbf{u_{\epsilon}}=0  & \text{in } D_{\epsilon}, \\ 
    
    \mathbf{u_{\epsilon}}|_- = \mathbf{u_{\epsilon}}|_+ \: \text{and} \: \left. \frac{\partial (\mathbf{u_{\epsilon}}, p_{\epsilon})}{\partial \nu_{(\infty, \widetilde{\mu})}} \right|_- = \left. \frac{\partial \mathbf{u_{\epsilon}}}{\partial \nu_{(\lambda, \mu)}} \right|_+ & \text{on } \partial D_{\epsilon}, \\ 
    
    \left. \frac{\partial \mathbf{u_{\epsilon}}}{\partial \nu_{(\lambda, \mu)}} \right|_{\partial \Omega} = g \in H_{\mathbb{R}}^{-\frac{1}{2}} (\partial \Omega) \quad\text{and} \quad \mathbf{u_{\epsilon}}|_{\partial \Omega} \in H_{\mathbb{R}}^{\frac{1}{2}} (\partial \Omega),
    \end{cases}
    \end{equation}
    where $\mathcal{L}_{\widetilde{\mu}}(\mathbf{u_{\epsilon}}, p_{\epsilon}) = \widetilde{\mu} \Delta \mathbf{u_{\epsilon}} + \nabla p_{\epsilon}$ denotes the Stokes operator with viscosity constant $\widetilde{\mu}$ and $p_{\epsilon}$ is the pressure field. Its exterior derivative is defined as $
\frac{\partial (u, p)}{\partial \nu_{(\infty, \mu)}} \bigg|_{-} := pN + 2\mu D(u)N.$

    We are interested in this Lame-Stokes system for the High contrast limit and the corresponding error estimation which was proved by Fu X. and Jing W~\citep{homogenization} .
$$\Vert \mathbf{u}_{lim}-\mathbf{u}_{\epsilon} \Vert \leq \frac{C}{\widetilde{\lambda}} \Vert g \Vert_{ H^{-\frac{1}{2}} (\partial \Omega)}
$$

\subsubsection{Summary of Proof Results}
We use our agent based on two LRMs DeepSeek-R1 and OpenAI o4-mini, and we get different experimental outputs.

AIM explores the energy control of multiple functions and derived several key conclusions by applying the variational method. The agent first establishes energy control mechanisms through rigorous mathematical derivation and analysis for both the solution $\mathbf{u{\epsilon}}$ and pressure term $p_{\epsilon}$ of the governing equations. Building on this theoretical foundation, the analysis further develops an error control framework by investigating perturbation propagation mechanisms. Through careful quantification of error accumulation patterns, AIM ultimately derives a global error estimate. 

Overall, this proof process has uncovered novel and correct conclusions, providing consistent global control over the solutions of the equation—a pleasantly surprising outcome. However, some notable issues can be observed. During intermediate steps, the model evidently could have derived stronger conclusions but prematurely halted its exploration. Instead, it reengaged in intricate reasoning to reattain the desired conclusions through redundant complexity.

When we use OpenAI o4-mini, the output is different. The proof process of this model is more open and comprehensive. The agent has taken into account specific spatial characteristics and homogenization properties. It is evident that the agent engaged in more extensive thinking and exploration, yielding additional intermediate conclusions. However, there are still instances of imprecision and errors that emerged during the process.

\subsubsection{Detailed Analysis of Selected Proofs Generated by DeepSeek-R1}
We firstly explain some results of AIM based on DeepSeek-R1 .

\begin{framed}
    \small
**Step 3: Corrected Test with \(\mathbf{v} = \nabla \phi\)**  
Substituting \(\mathbf{v}\) into the variational equation:
\[
-\widetilde{\lambda} \int_{D_\epsilon} (\operatorname{div} \mathbf{u}_\epsilon)^2 dx + 2\int_\Omega \mu(x) D\mathbf{u}_\epsilon : D(\nabla \phi) \, dx = \int_{\partial\Omega} g \cdot \nabla \phi \, ds. 
\]
**Step 3: Corrected Test with \(\mathbf{v} = \nabla \phi\)**  
Substituting \(\mathbf{v}\) into the variational equation:
\[
-\widetilde{\lambda} \int_{D_\epsilon} (\operatorname{div} \mathbf{u}_\epsilon)^2 dx + 2\int_\Omega \mu(x) D\mathbf{u}_\epsilon : D(\nabla \phi) \, dx = \int_{\partial\Omega} g \cdot \nabla \phi \, ds. \tag{2}
\]
Rearranging and taking absolute values:
\[
\widetilde{\lambda} \|\operatorname{div} \mathbf{u}_\epsilon\|_{L^2(D_\epsilon)}^2 \leq \left| 2\int_\Omega \mu D\mathbf{u}_\epsilon : D(\nabla \phi) dx \right| + \left| \int_{\partial\Omega} g \cdot \nabla \phi \, ds \right|.
\]
Using Hölder inequalities and trace duality:
\[
\left| 2\int_\Omega \mu D\mathbf{u}_\epsilon : D(\nabla \phi) dx \right| \leq C \|g\|_{H^{-1/2}} \|\operatorname{div} \mathbf{u}_\epsilon\|_{L^2(D_\epsilon)},
\]
\[
\left| \int_{\partial\Omega} g \cdot \nabla \phi \, ds \right| \leq C \|g\|_{H^{-1/2}} \|\operatorname{div} \mathbf{u}_\epsilon\|_{L^2(D_\epsilon)}.
\]

**Step 4: Final Estimate**  
Combining these bounds:
\[
\widetilde{\lambda} \|\operatorname{div} \mathbf{u}_\epsilon\|_{L^2(D_\epsilon)}^2 \leq 2C \|g\|_{H^{-1/2}} \|\operatorname{div} \mathbf{u}_\epsilon\|_{L^2(D_\epsilon)}.
\]
Dividing by \(\|\operatorname{div} \mathbf{u}_\epsilon\|_{L^2(D_\epsilon)}\) yields:
\[
\|\operatorname{div} \mathbf{u}_\epsilon\|_{L^2(D_\epsilon)} \leq \frac{2C}{\widetilde{\lambda}} \|g\|_{H^{-1/2}},
\]
completing the proof with modified constant \(C\).  

\textcolor{Green}{$\textbf{[Correct]}$ \small Use inequality techniques to derive the correct key conclusions from variational equations.}

\end{framed}

The agent investigated the properties of divergence, leveraging the framework of variational equations and rigorous mathematical derivations to establish a uniform upper bound on the $\operatorname{div} \mathbf{u}_\epsilon$. This deprivation process is correct and rigorous.

\begin{framed}
    \small
**Step 1: Variational Formulations**  
The original problem (9) has the weak form: Find \(\mathbf{u}_\epsilon \in H^1(\Omega)\) such that for all \(\mathbf{v} \in H^1(\Omega)\),
\[
\int_\Omega \left[ \lambda(x) \operatorname{div} \mathbf{u}_\epsilon \operatorname{div} \mathbf{v} + 2\mu(x) D(\mathbf{u}_\epsilon) : D(\mathbf{v}) \right] dx = \int_{\partial\Omega} g \cdot \mathbf{v} \, ds.
\]

\textcolor{Green}{$\textbf{[Correct]}$ \small Consider the weak form of this equation.}
\end{framed}

\begin{framed}
    \small
The limit problem (10) is a saddle-point problem: Find \((\mathbf{u}_{\text{lim}}, p_{\text{lim}}) \in H^1(\Omega) \times L^2(D_\epsilon)\) such that for all \(\mathbf{v} \in H^1(\Omega)\) and \(q \in L^2(D_\epsilon)\),
\[
\begin{cases}
\int_{\Omega \setminus D_\epsilon} \left[ \lambda \operatorname{div} \mathbf{u}_{\text{lim}} \operatorname{div} \mathbf{v} + 2\mu D(\mathbf{u}_{\text{lim}}) : D(\mathbf{v}) \right] dx + \int_{D_\epsilon} \left[ 2\widetilde{\mu} D(\mathbf{u}_{\text{lim}}) : D(\mathbf{v}) - p_{\text{lim}} \operatorname{div} \mathbf{v} \right] dx \\
=\int_{\partial\Omega} g \cdot \mathbf{v} \, ds, \\
\int_{D_\epsilon} q \operatorname{div} \mathbf{u}_{\text{lim}} \, dx = 0.
\end{cases}
\]

\textcolor{Blue}{$\textbf{[Vague]}$ \small Here consider the limited function, but it does not show the detailed deprivation of the the existence and
property of the function.}
\end{framed}
\begin{framed}
\small
**Step 2: Subtract the Weak Forms**  
Let \(\mathbf{w}_\epsilon = \mathbf{u}_\epsilon - \mathbf{u}_{\text{lim}}\). Subtract the weak form of the limit problem from the original problem, tested with \(\mathbf{v} = \mathbf{w}_\epsilon\):
\[
\begin{cases}
    \int_{\Omega \setminus D_\epsilon} \left[ \lambda (|\operatorname{div} \mathbf{w}_\epsilon|^2 + 2\mu |D(\mathbf{w}_\epsilon)|^2) \right] dx + \int_{D_\epsilon} \left[ \widetilde{\lambda} |\operatorname{div} \mathbf{u}_\epsilon|^2 + 2\widetilde{\mu} |D(\mathbf{w}_\epsilon)|^2 \right] dx \\
= \int_{D_\epsilon} p_{\text{lim}} \operatorname{div} \mathbf{w}_\epsilon \, dx.
\end{cases}
\]

Since \(\operatorname{div} \mathbf{u}_{\text{lim}} = 0\) in \(D_\epsilon\), substitute \(\operatorname{div} \mathbf{w}_\epsilon = \operatorname{div} \mathbf{u}_\epsilon\) in \(D_\epsilon\):
\[
\int_{D_\epsilon} p_{\text{lim}} \operatorname{div} \mathbf{w}_\epsilon \, dx = \int_{D_\epsilon} p_{\text{lim}} \operatorname{div} \mathbf{u}_\epsilon \, dx.
\]
\textcolor{Green}{$\textbf{[Correct]}$ \small According to the above weakly form and the setting condition of the problem, the model correctly obtains the above key conclusion.}

\end{framed}

This proof process demonstrates the agent's rigorous and precise mathematical reasoning capabilities. By analyzing the properties of limit functions, the agent exhibits a holistic understanding of the problem. However, the critical steps lack detailed derivations to substantiate the reasoning, which diminishes the transparency of the logical progression.

\begin{framed}
\small
    Under the given assumptions, the difference \(\mathbf{w}_\varepsilon = \mathbf{u}_{\text{lim}} - \mathbf{u}_\varepsilon\) between the solutions of the limit problem (10) and the original problem (9) satisfies the homogenized error estimate in the energy norm weighted by the Lamé parameters:
\[
\left( \int_\Omega \left[ \lambda(x) |\operatorname{div} \mathbf{w}_\varepsilon|^2 + 2\mu(x) |\mathcal{D}(\mathbf{w}_\varepsilon)|^2 \right] dx \right)^{1/2} \leq \frac{C}{\sqrt{\widetilde{\lambda}}} \|g\|_{H^{-\frac{1}{2}}(\partial\Omega)},
\]
where \(C\) is independent of \(\widetilde{\lambda}\). This implies that the error in the energy norm decays proportionally to \(\widetilde{\lambda}^{-\frac{1}{2}}\).

\textbf{correctness}: True
\end{framed}
The agent here incorporates a global control of errors, representing a novel conclusion in error control. Though the derived bound lacks optimal strength, it reflects the agent's valid mathematical exploration in establishing this conclusion.

\subsubsection{Detailed Analysis of Selected Proofs Generated by OpenAI o4-mini}
The following results are some output based on OpenAI o4-mini.
\begin{framed}
\small
Step 1. Weak formulation on $W \times Q$.  
Since $u_{\text{ext}}$ and $u_{\text{int}}$ must agree on $\partial D_\varepsilon$, we assemble them into a single global field $u \in H^1(\Omega)$ which on $\Omega_{\text{ext}}$ carries the elasticity operator and on $D_\varepsilon$ carries the incompressible Stokes operator. The space $W$ enforces both the $H^1$-regularity across $\partial D_\varepsilon$ (by matching traces) and the rigid-motion orthogonality on $\partial \Omega$. The pressure $p$ lives only in $D_\varepsilon$ and has zero mean.

The variational equations, obtained by integration by parts in $\Omega_{\text{ext}}$ and in each inclusion, read:  
\begin{align*}
(a_{\text{ext}} + a_{\text{int}})(u, v) + b(v, p) &= L(v) \quad \text{for all } v \in W, \\
b(u, q) &= 0 \quad \text{for all } q \in Q,
\end{align*}
and incorporate the continuity of tractions and of displacements automatically.

\textcolor{Green}{\textbf{[Correct]} \small Here, the agent actively construct the specific proof design.}
\end{framed}
\begin{framed}
\small
Step 2. Continuity of the bilinear forms.  
By Cauchy–Schwarz and standard trace estimates, there is $M > 0$ independent of $\varepsilon$ such that  
\[
|a_{\text{ext}}(u, v)| + |a_{\text{int}}(u, v)| + |b(v, p)| \leq M \|u\|_{H^1(\Omega)} \|v\|_{H^1(\Omega)} + M \|p\|_{L^2(D_\varepsilon)} \|v\|_{H^1(\Omega)}.
\]

Step 3. Coercivity of $A := a_{\text{ext}} + a_{\text{int}}$ on $\text{Ker}\, B$.  
The kernel of $B$ is $K = \{ v \in W : \text{div}\, v = 0 \text{ in } D_\varepsilon \}$. For any $v \in K$ we have  
\[
A(v, v) = \int_{\Omega_{\text{ext}}} [2\mu |D(v)|^2 + \lambda |\text{div}\, v|^2] \, dx + \int_{D_\varepsilon} 2\tilde{\mu} |D(v)|^2 \, dx.
\]
Define $\mu_{\text{min}} = \min\{\mu, \tilde{\mu}\} > 0$. Since $v \in H^1(\Omega)$ and $\Omega$ is Lipschitz, the global Korn inequality on $\Omega$ gives  
\[
\|v\|^2_{H^1(\Omega)} \leq C_K \|D(v)\|^2_{L^2(\Omega)}
\]
with $C_K$ independent of $\varepsilon$. Hence  
\[
A(v, v) \geq 2\mu_{\text{min}} \|D(v)\|^2_{L^2(\Omega)} \geq \left( \frac{2\mu_{\text{min}}}{C_K} \right) \|v\|^2_{H^1(\Omega)}.
\]
Thus $A$ is coercive on $K$ with constant $\alpha = 2\mu_{\text{min}} / C_K$, independent of $\varepsilon$.

\textcolor{Green}{\textbf{[Correct]} \small The derivation process presented herein is correct.}
\end{framed}
\begin{framed}
\small
Step 4. Inf–sup stability of $B$ on $W \times Q$.  
We must show  
\[
\inf_{q \in Q} \sup_{v \in W} \frac{b(v, q)}{\|v\|_{H^1(\Omega)} \|q\|_{L^2(D_\varepsilon)}} \geq \gamma > 0
\]
with $\gamma$ independent of $\varepsilon$.  
But $b(v, q) = -\int_{D_\varepsilon} q \, \text{div}\, v$, and $\text{div}$ acts only in $D_\varepsilon$. Restricting to test-functions $v$ supported in $D_\varepsilon$ we recover the classical Stokes inf–sup condition on the perforated region $D_\varepsilon$. Since $D_\varepsilon$ is a union of $\varepsilon$-scaled, well-separated Lipschitz inclusions, the local inf–sup constant on each inclusion is scale-invariant and the global constant $\gamma$ (the minimum over finitely many identical blocks) is independent of $\varepsilon$.

Step 5. Conclusion by Babuška–Brezzi.  
By Steps 2–4 the pair of forms $(A, B)$ on $W \times Q$ satisfies the continuity, coercivity on $\text{Ker}\, B$, and the inf–sup condition with constants $M, \alpha, \gamma$ independent of $\varepsilon$. Hence there is a unique solution $(u, p) \in W \times Q$ and the a priori bound  
\[
\|u\|_{H^1(\Omega)} + \|p\|_{L^2(D_\varepsilon)} \leq C \sup_{(v, q) \neq 0} \frac{L(v) + 0}{\|v\|_{H^1(\Omega)} + \|q\|_{L^2(D_\varepsilon)}} = C \sup_{v \in W} \frac{\int_{\partial \Omega} g \cdot v \, dS}{\|v\|_{H^1(\Omega)}} = C \|g\|_{H^{-1/2}(\partial \Omega)},
\]
holds with $C$ depending only on $d, \lambda, \mu, \tilde{\mu}$, the Lipschitz character of $\Omega$, and the reference inclusion $\omega$, but \textbf{not} on $\varepsilon$. This completes the proof.

\textcolor{Blue}{\textbf{[Vague]} \small Here, more detailed explanation is required for the deprivation and the application of the theorems.}
\end{framed}

The agent here provides clear proof steps and precise key conclusions. Although certain derivational details are omitted, the underlying reasoning framework and final conclusions remain rigorously accurate and align with the problem's requirements. And we can also find that the agent verifies the condition of theorems. This demonstrates that the agent has achieved a commendable level of understanding and exploratory capability in mathematical theory.

\begin{framed}
    \small
    
Next consider the difference
\[
  w := \mathbf u_{\epsilon}-\mathbf u_{\lim}
  \quad\text{on }D.
\]
Then \(w\in H^1(D)\) solves the compressible‐elastic Neumann problem
\[
  \widetilde\mu\,\Delta w
  +(\widetilde\lambda+\widetilde\mu)\,\nabla(\mathrm{div}\,w)
  =0\quad\text{in }D,
  \qquad
  \sigma_{(\widetilde\lambda,\widetilde\mu)}(w)\,N
  = J_\epsilon
  \quad\text{on }\partial D,
\]
with \(\int_{\partial D}J_\epsilon\cdot r=0\) for every rigid motion \(r\).  By Lemma 29 (Boundary‐Trace Rate–Improvement via First‐Order Asymptotics) on the \(C^{1,1}\) domain \(D\), there is \(C_4\) depending only on \(D,d,\widetilde\mu\) and the \(C^{1,1}\)‐character of \(\partial D\), such that
\[
  \|\,w\|_{H^{1/2}(\partial D)}
  \;\le\;
  C_4\,\widetilde\lambda^{-1}\,
  \|\,J_\epsilon\|_{H^{-1/2}(\partial D)}.
\]
Combining with \eqref{J-bound} yields
\[
  \|\mathbf u_{\epsilon}-\mathbf u_{\lim}\|_{H^{1/2}(\partial D)}
  =\|w\|_{H^{1/2}(\partial D)}
  \;\le\;
  C\,\widetilde\lambda^{-1}\,
  \|g\|_{H^{-1/2}(\partial\Omega)},
\]
with \(C=C_3\,C_4\) independent of \(\epsilon,\widetilde\lambda\).  Since there are finitely many inclusions, the same bound holds uniformly over each \(\partial D_\epsilon^i\).  This completes the proof.

\textcolor{Green}{\textbf{[Correct]} \small The conclusion presented herein is methodologically derived through an extension of previously established procedural frameworks.}
\end{framed}

Based on previously established conclusions, the AI agent conducts an extensive analysis of error control mechanisms. The derivation process and the modifications to the equations demonstrate methodological correctness. However, the absence of explicit derivational specifics warrants further elaboration.

\subsection{Homogenization Problem}
\subsubsection{Problem Description}
The setting of this problem is the same as the High Contrast Limit problem. But we concentrate on another open problem.
We still consider this transmission problem
    \begin{equation}
    \begin{cases} 
    \mathcal{L}_{\lambda, \mu} \mathbf{u_{\epsilon}} = 0 & \text{in } \Omega \setminus \overline{D_{\epsilon}}, \\ 
    \mathcal{L}_{\widetilde{\mu}}(\mathbf{u_{\epsilon}}, p_{\epsilon}) = 0 \:\text{and} \operatorname{div} \:\mathbf{u_{\epsilon}}=0  & \text{in } D_{\epsilon}, \\ 
    
    \mathbf{u_{\epsilon}}|_- = \mathbf{u_{\epsilon}}|_+ \: \text{and} \: \left. \frac{\partial (\mathbf{u_{\epsilon}}, p_{\epsilon})}{\partial \nu_{(\infty, \widetilde{\mu})}} \right|_- = \left. \frac{\partial \mathbf{u_{\epsilon}}}{\partial \nu_{(\lambda, \mu)}} \right|_+ & \text{on } \partial D_{\epsilon}, \\ 
    
    \left. \frac{\partial \mathbf{u_{\epsilon}}}{\partial \nu_{(\lambda, \mu)}} \right|_{\partial \Omega} = g \in H_{\mathbb{R}}^{-\frac{1}{2}} (\partial \Omega) \quad\text{and} \quad \mathbf{u_{\epsilon}}|_{\partial \Omega} \in H_{\mathbb{R}}^{\frac{1}{2}} (\partial \Omega),
    \end{cases}
    \end{equation}
    where $\mathcal{L}_{\widetilde{\mu}}(\mathbf{u_{\epsilon}}, p_{\epsilon}) = \widetilde{\mu} \Delta \mathbf{u_{\epsilon}} + \nabla p_{\epsilon}$ denotes the Stokes operator with viscosity constant $\widetilde{\mu}$ and $p_{\epsilon}$ is the pressure field. Its exterior derivative is defined as $
\frac{\partial (u, p)}{\partial \nu_{(\infty, \mu)}} \bigg|_{-} := pN + 2\mu D(u)N.$

   Now we let $\epsilon \to 0 $, so that the scale of the cell is tend to be zero. We wonder what homogenization equation does the limited solution $\mathbf{u}_{lim}$ satisfies. At the same time, whether the estimate between the original solution and the limited solution has the form $C{\epsilon}^{\alpha}\Vert g \Vert_{H^{-\frac{1}{2}} (\partial \Omega)}$ for some $\alpha \in (0,1)$ .
   
We use AIM based on OpenAI o4-mini to explore this open problem.

\subsubsection{Summary of Proof Results}

The exploration process outlined by the agent comprises two components: homogenization equations and error control. Guided by the proposed two-scale expansion, AIM conducts an analysis of the homogenization equations. Specifically, it rigorously establishes the convergence of the expanded solutions and explicitly characterizes the mathematical form of the homogenized equations.

Building upon the homogenized equations, the agent subsequently analyzes and derives error estimates through systematic reasoning. Throughout the experimental results, we observe that the agent has yielded valid conclusions while simultaneously revealing certain issues. In addressing this open-ended problem, the agent’s methodological approach overlooks crucial specific properties of the physical setting, leading to some improper process in the proofs of lemmas. Moreover, likely due to constraints inherent to the invoked model, the reasoning process remains deficient in derivational details and overrelies on conclusory statements, undermining the reproducibility and robustness of the arguments.

\subsubsection{Detailed Analysis of Selected Proofs}

\begin{framed}
    \small

1. (Cellwise Korn with uniform constant.)
Since $Y_f=Y\setminus\overline{\omega}$ is a bounded Lipschitz set, the classical Korn inequality on $Y_f$ gives
\begin{equation*}
\text{for all } v\in H^1(Y_f;\mathbb{R}^d),
\quad \|v\|_{H^1(Y_f)} \leq C_0 \left(\|v\|_{L^2(Y_f)} + \|\operatorname{sym} \nabla v\|_{L^2(Y_f)}\right),
\end{equation*}
with $C_0$ depending only on $Y_f$. By the scaling $x=\varepsilon y$ one deduces that on each $\varepsilon$-cell $\varepsilon(Y_f+k)$
\begin{equation*}
\text{for all } v\in H^1(\varepsilon(Y_f+k)),
\quad \|v\|_{H^1(\varepsilon(Y_f+k))} \leq C_0 \left(\|v\|_{L^2(\varepsilon(Y_f+k))} + \|\operatorname{sym} \nabla v\|_{L^2(\varepsilon(Y_f+k))}\right).
\end{equation*}

\textcolor{Blue}{$\textbf{[Vague]}$ \small Here the agent considers the Korn inequality but the derivation process is not detailed enough.}
\end{framed}
\begin{framed}
\small
2. (Partition of unity subordinate to the periodic tiling.)
Let $\{\psi_k\}_{k\in K_\varepsilon}$ be a smooth partition of unity on $\Omega$:
\begin{itemize}
\item each $\psi_k\in C_c^\infty(\varepsilon(k+Y))$,
\item $0\leq\psi_k\leq 1$, $\sum_k\psi_k\equiv 1$ on $\Omega$ except in a boundary strip of width $\leq 2\varepsilon$,
\item and $|\nabla \psi_k|\leq C/\varepsilon$.
\end{itemize}
We also choose a finite set of boundary functions $\{\psi_b\}$ supported in the $\varepsilon$-neighborhood of $\partial\Omega$ so that $\sum_k\psi_k+\sum_b\psi_b\equiv 1$ on $\Omega$. By standard Korn--Poincaré on the fixed boundary patches (of size $O(\varepsilon)$), one shows likewise
\begin{equation*}
\text{for each boundary patch with cut-off } \psi_b,
\quad \|\psi_b u\|_{H^1} \leq C \left(\|u\|_{L^2(\operatorname{supp} \psi_b)} + \|\operatorname{sym} \nabla u\|_{L^2(\operatorname{supp} \psi_b)}\right).
\end{equation*}
\end{framed}

Here, the agent has constructed a uniform estimate of the solution in the boundary region. It is evident that the mathematical theories of Korn's inequality and partition of unity were leveraged. It can also  be observed that the agent's handling of spatial configurations implicitly employs established mathematical frameworks. However, the detailed derivations underpinning these steps remain unwritten.

\begin{framed}
    \small
    2.  Extraction of limits.  
   By the reflexivity of the three spaces, there exist, along a subsequence,  
     \[ U^*(x, y) \in L^2(\Omega; H^1(Y_f)), \]  
     \[ W(x, y) \in L^2(\Omega \times Y_f; \mathbb{R}^{d \times d}) \]  
   such that as $\varepsilon \to 0$:  
     \[ T_\varepsilon u_\varepsilon \rightharpoonup U^* \text{ weakly in } L^2(\Omega; H^1(Y_f)), \]  
     \[ \nabla_x T_\varepsilon u_\varepsilon \rightharpoonup \nabla_x U^*, \quad  
     \nabla_y T_\varepsilon u_\varepsilon \rightharpoonup W \text{ weakly in } L^2(\Omega \times Y_f). \]  

\textcolor{Green}{$\textbf{[Correct]}$ \small The accurate convergence of the function spaces.}
\end{framed}
\begin{framed}
\small
3.  Independence of $U^*$ on $y$ and strong $L^2$-convergence.  
   Because $\|\nabla_y T_\varepsilon u_\varepsilon\|_{L^2(\Omega \times Y_f)}$ remains bounded but multiplied by $\varepsilon$ vanishes in the original variables, one shows by the unfolding-to-two-scale characterization that $\partial_{y_i} U^* = 0$ in the distributional sense. Hence $U^*(x, y) =: u_0(x)$ is independent of $y$, and $u_0 \in H^1(\Omega)$. Moreover, the standard compact embedding $H^1 \to L^2$ in $x$ implies  
     \[ T_\varepsilon u_\varepsilon \to u_0 \text{ strongly in } L^2(\Omega \times Y_f), \]  
   which by unfolding-to-two-scale equivalence yields  
     \[ u_\varepsilon \to u_0 \text{ strongly in } L^2(\Omega). \]  

\textcolor{Blue}{$\textbf{[Vague]}$  \small The process of the two-scale expansion here is not detailed enough. The proof of the convergence of the function lacks details.}
\end{framed}

This line of mathematical reasoning is fundamentally valid. The analysis here focuses on the convergence properties of the expanded solutions, but the agent fails to provide critical details regarding the structure and control of the expansion functions themselves. Specifically, there is a lack of norm-controlled estimates.

\begin{framed}
    \small
    Then as $\varepsilon\to0$ one has, up to a subsequence,  
\quad $u_\varepsilon \rightharpoonup u_0$ in $H^1(\Omega)$, \quad $u_\varepsilon\to u_0$ in $L^2(\Omega)$,  
\quad $p_\varepsilon \stackrel{\text{two-scale}}{\to} p_1(x,y)$ in $\Omega\times Y_i$,  
where $u_0\in H^1_R(\Omega;\mathbb{R}^d)$ is the unique solution of the homogenized Lam\'e system  
\quad $-\mathrm{div}_x[C^{\mathrm{hom}}\,D_x(u_0)] = 0$ in $\Omega$, \quad $C^{\mathrm{hom}}\,D_x(u_0)\cdot n = g$ on $\partial\Omega$,  
and the effective fourth-order tensor $C^{\mathrm{hom}}$ is given by the following periodic cell transmission problem: for each fixed symmetric $E\in\mathbb{R}^{d\times d}_{\mathrm{sym}}$ find  
\quad $(\chi_e^E,\chi_i^E,\pi^E)\in H^1_{\mathrm{per}}(Y_e;\mathbb{R}^d)\times H^1_{\mathrm{per}}(Y_i;\mathbb{R}^d)\times L^2_{\mathrm{per}}(Y_i)/\mathbb{R}$  
solving  
\begin{enumerate}
    \item[(1)] In the elastic cell $Y_e$:  
           $-\mathrm{div}_y[C\,(E + D_y\,\chi_e^E)] = 0$, \quad $\mathrm{div}_y(Ey+\chi_e^E)=0$,  
    \item[(2)] In the fluid cell $Y_i$:  
           $-\mu\,\Delta_y\,\chi_i^E + \nabla_y\,\pi^E = 0$, \quad $\mathrm{div}_y(Ey+\chi_i^E)=0$,  
    \item[(3)] On the material interface $\Gamma=\partial Y_i$:  
           $\chi_e^E = \chi_i^E$,  
           $[C\,(E+D_y\,\chi_e^E)\cdot n] = [2\mu\,D_y\,\chi_i^E - \pi^E\,I]\cdot n$,  
    \item[(4)] $(Ey+\chi^E)$ is $Y$--periodic, with zero-mean normalization on $\chi_e^E$ and $\chi_i^E$.  
\end{enumerate}

Then  
\quad $C^{\mathrm{hom}}\,E : E$  
\quad $= \int_{Y_e}(E+D_y\,\chi_e^E):C(E+D_y\,\chi_e^E)\,\mathrm{d}y$  
\quad $\quad + \int_{Y_i}2\mu\,D_y\,\chi_i^E : D_y\,\chi_i^E\,\mathrm{d}y$.
\end{framed}

This represents that the agent's conclusions on the cell problem. We observe that the agent correctly adopts the homogenization framework to derive the key homogenized equations. While the conclusions here are broadly reasonable, localized issues can be identified, likely due to the agent's insufficient clarity regarding the physical configuration of individual unit cells (e.g., boundary interactions, material periodicity assumptions). These ambiguities in microscale definitions may propagate into approximations at the macroscale.

\begin{framed}
    \small 4. Homogenized tensor and macroscopic equation  
By choosing in the cell system the test-function $\theta = \chi^E$, one derives the energy representation  
\[ C^{\mathrm{hom}} E:E  
= \int_{Y_e} (E + D_y \chi_e^E):C(E + D_y \chi_e^E) \, dy  
+ \int_{Y_i} 2\mu D_y \chi_i^E : D_y \chi_i^E \, dy. \]  
On the other hand, testing the original problem against $\varphi(x)$ alone, passing to the limit shows  
\[ \int_\Omega C^{\mathrm{hom}} D_x u_0 : D_x \varphi = \langle g, \varphi \rangle \quad \forall \varphi \in H^1_R(\Omega; \mathbb{R}^d), \]  
hence $u_0$ solves  
\[ -\mathrm{div}_x [C^{\mathrm{hom}} D_x u_0] = 0 \text{ in } \Omega, \]  
\[ C^{\mathrm{hom}} D_x u_0 \cdot n = g \text{ on } \partial \Omega. \]

\textcolor{Blue}{$\textbf{[Vague]}$ \small Here the agent gets the homogenization equation. But whether this conclusion is correct or not, more verification processes are needed.}
\end{framed}
\begin{framed}
\small
5. Boundary-layer estimates  
By constructing $\varepsilon$--tubular cut-off functions near $\partial D_\varepsilon$ and $\partial \Omega$ (supported in $O(\varepsilon)$--neighborhoods) and using the uniform $H^1$--bound together with trace and Poincar\'e inequalities, one shows that all interface and boundary-layer remainders vanish as $o(1)$ when $\varepsilon \to 0$ (see Lemma A.5 for details).

\textcolor{Blue}{$\textbf{[Vague]}$ \small The conditions for using the theorem need to be verified.}
\end{framed}
\begin{framed}
\small
6. Error estimate and strong convergence  
Define the two-scale corrector  
\[ u_\varepsilon^{\mathrm{app}}(x) = u_0(x) + \varepsilon \chi^{D_x u_0(x)}(x/\varepsilon) \text{ in } \Omega_\varepsilon, \]  
extended in each $D_\varepsilon$ by $Ey + \chi_i^E(y)$. A C\'ea-lemma--type argument in the perforated domain (again relying on uniform Korn and Stokes inf--sup, Lemma A.1--A.2) yields for some $\alpha \in (0,1)$  
\[ \|u_\varepsilon - u_\varepsilon^{\mathrm{app}}\|_{H^1(\Omega_\varepsilon)} \leq C \varepsilon^\alpha \|g\|_{H_R^{-1/2}(\partial \Omega)}. \]  
From this one deduces $u_\varepsilon \to u_0$ strongly in $L^2(\Omega)$ (and even in $H^1(\Omega_\varepsilon)$ up to the corrector).  

This completes a fully rigorous derivation of the homogenized elastic limit and the explicit formula for $C^{\mathrm{hom}}$.
\end{framed}

Based on the homogenized equations, the agent conducts an analysis of the solutions. Although critical derivation steps are omitted, the following can be inferred.
The agent employs the two-scale expansion structure consistent with homogenization theory.
And techniques like energy estimates, Korn's inequality, and weak convergence arguments are implicitly utilized to justify convergence. The final results align with theoretical expectations, suggesting the agent captures the essential mathematical machinery.
However, the absence of explicit error bounds for truncation terms and insufficient discussion on the regularity of expansion functions weaken the analytical completeness. The conclusions remain plausible but require rigorous verification under specific functional space assumptions.
\begin{framed}
\small\textbf{Step 2: Construction of the two--scale approximation.}
Let $\chi^E(y) = (\chi_e^E, \chi_i^E)(y)$ be the periodic cell--correctors of Lemma~3 for each macroscopic strain $E$. Define the two--scale ansatz
\[
    U_\varepsilon(x) = u_0(x) + \varepsilon \chi^{D_x u_0(x)}(x/\varepsilon)
\]
in $\Omega_\varepsilon$, extended by zero on $D_\varepsilon$. A standard calculation (see Lemma~4) shows that $U_\varepsilon$ nearly solves the $\varepsilon$--problem up to an $O(\varepsilon^\alpha)$ remainder in $H^{-1}$--norm.

\textbf{Step 3: Energy--error estimate.}
Set $w_\varepsilon = u_\varepsilon - U_\varepsilon$. Subtracting the weak forms of the $\varepsilon$--problem for $u_\varepsilon$ and the one nearly satisfied by $U_\varepsilon$ and testing with $w_\varepsilon$ itself yields, via coercivity and the uniform Korn bound,
\[
    \|w_\varepsilon\|_{H^1(\Omega_\varepsilon)}^2 \leq C \varepsilon^\alpha \|g\|_{H_R^{-1/2}(\partial \Omega)} \|w_\varepsilon\|_{H^1(\Omega_\varepsilon)}.
\]
Divide through by $\|w_\varepsilon\|_{H^1(\Omega_\varepsilon)}$ to conclude
\[
    \|u_\varepsilon - U_\varepsilon\|_{H^1(\Omega_\varepsilon)} \leq C \varepsilon^\alpha \|g\|_{H_R^{-1/2}(\partial \Omega)}.
\]

\textbf{Step 4: Passage to the limit.}
Since $U_\varepsilon \to u_0$ strongly in $H^1(\Omega_\varepsilon)$ (by periodicity and $u_0 \in H^2$), the triangle inequality immediately gives $u_\varepsilon \to u_0$ with the same rate.

This completes the proof of both the homogenized limit and the $O(\varepsilon^\alpha)$--rate.
    
\end{framed}

Finally, the agent consolidates the aforementioned lemmas to formulate complete proof steps and presents the key conclusions. But it does not get the real value of $\alpha$. It can be reasonably asserted that this reasoning framework adheres to normative mathematical logic and delineates potential avenues for addressing the problem.

\section{Discussion}

AIM is a simple agent framework that aims at autonomous mathematical research and discovery. With proper exploration and verification mechanisms, AIM already achieved impressive results in some open problems, and the same framework can be easily adopted for various research topics. However, our system still has many shortcomings in terms of reliability, that the proofs generated by AIM can not be directly accepted before manual review of an expert in math.

We also observed several drawbacks in our AIM system during experiments. These drawbacks may be alleviated through better prompt engineering or other techniques, but they might ultimately stem from the inherent limitations of current LLMs and will require substantial further research to be fully resolved.

\begin{itemize}
    \item \textbf{Repetitive Exploration}. Our agent often explores through the same direction and proposes a series of similar conjectures. Although in our experiments it can ultimately find the correct proof or obtain some useful findings, repetitive exploration still increases the cost, while reducing efficiency and the potential upper bound of performance. It is crucial to encourage the model to engage in more diverse exploration.
    \item \textbf{Deficiency in Comprehending Certain Mathematical Configurations}. Despite significant advancements in certain modalities, large models still demonstrate suboptimal comprehension of spatial configurations. Mathematical theories often involve intricate physical setups and precisely defined boundary conditions. Weak interpretative capability regarding these configurations results in analyses that lack clarity and mathematical precision.
    \item \textbf{Lack of Intermediate Steps}. Current LRMs do not tend to produce complete and rigorous proofs. This might not be a critical problem, but still impacts the proof quality. This may stem from their training methodology, as reinforcement learning primarily emphasizes the correctness of the final answer rather than the intermediate steps of the proof. This phenomenon can be mitigated through multiple PRV and refine processes, but this may not be an elegant solution and consumes a large amount of resources.
\end{itemize}
AIM has achieved some basic and important mathematical ability in mathematical research. It can achieve multiple functions in the research of mathematical theory. Our agent invokes foundational theorems within appropriate contexts and precisely applies the theorems. It also demonstrates the ability to recognize and interpret human-provided guidance and prompts, enabling it to execute multiple proof procedures as instructed. In the experiment, AIM executes logically coherent proof steps (e.g., energy estimates, asymptotic expansions) with formal consistency. Then it generates and proposes some hypotheses or conclusions aligned with empirical patterns and theoretical constraints. Our agent currently demonstrates the following functionalities in mathematical research:

\begin{itemize}
    \item  \textbf{Proof Assistant}. AIM can successfully complete proofs for various mathematical theories. It provides critical intermediate conclusions and core procedures, though certain derivation details may require supplementation. Mathematical researchers can utilize the agent's outputs to refine specific proof components, thereby accomplishing the comprehensive theoretical work.

    \item  \textbf{Verify and Prove Ideas}. In mathematical theory research, researchers often generate multiple proof concepts and conjectures. These potential approaches can be systematically input into AIM for automated proof attempt generation. Researchers can then evaluate the agent's output trajectories to determine the feasibility of validation methodologies, thereby refining their subsequent attempts to advance the theoretical construction.

    \item  \textbf{Exploration for Open Problem}. AIM can provide assistance in solving open mathematical problems. Our agent can directly attempt to solve these open problems and provide core conclusions, key processes, and reasonable inferences. Mathematicians can analyze the output results obtained, try more methods and ideas, and continue to use our agents to further explore the conclusions of this open problem.
\end{itemize}

\section{Future Works}

Our AIM is still in a very early stage. In addition to addressing the aforementioned limitations, we have also identified the following potential directions for future work.

\begin{itemize}
    \item \textbf{Memory Reflection Mechanism}. A proper memory reflection or summarize mechanism might help stabilize inference process and enable longer exploration trajectories. Also, this can help to retain more mathematical techniques for our agent to enhance its mathematical capacity.
    \item \textbf{Retrieve Augmentation}. For some frontier open problems our AIM agent need to retrieve through relevant papers to provide background context of it.
    \item \textbf{Model Capability Optimization}. Faced with the problems of the model itself, we can consider reducing or even solving its negative impact on mathematical theory problems. We can consider addressing the issue of insufficient detail in the model output process and unclear understanding of problem settings so that our agent can work out mathematical theory more rigorously and accurately.
    \item \textbf{Multi-Agent Collaboration}. We propose enhancing mathematical problem-solving capabilities by establishing a multi-agent collaborative framework. For instance, constructing dedicated agents specializing in evaluating the validity, rigor, and logical coherence of mathematical proofs. Such a structure would enable specialized agents to perform their designated roles during mathematical theorem exploration, achieving superior collaborative outcomes through division of labor.
    \item \textbf{Reinforcement Learning Methods}. Integrating reinforcement learning methodologies into proof derivation processes could significantly enhance system capabilities. We propose employing reinforcement learning techniques to fine-tune the models, thereby improving their deductive reasoning capacity in specific mathematical domains. This optimization would ultimately improve the overall performance of our intelligent agent system.
\end{itemize}

\section*{Acknowledgment}

 We would like to thank Zihan Tang, Zejin Lin, and Zhongwei Sun for their valuable assistance during the early exploration phase. Special thanks go to Associate Professor Wenjia Jing from Qiuzhen College, Tsinghua University, whose guidance in mathematics greatly supported this work.

\section*{Author Contributions}

Among the first three authors, Yanxing Huang was responsible for the system design and implementation. Yuanhang Liu reinvented some ideas after the implementation of the agent. Yuanhang Liu and Yanqiao Wang contributed to the early-stage experimental exploration. In the later stages,  Yanqiao Wang reviewed the quantum algorithm problem. Yuanhang Liu reviewed three major experimental results, the absorbing boundary condition problem, the high contrast limit problem and the homogenization problem. Peng Li and Yang Liu jointly supervised the project, and Yang Liu coined the term ``AI Mathematician''.

\bibliography{ref}

\begin{thebibliography}{19}
\providecommand{\natexlab}[1]{#1}
\providecommand{\url}[1]{\texttt{#1}}
\expandafter\ifx\csname urlstyle\endcsname\relax
  \providecommand{\doi}[1]{doi: #1}\else
  \providecommand{\doi}{doi: \begingroup \urlstyle{rm}\Url}\fi

\bibitem[An et~al.(2023)An, Liu, and Lin]{PhysRevLett.131.150603}
D.~An, J.-P. Liu, and L.~Lin.
\newblock Linear combination of {Hamiltonian} simulation for nonunitary
  dynamics with optimal state preparation cost.
\newblock \emph{Phys. Rev. Lett.}, 131:\penalty0 150603, Oct 2023.
\newblock \doi{10.1103/PhysRevLett.131.150603}.
\newblock URL \url{https://link.aps.org/doi/10.1103/PhysRevLett.131.150603}.

\bibitem[Cobbe et~al.(2021)Cobbe, Kosaraju, Bavarian, Chen, Jun, Kaiser,
  Plappert, Tworek, Hilton, Nakano, Hesse, and
  Schulman]{cobbe2021trainingverifierssolvemath}
K.~Cobbe, V.~Kosaraju, M.~Bavarian, M.~Chen, H.~Jun, L.~Kaiser, M.~Plappert,
  J.~Tworek, J.~Hilton, R.~Nakano, C.~Hesse, and J.~Schulman.
\newblock Training verifiers to solve math word problems, 2021.
\newblock URL \url{https://arxiv.org/abs/2110.14168}.

\bibitem[DeepSeek-AI et~al.(2025)DeepSeek-AI, Guo, Yang, Zhang, Song, Zhang,
  Xu, Zhu, Ma, Wang, Bi, Zhang, Yu, Wu, Wu, Gou, Shao, Li, Gao, Liu, Xue, Wang,
  Wu, Feng, Lu, Zhao, Deng, Zhang, Ruan, Dai, Chen, Ji, Li, Lin, Dai, Luo, Hao,
  Chen, Li, Zhang, Bao, Xu, Wang, Ding, Xin, Gao, Qu, Li, Guo, Li, Wang, Chen,
  Yuan, Qiu, Li, Cai, Ni, Liang, Chen, Dong, Hu, Gao, Guan, Huang, Yu, Wang,
  Zhang, Zhao, Wang, Zhang, Xu, Xia, Zhang, Zhang, Tang, Li, Wang, Li, Tian,
  Huang, Zhang, Wang, Chen, Du, Ge, Zhang, Pan, Wang, Chen, Jin, Chen, Lu,
  Zhou, Chen, Ye, Wang, Yu, Zhou, Pan, Li, Zhou, Wu, Ye, Yun, Pei, Sun, Wang,
  Zeng, Zhao, Liu, Liang, Gao, Yu, Zhang, Xiao, An, Liu, Wang, Chen, Nie,
  Cheng, Liu, Xie, Liu, Yang, Li, Su, Lin, Li, Jin, Shen, Chen, Sun, Wang,
  Song, Zhou, Wang, Shan, Li, Wang, Wei, Zhang, Xu, Li, Zhao, Sun, Wang, Yu,
  Zhang, Shi, Xiong, He, Piao, Wang, Tan, Ma, Liu, Guo, Ou, Wang, Gong, Zou,
  He, Xiong, Luo, You, Liu, Zhou, Zhu, Xu, Huang, Li, Zheng, Zhu, Ma, Tang,
  Zha, Yan, Ren, Ren, Sha, Fu, Xu, Xie, Zhang, Hao, Ma, Yan, Wu, Gu, Zhu, Liu,
  Li, Xie, Song, Pan, Huang, Xu, Zhang, and
  Zhang]{deepseekai2025deepseekr1incentivizingreasoningcapability}
DeepSeek-AI, D.~Guo, D.~Yang, H.~Zhang, J.~Song, R.~Zhang, R.~Xu, Q.~Zhu,
  S.~Ma, P.~Wang, X.~Bi, X.~Zhang, X.~Yu, Y.~Wu, Z.~F. Wu, Z.~Gou, Z.~Shao,
  Z.~Li, Z.~Gao, A.~Liu, B.~Xue, B.~Wang, B.~Wu, B.~Feng, C.~Lu, C.~Zhao,
  C.~Deng, C.~Zhang, C.~Ruan, D.~Dai, D.~Chen, D.~Ji, E.~Li, F.~Lin, F.~Dai,
  F.~Luo, G.~Hao, G.~Chen, G.~Li, H.~Zhang, H.~Bao, H.~Xu, H.~Wang, H.~Ding,
  H.~Xin, H.~Gao, H.~Qu, H.~Li, J.~Guo, J.~Li, J.~Wang, J.~Chen, J.~Yuan,
  J.~Qiu, J.~Li, J.~L. Cai, J.~Ni, J.~Liang, J.~Chen, K.~Dong, K.~Hu, K.~Gao,
  K.~Guan, K.~Huang, K.~Yu, L.~Wang, L.~Zhang, L.~Zhao, L.~Wang, L.~Zhang,
  L.~Xu, L.~Xia, M.~Zhang, M.~Zhang, M.~Tang, M.~Li, M.~Wang, M.~Li, N.~Tian,
  P.~Huang, P.~Zhang, Q.~Wang, Q.~Chen, Q.~Du, R.~Ge, R.~Zhang, R.~Pan,
  R.~Wang, R.~J. Chen, R.~L. Jin, R.~Chen, S.~Lu, S.~Zhou, S.~Chen, S.~Ye,
  S.~Wang, S.~Yu, S.~Zhou, S.~Pan, S.~S. Li, S.~Zhou, S.~Wu, S.~Ye, T.~Yun,
  T.~Pei, T.~Sun, T.~Wang, W.~Zeng, W.~Zhao, W.~Liu, W.~Liang, W.~Gao, W.~Yu,
  W.~Zhang, W.~L. Xiao, W.~An, X.~Liu, X.~Wang, X.~Chen, X.~Nie, X.~Cheng,
  X.~Liu, X.~Xie, X.~Liu, X.~Yang, X.~Li, X.~Su, X.~Lin, X.~Q. Li, X.~Jin,
  X.~Shen, X.~Chen, X.~Sun, X.~Wang, X.~Song, X.~Zhou, X.~Wang, X.~Shan, Y.~K.
  Li, Y.~Q. Wang, Y.~X. Wei, Y.~Zhang, Y.~Xu, Y.~Li, Y.~Zhao, Y.~Sun, Y.~Wang,
  Y.~Yu, Y.~Zhang, Y.~Shi, Y.~Xiong, Y.~He, Y.~Piao, Y.~Wang, Y.~Tan, Y.~Ma,
  Y.~Liu, Y.~Guo, Y.~Ou, Y.~Wang, Y.~Gong, Y.~Zou, Y.~He, Y.~Xiong, Y.~Luo,
  Y.~You, Y.~Liu, Y.~Zhou, Y.~X. Zhu, Y.~Xu, Y.~Huang, Y.~Li, Y.~Zheng, Y.~Zhu,
  Y.~Ma, Y.~Tang, Y.~Zha, Y.~Yan, Z.~Z. Ren, Z.~Ren, Z.~Sha, Z.~Fu, Z.~Xu,
  Z.~Xie, Z.~Zhang, Z.~Hao, Z.~Ma, Z.~Yan, Z.~Wu, Z.~Gu, Z.~Zhu, Z.~Liu, Z.~Li,
  Z.~Xie, Z.~Song, Z.~Pan, Z.~Huang, Z.~Xu, Z.~Zhang, and Z.~Zhang.
\newblock {DeepSeek-R1}: Incentivizing reasoning capability in {LLM}s via
  reinforcement learning, 2025.
\newblock URL \url{https://arxiv.org/abs/2501.12948}.

\bibitem[{DeepSeek}-{AI} et~al.(2025){DeepSeek}-{AI}, Liu, Feng, Xue, Wang, Wu,
  Lu, Zhao, Deng, Zhang, Ruan, Dai, Guo, Yang, Chen, Ji, Li, Lin, Dai, Luo,
  Hao, Chen, Li, Zhang, Bao, Xu, Wang, Zhang, Ding, Xin, Gao, Li, Qu, Cai,
  Liang, Guo, Ni, Li, Wang, Chen, Chen, Yuan, Qiu, Li, Song, Dong, Hu, Gao,
  Guan, Huang, Yu, Wang, Zhang, Xu, Xia, Zhao, Wang, Zhang, Li, Wang, Zhang,
  Zhang, Tang, Li, Tian, Huang, Wang, Zhang, Wang, Zhu, Chen, Du, Chen, Jin,
  Ge, Zhang, Pan, Wang, Xu, Zhang, Chen, Li, Lu, Zhou, Chen, Wu, Ye, Ye, Ma,
  Wang, Zhou, Yu, Zhou, Pan, Wang, Yun, Pei, Sun, Xiao, Zeng, Zhao, An, Liu,
  Liang, Gao, Yu, Zhang, Li, Jin, Wang, Bi, Liu, Wang, Shen, Chen, Zhang, Chen,
  Nie, Sun, Wang, Cheng, Liu, Xie, Liu, Yu, Song, Shan, Zhou, Yang, Li, Su,
  Lin, Li, Wang, Wei, Zhu, Zhang, Xu, Xu, Huang, Li, Zhao, Sun, Li, Wang, Yu,
  Zheng, Zhang, Shi, Xiong, He, Tang, Piao, Wang, Tan, Ma, Liu, Guo, Wu, Ou,
  Zhu, Wang, Gong, Zou, He, Zha, Xiong, Ma, Yan, Luo, You, Liu, Zhou, Wu, Ren,
  Ren, Sha, Fu, Xu, Huang, Zhang, Xie, Zhang, Hao, Gou, Ma, Yan, Shao, Xu, Wu,
  Zhang, Li, Gu, Zhu, Liu, Li, Xie, Song, Gao, and
  Pan]{deepseek-ai_deepseek-v3_2025}
{DeepSeek}-{AI}, A.~Liu, B.~Feng, B.~Xue, B.~Wang, B.~Wu, C.~Lu, C.~Zhao,
  C.~Deng, C.~Zhang, C.~Ruan, D.~Dai, D.~Guo, D.~Yang, D.~Chen, D.~Ji, E.~Li,
  F.~Lin, F.~Dai, F.~Luo, G.~Hao, G.~Chen, G.~Li, H.~Zhang, H.~Bao, H.~Xu,
  H.~Wang, H.~Zhang, H.~Ding, H.~Xin, H.~Gao, H.~Li, H.~Qu, J.~L. Cai,
  J.~Liang, J.~Guo, J.~Ni, J.~Li, J.~Wang, J.~Chen, J.~Chen, J.~Yuan, J.~Qiu,
  J.~Li, J.~Song, K.~Dong, K.~Hu, K.~Gao, K.~Guan, K.~Huang, K.~Yu, L.~Wang,
  L.~Zhang, L.~Xu, L.~Xia, L.~Zhao, L.~Wang, L.~Zhang, M.~Li, M.~Wang,
  M.~Zhang, M.~Zhang, M.~Tang, M.~Li, N.~Tian, P.~Huang, P.~Wang, P.~Zhang,
  Q.~Wang, Q.~Zhu, Q.~Chen, Q.~Du, R.~J. Chen, R.~L. Jin, R.~Ge, R.~Zhang,
  R.~Pan, R.~Wang, R.~Xu, R.~Zhang, R.~Chen, S.~S. Li, S.~Lu, S.~Zhou, S.~Chen,
  S.~Wu, S.~Ye, S.~Ye, S.~Ma, S.~Wang, S.~Zhou, S.~Yu, S.~Zhou, S.~Pan,
  T.~Wang, T.~Yun, T.~Pei, T.~Sun, W.~L. Xiao, W.~Zeng, W.~Zhao, W.~An, W.~Liu,
  W.~Liang, W.~Gao, W.~Yu, W.~Zhang, X.~Q. Li, X.~Jin, X.~Wang, X.~Bi, X.~Liu,
  X.~Wang, X.~Shen, X.~Chen, X.~Zhang, X.~Chen, X.~Nie, X.~Sun, X.~Wang,
  X.~Cheng, X.~Liu, X.~Xie, X.~Liu, X.~Yu, X.~Song, X.~Shan, X.~Zhou, X.~Yang,
  X.~Li, X.~Su, X.~Lin, Y.~K. Li, Y.~Q. Wang, Y.~X. Wei, Y.~X. Zhu, Y.~Zhang,
  Y.~Xu, Y.~Xu, Y.~Huang, Y.~Li, Y.~Zhao, Y.~Sun, Y.~Li, Y.~Wang, Y.~Yu,
  Y.~Zheng, Y.~Zhang, Y.~Shi, Y.~Xiong, Y.~He, Y.~Tang, Y.~Piao, Y.~Wang,
  Y.~Tan, Y.~Ma, Y.~Liu, Y.~Guo, Y.~Wu, Y.~Ou, Y.~Zhu, Y.~Wang, Y.~Gong,
  Y.~Zou, Y.~He, Y.~Zha, Y.~Xiong, Y.~Ma, Y.~Yan, Y.~Luo, Y.~You, Y.~Liu,
  Y.~Zhou, Z.~F. Wu, Z.~Z. Ren, Z.~Ren, Z.~Sha, Z.~Fu, Z.~Xu, Z.~Huang,
  Z.~Zhang, Z.~Xie, Z.~Zhang, Z.~Hao, Z.~Gou, Z.~Ma, Z.~Yan, Z.~Shao, Z.~Xu,
  Z.~Wu, Z.~Zhang, Z.~Li, Z.~Gu, Z.~Zhu, Z.~Liu, Z.~Li, Z.~Xie, Z.~Song,
  Z.~Gao, and Z.~Pan.
\newblock {DeepSeek-V3} technical report, 2025.
\newblock URL \url{http://arxiv.org/abs/2412.19437}.

\bibitem[{Google DeepMind}(2025)]{deepmind_alphaevolve_2025}
{Google DeepMind}.
\newblock {AlphaEvolve}: A {Gemini}-powered coding agent for designing advanced
  algorithms, 2025.
\newblock URL
  \url{https://deepmind.google/discover/blog/alphaevolve-a-gemini-powered-coding-agent-for-designing-advanced-algorithms/}.
\newblock Accessed: 2025-05-20.

\bibitem[J.(2003)]{abs}
S.~J.
\newblock Absorbing boundary conditions for reaction-diffusion equations.
\newblock \emph{IMA J. Appl.Math}, 68:\penalty0 167--184, 2003.
\newblock \doi{https://doi.org/10.1093/imamat/68.2.167}.

\bibitem[Liu et~al.(2024)Liu, Liu, Zhu, Lei, Yang, Zhang, Li, and
  Liu]{liu2024aigsgeneratingscienceaipowered}
Z.~Liu, K.~Liu, Y.~Zhu, X.~Lei, Z.~Yang, Z.~Zhang, P.~Li, and Y.~Liu.
\newblock {AIGS}: Generating science from {AI}-powered automated falsification,
  2024.
\newblock URL \url{https://arxiv.org/abs/2411.11910}.

\bibitem[Lu et~al.(2024)Lu, Lu, Lange, Foerster, Clune, and
  Ha]{lu2024aiscientistfullyautomated}
C.~Lu, C.~Lu, R.~T. Lange, J.~Foerster, J.~Clune, and D.~Ha.
\newblock The {AI} {Scientist}: Towards fully automated open-ended scientific
  discovery, 2024.
\newblock URL \url{https://arxiv.org/abs/2408.06292}.

\bibitem[Newell et~al.(1957)Newell, Shaw, and Simon]{newell1957logic}
A.~Newell, J.~C. Shaw, and H.~A. Simon.
\newblock Empirical explorations with the logic theory machine: A case study in
  heuristics.
\newblock In \emph{Proceedings of the Western Joint Computer Conference},
  volume~15, pages 218--239. ACM, 1957.
\newblock \doi{10.1145/1455567.1455605}.

\bibitem[OpenAI(2025)]{openai2025o3}
OpenAI.
\newblock {OpenAI} {o3} and {o4-mini} system card.
\newblock
  \url{https://cdn.openai.com/pdf/2221c875-02dc-4789-800b-e7758f3722c1/o3-and-o4-mini-system-card.pdf},
  2025.
\newblock Accessed: 2025-05-20.

\bibitem[OpenAI et~al.(2024)OpenAI, Jaech, Kalai, Lerer, Richardson, El-Kishky,
  Low, Helyar, Madry, Beutel, Carney, Iftimie, Karpenko, Passos, Neitz,
  Prokofiev, Wei, Tam, Bennett, Kumar, Saraiva, Vallone, Duberstein, Kondrich,
  Mishchenko, Applebaum, Jiang, Nair, Zoph, Ghorbani, Rossen, Sokolowsky,
  Barak, McGrew, Minaiev, Hao, Baker, Houghton, McKinzie, Eastman, Lugaresi,
  Bassin, Hudson, Li, de~Bourcy, Voss, Shen, Zhang, Koch, Orsinger, Hesse,
  Fischer, Chan, Roberts, Kappler, Levy, Selsam, Dohan, Farhi, Mely, Robinson,
  Tsipras, Li, Oprica, Freeman, Zhang, Wong, Proehl, Cheung, Mitchell, Wallace,
  Ritter, Mays, Wang, Such, Raso, Leoni, Tsimpourlas, Song, von Lohmann, Sulit,
  Salmon, Parascandolo, Chabot, Zhao, Brockman, Leclerc, Salman, Bao, Sheng,
  Andrin, Bagherinezhad, Ren, Lightman, Chung, Kivlichan, O'Connell, Osband,
  Gilaberte, Akkaya, Kostrikov, Sutskever, Kofman, Pachocki, Lennon, Wei, Harb,
  Twore, Feng, Yu, Weng, Tang, Yu, Candela, Palermo, Parish, Heidecke, Hallman,
  Rizzo, Gordon, Uesato, Ward, Huizinga, Wang, Chen, Xiao, Singhal, Nguyen,
  Cobbe, Shi, Wood, Rimbach, Gu-Lemberg, Liu, Lu, Stone, Yu, Ahmad, Yang, Liu,
  Maksin, Ho, Fedus, Weng, Li, McCallum, Held, Kuhn, Kondraciuk, Kaiser, Metz,
  Boyd, Trebacz, Joglekar, Chen, Tintor, Meyer, Jones, Kaufer, Schwarzer, Shah,
  Yatbaz, Guan, Xu, Yan, Glaese, Chen, Lampe, Malek, Wang, Fradin, McClay,
  Pavlov, Wang, Wang, Murati, Bavarian, Rohaninejad, McAleese, Chowdhury,
  Chowdhury, Ryder, Tezak, Brown, Nachum, Boiko, Murk, Watkins, Chao,
  Ashbourne, Izmailov, Zhokhov, Dias, Arora, Lin, Lopes, Gaon, Miyara, Leike,
  Hwang, Garg, Brown, James, Shu, Cheu, Greene, Jain, Altman, Toizer, Toyer,
  Miserendino, Agarwal, Hernandez, Baker, McKinney, Yan, Zhao, Hu, Santurkar,
  Chaudhuri, Zhang, Fu, Papay, Lin, Balaji, Sanjeev, Sidor, Broda, Clark, Wang,
  Gordon, Sanders, Patwardhan, Sottiaux, Degry, Dimson, Zheng, Garipov, Stasi,
  Bansal, Creech, Peterson, Eloundou, Qi, Kosaraju, Monaco, Pong, Fomenko,
  Zheng, Zhou, McCabe, Zaremba, Dubois, Lu, Chen, Cha, Bai, He, Zhang, Wang,
  Shao, and Li]{openai2024openaio1card}
OpenAI, A.~Jaech, A.~Kalai, A.~Lerer, A.~Richardson, A.~El-Kishky, A.~Low,
  A.~Helyar, A.~Madry, A.~Beutel, A.~Carney, A.~Iftimie, A.~Karpenko, A.~T.
  Passos, A.~Neitz, A.~Prokofiev, A.~Wei, A.~Tam, A.~Bennett, A.~Kumar,
  A.~Saraiva, A.~Vallone, A.~Duberstein, A.~Kondrich, A.~Mishchenko,
  A.~Applebaum, A.~Jiang, A.~Nair, B.~Zoph, B.~Ghorbani, B.~Rossen,
  B.~Sokolowsky, B.~Barak, B.~McGrew, B.~Minaiev, B.~Hao, B.~Baker,
  B.~Houghton, B.~McKinzie, B.~Eastman, C.~Lugaresi, C.~Bassin, C.~Hudson,
  C.~M. Li, C.~de~Bourcy, C.~Voss, C.~Shen, C.~Zhang, C.~Koch, C.~Orsinger,
  C.~Hesse, C.~Fischer, C.~Chan, D.~Roberts, D.~Kappler, D.~Levy, D.~Selsam,
  D.~Dohan, D.~Farhi, D.~Mely, D.~Robinson, D.~Tsipras, D.~Li, D.~Oprica,
  E.~Freeman, E.~Zhang, E.~Wong, E.~Proehl, E.~Cheung, E.~Mitchell, E.~Wallace,
  E.~Ritter, E.~Mays, F.~Wang, F.~P. Such, F.~Raso, F.~Leoni, F.~Tsimpourlas,
  F.~Song, F.~von Lohmann, F.~Sulit, G.~Salmon, G.~Parascandolo, G.~Chabot,
  G.~Zhao, G.~Brockman, G.~Leclerc, H.~Salman, H.~Bao, H.~Sheng, H.~Andrin,
  H.~Bagherinezhad, H.~Ren, H.~Lightman, H.~W. Chung, I.~Kivlichan,
  I.~O'Connell, I.~Osband, I.~C. Gilaberte, I.~Akkaya, I.~Kostrikov,
  I.~Sutskever, I.~Kofman, J.~Pachocki, J.~Lennon, J.~Wei, J.~Harb, J.~Twore,
  J.~Feng, J.~Yu, J.~Weng, J.~Tang, J.~Yu, J.~Q. Candela, J.~Palermo,
  J.~Parish, J.~Heidecke, J.~Hallman, J.~Rizzo, J.~Gordon, J.~Uesato, J.~Ward,
  J.~Huizinga, J.~Wang, K.~Chen, K.~Xiao, K.~Singhal, K.~Nguyen, K.~Cobbe,
  K.~Shi, K.~Wood, K.~Rimbach, K.~Gu-Lemberg, K.~Liu, K.~Lu, K.~Stone, K.~Yu,
  L.~Ahmad, L.~Yang, L.~Liu, L.~Maksin, L.~Ho, L.~Fedus, L.~Weng, L.~Li,
  L.~McCallum, L.~Held, L.~Kuhn, L.~Kondraciuk, L.~Kaiser, L.~Metz, M.~Boyd,
  M.~Trebacz, M.~Joglekar, M.~Chen, M.~Tintor, M.~Meyer, M.~Jones, M.~Kaufer,
  M.~Schwarzer, M.~Shah, M.~Yatbaz, M.~Y. Guan, M.~Xu, M.~Yan, M.~Glaese,
  M.~Chen, M.~Lampe, M.~Malek, M.~Wang, M.~Fradin, M.~McClay, M.~Pavlov,
  M.~Wang, M.~Wang, M.~Murati, M.~Bavarian, M.~Rohaninejad, N.~McAleese,
  N.~Chowdhury, N.~Chowdhury, N.~Ryder, N.~Tezak, N.~Brown, O.~Nachum,
  O.~Boiko, O.~Murk, O.~Watkins, P.~Chao, P.~Ashbourne, P.~Izmailov,
  P.~Zhokhov, R.~Dias, R.~Arora, R.~Lin, R.~G. Lopes, R.~Gaon, R.~Miyara,
  R.~Leike, R.~Hwang, R.~Garg, R.~Brown, R.~James, R.~Shu, R.~Cheu, R.~Greene,
  S.~Jain, S.~Altman, S.~Toizer, S.~Toyer, S.~Miserendino, S.~Agarwal,
  S.~Hernandez, S.~Baker, S.~McKinney, S.~Yan, S.~Zhao, S.~Hu, S.~Santurkar,
  S.~R. Chaudhuri, S.~Zhang, S.~Fu, S.~Papay, S.~Lin, S.~Balaji, S.~Sanjeev,
  S.~Sidor, T.~Broda, A.~Clark, T.~Wang, T.~Gordon, T.~Sanders, T.~Patwardhan,
  T.~Sottiaux, T.~Degry, T.~Dimson, T.~Zheng, T.~Garipov, T.~Stasi, T.~Bansal,
  T.~Creech, T.~Peterson, T.~Eloundou, V.~Qi, V.~Kosaraju, V.~Monaco, V.~Pong,
  V.~Fomenko, W.~Zheng, W.~Zhou, W.~McCabe, W.~Zaremba, Y.~Dubois, Y.~Lu,
  Y.~Chen, Y.~Cha, Y.~Bai, Y.~He, Y.~Zhang, Y.~Wang, Z.~Shao, and Z.~Li.
\newblock {OpenAI} {o1} system card, 2024.
\newblock URL \url{https://arxiv.org/abs/2412.16720}.

\bibitem[Sapkota et~al.(2025)Sapkota, Roumeliotis, and
  Karkee]{sapkota2025aiagentsvsagentic}
R.~Sapkota, K.~I. Roumeliotis, and M.~Karkee.
\newblock {AI} agents vs. agentic {AI}: A conceptual taxonomy, applications and
  challenges, 2025.
\newblock URL \url{https://arxiv.org/abs/2505.10468}.

\bibitem[Shao et~al.(2024)Shao, Wang, Zhu, Xu, Song, Bi, Zhang, Zhang, Li, Wu,
  and Guo]{shao2024deepseekmathpushinglimitsmathematical}
Z.~Shao, P.~Wang, Q.~Zhu, R.~Xu, J.~Song, X.~Bi, H.~Zhang, M.~Zhang, Y.~K. Li,
  Y.~Wu, and D.~Guo.
\newblock {DeepSeekMath}: Pushing the limits of mathematical reasoning in open
  language models, 2024.
\newblock URL \url{https://arxiv.org/abs/2402.03300}.

\bibitem[Sun et~al.(2025)Sun, Qiao, Guo, Fan, Hou, Jiang, Xie, Zhang, Huang,
  and Zhou]{sun2025zerosearchincentivizesearchcapability}
H.~Sun, Z.~Qiao, J.~Guo, X.~Fan, Y.~Hou, Y.~Jiang, P.~Xie, Y.~Zhang, F.~Huang,
  and J.~Zhou.
\newblock {ZeroSearch}: Incentivize the search capability of {LLMs} without
  searching, 2025.
\newblock URL \url{https://arxiv.org/abs/2505.04588}.

\bibitem[Team et~al.(2025)Team, Du, Gao, Xing, Jiang, Chen, Li, Xiao, Du, Liao,
  Tang, Wang, Zhang, Yuan, Lu, Tang, Sung, Wei, Lai, Guo, Zhu, Ding, Hu, Yang,
  Zhang, Yao, Zhao, Lu, Li, Yu, Gao, Zheng, Yuan, Chen, Guo, Su, Wang, Zhao,
  Zhang, Liu, Yan, Wu, Shi, Ye, Yu, Dong, Zhang, Ma, Pan, Gong, Liu, Ma, Wei,
  Cao, Huang, Jiang, Gao, Xiong, He, Huang, Wu, He, Wei, Jia, Wu, Xu, Zu, Zhou,
  Pan, Charles, Li, Hu, Liu, Chen, Wang, Liu, Qin, Liu, Yang, Bao, Du, Wu,
  Wang, Zhou, Wang, Li, Zhu, Zhang, Wang, Yang, Huang, Huang, Xu, and
  Yang]{kimiteam2025kimik15scalingreinforcement}
K.~Team, A.~Du, B.~Gao, B.~Xing, C.~Jiang, C.~Chen, C.~Li, C.~Xiao, C.~Du,
  C.~Liao, C.~Tang, C.~Wang, D.~Zhang, E.~Yuan, E.~Lu, F.~Tang, F.~Sung,
  G.~Wei, G.~Lai, H.~Guo, H.~Zhu, H.~Ding, H.~Hu, H.~Yang, H.~Zhang, H.~Yao,
  H.~Zhao, H.~Lu, H.~Li, H.~Yu, H.~Gao, H.~Zheng, H.~Yuan, J.~Chen, J.~Guo,
  J.~Su, J.~Wang, J.~Zhao, J.~Zhang, J.~Liu, J.~Yan, J.~Wu, L.~Shi, L.~Ye,
  L.~Yu, M.~Dong, N.~Zhang, N.~Ma, Q.~Pan, Q.~Gong, S.~Liu, S.~Ma, S.~Wei,
  S.~Cao, S.~Huang, T.~Jiang, W.~Gao, W.~Xiong, W.~He, W.~Huang, W.~Wu, W.~He,
  X.~Wei, X.~Jia, X.~Wu, X.~Xu, X.~Zu, X.~Zhou, X.~Pan, Y.~Charles, Y.~Li,
  Y.~Hu, Y.~Liu, Y.~Chen, Y.~Wang, Y.~Liu, Y.~Qin, Y.~Liu, Y.~Yang, Y.~Bao,
  Y.~Du, Y.~Wu, Y.~Wang, Z.~Zhou, Z.~Wang, Z.~Li, Z.~Zhu, Z.~Zhang, Z.~Wang,
  Z.~Yang, Z.~Huang, Z.~Huang, Z.~Xu, and Z.~Yang.
\newblock {Kimi k1.5}: Scaling reinforcement learning with {LLMs}, 2025.
\newblock URL \url{https://arxiv.org/abs/2501.12599}.

\bibitem[W.(2024)]{homogenization}
F.~X.~J. W.
\newblock Uniform convergence for linear elastostatic systems with periodic
  high contrast inclusions.
\newblock \emph{Partial Differ. Equ. Appl.}, 5\penalty0 (2), 2024.
\newblock \doi{https://doi.org/10.1007/s42985-024-00271-5}.

\bibitem[Wang et~al.(2024)Wang, Ma, Feng, Zhang, Yang, Zhang, Chen, Tang, Chen,
  Lin, Zhao, Wei, and Wen]{Wang_2024}
L.~Wang, C.~Ma, X.~Feng, Z.~Zhang, H.~Yang, J.~Zhang, Z.~Chen, J.~Tang,
  X.~Chen, Y.~Lin, W.~X. Zhao, Z.~Wei, and J.~Wen.
\newblock A survey on large language model based autonomous agents.
\newblock \emph{Frontiers of Computer Science}, 18\penalty0 (6), Mar. 2024.
\newblock ISSN 2095-2236.
\newblock \doi{10.1007/s11704-024-40231-1}.
\newblock URL \url{http://dx.doi.org/10.1007/s11704-024-40231-1}.

\bibitem[Wang et~al.(2025)Wang, Yang, Zeng, Ren, Liu, Peng, Cheng, He, Wang,
  Gao, Chen, Wang, Du, and Shen]{wang2025reinforcementlearningreasoninglarge}
Y.~Wang, Q.~Yang, Z.~Zeng, L.~Ren, L.~Liu, B.~Peng, H.~Cheng, X.~He, K.~Wang,
  J.~Gao, W.~Chen, S.~Wang, S.~S. Du, and Y.~Shen.
\newblock Reinforcement learning for reasoning in large language models with
  one training example, 2025.
\newblock URL \url{https://arxiv.org/abs/2504.20571}.

\bibitem[Zuo et~al.(2025)Zuo, Zhang, Sheng, Qu, Cui, Zhu, Li, Zhang, Long, Hua,
  Qi, Sun, Ma, Yuan, Ding, and Zhou]{zuo2025ttrltesttimereinforcementlearning}
Y.~Zuo, K.~Zhang, L.~Sheng, S.~Qu, G.~Cui, X.~Zhu, H.~Li, Y.~Zhang, X.~Long,
  E.~Hua, B.~Qi, Y.~Sun, Z.~Ma, L.~Yuan, N.~Ding, and B.~Zhou.
\newblock {TTRL}: Test-time reinforcement learning, 2025.
\newblock URL \url{https://arxiv.org/abs/2504.16084}.

\end{thebibliography}

%%%%%%%%%%%%%%%%%%%%%%%%%%%%%%%%%%%%%%%%%%%%%%%%%%%%%%%%%%%%
\newpage
\appendix
The materials presented in the appendix are the experimental results of our four theoretical mathematical problems. All content was generated by AIM, with only minor formatting adjustments made by us.
"{\bf type}" in the agent's output refers to the categorization of the following content; "{\bf correctness}" indicates the agent's judgment on the validity of the aforementioned conclusions; and "{\bf comment}" contains evaluative remarks regarding the proof process.

\section{Proof of Quantum Algorithm Problem Generated by AIM}
\textbf{type}: lemma
\begin{lemma}

The Black-Scholes-Merton (BSM) partial differential equation (PDE) can be transformed into the standard heat equation via a logarithmic substitution. Specifically, under the transformation \( x = \ln(S/K) + (r - \sigma^2/2)(T - t) \), \( \tau = \sigma^2(T - t)/2 \), and \( V(S,t) = e^{-r(T-t)}U(x,\tau) \), the BSM PDE reduces to:

\[
\frac{\partial U}{\partial \tau} = \frac{\partial^2 U}{\partial x^2}.
\]

Discretizing this PDE spatially yields a matrix equation \( \partial_\tau |U\rangle = -B|U\rangle \), where \( B = -\partial_x^2 \). The operator \( B \) is positive semi-definite and can be directly interpreted as \( B = L \), where \( L = -\partial_x^2 \). While the original Hermitian decomposition \( B = L + iH \) (with \( H = 0 \)) trivializes the LCHS lemma, the dissipative nature of \( B \) permits alternative quantum simulation methods tailored to parabolic equations.
\end{lemma}

\textbf{correctness}: True
\begin{proof}

**Step 1: Transform BSM to the Heat Equation**

Start with the BSM PDE:
\[
\frac{\partial V}{\partial t} + \frac{1}{2}\sigma^2S^2\frac{\partial^2 V}{\partial S^2} + rS\frac{\partial V}{\partial S} - rV = 0.
\]
Apply the substitution \( V(S,t) = e^{-r(T-t)}U(x,\tau) \), \( x = \ln(S/K) + (r - \sigma^2/2)(T - t) \), and \( \tau = \sigma^2(T - t)/2 \). Compute derivatives rigorously:
- \( \partial_t V = e^{-r(T-t)}\left(rU - \frac{\sigma^2}{2}\partial_\tau U - (r - \sigma^2/2)\partial_x U\right) \),
- \( \partial_S V = e^{-r(T-t)} \partial_x U \cdot \frac{1}{S} \),
- \( \partial_S^2 V = e^{-r(T-t)} \left(\partial_x^2 U - \partial_x U\right) \cdot \frac{1}{S^2} \).

Substitute into the BSM PDE:
\[
e^{-r(T-t)}\left[ rU - \frac{\sigma^2}{2}\partial_\tau U - (r - \sigma^2/2)\partial_x U + \frac{1}{2}\sigma^2(\partial_x^2 U - \partial_x U) + r\partial_x U - rU \right] = 0.
\]
Multiply through by \( e^{r(T-t)} \), cancel terms, and simplify:
\[
-\frac{\sigma^2}{2}\partial_\tau U + \frac{\sigma^2}{2}\partial_x^2 U = 0 \implies \partial_\tau U = \partial_x^2 U.
\]

**Step 2: Spatial Discretization**

Discretize \( x \) on a grid with spacing \( \Delta x \). Approximate \( \partial_x^2 U \) via finite differences:
\[
\partial_x^2 U \approx \frac{U_{j+1} - 2U_j + U_{j-1}}{\Delta x^2}.
\]
Let \( D_2 \) be the discretized second derivative matrix. The operator becomes:
\[
B = -D_2.
\]

**Step 3: Positive Semi-Definite Operator**

The matrix \( B = -D_2 \) is positive semi-definite because \( D_2 \) (a centered second derivative discretization) is negative semi-definite. This aligns with the dissipative nature of the heat equation.

**Step 4: Implications for Quantum Simulation**

While the original decomposition \( B = L + iH \) (with \( H = -ic\partial_x \)) is invalid due to the absence of a first-order term, the operator \( B \) itself is suitable for quantum simulation methods designed for parabolic equations. Specifically, the time evolution \( \partial_\tau |U\rangle = -B|U\rangle \) can be simulated using Trotterization or other dissipative quantum algorithms, bypassing the need for the LCHS lemma.
\end{proof}

\textbf{comment}: None
\textbf{type}: lemma
\\
\textcolor{Green}{\textbf{[Correct]} \small Variable substitution and equation transformation are correct. Discretization of the spatial variable is correct. And AIM finds the reduced PDE trivializes the LCHS lemma. It's right!}

\begin{lemma}

The integral representation in the LCHS lemma for the BSM model can be approximated with precision \(\epsilon\) using \(O(1/\epsilon^2)\) terms through an adaptive discretization of the \(k\)-integral, leveraging the rapid decay of the Cauchy kernel \(\eta(k) = \frac{1}{\pi(1 + k^2)}\).
\end{lemma}

\textbf{correctness}: True
\begin{proof}

**Truncation Error Analysis**:  
The integral \(I = \int_{-\infty}^{\infty} \eta(k) u(t, k) dk\) is truncated to \([-K, K]\). The tail error is bounded by:
\[
\int_{|k| > K} \eta(k) dk = \frac{2}{\pi} \int_{K}^{\infty} \frac{1}{1 + k^2} dk = \frac{2}{\pi} \left(\frac{\pi}{2} - \arctan(K)\right) \approx \frac{1}{\pi K} \quad \text{for } K \gg 1.
\]
Setting \(\frac{1}{\pi K} \leq \epsilon/2\) gives \(K \geq \frac{2}{\pi \epsilon}\). Thus, \(K = O(1/\epsilon)\).

**Adaptive Discretization**:  
1. **Central Interval \([-K, K]\)**: The entire truncated domain \([-K, K]\) must be discretized. To bound the discretization error, use a uniform grid with spacing \(\Delta k = \Theta(\epsilon)\). The number of points is \(N = O(K / \Delta k) = O(1/\epsilon^2)\). The discretization error using the trapezoidal rule scales as \(O(K \Delta k^2 \cdot \max |\partial_k^2 (\eta(k) u(t, k))|)\). Since \(\eta(k)\) and \(u(t, k)\) are smooth, \(\max |\partial_k^2 (\eta(k) u(t, k))|\) is bounded over \([-K, K]\). The total error becomes \(O(K \Delta k^2) = O\left(\frac{1}{\epsilon} \cdot \epsilon^2\right) = O(\epsilon)\).

**Outer Interval Contribution**:  
The truncated integral already includes all significant contributions, as the tail beyond \(K = O(1/\epsilon)\) contributes at most \(\epsilon/2\). The central interval \([-K, K]\) is approximated with \(O(1/\epsilon^2)\) points, ensuring the total discretization error is \(O(\epsilon)\). Additional adaptive methods in outer intervals (e.g., geometrically spaced grids) would only reduce logarithmic factors, which are asymptotically dominated by \(O(1/\epsilon^2)\).

**Total Terms**:  
The uniform grid over \([-K, K]\) requires \(O(1/\epsilon^2)\) terms. While adaptive grids in outer regions could marginally reduce constants, the dominant term remains \(O(1/\epsilon^2)\).

**Final Error**:  
Truncation error \(O(\epsilon/2)\) + discretization error \(O(\epsilon/2) = O(\epsilon)\). Thus, \(O(1/\epsilon^2)\) terms suffice for \(\epsilon\)-precision approximation.
\end{proof}

\textbf{comment}: None
\textbf{type}: lemma
\\ \textcolor{Green}{\textbf{[Correct]} \small Truncate and discretize the integral correctly.}

\begin{lemma}

The solution operator for the heat equation derived from the BSM model, \( U(\tau) = e^{\tau B} \), can be approximated via the LCHS lemma as a linear combination of Hamiltonian simulations. Specifically, the integral representation \(\int_{\mathbb{R}} \eta(k) e^{-i k L \tau} dk\) (where \( L = -B \)) can be discretized into \( M = O(1/\epsilon^2) \) terms with quadrature weights \( \eta(k_j)\Delta k_j \). Each term \( e^{-i k_j L \tau} \) is a Hamiltonian simulation of \( L \) scaled by \( k_j \tau \). The total query complexity to the \( L \)-oracle is \( \tilde{O}\left( \|L\| T / \epsilon^3 \right) \), where \( T = \frac{2\tau}{\sigma^2} \), and the gate complexity scales polynomially with \( \|L\| \), \( T \), \( 1/\epsilon \), and logarithmic factors.
\end{lemma}

\textbf{correctness}: True
\begin{proof}

**Step 1: Integral Representation via LCHS**
From Lemma-0, the solution operator for the heat equation is \( U(\tau) = e^{\tau B} \). Let \( L = -B \), a positive semi-definite operator. By the LCHS lemma with \( H = 0 \), we express:
\[
e^{\tau B} = \int_{\mathbb{R}} \eta(k) e^{-i k L \tau} dk,\quad \eta(k) = \frac{1}{\pi(1 + k^2)}.
\]

**Step 2: Quadrature Discretization**
Lemma-1 guarantees the integral can be discretized into \( M = O(1/\epsilon^2) \) terms with weights \( \eta(k_j)\Delta k \), yielding:
\[
e^{\tau B} \approx \sum_{j=1}^M \eta(k_j)\Delta k \cdot e^{-i k_j L \tau}.
\]
The truncation and discretization error is \( O(\epsilon) \) due to the \( O(1/k^2) \) decay of \( \eta(k) \).
\\ \textcolor{Green}{\textbf{[Correct]} \small Properly utilizing the LCHS lemma and memory lemmas for problem transformation.}

**Step 3: Hamiltonian Simulation Costs**
Each term \( e^{-i k_j L \tau} \) requires simulating \( L \) for time \( t_j = k_j \tau \). Using optimal Hamiltonian simulation algorithms (e.g., quantum signal processing or Taylor series) with query complexity \( O\left( \|L\| t_j + \log(1/\epsilon_{\text{sim}}) \right) \), set \( \epsilon_{\text{sim}} = \epsilon/M \) to bound the total simulation error by \( \epsilon \). Since \( k_j = O(1/\epsilon) \) dominates the quadrature, \( t_j = O(\tau / \epsilon) \). The per-term complexity becomes:
\[
O\left( \|L\| \tau / \epsilon + \log(M/\epsilon) \right) = \tilde{O}\left( \|L\| \tau / \epsilon \right).
\]

**Step 4: Total Query Complexity**
Summing over \( M = O(1/\epsilon^2) \) terms, the total queries to the \( L \)-oracle are:
\[
\tilde{O}\left( M \cdot \frac{\|L\| \tau}{\epsilon} \right) = \tilde{O}\left( \frac{\|L\| \tau}{\epsilon^3} \right).
\]
Expressing in terms of \( T = \frac{2\tau}{\sigma^2} \) and absorbing constants:
\[
\tilde{O}\left( \frac{\|L\| T}{\epsilon^3} \right).
\]

**Step 5: Ancilla and Gate Complexity**
The LCU framework requires \( \log M = O(\log(1/\epsilon)) \) ancilla qubits. Gate complexity inherits \( \tilde{O}(M) \)-scaling from the superposition state, augmented by \( O(\text{poly}(\log(\|L\| T / \epsilon))) \) factors from simulation subroutines. The dominant terms remain polynomial in \( \|L\| \), \( T \), and \( 1/\epsilon \), with polylogarithmic corrections.
\\ \textcolor{red}{\textbf{[Error]} \small There are some mistakes about complexity computing. And the calculation process lacks detail.}
\end{proof} 

\textbf{comment}: None
\textbf{type}: lemma

\begin{lemma}

The solution operator \( e^{-\tau B} \) for the heat equation derived from the BSM model can be approximated with error \( \epsilon \) using a quantum algorithm that implements a discretized version of the LCHS lemma. This algorithm requires \( M = O\left(\frac{1}{\epsilon^2}\right) \) terms in the quadrature approximation, and the total query complexity to the \( B \)-oracle (encoding the discretized Laplacian) is \( \tilde{O}\left(\frac{\|B\| T}{\epsilon^3}\right) \), where \( T = \frac{2\tau}{\sigma^2} \). The gate complexity and ancilla qubits scale polynomially with \( \log(1/\epsilon) \), \( \|B\| \), and \( T \).
\end{lemma}

\textbf{correctness}: True
\begin{proof}

**Step 1: Integral Representation via LCHS**

From the LCHS lemma, for a time-independent \( A = B \) (Hermitian and positive semi-definite), the operator \( e^{-\tau B} \) is expressed as:
\[
e^{-\tau B} = \int_{\mathbb{R}} \eta(k) e^{-i k B \tau} dk, \quad \eta(k) = \frac{1}{\pi(1 + k^2)}.
\]
This follows by decomposing \( A = L + iH \), where \( L = B \) and \( H = 0 \), directly applying the lemma.

**Step 2: Quadrature Discretization**

The integral is approximated by a finite sum:
\[
e^{-\tau B} \approx \sum_{j=1}^M \eta(k_j) \Delta k_j e^{-i k_j B \tau}.
\]
The error from truncating the integral to \( k \in [-k_{\text{max}}, k_{\text{max}}] \) and discretizing is bounded by \( \epsilon \). Since \( \eta(k) \) decays as \( O(1/k^2) \), truncating at \( k_{\text{max}} = O(1/\epsilon) \) ensures the tail contributes \( O(\epsilon) \). Choosing \( M = O\left(\frac{1}{\epsilon^2}\right) \) quadrature points achieves \( \Delta k_j = O(\epsilon) \), yielding total discretization error \( O(\epsilon) \).

**Step 3: Hamiltonian Simulation Cost**

Each term \( e^{-i k_j B \tau} \) is simulated using a Hamiltonian simulation algorithm (e.g., Trotter-Suzuki or QSP). The cost for each simulation scales as \( \tilde{O}\left( \|k_j B \tau \| \right) = \tilde{O}\left( k_j \tau \|B\| \right) \). The dominant contributions come from \( k_j = O(1/\epsilon) \), so each simulation costs \( \tilde{O}\left( \frac{\|B\| T}{\epsilon} \right) \) (since \( \tau = \sigma^2 T/2 \), and \( \sigma^2 \) is absorbed into constants).

**Step 4: Total Query Complexity**

Summing over \( M = O(1/\epsilon^2) \) terms, the total query complexity to the \( B \)-oracle is:
\[
\tilde{O}\left( M \cdot \frac{\|B\| T}{\epsilon} \right) = \tilde{O}\left( \frac{\|B\| T}{\epsilon^3} \right).
\]
The logarithmic factors from simulation algorithms are included in the \( \tilde{O} \).

**Step 5: Ancilla Qubits and Gate Complexity**

The LCU requires \( O(\log M) = O(\log(1/\epsilon)) \) ancilla qubits to index the \( M \) terms. The gate complexity for state preparation (encoding \( \eta(k_j)\Delta k_j \)) and controlled unitaries scales polynomially with \( \log(1/\epsilon) \), \( \|B\| \), and \( T \), consistent with standard LCU methods.
\end{proof}

\textbf{comment}: None
\textbf{type}: lemma
\begin{lemma}

The BSM model PDE can be solved via a quantum algorithm using the LCHS lemma with query complexity \( \tilde{O}\left(\frac{\|B\| T}{\epsilon^3}\right) \) to the \( B \)-oracle, where \( B = -\partial_x^2 \) is the discretized positive semi-definite Laplacian, \( T \) is the time horizon, and \( \epsilon \) is the approximation error. The algorithm approximates the LCHS integral with \( M = O(1/\epsilon^2) \) terms, each requiring Hamiltonian simulation of \( k_j B \) for time \( \tau = \sigma^2 T/2 \). The gate complexity and ancilla qubits scale polynomially with \( \log(1/\epsilon) \), \( \|B\| \), and \( T \).
\end{lemma}

\textbf{correctness}: True
\begin{proof}

**Step 1: Transform BSM to Heat Equation**  
Under the substitution \( x = \ln(S/K) + (r - \sigma^2/2)(T - t) \), \( \tau = \sigma^2(T - t)/2 \), and \( V(S,t) = e^{-r(T-t)}U(x,\tau) \), the BSM PDE reduces to the heat equation (Lemma-0):  
\[
\frac{\partial U}{\partial \tau} = \frac{\partial^2 U}{\partial x^2}.
\]  
Discretizing spatially yields \( \partial_\tau |U\rangle = -B|U\rangle \), where \( B = -\partial_x^2 \). The solution is \( |U(\tau)\rangle = e^{-\tau B}|U(0)\rangle \).

**Step 2: Apply LCHS Integral Representation**  
Using Lemma-2, express \( e^{-\tau B} \) as a discretized integral (Lemma-1):  
\[
e^{-\tau B} \approx \sum_{j=1}^M \eta(k_j)\Delta k_j \, e^{-i k_j B \tau},
\]  
where \( \eta(k) = \frac{1}{\pi(1 + k^2)} \), \( M = O(1/\epsilon^2) \), and quadrature error \( O(\epsilon) \).

**Step 3: Adaptive Quadrature and Sum Scaling**  
The truncation \( |k_j| \leq K = O(1/\epsilon) \) ensures tail error \( O(\epsilon) \). Crucially, adaptive quadrature weights nodes near \( k = 0 \), where \( \eta(k) \) dominates. This ensures \( \sum_{j=1}^M |k_j| = O(M) \) (Lemma-1), as high-\(k\) terms are sparse and contribute minimally to the sum.

**Step 4: Per-Term Simulation Complexity**  
Each term \( e^{-i k_j B \tau} \) simulates \( H_j = k_j B \) for time \( \tau \). The spectral norm \( \|H_j\| = |k_j|\|B\| \leq O(\|B\|/\epsilon) \). To bound the total simulation error by \( \epsilon \), each term requires precision \( \delta = O(\epsilon) \). Using high-order Trotter or QSP, the per-term query complexity is \( \tilde{O}\left(\frac{\|H_j\| \tau}{\delta}\right) = \tilde{O}\left(\frac{|k_j|\|B\| \tau}{\epsilon}\right) \).

**Step 5: Total Query Complexity**  
Summing over \( M \) terms and leveraging \( \sum_{j} |k_j| = O(M) \):  
\[
\tilde{O}\left(\frac{\|B\| \tau}{\epsilon} \cdot \sum_{j} |k_j|\right) = \tilde{O}\left(\frac{\|B\| \tau}{\epsilon} \cdot M\right).
\]  
Substituting \( M = O(1/\epsilon^2) \) and \( \tau = \sigma^2 T/2 \):  
\[
\tilde{O}\left(\frac{\|B\| T}{\epsilon^3}\right).
\]

**Step 6: Ancilla Qubits and Gate Complexity**  
Each simulation requires \( O(\log(\|B\| \tau / \epsilon)) \) ancillas. The LCU framework adds \( O(\log M) = O(\log(1/\epsilon)) \) ancillas. Total ancillas: \( O\left(\log\left(\frac{\|B\| T}{\epsilon}\right)\right) \). Gate complexity scales as \( \text{poly}\left(\log(1/\epsilon), \|B\|, T\right) \), inheriting polylog factors from simulation and LCU steps.
\end{proof}

\textbf{comment}: None
\textbf{type}: lemma
\begin{lemma}

The quantum algorithm for solving the BSM PDE using the LCHS lemma achieves an approximation error \(\epsilon\) with query complexity \(\tilde{O}\left( \frac{\|B\| T}{\epsilon^3} \right)\) to the operator oracle encoding \(B\), where \(B = -\partial_x^2\) is the discretized Laplacian, \(T\) is the time horizon, and \(\epsilon\) is the error tolerance. The ancilla qubit count is \(O\left(\log \frac{1}{\epsilon}\right)\), and the gate complexity scales as \(\text{poly}\left( \|B\|, T, \log \frac{1}{\epsilon}, \frac{1}{\epsilon} \right)\).
\end{lemma}

\textbf{correctness}: True
\begin{proof}

**Proof:**

**Step 1: Transform BSM PDE to Heat Equation**
Applying the transformations in Lemma-0, the BSM PDE is converted to the standard heat equation:
\[
\frac{\partial U}{\partial \tau} = \frac{\partial^2 U}{\partial x^2},
\]
with solution \(U(\tau) = e^{-\tau B} U(0)\), where \(B = -\partial_x^2\) is the positive semi-definite discretized Laplacian operator. The matrix equation derived from spatial discretization is:
\[
\partial_\tau |U\rangle = -B|U\rangle.
\]

**Step 2: Apply LCHS Lemma**
By Lemma LCHS, the solution operator \(e^{-\tau B}\) can be represented as:
\[
e^{-\tau B} = \int_{\mathbb{R}} \eta(k) e^{-i k B \tau} dk,
\]
where \(\eta(k) = \frac{1}{\pi(1 + k^2)}\). This integral is exact by the LCHS lemma, as verified via Fourier analysis.

**Step 3: Discretize the Integral**
To approximate the integral, truncate \(k\) to \([-K, K]\) with \(K = O(1/\epsilon)\) and use a quadrature rule (e.g., trapezoidal) with \(M = O(1/\epsilon^2)\) points \(k_j\) and weights \(w_j = \eta(k_j) \Delta k_j\). The truncation and quadrature errors contribute \(O(\epsilon)\) due to the rapid decay of \(\eta(k)\).

**Step 4: Implement Linear Combination of Unitaries (LCU)**
The discretized operator becomes:
\[
e^{-\tau B} \approx \sum_{j=1}^M w_j e^{-i k_j B \tau}.
\]
The LCU method prepares this linear combination using a control register of \(\log M\) qubits (ancillas) and applies each term \(e^{-i k_j B \tau}\) conditioned on the register. 

**Step 5: Hamiltonian Simulation Cost**
Each term \(e^{-i k_j B \tau}\) requires simulating Hamiltonian \(k_j B\) for time \(\tau\). The maximum \(|k_j|\) is \(O(1/\epsilon)\), leading to simulation time \(t_j = |k_j| \tau = O(\tau / \epsilon)\). Using quantum signal processing (QSP), the query complexity per simulation is:
\[
O\left( \frac{\|k_j B\| \tau}{\log(1/\epsilon)} \right) = O\left( \frac{\|B\| \tau / \epsilon}{\log(1/\epsilon)} \right).
\]
Total queries over \(M = O(1/\epsilon^2)\) terms yield:
\[
\tilde{O}\left( \frac{\|B\| \tau}{\epsilon^3} \right) = \tilde{O}\left( \frac{\|B\| T}{\epsilon^3} \right),
\]
since \(\tau = \sigma^2 T / 2 = O(T)\).

**Step 6: Ancilla Qubits and Gate Complexity**
The LCU requires \(O(\log M) = O(\log(1/\epsilon))\) ancilla qubits. Each Hamiltonian simulation uses additional \(O(1)\) ancillas for QSP. The gate complexity per simulation is polynomial in \(\log(1/\epsilon)\), and with \(M\) terms, the total gate complexity is \(\text{poly}(\|B\|, T, \log(1/\epsilon), 1/\epsilon)\).

**Conclusion:**
The corrected derivation aligns the initial solution operator with the LCHS lemma application, ensuring consistency. The algorithm’s query complexity, ancilla requirements, and gate complexity match the conjecture’s claims.
\end{proof}

\textbf{comment}: None
\textbf{type}: lemma
\begin{lemma}

The quantum algorithm for solving the BSM PDE using the LCHS lemma requires simulating the discretized Laplacian operator \( B = -\partial_x^2 \) for a total time \( \tau = \frac{\sigma^2 T}{2} \). The integral \( \int_{\mathbb{R}} \eta(k) e^{-i k B \tau} dk \) is approximated by \( M = O\left(\frac{1}{\epsilon^2}\right) \) quadrature terms. Each term involves Hamiltonian simulation of \( k_j B \), where \( |k_j| \leq O\left(\frac{1}{\epsilon}\right) \). The total query complexity to the \( B \)-oracle is \( \tilde{O}\left( \frac{\|B\| T}{\epsilon^3} \right) \), and the gate complexity scales as \( \text{poly}\left( \|B\|, T, \log \frac{1}{\epsilon}, \frac{1}{\epsilon} \right) \), with ancilla qubit count \( O\left(\log \frac{1}{\epsilon}\right) \).
\end{lemma}

\textbf{correctness}: True
\begin{proof}

**Step 1: Heat Equation Reduction**  
From Lemma-0, the BSM PDE transforms into the heat equation \( \partial_\tau U = B U \), where \( B = -\partial_x^2 \). The solution is \( U(\tau) = e^{-B \tau} U(0) \). The LCHS lemma provides the integral representation \( e^{-B \tau} = \int_{\mathbb{R}} \eta(k) e^{-i k B \tau} dk \), with \( \eta(k) = \frac{1}{\pi(1 + k^2)} \).

**Step 2: Quadrature Approximation**  
Using Lemma-1, we approximate the integral with \( M = O\left(\frac{1}{\epsilon^2}\right) \) terms such that \( \left\| e^{-B \tau} - \sum_{j=1}^M w_j e^{-i k_j B \tau} \right\| \leq \frac{\epsilon}{2} \). The quadrature nodes satisfy \( |k_j| \leq O\left(\frac{1}{\epsilon}\right) \), and weights \( w_j \) preserve \( \sum_j |w_j| \leq 1 \).

**Step 3: Hamiltonian Simulation Error Allocation**  
Each term \( e^{-i k_j B \tau} \) must be simulated with error \( \leq \frac{\epsilon}{2} \) in spectral norm. Since \( \|k_j B \| \tau \leq O\left(\frac{\|B\| \tau}{\epsilon}\right) \), optimal Hamiltonian simulation algorithms (qubitization or QSP) require \( \tilde{O}\left(\frac{\|B\| \tau}{\epsilon}\right) \) queries to the \( B \)-oracle per term. The logarithmic factors in simulation time and error are absorbed into \( \tilde{O} \)-notation.

**Step 4: Total Query Complexity**  
With \( M = O\left(\frac{1}{\epsilon^2}\right) \) terms, the total query complexity is \( \tilde{O}\left(\frac{\|B\| \tau}{\epsilon} \cdot \frac{1}{\epsilon^2}\right) = \tilde{O}\left(\frac{\|B\| T}{\epsilon^3}\right) \), where \( \tau = \frac{\sigma^2 T}{2} \). This matches the conjectured scaling, as overlapping logarithmic factors from quadrature and simulation are negligible.

**Step 5: Ancilla and Gate Complexity**  
State preparation for \( M \)-term superposition requires \( O(\log M) = O\left(\log \frac{1}{\epsilon}\right) \) ancilla qubits. Gate complexity for arithmetic operations (weighted sums, Hamiltonian simulation) scales polynomially in \( \|B\| \), \( T \), \( \log \frac{1}{\epsilon} \), and \( \frac{1}{\epsilon} \), consistent with standard quantum subroutines.
\end{proof}

\textbf{comment}: None
\textbf{type}: theorem
\\\textcolor{Blue}{\textbf{[Vague]} \small Lemmas 4-7 merely reiterate the results derived from Lemmas 1-3.}

\begin{theorem}
Lemma(Linear combination of Hamiltonian simulation, LCHS): For \( t \in [0, T] \), let \( A(t) \in \mathbb{C}^{N \times N} \) be decomposed into Hermitian and anti-Hermitian parts such that \( A(t) = L(t) + iH(t) \), where \( L(t) = \frac{1}{2}[A(t) + A^{\dagger}(t)] \) and \( H(t) = \frac{1}{2i}[A(t) - A^{\dagger}(t)] \). Assume that \( L(t) \) is positive semi-definite for all \( t \in [0, T] \). Denoting the time ordering operator by \( \mathcal{T} \), we have \( \mathcal{T}\exp\left\{-\int_{0}^{t}A(s)ds\right\} = \int_{\mathbb{R}}\eta(k)u(t, k)dk \), where \( u(t, k) \) is the propagator for a time-dependent Hamiltonian simulation problem such that \( u(t, k) = \mathcal{T}\exp\left\{-i\int_{0}^{t}[H(s)+kL(s)]ds\right\} \), and \( \eta(k) = \frac{1}{\pi(1+k^2)} \) is the kernel function with respect to \( k \). BSM model: The PDE in the BSM model is given by \[\frac{\partial V(S, t)}{\partial t} + \frac{1}{2}\sigma^2S^2\frac{\partial^2 V}{\partial S^2} + rS\frac{\partial V}{\partial S} - rV = 0.\]Boundary conditions are characterized by:\( V(0,t) = 0 \) for all \( 0 < t \leq T \); \( V(S,t) \rightarrow S - Ke^{-r(T-t)} \) for \( S \rightarrow \infty \); \( V(S,T) = \max\{S - K, 0\} \).  Your tasks are to: Use the idea of LCHS to deal with the BSM model. Design a quantum algorithm to solve the equation. And compute the complexity (query complexity to state preparation oracle and operator oracles, gate complexity, ancilla qubits, (parameters: norm of boundary and initial, T, approximation error \(\epsilon\), etc)).\end{theorem}

\textbf{correctness}: True
\begin{proof}

**Step 1: Transform BSM to Heat Equation**  
As per Lemma-0, apply the substitution \( x = \ln(S/K) + (r - \sigma^2/2)(T - t) \), \( \tau = \frac{\sigma^2}{2}(T - t) \), and \( V(S,t) = e^{-r(T-t)}U(x,\tau) \). The PDE reduces to \( \partial_\tau U = \partial_x^2 U \). Spatial discretization yields \( \partial_\tau |U\rangle = -B|U\rangle \), where \( B = -\partial_x^2 \) is positive semi-definite.

**Step 2: LCHS Integral Representation**  
From the LCHS lemma, \( e^{-B\tau} = \int_{\mathbb{R}} \eta(k) e^{-i k B \tau} dk \), where \( \eta(k) = \frac{1}{\pi(1 + k^2)} \). By Lemma-1, approximate the integral as a quadrature sum with \( M = O(1/\epsilon^2) \) terms:  
\[
e^{-B\tau} \approx \sum_{j=1}^M w_j e^{-i k_j B \tau},
\]  
where \( w_j = \eta(k_j)\Delta k_j \), and \( |k_j| \leq O(1/\epsilon) \). The quadrature error is \( \leq \epsilon/2 \).

**Step 3: Quantum Algorithm via LCU**  
Implement the linear combination \( \sum_{j=1}^M w_j e^{-i k_j B \tau} \) using the Linear Combination of Unitaries (LCU) method:  
1. **Ancilla Preparation**: Prepare an \( m \)-qubit ancilla register in the state \( \frac{1}{\sqrt{\beta}} \sum_{j=1}^M \sqrt{w_j} |j\rangle \), where \( \beta = \sum_j w_j = O(1) \). This requires \( O(\log M) = O(\log \frac{1}{\epsilon}) \) ancillas and \( \text{poly}(M) \) gates.  
2. **Controlled Hamiltonian Simulation**: For each ancilla state \( |j\rangle \), apply the controlled unitary \( |j\rangle\langle j| \otimes e^{-i k_j B \tau} \). Each \( e^{-i k_j B \tau} \) is simulated using a QSVT-based method with query complexity \( \tilde{O}\left( |k_j| \tau \|B\| \right) \) to the \( B \)-oracle.  
3. **Amplitude Amplification**: Due to the LCU normalization factor \( \beta \), repeat amplitude amplification \( O(\sqrt{\beta}) = O(1) \) times to project onto the success subspace.  

**Step 4: Error and Complexity Analysis**  
- **Quadrature Error**: \( \leq \epsilon/2 \) by Lemma-1.  
- **Simulation Error**: Each \( e^{-i k_j B \tau} \) is approximated with error \( \leq \epsilon/(2M) \), contributing total error \( \leq \epsilon/2 \).  
- **Total Query Complexity**:  
  Each \( |k_j| \leq O(1/\epsilon) \), and \( \tau = O(T) \). QSVT requires \( \tilde{O}(|k_j| \tau \|B\|) \) queries per term. Summing over \( M = O(1/\epsilon^2) \) terms gives total queries:  
  \[
  \tilde{O}\left( \frac{\|B\| T}{\epsilon^3} \right).
  \]  
- **Ancilla Qubits**:  
  - \( O(\log \frac{1}{\epsilon}) \) for quadrature indexing.  
  - \( O(1) \) for QSVT internal ancillas.  
- **Gate Complexity**: Dominated by QSVT, scaling as \( \text{poly}(\|B\|, T, \log \frac{1}{\epsilon}, \frac{1}{\epsilon}) \).  

**Step 5: State Preparation**  
The initial state \( |U(0)\rangle \), encoding \( \max\{e^x - 1, 0\} \), is prepared using quantum arithmetic circuits with \( \text{poly}(n) \) gates for \( n \)-qubit discretization, independent of \( \epsilon \).  

**Conclusion**  
The quantum algorithm solves the BSM model with total error \( \leq \epsilon \), adhering to the stated complexity bounds.  
\end{proof}

\newpage
\section{Proof of Absorbing Boundary Condition Generated by AIM}

\title{Explore Trajectory of MathAgent}
\author{MathAgent}
\date{\today}

\maketitle

\textbf{type}: conjecture
\begin{lemma}
  
**A priori energy estimate for the coupled system**  
Let \( u \) and \( \varphi_k \) (\( k = 1, \dots, m \)) satisfy the system (2.10) with \( u_0 \in L^2(\Omega) \) and \( \varphi_k(0) = 0 \). Then, there exists a constant \( C > 0 \) depending on \( T, \beta, \alpha_k, d_k \), but independent of \( u_0 \), such that:  
\[
\begin{aligned}
&\sup_{t \in [0,T]} \|u(t)\|_{L^2(\Omega)}^2 + \int_0^T \|\nabla u(t)\|_{L^2(\Omega)}^2 dt + \int_0^T \|u(t)\|_{L^2(\mathcal{S})}^2 dt \\
&+ \sum_{k=1}^m \left( \sup_{t \in [0,T]} \|\varphi_k(t)\|_{L^2(\mathcal{S})}^2 + \int_0^T \|\nabla_{\mathcal{S}} \varphi_k(t)\|_{L^2(\mathcal{S})}^2 dt \right) \leq C \|u_0\|_{L^2(\Omega)}^2.
\end{aligned}
\]
\end{lemma}

\textbf{correctness}: True
\begin{proof}
  
**Revised Proof of the a priori estimate:**  
Define the total energy functional:  
\[
E(t) = \frac{1}{2}\|u(t)\|_{L^2(\Omega)}^2 + \sum_{k=1}^m \frac{\alpha_k^2 d_k}{2\beta'}\|\varphi_k(t)\|_{L^2(\mathcal{S})}^2, \quad \beta' = \beta + \sum_{k=1}^m \alpha_k.
\]

\textcolor{Green}{\textbf{[Correct]} \small The agent consider the energy method, and show the correct energy functional form.}

**Energy estimate for \( u \):** Multiply the \( u \)-equation by \( u \) and integrate over \( \Omega \):  
\[
\frac{1}{2}\frac{d}{dt}\|u\|_{L^2(\Omega)}^2 + \|\nabla u\|_{L^2(\Omega)}^2 + \beta' \|u\|_{L^2(\mathcal{S})}^2 = \sum_{k=1}^m \alpha_k d_k \int_{\mathcal{S}} u \varphi_k \, dS.
\]
Apply Young's inequality to the right-hand side:  
\[
\sum_{k=1}^m \alpha_k d_k \int_{\mathcal{S}} u \varphi_k \, dS \leq \frac{\beta'}{2}\|u\|_{L^2(\mathcal{S})}^2 + \sum_{k=1}^m \frac{\alpha_k^2 d_k^2}{2\beta'}\|\varphi_k\|_{L^2(\mathcal{S})}^2.
\]
Subtract the Young's bound and rearrange:  
\[
\frac{1}{2}\frac{d}{dt}\|u\|_{L^2(\Omega)}^2 + \|\nabla u\|_{L^2(\Omega)}^2 + \frac{\beta'}{2}\|u\|_{L^2(\mathcal{S})}^2 \leq \sum_{k=1}^m \frac{\alpha_k^2 d_k^2}{2\beta'}\|\varphi_k\|_{L^2(\mathcal{S})}^2.
\]

\textcolor{Green}{\textbf{[Correct]} \small Here, the agent gets the accurate result with the energy method and the application of inequality techniques. }

**Energy estimate for \( \varphi_k \):** Multiply the \( \varphi_k \)-equation by \( \frac{\alpha_k^2 d_k}{\beta'} \varphi_k \) and integrate:  
\[
\frac{\alpha_k^2 d_k}{2\beta'} \frac{d}{dt}\|\varphi_k\|_{L^2(\mathcal{S})}^2 + \frac{\alpha_k^2 d_k}{\beta'} \|\nabla_{\mathcal{S}} \varphi_k\|_{L^2(\mathcal{S})}^2 + \frac{\alpha_k^2 d_k^2}{2\beta'} \|\varphi_k\|_{L^2(\mathcal{S})}^2 \leq \frac{\alpha_k^2}{2\beta'} \|u\|_{L^2(\mathcal{S})}^2.
\]
Summing over \( k \), the right-hand side becomes:  
\[
\sum_{k=1}^m \frac{\alpha_k^2}{2\beta'} \|u\|_{L^2(\mathcal{S})}^2 \leq \sum_{k=1}^m \frac{\alpha_k}{2} \|u\|_{L^2(\mathcal{S})}^2 \leq \frac{\beta' - \beta}{2} \|u\|_{L^2(\mathcal{S})}^2.
\]

\textcolor{Green}{\textbf{[Correct]} \small Similarly, the same approach is applied for the assistant function.}

**Combine energy estimates:** Add the \( u \)-estimate and modified \( \varphi_k \)-estimates:  
\[
\frac{dE}{dt} + \|\nabla u\|_{L^2(\Omega)}^2 + \frac{\beta}{2}\|u\|_{L^2(\mathcal{S})}^2 + \sum_{k=1}^m \frac{\alpha_k^2 d_k}{\beta'} \|\nabla_{\mathcal{S}} \varphi_k\|_{L^2(\mathcal{S})}^2 \leq 0.
\]
Integrate from \( 0 \) to \( T \):  
\[
E(T) + \int_0^T \left( \|\nabla u\|_{L^2(\Omega)}^2 + \frac{\beta}{2}\|u\|_{L^2(\mathcal{S})}^2 + \sum_{k=1}^m \frac{\alpha_k^2 d_k}{\beta'} \|\nabla_{\mathcal{S}} \varphi_k\|_{L^2(\mathcal{S})}^2 \right) dt \leq E(0).
\]
Since \( E(0) = \frac{1}{2}\|u_0\|_{L^2(\Omega)}^2 \) and \( \frac{\alpha_k^2 d_k}{\beta'} \geq \frac{\alpha_{\text{min}}^2 d_{\text{min}}}{\beta + \sum \alpha_k} \), the coefficients are absorbed into \( C \), yielding:  
\[
\sup_{t \in [0,T]} \|u(t)\|_{L^2(\Omega)}^2 + \int_0^T \|\nabla u(t)\|_{L^2(\Omega)}^2 dt + \int_0^T \|u(t)\|_{L^2(\mathcal{S})}^2 dt + \sum_{k=1}^m \left( \sup_{t \in [0,T]} \|\varphi_k(t)\|_{L^2(\mathcal{S})}^2 + \int_0^T \|\nabla_{\mathcal{S}} \varphi_k(t)\|_{L^2(\mathcal{S})}^2 dt \right) \leq C \|u_0\|_{L^2(\Omega)}^2,
\]
where \( C \) depends on \( T, \beta, \alpha_k, d_k \), but not on \( u_0 \).  

\textcolor{Green}{\textbf{[Correct]} \small Finally, by integral over time, the agent proves the energy estimate conclusion.}
\end{proof}

\textbf{comment}: None

\textbf{type}: conjecture
\begin{lemma}
  
**Uniqueness of solutions**  
The system (2.10) admits at most one solution \( (u, \{\varphi_k\}) \) in the spaces specified.  
\end{lemma}

\textbf{correctness}: True
\begin{proof}
  
**Proof of uniqueness:** Suppose two solutions \( (u_1, \{\varphi_{k1}\}) \) and \( (u_2, \{\varphi_{k2}\}) \) exist with the same initial data \( u_0 \). Define \( w = u_1 - u_2 \) and \( \psi_k = \varphi_{k1} - \varphi_{k2} \). Then \( w \) and \( \psi_k \) satisfy:  
\[
\begin{cases}
\mathcal{L}w = 0, & \text{on } [0,T] \times \Omega, \\
\frac{\partial w}{\partial \nu} = -\beta w - \sum_{k=1}^m \alpha_k (\partial_t - \Delta_{\mathcal{S}}) \psi_k, & \text{on } [0,T] \times \mathcal{S}, \\
w(0) = 0, \\
(\partial_t - \Delta_{\mathcal{S}} + d_k) \psi_k = w, & \psi_k(0) = 0.
\end{cases}
\]

\textcolor{Green}{\textbf{[Correct]} \small The agent takes into account the equations that the difference function satisfies. }

**Energy estimate for \( w \):** Follow the same steps as in the a priori estimate, but with initial data zero. The energy equation becomes:  
\[
\frac{1}{2} \frac{d}{dt} \|w\|_{L^2(\Omega)}^2 + \|\nabla w\|_{L^2(\Omega)}^2 + \beta \|w\|_{L^2(\mathcal{S})}^2 + \sum_{k=1}^m \alpha_k \|w\|_{L^2(\mathcal{S})}^2 = \sum_{k=1}^m \alpha_k d_k \int_{\mathcal{S}} w \psi_k \, dS.
\]  
Similarly, the equation for \( \psi_k \):  
\[
\frac{1}{2} \frac{d}{dt} \|\psi_k\|_{L^2(\mathcal{S})}^2 + \|\nabla_{\mathcal{S}} \psi_k\|_{L^2(\mathcal{S})}^2 + d_k \|\psi_k\|_{L^2(\mathcal{S})}^2 = \int_{\mathcal{S}} w \psi_k \, dS.
\]  

\textcolor{Green}{\textbf{[Correct]} \small The equation for estimating the energy that the difference function satisfies.}

**Combine and apply Grönwall:** Substitute bounds for \( \int w \psi_k \, dS \) using Young's inequality. Due to the zero initial data, integrating over \( t \) and applying Grönwall's inequality implies \( \|w(t)\| = 0 \) and \( \|\psi_k(t)\| = 0 \) for all \( t \in [0,T] \). Thus, \( u_1 = u_2 \) and \( \varphi_{k1} = \varphi_{k2} \), proving uniqueness.  
\end{proof}

\textcolor{Blue}{\textbf{[Vague]} \small By Grönwall's inequality, the agent gets the conclusion. But the process is not detailed.}

\textbf{comment}: None
\textbf{type}: conjecture
\begin{lemma}

**Existence of Approximate Solutions via Galerkin Method**
For each integer \( N \geq 1 \), there exists a Galerkin approximation \( (u^N, \{\varphi_k^N\}_{k=1}^m) \) in finite-dimensional subspaces \( V_N \subset H^1(\Omega) \) and \( W_{k,N} \subset H^1(\mathcal{S}) \) such that:
1. \( u^N(t,x) = \sum_{j=1}^N a_j(t) w_j(x) \), where \( \{w_j\} \) is a basis of \( H^1(\Omega) \).
2. \( \varphi_k^N(t,x) = \sum_{l=1}^N b_{k,l}(t) \theta_{k,l}(x) \), where \( \{\theta_{k,l}\} \) is a basis of \( H^1(\mathcal{S}) \).
3. The approximations satisfy the projected weak form of (2.10) and the uniform energy estimate from the lemma.
4. Subsequences of \( \{u^N\} \) and \( \{\varphi_k^N\} \) converge weakly to a solution \( (u, \{\varphi_k\}) \) as \( N \to \infty \).
\end{lemma}

\textbf{correctness}: True
\begin{proof}

**Step 1: Construct \( H^1 \)-Conforming Subspaces**  
Let \( \{w_j\} \subset H^1(\Omega) \) and \( \{\theta_{k,l}\} \subset H^1(\mathcal{S}) \) be eigenfunctions of the Laplace-Beltrami operators \( -\Delta_\Omega \) and \( -\Delta_{\mathcal{S}} \) with Neumann/Dirichlet boundary conditions (as appropriate). These form orthonormal bases in \( L^2(\Omega) \) and \( L^2(\mathcal{S}) \), respectively, and are dense in \( H^1(\Omega) \) and \( H^1(\mathcal{S}) \). Define \( V_N = \text{span}\{w_1, \dots, w_N\} \) and \( W_{k,N} = \text{span}\{\theta_{k,1}, \dots, \theta_{k,N}\} \).

\textcolor{Green}{\textbf{[Correct]} \small The agent applies the Galerkin method and construct the function spaces.}

**Step 2: Formulate the Projected Equations**  
For test functions \( v \in V_N \) and \( \psi \in W_{k,N} \), the Galerkin approximations satisfy the projected weak form:
\[
\begin{aligned}
&\int_\Omega \partial_t u^N v \, dx + \int_\Omega \nabla u^N \cdot \nabla v \, dx + \int_\Omega \rho u^N v \, dx \\
&\quad + \beta \int_{\mathcal{S}} u^N v \, ds + \sum_{k=1}^m \alpha_k \int_{\mathcal{S}} (\partial_t \varphi_k^N - \Delta_{\mathcal{S}} \varphi_k^N) v \, ds = 0, \\
&\int_{\mathcal{S}} \partial_t \varphi_k^N \psi \, ds + \int_{\mathcal{S}} \nabla_{\mathcal{S}} \varphi_k^N \cdot \nabla_{\mathcal{S}} \psi \, ds + d_k \int_{\mathcal{S}} \varphi_k^N \psi \, ds = \int_{\mathcal{S}} u^N \psi \, ds.
\end{aligned}
\]

**Step 3: Solve the ODE System**  
Substituting \( u^N = \sum_{j=1}^N a_j(t) w_j \) and \( \varphi_k^N = \sum_{l=1}^N b_{k,l}(t) \theta_{k,l} \) into the projected equations yields a linear ODE system for \( a_j(t) \) and \( b_{k,l}(t) \). The coefficients are smooth (as eigenfunctions are \( C^\infty \)), ensuring continuity. By the Cauchy-Lipschitz theorem, unique solutions exist on \( [0,T] \).

\textcolor{Green}{\textbf{[Correct]} \small The agent gets the ODE satisfied with the approximated solution.}

\textcolor{Blue}{\textbf{[Vague]} \small It is better to have more detailed process.}

**Step 4: Uniform Energy Estimates**  
Testing with \( v = u^N \) and \( \psi = \varphi_k^N \), and using Gronwall’s inequality, we derive:
\[
\begin{aligned}
&\sup_{t \in [0,T]} \|u^N(t)\|_{L^2(\Omega)}^2 + \int_0^T \|\nabla u^N\|_{L^2(\Omega)}^2 dt + \int_0^T \|u^N\|_{L^2(\mathcal{S})}^2 dt \\
&\quad + \sum_{k=1}^m \left( \sup_{t \in [0,T]} \|\varphi_k^N(t)\|_{L^2(\mathcal{S})}^2 + \int_0^T \|\nabla_{\mathcal{S}} \varphi_k^N\|_{L^2(\mathcal{S})}^2 dt \right) \leq C \|u_0\|_{L^2(\Omega)}^2,
\end{aligned}
\]
uniformly in \( N \).

**Step 5: Weak and Strong Compactness**  
From the energy estimates, extract subsequences (still denoted \( u^N \), \( \varphi_k^N \)) such that:
- \( u^N \rightharpoonup u \) weakly-* in \( L^\infty(0,T;L^2(\Omega)) \) and weakly in \( L^2(0,T;H^1(\Omega)) \),
- \( \partial_t u^N \rightharpoonup \partial_t u \) weakly in \( L^2(0,T;H^{-1}(\Omega)) \),
- \( \varphi_k^N \rightharpoonup \varphi_k \) analogously in \( L^\infty(0,T;L^2(\mathcal{S})) \cap L^2(0,T;H^1(\mathcal{S})) \),
- By the compact trace embedding \( H^1(\Omega) \hookrightarrow L^2(\mathcal{S}) \), \( u^N|_{\mathcal{S}} \to u|_{\mathcal{S}} \) strongly in \( L^2(0,T;L^2(\mathcal{S})) \).

\textcolor{Green}{\textbf{[Correct]} \small Here the agents considers the weak convergence to get the limited function.}

**Step 6: Passage to the Limit**  
For fixed \( v \in V_{N_0} \) and \( \psi \in W_{k,N_0} \), the projected equations hold for all \( N \geq N_0 \). As \( N \to \infty \):
- Weak convergence handles volume integrals (e.g., \( \int \nabla u^N \cdot \nabla v \, dx \to \int \nabla u \cdot \nabla v \, dx \)).
- Strong trace convergence ensures \( \int_{\mathcal{S}} u^N v \, ds \to \int_{\mathcal{S}} u v \, ds \) and similar boundary terms.
- The density of \( \bigcup_{N} V_N \) in \( H^1(\Omega) \) and \( \bigcup_{N} W_{k,N} \) in \( H^1(\mathcal{S}) \) extends the equations to all \( v \in H^1(\Omega) \), \( \psi \in H^1(\mathcal{S}) \).

Thus, \( (u, \{\varphi_k\}) \) satisfies (2.10) in the weak sense. Uniqueness (by lemma) confirms the full sequence converges.

\textcolor{Blue}{\textbf{[Vague]} \small Here the regularity should have more deprivation process to make the proof sufficient accurate.}

\textcolor{Red}{\textbf{[Error]} \small  The uniqueness should be prove by the difference function rather than the sequence convergence.}
\end{proof}

\textbf{comment}: None

\textbf{type}: conjecture
\begin{lemma}

**Regularity of the Weak Solution**
The solution \( u \) belongs to \( C([0,T];L^2(\Omega)) \cap H^1([0,T];H^{-1}(\Omega)) \cap L^2([0,T];H^1(\Omega)) \), and each \( \varphi_k \) belongs to \( C([0,T];L^2(\mathcal{S})) \cap H^1([0,T];H^{-1}(\mathcal{S})) \cap L^2([0,T];H^1(\mathcal{S})) \).
\end{lemma}

\textbf{correctness}: True
\begin{proof}

**Step 1: Continuity in Time**  
From the energy estimate, \( u \in L^\infty(0,T;L^2(\Omega)) \) and \( \partial_t u \in L^2(0,T;H^{-1}(\Omega)) \). By the Lions-Magenes lemma, this implies \( u \in C([0,T];L^2(\Omega)) \). Similarly, \( \varphi_k \in C([0,T];L^2(\mathcal{S})) \).

**Step 2: Sobolev Regularity**  
The energy estimate directly gives \( u \in L^2(0,T;H^1(\Omega)) \). To verify \( u \in H^1([0,T];H^{-1}(\Omega)) \), observe that the PDE implies \( \partial_t u = \Delta u - \rho u - \text{boundary terms} \). Since \( \Delta u \in L^2(0,T;H^{-1}(\Omega)) \), \( \rho u \in L^2(0,T;L^2(\Omega)) \hookrightarrow L^2(0,T;H^{-1}(\Omega)) \), and the boundary terms (involving \( \varphi_k \)) are in \( L^2(0,T;H^{-1/2}(\mathcal{S})) \hookrightarrow L^2(0,T;H^{-1}(\Omega)) \), it follows that \( \partial_t u \in L^2(0,T;H^{-1}(\Omega)) \). Hence, \( u \in H^1([0,T];H^{-1}(\Omega)) \).

**Step 3: Regularity of \( \varphi_k \)**  
Analogously, \( \varphi_k \in L^2(0,T;H^1(\mathcal{S})) \cap H^1([0,T];H^{-1}(\mathcal{S})) \) follows from their PDE and the energy estimate. Continuity in time is similarly established.
\end{proof}

\textcolor{Blue}{\textbf{[Vague]} \small The proof details of the agent here are incomplete and detailed.}

\textbf{comment}: None
\textbf{type}: conjecture
\begin{lemma}

Please solve the following mathematical problem. You can use the galerkin method to get the approximated function series and try to prove the convergence and regularity by energy method so that finishing the theorem proving. Let  \(u\)  be a function defined on \((t,x)\in[0,\infty]\times\mathbb{R}^n\)  and satisfy the following equations:

\[
\begin{cases}
\mathcal{L}u=\partial_tu - \Delta u+\rho(x)u = 0, & t > 0, x\in\mathbb{R}^n\\
u = u_0, & t = 0
\end{cases}
\] 
where the operator \(\mathcal{L}=\partial_t-\Delta+\rho(x)\), \(\Delta\) is the Laplace operator, and \(\rho(x)\) is a non-negative function of \(x\).

 We focus on a bounded open set \(\Omega\) in \(\mathbb{R}^n\), and denote \(\partial\Omega=\mathcal{S}\). By restricting the function\(u\) to the bounded open set \(\Omega\), we investigate the behavior of the solution within this bounded open set. The absorbing boundary conditions are as follows 1):
\[
\begin{cases}
\mathcal{L}u = 0, & [0,\infty)\times\Omega\\
\frac{\partial u}{\partial\nu}=-\beta u-\sum_{k = 1}^{m}\alpha_k(\partial_t - \Delta_{\mathcal{S}})\varphi_k=\mathcal{F}_1u, & [0,\infty)\times\mathcal{S}\\
u = u_0, & t = 0
\end{cases}
\] (2.10)
\[
\begin{cases}
(\partial_t-\Delta_{\mathcal{S}} + d_k)\varphi_k = u, & t\geq0\\
\varphi_k = 0, & t = 0
\end{cases}
\]

Can you prove that for any \(u_0\in L^2(\Omega)\), there exists a unique \(u\in C([0,T];L^2(\Omega))\cap H^1([0,T];H^{-1}(\Omega))\cap L^2([0,T];H^1(\Omega))\), \(\varphi_k\in C([0,T];L^2(\mathcal{S}))\cap H^1([0,T];H^{-1}(\mathcal{S}))\cap L^2([0,T];H^1(\mathcal{S}))\) that satisfies the equations.
\end{lemma}

\textbf{correctness}: True
\begin{proof}

**Existence and Uniqueness Theorem**  
By the previous conjectures and the uniqueness lemma:  
1. **Existence:** The Galerkin method constructs approximations \( (u^N, \{\varphi_k^N\}) \) with uniform energy bounds. Weak limits \( (u, \{\varphi_k\}) \) satisfy (2.10) and the regularity follows from Conjecture 2.  
2. **Uniqueness:** The lemma ensures at most one solution.  
Thus, there exists a unique solution \( (u, \{\varphi_k\}) \) in the specified spaces.
\end{proof}

\textbf{comment}: None

\newpage
\section{Proof of High Contrast Limit Generated by AIM Based on DeepSeek-R1}

\textbf{type}: lemma
\begin{lemma}

The pressure field \( p_{\text{lim}} \) in the Stokes inclusion satisfies the uniform bound:
\[
\| p_{\text{lim}} \|_{L^2(D_{\epsilon})} \leq C \| g \|_{H^{-\frac{1}{2}}(\partial \Omega)},
\]
where \( C \) depends only on \( \lambda_0 \), the domain \( \Omega \), and the reference cell \( Y \), but is independent of \( \widetilde{\lambda} \geq \lambda_0 > 0 \) and \( \varepsilon \).
\end{lemma}

\textbf{correctness}: True
\begin{proof}

**Step 1: Two-Scale Formulation and Cell Problem Analysis**  
Consider the Stokes system in \( \Omega_\varepsilon \cup D_\varepsilon \) with viscosity ratio \( \widetilde{\lambda} \geq \lambda_0 \). Let \( \mathbf{u}_\varepsilon \) and \( p_\varepsilon \) satisfy:
\[
\begin{cases}
-2\mu\nabla\cdot(\nabla^s \mathbf{u}_\varepsilon) + \nabla p_\varepsilon = 0 & \text{in } \Omega_\varepsilon \cup D_\varepsilon, \\
\text{div}\, \mathbf{u}_\varepsilon = 0 & \text{in } \Omega, \\
\mathbf{u}_\varepsilon = 0 & \text{on } \partial D_\varepsilon, \\
\sigma(\mathbf{u}_\varepsilon, p_\varepsilon)\mathbf{n} = g & \text{on } \partial\Omega,
\end{cases}
\]
where \( \mu = 1 \) in \( \Omega_\varepsilon \) and \( \mu = \widetilde{\lambda} \) in \( D_\varepsilon \). Using two-scale convergence, assume \( \mathbf{u}_\varepsilon \twoheadrightarrow \mathbf{u}_0(x) \), \( p_\varepsilon \twoheadrightarrow p_0(x) \), with correctors \( \mathbf{u}_1(x,y) \), \( p_1(x,y) \).

\textcolor{Blue}{\textbf{[Vague]} \small The proof process is lack of the detail of two-scale convergence.}

**Step 2: Cell Problem and Effective Viscosity**  
The cell problem for \( \mathbf{w}^j(y) \) and \( \pi^j(y) \) in \( Y \) is:
\[
\begin{cases}
-2\widetilde{\lambda}\Delta_y \mathbf{w}^j + \nabla_y \pi^j = \mathbf{e}_j & \text{in } \omega, \\
-2\Delta_y \mathbf{w}^j + \nabla_y \pi^j = \mathbf{e}_j & \text{in } Y_f, \\
\text{div}_y \mathbf{w}^j = 0, \quad \mathbf{w}^j = 0 & \text{on } \partial\omega, \\
\text{periodic in } Y.
\end{cases}
\]
Testing with \( \mathbf{w}^j \), we derive the energy identity:
\[
2\widetilde{\lambda}\|\nabla_y \mathbf{w}^j\|^2_{L^2(\omega)} + 2\|\nabla_y \mathbf{w}^j\|^2_{L^2(Y_f)} = \int_Y \mathbf{w}^j \cdot \mathbf{e}_j \, dy.
\]
Using Poincaré inequality in \( \omega \) and \( Y_f \), we bound \( \|\mathbf{w}^j\|_{H^1(Y)} \leq C \max(1, \widetilde{\lambda}^{-1}) \). For \( \widetilde{\lambda} \geq \lambda_0 \), this becomes \( \|\nabla \mathbf{w}^j\|_{L^2(Y)} \leq C(\lambda_0) \), uniformly in \( \widetilde{\lambda} \). The effective viscosity \( \mu_{\text{eff}} = \langle \nabla_y \mathbf{w}^j + (\nabla_y \mathbf{w}^j)^T \rangle_Y \) satisfies \( \mu_{\text{eff}} \geq C\lambda_0 \) due to the coercivity of the cell problem.

\textcolor{Green}{\textbf{[Correct]} \small The proof process here is correct. The agent employs energy identities and some inequality techniques.}

**Step 3: Uniform Stability Estimates**  
The inf-sup condition for the homogenized system follows from the geometric regularity and the uniform bounds on \( \mathbf{w}^j \). For all \( \varepsilon > 0 \):
\[
\|p_\varepsilon\|_{L^2(\Omega)} \leq C \sup_{\mathbf{v} \in H^1_0(\Omega)^d} \frac{\int_\Omega p_\varepsilon \, \text{div}\, \mathbf{v}\, dx}{\|\mathbf{v}\|_{H^1}},
\]
with \( C \) independent of \( \varepsilon \) and \( \widetilde{\lambda} \geq \lambda_0 \). This is ensured by the uniform ellipticity of the homogenized operator and the \( \widetilde{\lambda} \)-independent bounds on the correctors.

**Step 4: Two-Scale Pressure Decomposition**  
The pressure splits as \( p_\varepsilon(x) = p_0(x) + \varepsilon p_1(x, x/\varepsilon) + \mathcal{O}(\varepsilon^2) \), where \( p_0 \) solves the homogenized Stokes system:
\[
-\text{div}(\mu_{\text{eff}} \nabla \mathbf{u}_0) + \nabla p_0 = 0, \quad \text{div}\, \mathbf{u}_0 = 0.
\]
Standard Stokes estimates give \( \|p_0\|_{L^2(\Omega)} \leq C\|g\|_{H^{-1/2}} \). The corrector \( p_1 \) satisfies \( \|p_1\|_{L^2(Y)} \leq C(\lambda_0) \|\nabla_y \mathbf{w}^j\|_{L^2(Y)} \leq C(\lambda_0) \).

\textcolor{Blue}{\textbf{[Vague]} \small It is necessary to write out the equations satisfied by the expanded functions here and combine them with the properties of the specific boundary.}

**Step 5: Oscillatory Pressure Contribution**  
In \( D_\varepsilon = \bigcup \omega^\mathbf{n}_\varepsilon \), the oscillatory term \( \varepsilon p_1(x, x/\varepsilon) \) averages over inclusions:
\[
\| \varepsilon p_1 \|_{L^2(D_\varepsilon)} \leq \varepsilon \|p_1\|_{L^2(D_\varepsilon)} \leq C\varepsilon \|p_1\|_{L^\infty(Y)} |D_\varepsilon|^{1/2} \leq C\varepsilon \cdot \varepsilon^{-d/2} \cdot \varepsilon^{d/2} = C\varepsilon \to 0.
\]
Thus, \( \|p_{\text{lim}}\|_{L^2(D_\varepsilon)} \) is dominated by \( \|p_0\|_{L^2(\Omega)} \cdot \sqrt{|\omega|/|Y|} + o(1) \), yielding the uniform bound as \( \varepsilon \to 0 \).

\textcolor{Green}{\textbf{[Correct]} \small The derivation process by the agent here is correct, and it has obtained the correct conclusion.}

**Conclusion**: The uniform bounds on the cell problem correctors, homogenized pressure estimates, and vanishing oscillations ensure \( \|p_{\text{lim}}\|_{L^2(D_\varepsilon)} \leq C(\lambda_0, \Omega, Y)\|g\|_{H^{-1/2}} \), independent of \( \varepsilon \) and \( \widetilde{\lambda} \geq \lambda_0 \).
\end{proof}

\textbf{comment}: None

\textbf{type}: lemma
\begin{lemma}

For the solutions \( u_{\epsilon} \) and \( u_{\mathrm{lim}} \) of problems (9) and (10), respectively, the divergence of \( u_{\epsilon} \) inside the inclusion \( D_{\epsilon} \) satisfies the estimate:
\[
\| \text{div } \mathbf{u}_\epsilon \|_{L^2(D_\epsilon)} \leq \frac{C}{\sqrt{\widetilde{\lambda}}} \| g \|_{H^{-\frac{1}{2}}(\partial \Omega)},
\]
where \( C \) is a constant independent of \( \widetilde{\lambda} \).
\end{lemma}

\textbf{correctness}: True
\begin{proof}

Consider the variational formulation of problem (9). For any test function $v \in H^1(\Omega) $:
\[
\int_{\Omega} \left[ \lambda(x) \text{div } \mathbf{u}_\epsilon \text{div } \mathbf{v} + 2\mu(x) D(\mathbf{u}_\epsilon) : D(\mathbf{v}) \right] dx = \langle g, \mathbf{v} \rangle_{\partial \Omega}.
\]
Testing with \( v= u_\epsilon \) yields:
\[
\int_{\Omega} \lambda(x) |\text{div } \mathbf{u}_\epsilon|^2 dx + 2\int_{\Omega} \mu(x) |D(\mathbf{u}_\epsilon)|^2 dx = \langle g, \mathbf{u}_\epsilon \rangle_{\partial \Omega}.
\]
Dropping the non-negative \( \mu \)-term and splitting \( \lambda(x) \):
\[
\widetilde{\lambda} \int_{D_\epsilon} |\text{div } \mathbf{u}_\epsilon|^2 dx \leq \langle g, \mathbf{u}_\epsilon \rangle \leq \| g \|_{H^{-1/2}} \| \mathbf{u}_\epsilon \|_{H^{1/2}}.
\]
By the trace theorem, \( \| \mathbf{u}_\epsilon \|_{H^{1/2}} \leq C_T \| \mathbf{u}_\epsilon \|_{H^1} \). To bound \( \| \mathbf{u}_\epsilon \|_{H^1} \), use the coercivity from the \( \mu \)-term:
\[
2\mu_0 \| D(\mathbf{u}_\epsilon) \|_{L^2}^2 \leq \langle g, \mathbf{u}_\epsilon \rangle \leq C_T \| g \|_{H^{-1/2}} \| \mathbf{u}_\epsilon \|_{H^1}.
\]

\textcolor{Blue}{\textbf{[Vague]} \small When applying trace theorem, clearly specifying the normed space is crucial for ensuring the rigor and correctness of the derivation.}

Korn’s inequality implies \( \| \mathbf{u}_\epsilon \|_{H^1} \leq C_K \| D(\mathbf{u}_\epsilon) \|_{L^2} \). Substituting:
\[
\| D(\mathbf{u}_\epsilon) \|_{L^2} \leq \frac{C_T C_K}{2\mu_0} \| g \|_{H^{-1/2}}.
\]
Thus, \( \| \mathbf{u}_\epsilon \|_{H^1} \leq C \| g \|_{H^{-1/2}} \). Substituting back:
\[
\widetilde{\lambda} \| \text{div } \mathbf{u}_\epsilon \|_{L^2(D_\epsilon)}^2 \leq C \| g \|_{H^{-1/2}}^2.
\]
Dividing by \( \widetilde{\lambda} \) and taking the square root:
\[
\| \text{div } \mathbf{u}_\epsilon \|_{L^2(D_\epsilon)} \leq \frac{C}{\sqrt{\widetilde{\lambda}}} \| g \|_{H^{-1/2}}.
\]

\textcolor{Green}{\textbf{[Correct]} \small The deprivation here is accurate.}
\end{proof}

\textbf{comment}: None

\textbf{type}: lemma
\begin{lemma}

The pressure field \( p_{\epsilon} \) in the Stokes inclusion of the original problem (9) satisfies the estimate:
\[
\| p_{\epsilon} + (\widetilde{\lambda} + \mu) \text{div } \mathbf{u}_{\epsilon} \|_{L^2(D_{\epsilon})} \leq \frac{C}{\sqrt{\widetilde{\lambda}}} \| g \|_{H^{-\frac{1}{2}}(\partial \Omega)},
\]
where \( C \) is independent of \( \widetilde{\lambda} \).
\end{lemma}

\textbf{correctness}: True
\begin{proof}

**Step 1: Stress Tensor Comparison and Pressure Definition**

In the inclusion domain \( D_{\epsilon} \), the stress tensor for problem (9) is:
\[
\sigma(\mathbf{u}_{\epsilon}) = \widetilde{\lambda} (\text{div } \mathbf{u}_{\epsilon}) I + 2\widetilde{\mu} \mathcal{D}(\mathbf{u}_{\epsilon}).
\]
By comparing with the stress tensor \( \sigma_{\text{lim}} = -p_{\text{lim}} I + 2\widetilde{\mu} \mathcal{D}(\mathbf{u}_{\text{lim}}) \) from the limit problem (10), we recognize that \( \widetilde{\lambda} \text{div } \mathbf{u}_{\epsilon} \) approximates \( -p_{\text{lim}} \) as \( \widetilde{\lambda} \to \infty \). Thus, define the pressure in \( D_{\epsilon} \) as:
\[
p_{\epsilon} := -\widetilde{\lambda} \text{div } \mathbf{u}_{\epsilon}.
\]

**Step 2: Divergence Estimate via Energy Method**

From the given lemma (validated by the reviewer), the divergence of \( \mathbf{u}_{\epsilon} \) satisfies:
\[
\| \text{div } \mathbf{u}_{\epsilon} \|_{L^2(D_{\epsilon})} \leq \frac{C}{\sqrt{\widetilde{\lambda}}} \| g \|_{H^{-\frac{1}{2}}(\partial \Omega)}.
\]
This result follows from the energy estimate derived by testing the weak formulation with \( \mathbf{u}_{\epsilon} \), applying Korn's inequality, and leveraging the high contrast parameter \( \widetilde{\lambda} \). The trace theorem and boundedness of \( \sigma(\mathbf{u}_{\epsilon}) \mathbf{n} \) on \( \partial D_{\epsilon} \) ensure the scaling \( 1/\sqrt{\widetilde{\lambda}} \), not \( 1/\widetilde{\lambda} \).

**Step 3: Pressure-Divergence Combination**

Substituting \( p_{\epsilon} = -\widetilde{\lambda} \text{div } \mathbf{u}_{\epsilon} \) into the target expression:
\[
p_{\epsilon} + (\widetilde{\lambda} + \mu) \text{div } \mathbf{u}_{\epsilon} = \mu \text{div } \mathbf{u}_{\epsilon}.
\]
Taking the \( L^2 \)-norm and applying the divergence estimate:
\[
\| p_{\epsilon} + (\widetilde{\lambda} + \mu) \text{div } \mathbf{u}_{\epsilon} \|_{L^2(D_{\epsilon})} = \| \mu \text{div } \mathbf{u}_{\epsilon} \|_{L^2(D_{\epsilon})} \leq \frac{C \mu}{\sqrt{\widetilde{\lambda}}} \| g \|_{H^{-\frac{1}{2}}(\partial \Omega)}.
\]
Absorbing \( \mu \) into the constant \( C \) (which depends only on domain geometry and reference parameters) yields the refined estimate:
\[
\| p_{\epsilon} + (\widetilde{\lambda} + \mu) \text{div } \mathbf{u}_{\epsilon} \|_{L^2(D_{\epsilon})} \leq \frac{C}{\sqrt{\widetilde{\lambda}}} \| g \|_{H^{-\frac{1}{2}}(\partial \Omega)}.
\]
\end{proof}

\textbf{comment}: None

\textbf{type}: theorem
\begin{theorem}
 The problem construction is following. 
 A real pair \((\lambda,\mu)\) is called an admissible Lamé parameter pair if it satisfies:  \[\mu>0\quad\text{and}\quad d\lambda+2\mu>0\] (1)  The static elasticity system (Lamé system) corresponding to the Lamé parameter pair \((\lambda,\mu)\) is:  \[\mathcal{L}_{\lambda,\mu}{\bf u}:=\mu\Delta{\bf u}+(\lambda+\mu)\nabla{\rm div}\,{\bf u}\] (2)  where the displacement field is denoted as \({\bf u}=(u^{1},\ldots,u^{d})\), and its divergence is:  \[{\rm div}\,{\bf u}=\sum_{i=1}^{d}\frac{\partial u^{i}}{\partial x_{i}}\]  The admissibility condition ensures the ellipticity of the Lamé operator, and physical laws guarantee that natural materials always satisfy this condition.  The Lamé operator can be expressed as \(\nabla\cdot\sigma({\bf u})\), where the stress tensor is:  \[\sigma({\bf u}):=\lambda(\nabla\cdot{\bf u})I_{d}+2\mu\mathcal{D}({\bf u})\] (3)  \[\mathcal{D}({\bf u})=\frac{1}{2}(\nabla+\nabla^{T}){\bf u}=\frac{1}{2}(\partial_{i}u^{j}+\partial_{j}u^{i})_{ij}\] (4)  Here, \(I_{d}\) is the identity matrix of order \(d\), and \(\mathcal{D}\) represents the symmetrized differential operator.  The normal derivative (boundary traction) on the boundary of a region \(E\) is defined as:  \[\left.\frac{\partial{\bf u}}{\partial\nu_{\lambda,\mu}}\right|_{\partial E}:=\sigma({\bf u})N=\lambda({\rm div}\,\,{\bf u})N+2\mu\mathcal{D}({\bf u})N\quad\text{on }\partial E\] (5)  where \(N\) is the unit outward normal vector on the boundary.  We use the standard Sobolev space \(H^{1}(\Omega)\) and its trace space \(H^{\frac{1}{2}}(\partial E)\) on the boundary, with the dual space \(H^{-\frac{1}{2}}(\partial E)\). Consider the space of rigid body motions in \(R^{d}\):  \[R:=\left\{{\bf r}=(r_{1},\ldots,r_{d})^{T}:\mathcal{D}({\bf r})=0\,\,\text{in }\,R^{d}\right\}\] (6)  The dimension of \(R\) is \(\frac{d(d+1)}{2}\), spanned by the following basis vectors:  \[\mathbf{e}_{1},\ldots,\mathbf{e}_{d},\,x_{j}\mathbf{e}_{i}-x_{i}\mathbf{e}_{j},\text{ for }1\leq i<j\leq d\]  where \(\mathbf{e}_{i}\) denotes the standard basis vector. These basis vectors are denoted as \(\mathbf{r}_{j},\,j=1,\ldots,\frac{d(d+1)}{2}\).  Define the space orthogonal to rigid body motions:  \[H_{R}^{-\frac{1}{2}}(\partial D_{\epsilon}):=\left\{\phi\in H^{-1}(\partial D_{\epsilon}):(\phi,\mathbf{r})_{(H^{\frac{1}{2}}(\partial D_{\epsilon i}),H^{-\frac{1}{2}}(\partial D_{\epsilon i}))}=0,\forall\mathbf{r}\in R\text{ and }1 \leq i\leq N\right\}\] (7)  Similarly, define \(H_{R}^{\frac{1}{2}}(\partial D_{\epsilon})\) and boundary spaces.  
 Consider the following transmission problem: \[\begin{cases}
\mathcal{L}_{\lambda,\mu}\mathbf{u}_{\epsilon} = 0 & \text{in } \Omega \setminus \overline{D_{\epsilon}}, \\
\mathcal{L}_{\widetilde{\lambda},\widetilde{\mu}}\mathbf{u}_{\epsilon} = 0 & \text{in } D_{\epsilon}, \\
\mathbf{u}_{\epsilon}|_{-} = \mathbf{u}_{\epsilon}|_{+} \text{ and } 
\left.\frac{\partial\mathbf{u}_{\epsilon}}{\partial\nu_{(\lambda,\widetilde{\mu})}}\right|_{-} = 
\left.\frac{\partial\mathbf{u}_{\epsilon}}{\partial\nu_{(\lambda,\mu)}}\right|_{+} & \text{on } \partial D_{\epsilon}, \\
\left.\frac{\partial\mathbf{u}_{\epsilon}}{\partial\nu_{(\lambda,\mu)}}\right|_{\partial\Omega} = g \in H_{R}^{-\frac{1}{2}}(\partial\Omega) \text{ and } 
\mathbf{u}_{\epsilon}|_{\partial\Omega} \in H_{R}^{\frac{1}{2}}(\partial\Omega).
\end{cases} \tag{8}\]

This is equivalent to:
\[
\begin{cases}
\mathcal{L}_{\lambda(x),\mu(x)}\mathbf{u}_{\epsilon} = \nabla \cdot [\lambda(x)(\nabla \cdot \mathbf{u}_{\epsilon})I_{d} + 2\mu(x)D(\mathbf{u}_{\epsilon})] = 0 & \text{in } \Omega, \\
\left.\frac{\partial\mathbf{u}_{\epsilon}}{\partial\nu_{\lambda,\mu}}\right|_{\partial\Omega} = g \in H_{R}^{-\frac{1}{2}}(\partial\Omega), \quad 
\mathbf{u}_{\epsilon}|_{\partial\Omega} \in H_{R}^{\frac{1}{2}}(\partial\Omega),
\end{cases} \tag{9}
\]
where $\lambda(x) = \lambda I_{\Omega \setminus D_{\epsilon}} + \widetilde{\lambda} I_{D_{\epsilon}}$, 
$\mu(x) = \mu \chi_{\Omega \setminus D_{\epsilon}} + \widetilde{\mu} \chi_{D_{\epsilon}}$

Suppose the solution of equation (9) above is $\mathbf{u}_{\epsilon}$. The existence and uniqueness of the solution $\mathbf{u}_{\epsilon}$ are guaranteed by the ellipticity condition (1) and the Lax-Milgram theorem.

Finally, the homogenization needed to be solved is the following. When $\widetilde{\lambda} \to \infty$ and $\widetilde{\mu}$ is fixed, the coupled system in the \textbf{incompressible inclusion limit} is obtained:
\[
\begin{cases}
\mathcal{L}_{\lambda,\mu}\mathbf{u}_{\epsilon} = 0 & \text{in } \Omega \setminus \overline{D_{\epsilon}}, \\
\mathcal{L}_{\widetilde{\mu}}(\mathbf{u}_{\epsilon}, p_{\epsilon}) = 0 \text{ and } \mathrm{div}\, \mathbf{u}_{\epsilon} = 0 & \text{in } D_{\epsilon}, \\
\mathbf{u}_{\epsilon}|_{-} = \mathbf{u}_{\epsilon}|_{+} \text{ and } 
\left.\frac{\partial(\mathbf{u}_{\epsilon}, p_{\epsilon})}{\partial\nu_{(\infty,\widetilde{\mu})}}\right|_{-} = 
\left.\frac{\partial\mathbf{u}_{\epsilon}}{\partial\nu_{(\lambda,\mu)}}\right|_{+} & \text{on } \partial D_{\epsilon}, \\
\left.\frac{\partial\mathbf{u}_{\epsilon}}{\partial\nu_{(\lambda,\mu)}}\right|_{\partial\Omega} = g \in H_{R}^{-\frac{1}{2}}(\partial\Omega) \text{ and } 
\mathbf{u}_{\epsilon}|_{\partial\Omega} \in H_{R}^{\frac{1}{2}}(\partial\Omega),
\end{cases} \tag{10}
\]
where $\mathcal{L}_{\widetilde{\mu}}(\mathbf{u}_{\epsilon}, p_{\epsilon}) = \widetilde{\mu}\Delta\mathbf{u}_{\epsilon} + \nabla p_{\epsilon}$ represents the Stokes operator with viscosity coefficient $\widetilde{\mu}$, and $p_{\epsilon}$ is the pressure field. Its normal derivative is defined as 
\[
\left.\frac{\partial(\mathbf{u},p)}{\partial\nu_{(\infty,\mu)}}\right|_{-} := pN + 2\mu D(\mathbf{u})N.
\]

We suppose the solution of this limit problem (10) is $\mathbf{u}_{\mathrm{lim}}$. Please prove that for some constant $C$, there holds
\[
\Vert \mathbf{u}_{\mathrm{lim}} - \mathbf{u}_{\epsilon} \Vert \leq \frac{C}{\widetilde{\lambda}} \Vert g \Vert_{H^{-\frac{1}{2}}(\partial \Omega)}.
\]

\end{theorem}
\textbf{correctness}: True
\begin{proof}

To establish the estimate \[
\Vert \mathbf{u}_{\text{lim}} - \mathbf{u}_{\epsilon} \Vert_{H^1(\Omega)} \leq \frac{C}{\widetilde{\lambda}} \Vert g \Vert_{H^{-\frac{1}{2}}(\partial \Omega)},
\]
we proceed with an energy estimate comparing the solutions \(\mathbf{u}_{\epsilon}\) and \(\mathbf{u}_{\text{lim}}\), leveraging the provided lemmas.

\textcolor{Green}{\textbf{[Correct]} \small The agent has constructed the steps and methods of the proof. }

**Step 1: Define the Difference and Apply Variational Formulation**

Let \(\mathbf{w} = \mathbf{u}_{\text{lim}} - \mathbf{u}_{\epsilon}\). Subtracting the variational equations for \(\mathbf{u}_{\epsilon}\) (original problem) and \(\mathbf{u}_{\text{lim}}\) (limit problem) tested with \(\mathbf{v} = \mathbf{w}\) yields:

\[
\int_{\Omega \setminus D_{\epsilon}} \left[ \lambda (\text{div } \mathbf{w})^2 + 2\mu |D(\mathbf{w})|^2 \right] dx + \int_{D_{\epsilon}} \left[ 2\widetilde{\mu} |D(\mathbf{w})|^2 + \widetilde{\lambda} (\text{div } \mathbf{u}_{\epsilon})^2 \right] dx = \int_{D_{\epsilon}} \mathcal{R} \, dx,
\]

where the residual term \(\mathcal{R}\) arises from the mismatch in the equations and pressures. Using Lemma 3, \(\mathcal{R}\) is bounded by \(C / \widetilde{\lambda} \Vert g \Vert^2\).

\textcolor{Green}{\textbf{[Correct]} \small The result of the above lemma is applied here, and the proof process is accurate. This step is well completed.}

**Step 2: Utilize the Lemmas to Bound Residual Terms**

From Lemma 2, \(\|\text{div } \mathbf{u}_{\epsilon}\|_{L^2(D_{\epsilon})} \leq C / \sqrt{\widetilde{\lambda}} \Vert g \Vert\), and Lemma 3 implies \(p_{\epsilon} + (\widetilde{\lambda} + \mu) \text{div } \mathbf{u}_{\epsilon}\) is bounded similarly. Substituting these into the residual term:

\[
\int_{D_{\epsilon}} \widetilde{\lambda} (\text{div } \mathbf{u}_{\epsilon})^2 dx \leq C \Vert g \Vert^2,
\]

while the Stokes pressure \(p_{\text{lim}}\) is bounded by Lemma 1. The remaining terms involving \(D(\mathbf{w})\) are controlled by the ellipticity of the Lamé and Stokes systems.

**Step 3: Apply Korn's Inequality and Ellipticity**

Korn's inequality ensures that \(\| \mathbf{w} \|_{H^1(\Omega)} \leq C \left( \| D(\mathbf{w}) \|_{L^2(\Omega)} + \| \mathbf{w} \|_{L^2(\Omega)} \right)\). The boundary conditions and orthogonality to rigid motions eliminate the \(L^2\) term, yielding:

\[
\| \mathbf{w} \|_{H^1(\Omega)} \leq C \| D(\mathbf{w}) \|_{L^2(\Omega)}.
\]

Combining this with the energy estimate and Lemma 2-3 results in:

\[
\| \mathbf{w} \|_{H^1(\Omega)} \leq \frac{C}{\widetilde{\lambda}} \Vert g \Vert_{H^{-\frac{1}{2}}(\partial \Omega)}.
\]

Thus, the desired estimate is proven.

\textcolor{Blue}{\textbf{[Vague]} \small 
Finally, the model employs inequality techniques and the previously obtained conclusions to complete the final proof process. However, there are still some missing details in the proof here.}

\end{proof}

\textbf{comment}: None

\textbf{type}: conjecture
\begin{lemma}

For the solution \(\mathbf{u}_\epsilon\) of the original problem (9) with large \(\widetilde{\lambda}\), the divergence in the inclusion region satisfies the uniform bound:
\[
\|\operatorname{div} \mathbf{u}_\epsilon\|_{L^2(D_\epsilon)} \leq \frac{C}{\widetilde{\lambda}} \|g\|_{H^{-1/2}(\partial\Omega)},
\]
where \(C\) is independent of \(\widetilde{\lambda}\).
\end{lemma}

\textbf{correctness}: True
\begin{proof}

**Step 1: Energy Estimate**  
Testing the variational formulation with \(\mathbf{v} = \mathbf{u}_\epsilon\) gives:
\[
\int_\Omega \left[\widetilde{\lambda} \chi_{D_\epsilon} |\operatorname{div} \mathbf{u}_\epsilon|^2 + 2\mu(x) |D\mathbf{u}_\epsilon|^2 \right] dx = \int_{\partial\Omega} g \cdot \mathbf{u}_\epsilon \, ds.
\]
By Korn's inequality and elliptic regularity, \(\|\mathbf{u}_\epsilon\|_{H^1(\Omega)} \leq C \|g\|_{H^{-1/2}(\partial\Omega)}\). This implies:
\[
\widetilde{\lambda} \|\operatorname{div} \mathbf{u}_\epsilon\|_{L^2(D_\epsilon)}^2 + \|D\mathbf{u}_\epsilon\|_{L^2(\Omega)}^2 \leq C \|g\|_{H^{-1/2}}^2. \tag{1}
\]

\textcolor{Green}{\textbf{[Correct]} \small The agent considers variational formulation method to get the energy estimate and uses some inequality skills to deduce the result.}

**Step 2: Global Duality Argument**  
Let \(\phi \in H_0^1(\Omega)\) solve the Poisson equation:
\[
-\Delta \phi = \chi_{D_\epsilon} \operatorname{div} \mathbf{u}_\epsilon \quad \text{in } \Omega, \quad \phi = 0 \ \text{on } \partial\Omega.
\]
By elliptic regularity, \(\|\nabla \phi\|_{H^1(\Omega)} \leq C \|\operatorname{div} \mathbf{u}_\epsilon\|_{L^2(D_\epsilon)}\). Define \(\mathbf{v} = \nabla \phi \in H^1(\Omega)\), which inherits regularity from \(\phi \in H^2(\Omega)\).

\textcolor{Blue}{\textbf{[Vague]} \small Here the agent considers the elliptic regularity, but it is better to have a more detailed specific process.}
 
**Step 3: Corrected Test with \(\mathbf{v} = \nabla \phi\)**  
Substituting \(\mathbf{v}\) into the variational equation:
\[
-\widetilde{\lambda} \int_{D_\epsilon} (\operatorname{div} \mathbf{u}_\epsilon)^2 dx + 2\int_\Omega \mu(x) D\mathbf{u}_\epsilon : D(\nabla \phi) \, dx = \int_{\partial\Omega} g \cdot \nabla \phi \, ds. \tag{2}
\]
Rearranging and taking absolute values:
\[
\widetilde{\lambda} \|\operatorname{div} \mathbf{u}_\epsilon\|_{L^2(D_\epsilon)}^2 \leq \left| 2\int_\Omega \mu D\mathbf{u}_\epsilon : D(\nabla \phi) dx \right| + \left| \int_{\partial\Omega} g \cdot \nabla \phi \, ds \right|.
\]
Using Hölder inequalities and trace duality:
\[
\left| 2\int_\Omega \mu D\mathbf{u}_\epsilon : D(\nabla \phi) dx \right| \leq C \|g\|_{H^{-1/2}} \|\operatorname{div} \mathbf{u}_\epsilon\|_{L^2(D_\epsilon)},
\]
\[
\left| \int_{\partial\Omega} g \cdot \nabla \phi \, ds \right| \leq C \|g\|_{H^{-1/2}} \|\operatorname{div} \mathbf{u}_\epsilon\|_{L^2(D_\epsilon)}.
\]

**Step 4: Final Estimate**  
Combining these bounds:
\[
\widetilde{\lambda} \|\operatorname{div} \mathbf{u}_\epsilon\|_{L^2(D_\epsilon)}^2 \leq 2C \|g\|_{H^{-1/2}} \|\operatorname{div} \mathbf{u}_\epsilon\|_{L^2(D_\epsilon)}.
\]
Dividing by \(\|\operatorname{div} \mathbf{u}_\epsilon\|_{L^2(D_\epsilon)}\) yields:
\[
\|\operatorname{div} \mathbf{u}_\epsilon\|_{L^2(D_\epsilon)} \leq \frac{2C}{\widetilde{\lambda}} \|g\|_{H^{-1/2}},
\]
completing the proof with modified constant \(C\).  

\textcolor{Green}{\textbf{[Correct]} \small Use inequality techniques to derive the correct key conclusions from variational equations.}

\end{proof}

\textbf{comment}: None

\textbf{type}: conjecture
\begin{lemma}

The pressure \(p_\epsilon = -\widetilde{\lambda} \operatorname{div} \mathbf{u}_\epsilon\) in \(D_\epsilon\) satisfies:
\[
\|p_\epsilon\|_{L^2(D_\epsilon)} \leq C \|g\|_{H^{-1/2}(\partial\Omega)}.
\]
\end{lemma}

\textbf{correctness}: True
\begin{proof}

From the definition \(p_\epsilon = -\widetilde{\lambda} \operatorname{div} \mathbf{u}_\epsilon\), using the previous conjecture:
\[
\|p_\epsilon\|_{L^2(D_\epsilon)} = \widetilde{\lambda} \|\operatorname{div} \mathbf{u}_\epsilon\|_{L^2(D_\epsilon)} \leq C \|g\|_{H^{-1/2}}.
\]
The boundedness follows from compensating the \(\widetilde{\lambda}\) factor with the decay of \(\operatorname{div} \mathbf{u}_\epsilon\).
\end{proof}

\textcolor{Green}{\textbf{[Correct]} \small Based on the lemma above, it gets a norm bound concerning the pressure term.}

\textbf{comment}: None

\textbf{type}: conjecture
\begin{lemma}

The difference \(\mathbf{w}_\epsilon = \mathbf{u}_\epsilon - \mathbf{u}_{\text{lim}}\) between the solutions of the original problem (9) and the limit problem (10) satisfies the energy estimate:
\[
\int_\Omega \left[ \lambda(x) |\operatorname{div} \mathbf{w}_\epsilon|^2 + 2\mu(x) |D(\mathbf{w}_\epsilon)|^2 \right] dx \leq \frac{C}{\widetilde{\lambda}} \|g\|_{H^{-1/2}(\partial\Omega)}^2,
\]
where \(C\) is independent of \(\widetilde{\lambda}\). Consequently, by Korn's inequality, \(\|\mathbf{w}_\epsilon\|_{H^1(\Omega)} \leq \frac{C}{\sqrt{\widetilde{\lambda}}} \|g\|_{H^{-1/2}(\partial\Omega)}\).
\end{lemma}

\textbf{correctness}: True
\begin{proof}

**Step 1: Variational Formulations**  
The original problem (9) has the weak form: Find \(\mathbf{u}_\epsilon \in H^1(\Omega)\) such that for all \(\mathbf{v} \in H^1(\Omega)\),
\[
\int_\Omega \left[ \lambda(x) \operatorname{div} \mathbf{u}_\epsilon \operatorname{div} \mathbf{v} + 2\mu(x) D(\mathbf{u}_\epsilon) : D(\mathbf{v}) \right] dx = \int_{\partial\Omega} g \cdot \mathbf{v} \, ds.
\]

\textcolor{Green}{\textbf{[Correct]}  \small Consider the weak form of this equation.}

The limit problem (10) is a saddle-point problem: Find \((\mathbf{u}_{\text{lim}}, p_{\text{lim}}) \in H^1(\Omega) \times L^2(D_\epsilon)\) such that for all \(\mathbf{v} \in H^1(\Omega)\) and \(q \in L^2(D_\epsilon)\),
\[
\begin{cases}
\int_{\Omega \setminus D_\epsilon} \left[ \lambda \operatorname{div} \mathbf{u}_{\text{lim}} \operatorname{div} \mathbf{v} + 2\mu D(\mathbf{u}_{\text{lim}}) : D(\mathbf{v}) \right] dx + \int_{D_\epsilon} \left[ 2\widetilde{\mu} D(\mathbf{u}_{\text{lim}}) : D(\mathbf{v}) - p_{\text{lim}} \operatorname{div} \mathbf{v} \right] dx = \int_{\partial\Omega} g \cdot \mathbf{v} \, ds, \\
\int_{D_\epsilon} q \operatorname{div} \mathbf{u}_{\text{lim}} \, dx = 0.
\end{cases}
\]

\textcolor{Blue}{\textbf{[Vague]} \small Although considering the limited function, but it does not show the detailed deprivation of the existence and the property of this solution.}

**Step 2: Subtract the Weak Forms**  
Let \(\mathbf{w}_\epsilon = \mathbf{u}_\epsilon - \mathbf{u}_{\text{lim}}\). Subtract the weak form of the limit problem from the original problem, tested with \(\mathbf{v} = \mathbf{w}_\epsilon\):
\[
\int_{\Omega \setminus D_\epsilon} \left[ \lambda (|\operatorname{div} \mathbf{w}_\epsilon|^2 + 2\mu |D(\mathbf{w}_\epsilon)|^2) \right] dx + \int_{D_\epsilon} \left[ \widetilde{\lambda} |\operatorname{div} \mathbf{u}_\epsilon|^2 + 2\widetilde{\mu} |D(\mathbf{w}_\epsilon)|^2 \right] dx = \int_{D_\epsilon} p_{\text{lim}} \operatorname{div} \mathbf{w}_\epsilon \, dx.
\]
Since \(\operatorname{div} \mathbf{u}_{\text{lim}} = 0\) in \(D_\epsilon\), substitute \(\operatorname{div} \mathbf{w}_\epsilon = \operatorname{div} \mathbf{u}_\epsilon\) in \(D_\epsilon\):
\[
\int_{D_\epsilon} p_{\text{lim}} \operatorname{div} \mathbf{w}_\epsilon \, dx = \int_{D_\epsilon} p_{\text{lim}} \operatorname{div} \mathbf{u}_\epsilon \, dx.
\]
\textcolor{Green}{\textbf{[Correct]} \small According to the above weakly form and the setting condition of the problem, the model correctly obtains the above key conclusion.}

**Step 3: Bound \( p_{\text{lim}} \)**  
From the limit problem’s variational equation, choose test functions \(\mathbf{v} \in H^1_0(D_\epsilon)\) satisfying \(\operatorname{div} \mathbf{v} = p_{\text{lim}}\) with \(\|\mathbf{v}\|_{H^1(D_\epsilon)} \leq C \|p_{\text{lim}}\|_{L^2(D_\epsilon)}\) (by the inf-sup condition). Substituting \(\mathbf{v}\) into the limit problem and using Korn’s inequality on \(\mathbf{u}_{\text{lim}}\), we derive:
\[
\|p_{\text{lim}}\|_{L^2(D_\epsilon)} \leq C \|g\|_{H^{-1/2}(\partial\Omega)}.
\]
By Lemma 1, \(\|\operatorname{div} \mathbf{u}_\epsilon\|_{L^2(D_\epsilon)} \leq \frac{C}{\widetilde{\lambda}} \|g\|_{H^{-1/2}(\partial\Omega)}\). Applying Cauchy-Schwarz:
\[
\left| \int_{D_\epsilon} p_{\text{lim}} \operatorname{div} \mathbf{u}_\epsilon \, dx \right| \leq \|p_{\text{lim}}\|_{L^2(D_\epsilon)} \|\operatorname{div} \mathbf{u}_\epsilon\|_{L^2(D_\epsilon)} \leq \frac{C}{\widetilde{\lambda}} \|g\|_{H^{-1/2}(\partial\Omega)}^2.
\]

\textcolor{Green}{\textbf{[Correct]} \small The model correctly applied the above conclusions and mathematical techniques.}

**Step 4: Energy Estimate**  
Combining results:
\[
\int_{\Omega} \left[ \lambda(x) |\operatorname{div} \mathbf{w}_\epsilon|^2 + 2\mu(x) |D(\mathbf{w}_\epsilon)|^2 \right] dx \leq \frac{C}{\widetilde{\lambda}} \|g\|_{H^{-1/2}(\partial\Omega)}^2.
\]

**Step 5: Apply Korn's Inequality**  
Using Korn’s inequality (applicable due to boundary conditions suppressing rigid motions) and the energy estimate:
\[
\|\mathbf{w}_\epsilon\|_{H^1(\Omega)}^2 \leq C \int_\Omega |D(\mathbf{w}_\epsilon)|^2 dx \leq \frac{C}{\mu_0} \int_\Omega 2\mu(x)|D(\mathbf{w}_\epsilon)|^2 dx \leq \frac{C}{\mu_0 \widetilde{\lambda}} \|g\|_{H^{-1/2}(\partial\Omega)}^2,
\]
where \(\mu_0 = \inf_\Omega \mu(x) > 0\). Taking square roots:
\[
\|\mathbf{w}_\epsilon\|_{H^1(\Omega)} \leq \frac{C}{\sqrt{\widetilde{\lambda}}} \|g\|_{H^{-1/2}(\partial\Omega)}.
\]

\textcolor{Green}{\textbf{[Correct]} \small Finally, the final result is obtained through the application of Korn's inequality and finish this proof.}

\end{proof}

\textbf{comment}: None

\textbf{type}: conjecture
\begin{lemma}

The difference \(\mathbf{w}_\epsilon = \mathbf{u}_\text{lim} - \mathbf{u}_\epsilon\) between the solutions of the limit problem (10) and the original problem (9) satisfies the energy estimate:
\[
\int_\Omega \left[ \lambda(x) |\operatorname{div} \mathbf{w}_\epsilon|^2 + 2\mu(x) |D(\mathbf{w}_\epsilon)|^2 \right] dx \leq \frac{C}{\widetilde{\lambda}} \|g\|_{H^{-1/2}(\partial\Omega)}^2,
\]
leading to \(\|\mathbf{w}_\epsilon\|_{H^1(\Omega)} \leq \frac{C}{\sqrt{\widetilde{\lambda}}} \|g\|_{H^{-1/2}(\partial\Omega)}\).
\end{lemma}

\textbf{correctness}: True

\textcolor{Green}{\textbf{[Correct]} \small This is the same conclusion as the former lemma.}

\begin{proof}

1. **Variational Equations**: Let \(\mathbf{u}_\epsilon\) solve (9) and \(\mathbf{u}_\text{lim}\) solve (10). For any \(\mathbf{v} \in H^1(\Omega)\):
   \[
   \int_\Omega \left[ \lambda \operatorname{div}\mathbf{u}_\epsilon \operatorname{div}\mathbf{v} + 2\mu D(\mathbf{u}_\epsilon):D(\mathbf{v}) \right] dx = \int_{\partial\Omega} g \cdot \mathbf{v} \, ds,
   \]
   \[
   \int_\Omega \left[ \lambda \operatorname{div}\mathbf{u}_\text{lim} \operatorname{div}\mathbf{v} + 2\mu D(\mathbf{u}_\text{lim}):D(\mathbf{v}) \right] dx + \int_{D_\epsilon} p_\text{lim} \operatorname{div}\mathbf{v} \, dx = \int_{\partial\Omega} g \cdot \mathbf{v} \, ds.
   \]

\textcolor{Green}{}
   
2. **Subtraction**: Taking \(\mathbf{v} = \mathbf{w}_\epsilon = \mathbf{u}_\text{lim} - \mathbf{u}_\epsilon\):
   \[
   \int_\Omega \left[ \lambda |\operatorname{div}\mathbf{w}_\epsilon|^2 + 2\mu |D(\mathbf{w}_\epsilon)|^2 \right] dx = \int_{D_\epsilon} (p_\text{lim} + \widetilde{\lambda} \operatorname{div}\mathbf{u}_\epsilon) \operatorname{div}\mathbf{w}_\epsilon \, dx.
   \]
   
3. **Pressure Analysis**: By Lemma 2, \(\|\widetilde{\lambda} \operatorname{div}\mathbf{u}_\epsilon\|_{L^2(D_\epsilon)} \leq C\|g\|_{H^{-1/2}}\). From Stokes regularity, \(\|p_\text{lim}\|_{L^2(D_\epsilon)} \leq C\|g\|_{H^{-1/2}}\). Thus:
   \[
   \|p_\text{lim} + \widetilde{\lambda} \operatorname{div}\mathbf{u}_\epsilon\|_{L^2(D_\epsilon)} \leq C\|g\|_{H^{-1/2}}.
   \]
   
4. **Cauchy-Schwarz Inequality**:
   \[
   \left| \int_{D_\epsilon} (p_\text{lim} + \widetilde{\lambda} \operatorname{div}\mathbf{u}_\epsilon) \operatorname{div}\mathbf{w}_\epsilon \, dx \right| \leq C\|g\|_{H^{-1/2}} \|\operatorname{div}\mathbf{w}_\epsilon\|_{L^2(D_\epsilon)}.
   \]
   
5. **Divergence Bound**: From Lemma 1, \(\|\operatorname{div}\mathbf{w}_\epsilon\|_{L^2(D_\epsilon)} \leq \|\operatorname{div}\mathbf{u}_\epsilon\|_{L^2(D_\epsilon)} \leq C\widetilde{\lambda}^{-1}\|g\|_{H^{-1/2}}\). Substituting:
   \[
   \text{RHS} \leq C\|g\|_{H^{-1/2}} \cdot (C\widetilde{\lambda}^{-1}\|g\|_{H^{-1/2}}) = \frac{C}{\widetilde{\lambda}}\|g\|_{H^{-1/2}}^2.
   \]
   
6. **Korn's Inequality**: The left-hand side is coercive via Korn's inequality:
   \[
   \|\mathbf{w}_\epsilon\|_{H^1(\Omega)}^2 \leq C \int_\Omega \left[ \lambda|\operatorname{div}\mathbf{w}_\epsilon|^2 + 2\mu|D(\mathbf{w}_\epsilon)|^2 \right] dx \leq \frac{C}{\widetilde{\lambda}}\|g\|_{H^{-1/2}}^2,
   \]
   yielding \(\|\mathbf{w}_\epsilon\|_{H^1(\Omega)} \leq C\widetilde{\lambda}^{-1/2}\|g\|_{H^{-1/2}}\).  
\end{proof}

\textbf{comment}: None

\textbf{type}: conjecture
\begin{lemma}

The difference \(\mathbf{w}_\varepsilon = \mathbf{u}_{\text{lim}} - \mathbf{u}_\varepsilon\) satisfies the \(L^2(\Omega)\) error estimate:
\[
\|\mathbf{w}_\varepsilon\|_{L^2(\Omega)} \leq \frac{C}{\sqrt{\widetilde{\lambda}}} \|g\|_{H^{-1/2}(\partial\Omega)},
\]
where \(C\) is independent of \(\widetilde{\lambda}\).
\end{lemma}

\textbf{correctness}: True
\begin{proof}

**Revised Proof:**

Let \(\mathbf{w}_\varepsilon = \mathbf{u}_{\text{lim}} - \mathbf{u}_\varepsilon\). Consider the adjoint problem for the coupled Lamé-Stokes system: For \(\mathbf{f} \in L^2(\Omega)\), find \(\mathbf{z} \in H^1(\Omega)\) and pressure \(q \in L^2(D_\varepsilon)\) satisfying
\[
\begin{cases}
\mathcal{L}_{\lambda, \mu} \mathbf{z} = \mathbf{f} & \text{in } \Omega \setminus \overline{D_\varepsilon}, \\
-\widetilde{\mu} \Delta \mathbf{z} + \nabla q = \mathbf{f}, \quad \mathrm{div}\, \mathbf{z} = 0 & \text{in } D_\varepsilon, \\
\mathbf{z}|_- = \mathbf{z}|_+, \quad \left(\widetilde{\mu}\frac{\partial \mathbf{z}}{\partial \nu} - q\nu\right)\Big|_- = \left.\frac{\partial \mathbf{z}}{\partial \nu_{(\lambda, \mu)}}\right|_+ & \text{on } \partial D_\varepsilon, \\
\left.\frac{\partial \mathbf{z}}{\partial \nu_{\lambda, \mu}}\right|_{\partial\Omega} = 0, \quad \mathbf{z}|_{\partial\Omega} \in H^{1/2}_R(\partial\Omega).
\end{cases}
\]
By elliptic regularity, \(\mathbf{z} \in H^2(\Omega)\) with:
\[
\|\mathbf{z}\|_{H^2(\Omega)} + \|q\|_{H^1(D_\varepsilon)} \leq C \|\mathbf{f}\|_{L^2(\Omega)}.
\]

Using the duality argument, we test \(\mathbf{w}_\varepsilon\) against \(\mathbf{z}\):
\[
\int_\Omega \mathbf{w}_\varepsilon \cdot \mathbf{f} \, dx = B(\mathbf{w}_\varepsilon, \mathbf{z}) - \int_{D_\varepsilon} q \, \mathrm{div}\, \mathbf{w}_\varepsilon \, dx,
\]
where \(B(\cdot, \cdot)\) is the Lamé bilinear form. Since \(\mathbf{u}_{\text{lim}}\) is incompressible in \(D_\varepsilon\), \(\mathrm{div}\, \mathbf{w}_\varepsilon = -\mathrm{div}\, \mathbf{u}_\varepsilon\). From Lemma 1:
\[
\|\mathrm{div}\, \mathbf{u}_\varepsilon\|_{L^2(D_\varepsilon)} \leq \frac{C}{\widetilde{\lambda}} \|g\|_{H^{-1/2}(\partial\Omega)}.
\]

\textcolor{Blue}{\textbf{[Vague]} \small Here the elliptic regularity and the bilinear form should have the specific definition and form. Therefore, in mathematical proofs, this is not rigorous enough.}

**Pressure Term:** Substitute \(p_\varepsilon = -\widetilde{\lambda} \, \mathrm{div}\, \mathbf{u}_\varepsilon\) (Lemma 2) and use Hölder's inequality:
\[
\left| \int_{D_\varepsilon} q \, \mathrm{div}\, \mathbf{w}_\varepsilon \, dx \right| = \left| \int_{D_\varepsilon} q \, \mathrm{div}\, \mathbf{u}_\varepsilon \, dx \right| \leq \frac{1}{\widetilde{\lambda}} \|q\|_{L^2(D_\varepsilon)} \|p_\varepsilon\|_{L^2(D_\varepsilon)} \leq \frac{C}{\widetilde{\lambda}} \|\mathbf{f}\|_{L^2(\Omega)} \|g\|_{H^{-1/2}(\partial\Omega)}.
\]

**Bilinear Form:** From Lemma 4, \(\|\mathbf{w}_\varepsilon\|_{H^1(\Omega)} \leq \frac{C}{\sqrt{\widetilde{\lambda}}} \|g\|_{H^{-1/2}(\partial\Omega)}\). Using continuity of \(B(\cdot, \cdot)\) in \(H^1(\Omega)\):
\[
|B(\mathbf{w}_\varepsilon, \mathbf{z})| \leq C \|\mathbf{w}_\varepsilon\|_{H^1(\Omega)} \|\mathbf{z}\|_{H^1(\Omega)} \leq \frac{C}{\sqrt{\widetilde{\lambda}}} \|g\|_{H^{-1/2}(\partial\Omega)} \|\mathbf{f}\|_{L^2(\Omega)}.
\]

\textcolor{Blue}{\textbf{[Vague]} \small The same as the former, it should show the continuity of the bilinear form \(B(\cdot, \cdot)\) to get this conclusion.}

Combining both terms:
\[
\left| \int_\Omega \mathbf{w}_\varepsilon \cdot \mathbf{f} \, dx \right| \leq \left( \frac{C}{\sqrt{\widetilde{\lambda}}} + \frac{C}{\widetilde{\lambda}} \right) \|g\|_{H^{-1/2}(\partial\Omega)} \|\mathbf{f}\|_{L^2(\Omega)}.
\]

For \(\widetilde{\lambda} \gg 1\), the dominant term is \(O(1/\sqrt{\widetilde{\lambda}})\). Choosing \(\mathbf{f} = \mathbf{w}_\varepsilon\) and applying duality:
\[
\|\mathbf{w}_\varepsilon\|_{L^2(\Omega)}^2 \leq \frac{C}{\sqrt{\widetilde{\lambda}}} \|g\|_{H^{-1/2}(\partial\Omega)} \|\mathbf{w}_\varepsilon\|_{L^2(\Omega)},
\]
which simplifies to:
\[
\|\mathbf{w}_\varepsilon\|_{L^2(\Omega)} \leq \frac{C}{\sqrt{\widetilde{\lambda}}} \|g\|_{H^{-1/2}(\partial\Omega)}.
\]

\textcolor{Green}{\textbf{[Correct]} \small Here, the model correctly applies mathematical techniques and comes to the right conclusions. And we can see the clear mathematical thought here.}

\end{proof}

\textbf{comment}: None

\textbf{type}: conjecture
\begin{lemma}

The difference \(\mathbf{w}_\varepsilon = \mathbf{u}_{\text{lim}} - \mathbf{u}_\varepsilon\) between the solutions of the limit problem (10) and the original problem (9) satisfies the improved energy estimate:
\[
\|\mathbf{w}_\varepsilon\|_{H^1(\Omega)} \leq \frac{C}{\sqrt{\widetilde{\lambda}}} \|g\|_{H^{-1/2}(\partial\Omega)},
\]
where \(C\) is independent of \(\widetilde{\lambda}\).
\end{lemma}

\textbf{correctness}: True
\begin{proof}

1. **Energy Identity**: Subtract the weak formulations of (9) and (10) and test with \(\mathbf{w}_\varepsilon\). This gives:
\[
\int_\Omega \left[ \lambda(x) |\operatorname{div} \mathbf{w}_\varepsilon|^2 + 2\mu(x) |D(\mathbf{w}_\varepsilon)|^2 \right] dx + \int_{D_\varepsilon} \widetilde{\lambda} (\operatorname{div} \mathbf{u}_\varepsilon)(\operatorname{div} \mathbf{w}_\varepsilon) dx = 0.
\]

2. **Divergence Relationship**: Since \(\mathbf{u}_{\text{lim}}\) is incompressible in \(D_\varepsilon\) (from (10)), we have \(\operatorname{div} \mathbf{w}_\varepsilon = -\operatorname{div} \mathbf{u}_\varepsilon\) in \(D_\varepsilon\). Substitute this into the energy identity:
\[
\int_\Omega \left[ \lambda(x) |\operatorname{div} \mathbf{w}_\varepsilon|^2 + 2\mu(x) |D(\mathbf{w}_\varepsilon)|^2 \right] dx = \widetilde{\lambda} \|\operatorname{div} \mathbf{u}_\varepsilon\|_{L^2(D_\varepsilon)}^2.
\]

3. **Lemma 1 Application**: By Lemma 1, \(\|\operatorname{div} \mathbf{u}_\varepsilon\|_{L^2(D_\varepsilon)} \leq \frac{C}{\widetilde{\lambda}} \|g\|_{H^{-1/2}(\partial\Omega)}\). Thus:
\[
\text{Energy integral} \leq \frac{C}{\widetilde{\lambda}} \|g\|_{H^{-1/2}(\partial\Omega)}^2.
\]

4. **Korn's Inequality**: Apply Korn's inequality to \(\mathbf{w}_\varepsilon\). Under boundary conditions orthogonal to rigid motions, there exists \(C_K > 0\) such that:
\[
\|\mathbf{w}_\varepsilon\|_{H^1(\Omega)}^2 \leq C_K \int_\Omega \left[ |D(\mathbf{w}_\varepsilon)|^2 + |\operatorname{div} \mathbf{w}_\varepsilon|^2 \right] dx.
\]
Using the energy bound:
\[
\|\mathbf{w}_\varepsilon\|_{H^1(\Omega)}^2 \leq \frac{C_K C}{\widetilde{\lambda}} \|g\|_{H^{-1/2}(\partial\Omega)}^2.
\]
Taking square roots completes the proof:
\[
\|\mathbf{w}_\varepsilon\|_{H^1(\Omega)} \leq \frac{C}{\sqrt{\widetilde{\lambda}}} \|g\|_{H^{-1/2}(\partial\Omega)}.
\]
\end{proof}

\textbf{comment}: None

\textbf{type}: conjecture
\begin{lemma}

The difference \(\mathbf{w}_\varepsilon = \mathbf{u}_{\text{lim}} - \mathbf{u}_\varepsilon\) between the solutions of the limit problem (10) and the original problem (9) satisfies the homogenized error estimate in the energy norm:
\[
\left( \int_\Omega \left[ \lambda(x) |\operatorname{div} \mathbf{w}_\varepsilon|^2 + 2\mu(x) |\mathcal{D}(\mathbf{w}_\varepsilon)|^2 \right] dx \right)^{1/2} \leq \frac{C}{\sqrt{\widetilde{\lambda}}} \|g\|_{H^{-1/2}(\partial\Omega)},
\]
where \(C\) is independent of \(\widetilde{\lambda}\). Consequently, in the natural energy norm associated with the Lamé system, the error is of order \(\mathcal{O}(\widetilde{\lambda}^{-1/2})\). 
\end{lemma}

\textbf{correctness}: True
\begin{proof}

**Step 1: Variational Formulations and Key Bounds**

The original problem (9) and limit problem (10) yield solutions \(\mathbf{u}_\varepsilon\) and \(\mathbf{u}_{\text{lim}}\), respectively. By Lemma 2, the divergence in \(D_\varepsilon\) satisfies:
\[
\|\operatorname{div} \mathbf{u}_\varepsilon\|_{L^2(D_\varepsilon)} \leq \frac{C}{\widetilde{\lambda}} \|g\|_{H^{-1/2}(\partial\Omega)}.
\]
The pressure \(p_\varepsilon = -\widetilde{\lambda} \operatorname{div} \mathbf{u}_\varepsilon\) is bounded in \(L^2(D_\varepsilon)\) by Lemma 2.

**Step 2: Energy Norm Decomposition**

Define \(\mathbf{w}_\varepsilon = \mathbf{u}_{\text{lim}} - \mathbf{u}_\varepsilon\). Let \(\lambda(x) = \widetilde{\lambda}\) in \(D_\varepsilon\) and \(\lambda(x) = \lambda_0\) (fixed) in \(\Omega \setminus D_\varepsilon\). The energy norm splits as:
\[
\int_\Omega \lambda(x)|\operatorname{div} \mathbf{w}_\varepsilon|^2 dx = \underbrace{\int_{D_\varepsilon} \widetilde{\lambda}|\operatorname{div} \mathbf{w}_\varepsilon|^2 dx}_{I_1} + \underbrace{\int_{\Omega \setminus D_\varepsilon} \lambda_0|\operatorname{div} \mathbf{w}_\varepsilon|^2 dx}_{I_2}.
\]
In \(D_\varepsilon\), \(\operatorname{div} \mathbf{w}_\varepsilon = -\operatorname{div} \mathbf{u}_\varepsilon\) (since \(\mathbf{u}_{\text{lim}}\) is divergence-free there). Thus:
\[
I_1 = \widetilde{\lambda} \|\operatorname{div} \mathbf{u}_\varepsilon\|_{L^2(D_\varepsilon)}^2 \leq \frac{C}{\widetilde{\lambda}} \|g\|_{H^{-1/2}(\partial\Omega)}^2.
\]
For \(I_2\), use the \(H^1\)-bound from Lemma 4:
\[
I_2 \leq \lambda_0 \|\operatorname{div} \mathbf{w}_\varepsilon\|_{L^2(\Omega)}^2 \leq \lambda_0 C \|\mathbf{w}_\varepsilon\|_{H^1(\Omega)}^2 \leq \frac{C}{\widetilde{\lambda}} \|g\|_{H^{-1/2}(\partial\Omega)}^2,
\]
where \(\lambda_0\) is absorbed into \(C\).

\textcolor{Green}{\textbf{[Correct]} \small The energy terms are clearly split and the inequality technique is correctly applied to control the norms of the divergence of error.}

**Step 3: Symmetric Gradient via Korn’s Inequality**

By Lemma 4 and Korn’s inequality:
\[
\int_\Omega 2\mu|\mathcal{D}(\mathbf{w}_\varepsilon)|^2 dx \leq C \|\mathbf{w}_\varepsilon\|_{H^1(\Omega)}^2 \leq \frac{C}{\widetilde{\lambda}} \|g\|_{H^{-1/2}(\partial\Omega)}^2.
\]

**Step 4: Final Energy Estimate**

Combining all contributions:
\[
\int_\Omega \left[ \lambda|\operatorname{div} \mathbf{w}_\varepsilon|^2 + 2\mu|\mathcal{D}(\mathbf{w}_\varepsilon)|^2 \right] dx \leq \frac{C}{\widetilde{\lambda}} \|g\|^2 + \frac{C}{\widetilde{\lambda}} \|g\|^2 = \frac{C}{\widetilde{\lambda}} \|g\|^2.
\]
Taking the square root confirms:
\[
\left( \int_\Omega \left[ \lambda|\operatorname{div} \mathbf{w}_\varepsilon|^2 + 2\mu|\mathcal{D}(\mathbf{w}_\varepsilon)|^2 \right] dx \right)^{1/2} \leq \frac{C}{\sqrt{\widetilde{\lambda}}} \|g\|_{H^{-1/2}(\partial\Omega)}.
\]
The error is rigorously shown to be \(\mathcal{O}(\widetilde{\lambda}^{-1/2})\) in the energy norm. 

\textcolor{Blue}{\textbf{[Vague]} \small Here the Korn's inequality is considered again, proving the final conclusion. But there is a stronger conclusion to be drawn here, but our agent is  not aware of that.}

\end{proof}

\textbf{comment}: None

\textbf{type}: conjecture
\begin{lemma}

Under the given assumptions, the difference \(\mathbf{w}_\varepsilon = \mathbf{u}_{\text{lim}} - \mathbf{u}_\varepsilon\) between the solutions of the limit problem (10) and the original problem (9) satisfies the homogenized error estimate in the energy norm weighted by the Lamé parameters:
\[
\left( \int_\Omega \left[ \lambda(x) |\operatorname{div} \mathbf{w}_\varepsilon|^2 + 2\mu(x) |\mathcal{D}(\mathbf{w}_\varepsilon)|^2 \right] dx \right)^{1/2} \leq \frac{C}{\sqrt{\widetilde{\lambda}}} \|g\|_{H^{-\frac{1}{2}}(\partial\Omega)},
\]
where \(C\) is independent of \(\widetilde{\lambda}\). This implies that the error in the energy norm decays proportionally to \(\widetilde{\lambda}^{-\frac{1}{2}}\).
\end{lemma}

\textbf{correctness}: True
\begin{proof}

To establish the energy estimate for \(\mathbf{w}_\varepsilon\), consider the variational formulations of problems (9) and (10). For any test function \(\mathbf{v} \in H^1(\Omega)\), the original problem satisfies:
\[
\int_\Omega \left[ \lambda(x) \operatorname{div} \mathbf{u}_\varepsilon \operatorname{div} \mathbf{v} + 2\mu(x) \mathcal{D}(\mathbf{u}_\varepsilon):\mathcal{D}(\mathbf{v}) \right] dx = \langle g, \mathbf{v} \rangle_{\partial\Omega},
\]
while the limit problem satisfies:
\[
\int_\Omega \left[ \lambda(x) \operatorname{div} \mathbf{u}_{\text{lim}} \operatorname{div} \mathbf{v} + 2\mu(x) \mathcal{D}(\mathbf{u}_{\text{lim}}):\mathcal{D}(\mathbf{v}) \right] dx + \int_{D_\varepsilon} p \operatorname{div} \mathbf{v} \, dx = \langle g, \mathbf{v} \rangle_{\partial\Omega},
\]
where \(p \in L^2(D_\varepsilon)\) is the Lagrange multiplier enforcing the incompressibility constraint (\(\operatorname{div} \mathbf{u}_{\text{lim}} = 0\) in \(D_\varepsilon\)).

Subtracting these equations and letting \(\mathbf{v} = \mathbf{w}_\varepsilon = \mathbf{u}_{\text{lim}} - \mathbf{u}_\varepsilon\), we obtain:
\[
\int_\Omega \left[ \lambda(x) |\operatorname{div} \mathbf{w}_\varepsilon|^2 + 2\mu(x) |\mathcal{D}(\mathbf{w}_\varepsilon)|^2 \right] dx + \int_{D_\varepsilon} p \operatorname{div} \mathbf{w}_\varepsilon \, dx = 0.
\]

In \(D_\varepsilon\), \(\operatorname{div} \mathbf{u}_{\text{lim}} = 0\), so \(\operatorname{div} \mathbf{w}_\varepsilon = -\operatorname{div} \mathbf{u}_\varepsilon\). Substituting \(p_\varepsilon = -\widetilde{\lambda} \operatorname{div} \mathbf{u}_\varepsilon\) (from the original problem's constitutive law) into the pressure term:
\[
\int_{D_\varepsilon} p \operatorname{div} \mathbf{w}_\varepsilon \, dx = -\int_{D_\varepsilon} p \operatorname{div} \mathbf{u}_\varepsilon \, dx = \frac{1}{\widetilde{\lambda}} \int_{D_\varepsilon} p \, p_\varepsilon \, dx.
\]

By the Cauchy-Schwarz inequality and Lemma 2 (\(\|p_\varepsilon\|_{L^2(D_\varepsilon)} \leq C \|g\|_{H^{-1/2}(\partial\Omega)}\)), along with the stability of the limit problem (\(\|p\|_{L^2(D_\varepsilon)} \leq C \|g\|_{H^{-1/2}(\partial\Omega)}\) via the inf-sup condition), we obtain:
\[
\frac{1}{\widetilde{\lambda}} \int_{D_\varepsilon} p \, p_\varepsilon \, dx \leq \frac{C}{\widetilde{\lambda}} \|p\|_{L^2(D_\varepsilon)} \|p_\varepsilon\|_{L^2(D_\varepsilon)} \leq \frac{C}{\widetilde{\lambda}} \|g\|_{H^{-1/2}(\partial\Omega)}^2.
\]

Thus, the energy integral satisfies:
\[
\int_\Omega \left[ \lambda(x) |\operatorname{div} \mathbf{w}_\varepsilon|^2 + 2\mu(x) |\mathcal{D}(\mathbf{w}_\varepsilon)|^2 \right] dx \leq \frac{C}{\widetilde{\lambda}} \|g\|_{H^{-1/2}(\partial\Omega)}^2.
\]

Taking the square root yields the refined estimate:
\[
\left( \int_\Omega \lambda(x) |\operatorname{div} \mathbf{w}_\varepsilon|^2 + 2\mu(x) |\mathcal{D}(\mathbf{w}_\varepsilon)|^2 dx \right)^{1/2} \leq \frac{C}{\sqrt{\widetilde{\lambda}}} \|g\|_{H^{-1/2}(\partial\Omega)},
\]
confirming the \(\widetilde{\lambda}^{-\frac{1}{2}}\) decay rate in the energy norm.

\textcolor{Green}{\textbf{[Correct]} \small The proof of the final conclusion is a detailed proof, correct mathematical derivation and analysis. According to the above conclusions, the error control in global space is obtained, and the results are correct and accurate.}

\textcolor{Red}{\textbf{[Error]} \small But this conclusion is not strong enough, and this is not the needed result.}
\end{proof}

\textbf{comment}: None

\newpage
\section{Proof of High Contrast Limit Generated by AIM Based on OpenAI o4-mini}

\title{Explore Trajectory of AIM}
\author{AI Mathematician}
\date{\today}

\maketitle

\textbf{type}: assumption
\textbf{content}: Consider a region \(D\subset\Omega\subset \mathbb{R}^{d}\,(d\geq 2)\), where \(\Omega\) represents the domain of an elastic material, and \(D\) is the inclusion region with high contrast.
\textbf{type}: assumption
\textbf{content}: 
1. $\Omega$ is a bounded open domain with a connected Lipschitz boundary $\partial\Omega$.  
2. $D$ is an open domain with a finite number of connected components, each having a Lipschitz boundary $\partial D$.  
3. $\Omega\setminus D$ is connected and has a Lipschitz boundary $\partial\Omega\cup\partial D$.  
The connected components of $D$ are denoted as $D_{i}$, $i=1,\ldots,N$, where $N$ is finite.

\textbf{type}: assumption
\textbf{content}: 
Given \(\varepsilon\in(0,1)\), \(D=D_{\varepsilon}\) consists of small inclusions arranged in an \(\varepsilon\)-periodic pattern. The construction steps are as follows:  
1. Take the unit cell \(Y=(-\frac{1}{2},\frac{1}{2})^{d}\).  
2. \(\omega\subset Y\) is a simply connected open subset with a Lipschitz boundary and satisfies \(\text{dist}(\omega,\partial Y)>0\). Here, \(Y_{f}=Y\setminus\overline{\omega}\) represents the model environment at the unit scale.  
3. For a given \(\varepsilon>0\) and \(\mathbf{n}\in \mathbb{Z}^{d}\), denote \(\varepsilon(\mathbf{n}+Y)\) and \(\varepsilon(\mathbf{n}+\omega)\) as \(Y^{\mathbf{n}}_{\varepsilon}\) and \(\omega^{\mathbf{n}}_{\varepsilon}\), respectively. Define the lattice point set \(\Pi_{\varepsilon}\) as the index set satisfying \(\overline{Y^{\mathbf{n}}_{\varepsilon}}\subset\Omega\):  
\[
\Pi_{\varepsilon}:=\left\{\mathbf{n}\in \mathbb{Z}^{d}:\overline{Y^{\mathbf{n}}_{\varepsilon}}\subset\Omega\right\}
\]  
The inclusion region \(D=D_{\varepsilon}\) and the matrix region \(\Omega_{\varepsilon}\) are defined as:  
\[
D_{\varepsilon}:=\bigcup_{\mathbf{n}\in\Pi_{\varepsilon}}\omega^{\mathbf{n}}_{\varepsilon},\quad\Omega_{\varepsilon}:=\Omega\setminus\overline{D_{\varepsilon}}
\]  
For each fixed \(\varepsilon\), the number of connected components of \(D_{\varepsilon}\) is \(N=|\Pi_{\varepsilon}|\).  
4. Define \(Y_{\varepsilon}\) as the union of all \(\varepsilon\)-cells in \(\Omega\), and \(K_{\varepsilon}\) as the buffer region:  
\[
K_{\varepsilon}=\Omega\setminus\left(\bigcup_{\mathbf{n}\in\Pi_{\varepsilon}}\overline{Y^{\mathbf{n}}_{\varepsilon}}\right),\quad Y_{\varepsilon}=\Omega\setminus\overline{K_{\varepsilon}}
\]  
For \(D_{\varepsilon}\) constructed according to Assumption 2, it can be verified that all conditions of Assumption 1 are satisfied.

\textbf{type}: assumption
\textbf{content}: Consider a region \(D\subset\Omega\subset \mathbb{R}^{d}\,(d\geq 2)\), where \(\Omega\) represents the domain of an elastic material, and \(D\) is the inclusion region with high contrast.
\textbf{type}: assumption
\textbf{content}: 
1. \(\Omega\) is a bounded open domain with a connected Lipschitz boundary \(\partial\Omega\).
2. \(D\) is an open domain with a finite number of connected components, each having a Lipschitz boundary \(\partial D\).
3. \(\Omega\setminus D\) is connected and has a Lipschitz boundary \(\partial\Omega\cup\partial D\). The connected components of \(D\) are denoted as \(D_{i}\), \(i=1,\ldots,N\), where \(N\) is finite.

\textbf{type}: assumption
\textbf{content}: 
Given \(\varepsilon\in(0,1)\), \(D=D_{\varepsilon}\) consists of small inclusions arranged in an \(\varepsilon\)-periodic pattern. The construction steps are as follows:  
1. Take the unit cell \(Y=(-\frac{1}{2},\frac{1}{2})^{d}\).  
2. \(\omega\subset Y\) is a simply connected open subset with a Lipschitz boundary and satisfies \(\text{dist}(\omega,\partial Y)>0\). Here, \(Y_{f}=Y\setminus\overline{\omega}\) represents the model environment at the unit scale.  
3. For a given \(\varepsilon>0\) and \(\mathbf{n}\in \mathbb{Z}^{d}\), denote \(\varepsilon(\mathbf{n}+Y)\) and \(\varepsilon(\mathbf{n}+\omega)\) as \(Y^{\mathbf{n}}_{\varepsilon}\) and \(\omega^{\mathbf{n}}_{\varepsilon}\), respectively. Define the lattice point set \(\Pi_{\varepsilon}\) as the index set satisfying \(\overline{Y^{\mathbf{n}}_{\varepsilon}}\subset\Omega\):  
\[\Pi_{\varepsilon}:=\left\{\mathbf{n}\in \mathbb{Z}^{d}:\overline{Y^{\mathbf{n}}_{\varepsilon}}\subset\Omega\right\}\]  
The inclusion region \(D=D_{\varepsilon}\) and the matrix region \(\Omega_{\varepsilon}\) are defined as:  
\[D_{\varepsilon}:=\bigcup_{\mathbf{n}\in\Pi_{\varepsilon}}\omega^{\mathbf{n}}_{\varepsilon},\quad\Omega_{\varepsilon}:=\Omega\setminus\overline{D_{\varepsilon}}\]  
For each fixed \(\varepsilon\), the number of connected components of \(D_{\varepsilon}\) is \(N=|\Pi_{\varepsilon}|\).  
4. Define \(Y_{\varepsilon}\) as the union of all \(\varepsilon\)-cells in \(\Omega\), and \(K_{\varepsilon}\) as the buffer region:  
\[K_{\varepsilon}=\Omega\setminus\left(\bigcup_{\mathbf{n}\in\Pi_{\varepsilon}}\overline{Y^{\mathbf{n}}_{\varepsilon}}\right),\quad Y_{\varepsilon}=\Omega\setminus\overline{K_{\varepsilon}}\]  
For \(D_{\varepsilon}\) constructed according to Assumption 2, it can be verified that all conditions of Assumption 1 are satisfied.

\textbf{type}: assumption
\textbf{content}: Consider a region \(D\subset\Omega\subset \mathbb{R}^{d}\,(d\geq 2)\), where \(\Omega\) represents the domain of an elastic material, and \(D\) is the inclusion region with high contrast.
\textbf{type}: assumption
\textbf{content}: 
1. \(\Omega\) is a bounded open domain with a connected Lipschitz boundary \(\partial\Omega\).  
2. \(D\) is an open domain with a finite number of connected components, each having a Lipschitz boundary \(\partial D\).  
3. \(\Omega\setminus D\) is connected and has a Lipschitz boundary \(\partial\Omega\cup\partial D\).  
The connected components of \(D\) are denoted as \(D_{i}\), \(i=1,\ldots,N\), where \(N\) is finite.

\textbf{type}: assumption
\textbf{content}: 
Given \(\varepsilon\in(0,1)\), \(D=D_{\varepsilon}\) consists of small inclusions arranged in an \(\varepsilon\)-periodic pattern. The construction steps are as follows:  
1. Take the unit cell \(Y=(-\frac{1}{2},\frac{1}{2})^{d}\).  
2. \(\omega\subset Y\) is a simply connected open subset with a Lipschitz boundary and satisfies \(\text{dist}(\omega,\partial Y)>0\).  
Here, \(Y_{f}=Y\setminus\overline{\omega}\) represents the model environment at the unit scale.  
3. For a given \(\varepsilon>0\) and \(\mathbf{n}\in \mathbb{Z}^{d}\), denote \(\varepsilon(\mathbf{n}+Y)\) and \(\varepsilon(\mathbf{n}+\omega)\) as \(Y^{\mathbf{n}}_{\varepsilon}\) and \(\omega^{\mathbf{n}}_{\varepsilon}\), respectively. Define the lattice point set \(\Pi_{\varepsilon}\) as the index set satisfying \(\overline{Y^{\mathbf{n}}_{\varepsilon}}\subset\Omega\):  
\[\Pi_{\varepsilon}:=\left\{\mathbf{n}\in \mathbb{Z}^{d}:\overline{Y^{\mathbf{n}}_{\varepsilon}}\subset\Omega\right\}\]  
The inclusion region \(D=D_{\varepsilon}\) and the matrix region \(\Omega_{\varepsilon}\) are defined as:  
\[D_{\varepsilon}:=\bigcup_{\mathbf{n}\in\Pi_{\varepsilon}}\omega^{\mathbf{n}}_{\varepsilon},\quad\Omega_{\varepsilon}:=\Omega\setminus\overline{D_{\varepsilon}}\]  
For each fixed \(\varepsilon\), the number of connected components of \(D_{\varepsilon}\) is \(N=|\Pi_{\varepsilon}|\).  
4. Define \(Y_{\varepsilon}\) as the union of all \(\varepsilon\)-cells in \(\Omega\), and \(K_{\varepsilon}\) as the buffer region:  
\[K_{\varepsilon}=\Omega\setminus\left(\bigcup_{\mathbf{n}\in\Pi_{\varepsilon}}\overline{Y^{\mathbf{n}}_{\varepsilon}}\right),\quad Y_{\varepsilon}=\Omega\setminus\overline{K_{\varepsilon}}\]  
For \(D_{\varepsilon}\) constructed according to Assumption 2, it can be verified that all conditions of Assumption 1 are satisfied.

\textbf{type}: lemma
\begin{lemma}

Energy estimate for the difference in the matrix region.  
Let \( u_{\mathrm{lim}} \) solve the limit (Stokes) problem in \( D_{\epsilon} \) and elasticity in \( \Omega_{\epsilon} \), and let \( u_{\epsilon} \) solve the full Lam\'e problem with large first Lam\'e parameter \( \tilde{\lambda} \gg 1 \) in \( D_{\epsilon} \). Then the ``error'' \( w := u_{\epsilon} - u_{\mathrm{lim}} \) satisfies in \( \Omega_{\epsilon} \) the homogeneous Lam\'e system with zero Neumann data on \( \partial \Omega \) and a ``jump--residue'' source on \( \partial D_{\epsilon} \), and one has the a priori bound
\[
\|w\|_{H^1(\Omega_{\epsilon})} \leq C \|\text{[traction jump]}\|_{H^{-\frac{1}{2}}(\partial D_{\epsilon})}.
\]
The constant \( C \) depends only on the background Lam\'e parameters \( (\lambda, \mu) \) and the geometry, not on \( \epsilon \) or \( \tilde{\lambda} \).
\end{lemma}

\textbf{correctness}: True
\begin{proof}

\begin{proof}
1. (Equation for $w$.) By subtracting the two variational formulations one finds that $w \in H^1(\Omega_e)$ satisfies
\[
\int_{\Omega_e} \left(2\mu\,\mathcal{D}(w):\mathcal{D}(\varphi) + (\lambda+\mu)\,\mathrm{div}\,w \cdot \mathrm{div}\,\varphi\right) dx
= -\langle[\text{traction jump}], \varphi \rangle_{\partial D_e},
\]
for all $\varphi \in H^1(\Omega_e)$ vanishing on $\partial\Omega$.

2. (Coercivity.) By the uniform Korn inequality (Conjecture 1) and the ellipticity $d\lambda+2\mu > 0$ we conclude the bilinear form in the left-hand side is coercive on $H^1_0(\Omega_e)$, namely
\[
a(w,w) \geq c_0 \|w\|_{H^1(\Omega_e)}^2.
\]

3. (Duality bound.) The right-hand side is bounded by the standard trace-inequality:
\[
|\langle[\text{jump}], w \rangle| \leq \|[\text{jump}]\|_{H^{-1/2}(\partial D_e)} \|w\|_{H^{1/2}(\partial D_e)}
\leq C \|[\text{jump}]\|_{H^{-1/2}(\partial D_e)} \|w\|_{H^1(\Omega_e)}.
\]

4. (Conclusion.) By Lax--Milgram,
\[
c_0 \|w\|_{H^1(\Omega_e)}^2 \leq C \|[\text{jump}]\|_{H^{-1/2}(\partial D_e)} \|w\|_{H^1(\Omega_e)},
\]
so that
\[
\|w\|_{H^1(\Omega_e)} \leq (C/c_0) \|[\text{jump}]\|_{H^{-1/2}(\partial D_e)},
\]
as claimed.
\end{proof}
\end{proof}

\textbf{comment}: None
\textbf{type}: lemma
\begin{lemma}

(Uniform Coercivity of the Elasticity Bilinear Form)  
Let $(\lambda,\mu)$ satisfy the admissibility $\lambda + \frac{2\mu}{d} > 0$, $\mu > 0$. For each perforated domain $\Omega_{\varepsilon}$ as above, define the bilinear form on $V_{\varepsilon}$ by  
\[
a_{\varepsilon}(u,v) := \int_{\Omega_{\varepsilon}} \left[2\mu D(u):D(v) + \lambda (\operatorname{div} u)(\operatorname{div} v)\right] \, dx.  
\]  
Then there exists $\alpha > 0$, independent of $\varepsilon$, such that for all $u \in V_{\varepsilon}$,  
\[
a_{\varepsilon}(u,u) \geq \alpha \|u\|^2_{H^1(\Omega_{\varepsilon})}.  
\]  
\end{lemma}

\textcolor{Blue}{\textbf{[Vague]}  \small Here the agent gives the energy estimate of the error, but the process form in the proof is not detailed.}

\textbf{correctness}: True
\begin{proof}

\begin{proof}
We combine the uniform Korn inequality just proved with the ellipticity of the Lam\'e form.

1. Ellipticity on Symmetric Gradients.  
By the admissibility condition ($d\lambda+2\mu>0$, $\mu>0$), the quadratic form on symmetric matrices $E\in\mathbb{R}^{d\times d}_{\mathrm{sym}}$,  
\[
Q(E) := 2\mu\,E:E + \lambda\,(\mathrm{tr}\, E)^2  
\]  
satisfies the pointwise bound  
\[
Q(E) \geq c_0\,|E|^2,  
\]  
where one checks easily that $c_0 = \min\{2\mu, d\lambda+2\mu\}>0$.

2. Lower Bound on $a^\varepsilon(u,u)$.  
Hence for any $u\in H^1(\Omega^\varepsilon;\mathbb{R}^d)$,  
\[
a^\varepsilon(u,u) = \int_{\Omega^\varepsilon} Q(D(u))\,dx \geq c_0\,\int_{\Omega^\varepsilon} |D(u)|^2\,dx.
\]

3. Apply Korn--Poincar\'e.  
By Conjecture 1 above, for all $u\in V^\varepsilon$,  
\[
\|u\|^2_{H^1(\Omega^\varepsilon)} \leq C^2\,\|D(u)\|^2_{L^2(\Omega^\varepsilon)},  
\]  
with $C$ independent of $\varepsilon$. Hence  
\[
a^\varepsilon(u,u) \geq c_0\,\|D(u)\|^2_{L^2(\Omega^\varepsilon)} \geq (c_0/C^2)\,\|u\|^2_{H^1(\Omega^\varepsilon)}.  
\]

4. Set $\alpha = c_0/C^2$.  
This $\alpha$ depends only on $\mu$, $\lambda$ and the geometry (through $C$), but is uniform in $\varepsilon$. Thus the form is uniformly coercive on $V^\varepsilon$.  
\qedhere
\end{proof}
\end{proof}
\textcolor{Blue}{\textbf{[Vague]} \small The conclusion here is basically correct, but the process requires more proof details,
which are not given here.}

\textbf{comment}: None
\textbf{type}: lemma
\begin{lemma}

(Uniform Korn Inequality on Periodically Perforated Domains, corrected)  
Let $\Omega \subset \mathbb{R}^d$ ($d \geq 2$) be a bounded connected Lipschitz domain. Fix a simply-connected Lipschitz inclusion $\omega \subset (-\frac{1}{2}, \frac{1}{2})^d$ with $\operatorname{dist}(\omega, \partial(-\frac{1}{2}, \frac{1}{2})^d) > 0$, and for $\varepsilon \in (0,1)$ let $D_\varepsilon \subset \Omega$ be the $\varepsilon$--periodic union of translated copies of $\varepsilon \cdot \omega$ as in Assumption~2. Assume in particular that each hole $\varepsilon(\omega + \mathbb{Z}^d \cap \Omega)$ is at distance $\geq c\varepsilon$ from $\partial \Omega$ and from its neighbors, and that $\Omega_\varepsilon := \Omega \setminus \overline{D_\varepsilon}$ is connected and Lipschitz with constants independent of $\varepsilon$. Define  
\[
V_\varepsilon := \left\{ u \in H^1(\Omega_\varepsilon; \mathbb{R}^d) : \int_{\partial \Omega} u \cdot r = 0 \text{ for every rigid motion } r \right\}.
\]
Then there exists a constant $C > 0$, independent of $\varepsilon \in (0,1)$, such that for all $u \in V_\varepsilon$  
\[
\|u\|_{H^1(\Omega_\varepsilon)} \leq C \|D(u)\|_{L^2(\Omega_\varepsilon)},
\]
where $D(u) = \frac{\nabla u + (\nabla u)^\top}{2}$ is the symmetrized gradient.  
\end{lemma}

\textbf{correctness}: True
\begin{proof}

\begin{proof}
We follow the standard three-step strategy of ``uniform extension + Korn on the fixed domain + pull-back,'' paying attention to the correct directions of the estimates.

1. Uniform extension from $\Omega_\varepsilon$ to $\Omega$.  
   By the periodicity, the uniform separation of holes, and the Lipschitz regularity of $\partial\Omega$, one constructs (e.g.\ via the methods of Oleinik--Shamaev--Yosifian or Cioranescu--Donato) a linear extension operator  
     \[
     E_\varepsilon: H^1(\Omega_\varepsilon;\mathbb{R}^d) \to H^1(\Omega;\mathbb{R}^d)
     \]
   such that for some constant $C_{\text{ext}}$ independent of $\varepsilon$, and for all $u\in H^1(\Omega_\varepsilon;\mathbb{R}^d)$,  
     \[
     \|E_\varepsilon u\|_{L^2(\Omega)} \leq C_{\text{ext}} \|u\|_{L^2(\Omega_\varepsilon)},  
     \]
     \[
     \|\nabla E_\varepsilon u\|_{L^2(\Omega)} \leq C_{\text{ext}} \|\nabla u\|_{L^2(\Omega_\varepsilon)},  
     \]
   hence also  
     \[
     \|D(E_\varepsilon u)\|_{L^2(\Omega)} \leq C_{\text{ext}} \|D(u)\|_{L^2(\Omega_\varepsilon)}.  
     \]
   Moreover one arranges $E_\varepsilon$ so that it preserves the boundary traces on $\partial\Omega$, hence if $u\in V_\varepsilon$ then $U:=E_\varepsilon u$ lies in  
     \[
     V := \{ U\in H^1(\Omega;\mathbb{R}^d): \textstyle\int_{\partial\Omega} U\cdot r = 0 \ \forall \text{ rigid motions } r \}.  
     \]

\textcolor{Green}{\textbf{[Correct]} \small Here the agent correctly applied the extension theorem.}

\textcolor{Blue}{\textbf{[Vague]} \small For this problem, this weak solution function space may not be accurate.}

2. Korn's inequality on the fixed domain $\Omega$.  
   Since $\Omega$ is a fixed Lipschitz domain, the classical Korn inequality on the rigid-motion-free subspace $V$ gives a constant $C_{\text{Korn}}$ independent of $\varepsilon$ such that for every $U\in V$,  
     \[
     \|U\|_{H^1(\Omega)} \leq C_{\text{Korn}} \|D(U)\|_{L^2(\Omega)}.  
     \]

3. Conclusion on $\Omega_\varepsilon$.  
   Take any $u\in V_\varepsilon$ and set $U:=E_\varepsilon u\in V$. Then by Step 2,  
     \[
     \|U\|_{H^1(\Omega)} \leq C_{\text{Korn}} \|D(U)\|_{L^2(\Omega)}.  
     \]
   On the other hand, by construction of $E_\varepsilon$,  
     \[
     \|D(U)\|_{L^2(\Omega)} \leq C_{\text{ext}} \|D(u)\|_{L^2(\Omega_\varepsilon)}.  
     \]
   Finally, since $u$ is the restriction of $U$ to $\Omega_\varepsilon$, we trivially have  
     \[
     \|u\|_{H^1(\Omega_\varepsilon)} \leq \|U\|_{H^1(\Omega)}.  
     \]
   Chaining these three inequalities yields  
     \[
     \|u\|_{H^1(\Omega_\varepsilon)}
     \leq \|U\|_{H^1(\Omega)}
     \leq C_{\text{Korn}} \|D(U)\|_{L^2(\Omega)}
     \leq C_{\text{Korn}} C_{\text{ext}} \|D(u)\|_{L^2(\Omega_\varepsilon)}.  
     \]
   Hence the desired estimate holds with $C:=C_{\text{Korn}}\cdot C_{\text{ext}}$, which is independent of $\varepsilon$.
   
\end{proof}
\end{proof}

\textcolor{Green}{\textbf{[Correct]} \small Here, the agent correctly employed mathematical inequality techniques to complete the derivation. }
\textbf{comment}: None
\textbf{type}: lemma
\begin{lemma}

Let $D \subset \mathbb{R}^d$ be a bounded Lipschitz domain and let $(\tilde{\lambda}, \tilde{\mu})$ satisfy $\tilde{\mu} > 0$ and $\tilde{\lambda} + 2\tilde{\mu}/d > 0$. On $D$ consider  

\quad (1) the compressible-elastic system  
\quad \quad $\tilde{\mu} \, \Delta \mathbf{u}_e + (\tilde{\lambda} + \tilde{\mu}) \, \nabla(\mathrm{div} \, \mathbf{u}_e) = 0$,  

\quad (2) the incompressible Stokes system  
\quad \quad $\tilde{\mu} \, \Delta \mathbf{u}_{lim} + \nabla p_{lim} = 0$, \quad $\mathrm{div} \, \mathbf{u}_{lim} = 0$,  

both with the same Dirichlet boundary trace on $\partial D$. Define  
\[
\mathbf{w} := \mathbf{u}_e - \mathbf{u}_{lim}, \quad \varphi := \mathrm{div} \, \mathbf{u}_e.
\]  

Then the following exact energy identity holds:  
\[
\tilde{\mu} \int_D |\nabla \mathbf{w}|^2 \, dx + (\tilde{\lambda} + \tilde{\mu}) \int_D |\varphi|^2 \, dx = \int_D p_{lim} \, \varphi \, dx.
\]
\end{lemma}

\textbf{correctness}: True
\begin{proof}

\begin{proof}
Since $u_e$ and $u_{lim}$ share the same trace on $\partial D$, $w$ vanishes on $\partial D$. Subtracting the Stokes equation from the elasticity equation gives in $D$  
\quad $\tilde{\mu} \Delta w + (\tilde{\lambda}+\tilde{\mu}) \nabla(\mathrm{div}\, u_e) - \nabla p_{lim} = 0$.  
But $\mathrm{div}\, u_e=\varphi$, so  
\quad $(*)\quad \tilde{\mu} \Delta w + (\tilde{\lambda}+\tilde{\mu}) \nabla\varphi = \nabla p_{lim}$.  

We now take the $L^2(D;\mathbb{R}^d)$ inner-product of $(*)$ with $w$ and integrate by parts. All boundary terms vanish because $w|_{\partial D}=0$. Thus  

\quad $\int_D \tilde{\mu} \Delta w \cdot w \, dx + \int_D (\tilde{\lambda}+\tilde{\mu})\nabla\varphi \cdot w \, dx$  
\quad $\quad = \int_D \nabla p_{lim} \cdot w \, dx$.  

We handle each term in turn.

1. By integration by parts and $w|_{\partial D}=0$,  
\quad $\int_D \tilde{\mu} \Delta w \cdot w \, dx$  
\quad $\; = - \tilde{\mu} \int_D \nabla w : \nabla w \, dx$  
\quad $\; = - \tilde{\mu} \int_D |\nabla w|^2 \, dx$.

2. Again integrating by parts,  
\quad $\int_D (\tilde{\lambda}+\tilde{\mu})\nabla\varphi \cdot w \, dx$  
\quad $\; = -(\tilde{\lambda}+\tilde{\mu})\int_D \varphi \, \mathrm{div}\, w \, dx$  
\quad $\; = -(\tilde{\lambda}+\tilde{\mu})\int_D \varphi \cdot \varphi \, dx$  
\quad $\; = -(\tilde{\lambda}+\tilde{\mu}) \int_D |\varphi|^2 \, dx$,  
since $\mathrm{div}\, w = \mathrm{div}\, u_e - \mathrm{div}\, u_{lim} = \varphi - 0$.

3. Finally,  
\quad $\int_D \nabla p_{lim} \cdot w \, dx$  
\quad $\; = -\int_D p_{lim} \, \mathrm{div}\, w \, dx$  
\quad $\; = -\int_D p_{lim} \, \varphi \, dx$.

Putting these together, the left-hand side becomes  
\quad $-\tilde{\mu} \int_D |\nabla w|^2 - (\tilde{\lambda}+\tilde{\mu}) \int_D |\varphi|^2$  
and the right-hand side is  
\quad $-\int_D p_{lim} \varphi$.  
Multiplying by $-1$ yields exactly  

\quad $\tilde{\mu} \int_D |\nabla w|^2 \, dx + (\tilde{\lambda}+\tilde{\mu}) \int_D |\varphi|^2 \, dx = \int_D p_{lim} \varphi \, dx$,

which is the desired energy identity.
\end{proof}
\end{proof}

\textcolor{Green}{\textbf{[Correct]} \small  Here’s the accurate derivation process.}

\textbf{comment}: None
\textbf{type}: lemma
\begin{lemma}

Rate-of-Convergence Estimate inside the Inclusion  
Under the same hypotheses as above, define $w$ and $\varphi$ as in the previous conjecture. Then there is a constant $C$ (depending only on $d$, $\tilde{\mu}$ and the geometry of $D$) such that  
\[
\|\mathrm{div}\, u_e\|_{L^2(D)} \leq C \tilde{\lambda}^{-1} \|p_{lim}\|_{L^2(D)},  
\]  
\[
\|w\|_{H^1(D)} \leq C \tilde{\lambda}^{-1/2} \|p_{lim}\|_{L^2(D)}.  
\]  
In particular both $\mathrm{div}\, u_e \to 0$ and $u_e \to u_{lim}$ in $H^1(D)$ as $\tilde{\lambda} \to \infty$, at the above rates.  

\end{lemma}

\textbf{correctness}: True
\begin{proof}

\begin{proof}
Starting from the energy identity proved above,
\[
2\tilde{\mu} \int_D |D(w)|^2 + \tilde{\lambda} \int_D |\varphi|^2 = \int_D p_{\mathrm{lim}} \varphi.
\]
We estimate the right-hand side by Cauchy--Schwarz and then Young's inequality: for any $\varepsilon>0$,
\[
\left| \int_D p_{\mathrm{lim}} \varphi \right| 
\leq \|p_{\mathrm{lim}}\|_{L^2(D)} \|\varphi\|_{L^2(D)} 
\leq \frac{\varepsilon \|\varphi\|^2_{L^2(D)} + \varepsilon^{-1} \|p_{\mathrm{lim}}\|^2_{L^2(D)}}{2}.
\]
Choose $\varepsilon = \tilde{\lambda}$. Then
\[
\int_D p_{\mathrm{lim}} \varphi 
\leq \frac{\tilde{\lambda} \|\varphi\|^2 + \tilde{\lambda}^{-1}\|p_{\mathrm{lim}}\|^2}{2}.
\]
Substitute into the energy identity:
\[
2\tilde{\mu} \|D(w)\|^2_{L^2(D)} + \tilde{\lambda}\|\varphi\|^2_{L^2(D)} 
\leq \frac{\tilde{\lambda}\|\varphi\|^2_{L^2(D)} + \tilde{\lambda}^{-1}\|p_{\mathrm{lim}}\|^2_{L^2(D)}}{2}.
\]
Rearrange:
\[
2\tilde{\mu}\|D(w)\|^2 + \frac{\tilde{\lambda}}{2}\|\varphi\|^2 
\leq \frac{\tilde{\lambda}^{-1}}{2}\|p_{\mathrm{lim}}\|^2.
\]
In particular,
\[
\tilde{\lambda}\|\varphi\|^2 \leq \tilde{\lambda}^{-1}\|p_{\mathrm{lim}}\|^2 
\implies \|\varphi\|_{L^2(D)} \leq \tilde{\lambda}^{-1} \|p_{\mathrm{lim}}\|_{L^2(D)}.
\]
Also,
\[
\|D(w)\|^2 \leq \frac{1}{4\tilde{\mu}} \tilde{\lambda}^{-1}\|p_{\mathrm{lim}}\|^2 
\implies \|D(w)\|_{L^2(D)} \leq C \tilde{\lambda}^{-1/2} \|p_{\mathrm{lim}}\|_{L^2(D)}.
\]
Finally, by Korn's inequality on $D$ one has
\[
\|w\|_{H^1(D)} \leq C (\|D(w)\|_{L^2(D)} + \|w\|_{L^2(D)}).
\]
But $w$ vanishes on $\partial D$, so Poincar\'e's inequality gives $\|w\|_{L^2(D)} \leq C\|\nabla w\|$ hence also $\leq C\|D(w)\|$. Altogether,
\[
\|w\|_{H^1(D)} \leq C \|D(w)\|_{L^2(D)} \leq C \tilde{\lambda}^{-1/2} \|p_{\mathrm{lim}}\|_{L^2(D)}.
\]
This completes the proof of the convergence rates.
\end{proof}
\end{proof}

\textcolor{Green}{\textbf{[Correct]} \small Here is the proof of the Poincare inequality in the spatial setting of our problem,
which is basically accurate. And the derivation process presented herein is correct.}

\textbf{comment}: None
\textbf{type}: lemma
\begin{lemma}

(Neumann--Trace Continuity on a Lipschitz Domain)  
Let $D \subset \mathbb{R}^d$ be a bounded Lipschitz domain, $d \geq 2$, with outward unit normal $N$. For any $u \in H^1(D;\mathbb{R}^d)$, define the stress tensor $\sigma(u) = \lambda(\operatorname{div} u)I + 2\mu D(u)$, where $D(u) = (\nabla u + (\nabla u)^\top)/2$. Then the boundary traction $t := \sigma(u) N$ is a well-defined element of $H^{-\frac{1}{2}}(\partial D;\mathbb{R}^d)$, and there exists a constant $C$ (depending only on $d$ and the Lipschitz character of $D$) such that  
\[
\|t\|_{H^{-\frac{1}{2}}(\partial D)} \leq C \left[ \lambda \|\operatorname{div} u\|_{L^2(D)} + \mu \|\nabla u\|_{L^2(D)} \right].  
\]
\end{lemma}

\textbf{correctness}: True
\begin{proof}
We prove continuity of the Neumann trace via duality and integration by parts.  

1.  Preliminaries.  
  Let $V=D$ and set $V^*=H^{-1}(D)$.  The trace space $H^{1/2}(\partial D)$ is the restriction of $H^1(D)$ and its dual is $H^{-1/2}(\partial D)$.  We will show that the linear functional  
    \[ L:\,H^{1/2}(\partial D;\mathbb{R}^d) \to \mathbb{R}, \quad L(\varphi) := \int_{\partial D} (\sigma(u)\,N)\cdot\varphi\,dS \]  
  extends continuously with the stated bound.  

2.  Extension of boundary data.  
  Given $\varphi\in H^{1/2}(\partial D;\mathbb{R}^d)$, there exists an extension $\Phi\in H^1(D;\mathbb{R}^d)$ such that $\Phi|_{\partial D}=\varphi$ and  
    \[ \|\Phi\|_{H^1(D)} \leq C_1\,\|\varphi\|_{H^{1/2}(\partial D)}, \]  
  where $C_1$ depends only on the Lipschitz character of $D$ (standard trace-extension theorem).

\textcolor{Blue}{\textbf{[Vague]} \small Here, more detailed explanation is required for the application of boundary properties and theorems.}

3.  Integration by parts.  
  Since $u\in H^1(D;\mathbb{R}^d)$ and $\Phi\in H^1(D;\mathbb{R}^d)$, both vanish of course for test supports in $D$, we may integrate by parts in the weak sense:  
    \[ \int_{\partial D} (\sigma(u)\,N)\cdot\varphi\,dS \]  
    \[ = \int_{\partial D} \sigma(u):[\Phi\otimes N]\,dS \]  
    \[ = \int_{D} \nabla\cdot\sigma(u)\cdot\Phi\,dx + \int_{D} \sigma(u):\nabla\Phi\,dx, \]  
  where we used the identity $\mathrm{div}(\sigma(u))\cdot\Phi + \sigma(u):\nabla\Phi = \mathrm{div}(\sigma(u)\,\Phi)$.  In fact $u$ has only $H^1$ regularity, so both integrals are interpreted distributionally.  Since $\sigma(u)\in L^2(D)^{d\times d}$, the boundary term is well-defined.  

4.  Estimation of the volume integrals.  
  (i)  The first term $\int_D \nabla\cdot\sigma(u)\cdot\Phi\,dx$:  in elasticity it holds in the sense of distributions that $\nabla\cdot\sigma(u)\in H^{-1}(D)$ with  
      \[ \|\nabla\cdot\sigma(u)\|_{H^{-1}(D)} \leq C_2\,[\,\lambda\,\|\mathrm{div}\,u\|_{L^2} + \mu\,\|\nabla u\|_{L^2}\,], \]  
    by the definition of $\sigma(u)$.  Hence  
      \[ \left|\int_D \nabla\cdot\sigma(u)\cdot\Phi\,dx\right| \]  
      \[ \leq \|\nabla\cdot\sigma(u)\|_{H^{-1}(D)}\,\|\Phi\|_{H^1(D)} \]  
      \[ \leq C_2\,[\lambda\,\|\mathrm{div}\,u\| + \mu\,\|\nabla u\|]\cdot C_1\,\|\varphi\|_{H^{1/2}}. \]  

  (ii)  The second term $\int_D \sigma(u):\nabla\Phi\,dx$:  since $\sigma(u)\in L^2(D)$, $\nabla\Phi\in L^2(D)$,  
      \[ \left|\int_D \sigma(u):\nabla\Phi\,dx\right| \]  
      \[ \leq \|\sigma(u)\|_{L^2(D)}\,\|\nabla\Phi\|_{L^2(D)}. \]  
    But  
      \[ \|\sigma(u)\|_{L^2} \]  
      \[ \leq \lambda\,\|\mathrm{div}\,u\|_{L^2} + 2\mu\,\|D(u)\|_{L^2} \]  
      \[ \leq \lambda\,\|\mathrm{div}\,u\| + \mu\,\|\nabla u\|, \]  
    and $\|\nabla\Phi\|\leq\|\Phi\|_{H^1}\leq C_1\|\varphi\|_{H^{1/2}}$.  Hence  
      \[ \left|\int_D \sigma(u):\nabla\Phi\right| \]  
      \[ \leq (\lambda\|\mathrm{div}\,u\| + \mu\|\nabla u\|)\cdot C_1\|\varphi\|_{H^{1/2}}. \]  

5.  Conclusion.  
  Combining the two estimates,  
    \[ |L(\varphi)| \leq C'\,[\lambda\|\mathrm{div}\,u\| + \mu\|\nabla u\|]\cdot\|\varphi\|_{H^{1/2}(\partial D)}, \]  
  so by duality  
    \[ \|\sigma(u)N\|_{H^{-1/2}(\partial D)} = \sup_{\varphi\neq 0} \frac{|L(\varphi)|}{\|\varphi\|_{H^{1/2}}} \]  
    \[ \leq C\,[\lambda\|\mathrm{div}\,u\|_{L^2(D)} + \mu\|\nabla u\|_{L^2(D)}], \]  
  as claimed.  
\end{proof}

\textcolor{Green}{\textbf{[Correct]} \small This proof process uses the basic idea of energy estimation, and the derivation process
and the application of the formula are accurate.}

\textbf{comment}: None
\textbf{type}: lemma
\begin{lemma}

(Uniform $H^1$--Stability for the Coupled Stokes--Elastic Limit Problem, corrected)  
Let $\Omega \subset \mathbb{R}^d$ ($d \geq 2$) be a bounded connected Lipschitz domain, and let $D_\varepsilon \subset \Omega$ be the $\varepsilon$--periodic union of small inclusions as in Assumptions 1--2. Assume the prescribed boundary traction $g \in H^{-\frac{1}{2}}(\partial \Omega; \mathbb{R}^d)$ satisfies  
\[
\int_{\partial \Omega} g \cdot r \, dS = 0 \quad \text{for every rigid motion } r.
\]  
Let $\Omega_{\text{ext}} = \Omega \setminus \overline{D_\varepsilon}$ and $D_\varepsilon$ be as above. Define the trial space  
\[
W := \left\{ u \in H^1(\Omega; \mathbb{R}^d) : \int_{\partial \Omega} u \cdot r \, dS = 0 \text{ for all rigid motions } r \right\},
\]  
and the pressure space  
\[
Q := L^2_0(D_\varepsilon) = \left\{ q \in L^2(D_\varepsilon) : \int_{D_\varepsilon} q \, dx = 0 \right\}.
\]  
We seek a pair $(u, p) \in W \times Q$ satisfying the weak transmission-Stokes--elastic system:  
\[
\text{for all } (v, q) \in W \times Q,
\]  
\[
a_{\text{ext}}(u, v) + a_{\text{int}}(u, v) + b(v, p) = L(v),
\]  
\[
b(u, q) = 0,
\]  
where  
\[
a_{\text{ext}}(u, v) = \int_{\Omega_{\text{ext}}} \left[ 2\mu D(u) : D(v) + \lambda (\text{div } u)(\text{div } v) \right] dx,
\]  
\[
a_{\text{int}}(u, v) = \int_{D_\varepsilon} 2\tilde{\mu} D(u) : D(v) \, dx,
\]  
\[
b(v, p) = -\int_{D_\varepsilon} p (\text{div } v) \, dx,
\]  
\[
L(v) = \int_{\partial \Omega} g \cdot v \, dS.
\]  
Then there exists $C > 0$, independent of $\varepsilon$, such that the unique solution $(u, p)$ satisfies  
\[
\|u\|_{H^1(\Omega)} + \|p\|_{L^2(D_\varepsilon)} \leq C \|g\|_{H^{-\frac{1}{2}}(\partial \Omega)}.
\]  
\end{lemma}

\textbf{correctness}: True
\begin{proof}

We rewrite the coupled transmission problem as a single saddle-point system in $W \times Q$ and verify the Babuška–Brezzi conditions with constants independent of $\varepsilon$.

Step 1. Weak formulation on $W \times Q$.  
Since $u_{\text{ext}}$ and $u_{\text{int}}$ must agree on $\partial D_\varepsilon$, we assemble them into a single global field $u \in H^1(\Omega)$ which on $\Omega_{\text{ext}}$ carries the elasticity operator and on $D_\varepsilon$ carries the incompressible Stokes operator. The space $W$ enforces both the $H^1$-regularity across $\partial D_\varepsilon$ (by matching traces) and the rigid-motion orthogonality on $\partial \Omega$. The pressure $p$ lives only in $D_\varepsilon$ and has zero mean.

The variational equations, obtained by integration by parts in $\Omega_{\text{ext}}$ and in each inclusion, read:  
\begin{align*}
(a_{\text{ext}} + a_{\text{int}})(u, v) + b(v, p) &= L(v) \quad \text{for all } v \in W, \\
b(u, q) &= 0 \quad \text{for all } q \in Q,
\end{align*}
and incorporate the continuity of tractions and of displacements automatically.

\textcolor{Green}{\textbf{[Correct]} \small Here, the agent actively construct the specific proof design.}

Step 2. Continuity of the bilinear forms.  
By Cauchy–Schwarz and standard trace estimates, there is $M > 0$ independent of $\varepsilon$ such that  
\[
|a_{\text{ext}}(u, v)| + |a_{\text{int}}(u, v)| + |b(v, p)| \leq M \|u\|_{H^1(\Omega)} \|v\|_{H^1(\Omega)} + M \|p\|_{L^2(D_\varepsilon)} \|v\|_{H^1(\Omega)}.
\]

Step 3. Coercivity of $A := a_{\text{ext}} + a_{\text{int}}$ on $\text{Ker}\, B$.  
The kernel of $B$ is $K = \{ v \in W : \text{div}\, v = 0 \text{ in } D_\varepsilon \}$. For any $v \in K$ we have  
\[
A(v, v) = \int_{\Omega_{\text{ext}}} [2\mu |D(v)|^2 + \lambda |\text{div}\, v|^2] \, dx + \int_{D_\varepsilon} 2\tilde{\mu} |D(v)|^2 \, dx.
\]
Define $\mu_{\text{min}} = \min\{\mu, \tilde{\mu}\} > 0$. Since $v \in H^1(\Omega)$ and $\Omega$ is Lipschitz, the global Korn inequality on $\Omega$ gives  
\[
\|v\|^2_{H^1(\Omega)} \leq C_K \|D(v)\|^2_{L^2(\Omega)}
\]
with $C_K$ independent of $\varepsilon$. Hence  
\[
A(v, v) \geq 2\mu_{\text{min}} \|D(v)\|^2_{L^2(\Omega)} \geq \left( \frac{2\mu_{\text{min}}}{C_K} \right) \|v\|^2_{H^1(\Omega)}.
\]
Thus $A$ is coercive on $K$ with constant $\alpha = 2\mu_{\text{min}} / C_K$, independent of $\varepsilon$.

\textcolor{Green}{\textbf{[Correct]} \small The derivation process presented herein is correct.}

Step 4. Inf–sup stability of $B$ on $W \times Q$.  
We must show  
\[
\inf_{q \in Q} \sup_{v \in W} \frac{b(v, q)}{\|v\|_{H^1(\Omega)} \|q\|_{L^2(D_\varepsilon)}} \geq \gamma > 0
\]
with $\gamma$ independent of $\varepsilon$.  
But $b(v, q) = -\int_{D_\varepsilon} q \, \text{div}\, v$, and $\text{div}$ acts only in $D_\varepsilon$. Restricting to test-functions $v$ supported in $D_\varepsilon$ we recover the classical Stokes inf–sup condition on the perforated region $D_\varepsilon$. Since $D_\varepsilon$ is a union of $\varepsilon$-scaled, well-separated Lipschitz inclusions, the local inf–sup constant on each inclusion is scale-invariant and the global constant $\gamma$ (the minimum over finitely many identical blocks) is independent of $\varepsilon$.

Step 5. Conclusion by Babuška–Brezzi.  
By Steps 2–4 the pair of forms $(A, B)$ on $W \times Q$ satisfies the continuity, coercivity on $\text{Ker}\, B$, and the inf–sup condition with constants $M, \alpha, \gamma$ independent of $\varepsilon$. Hence there is a unique solution $(u, p) \in W \times Q$ and the a priori bound  
\[
\|u\|_{H^1(\Omega)} + \|p\|_{L^2(D_\varepsilon)} \leq C \sup_{(v, q) \neq 0} \frac{L(v) + 0}{\|v\|_{H^1(\Omega)} + \|q\|_{L^2(D_\varepsilon)}} = C \sup_{v \in W} \frac{\int_{\partial \Omega} g \cdot v \, dS}{\|v\|_{H^1(\Omega)}} = C \|g\|_{H^{-1/2}(\partial \Omega)},
\]
holds with $C$ depending only on $d, \lambda, \mu, \tilde{\mu}$, the Lipschitz character of $\Omega$, and the reference inclusion $\omega$, but \textbf{not} on $\varepsilon$. This completes the proof.
\end{proof}

\textcolor{Blue}{\textbf{[Vague]} \small Here, more detailed explanation is required for the deprivation and the application of the theorems.}

\textbf{comment}: None
\textbf{type}: lemma
\begin{lemma}

(Estimate of the Traction--Jump Remainder in the High--Contrast Limit)  
Let $u_\varepsilon$ solve the full compressible--elastic system in $D_\varepsilon$ with Lam\'e parameters $(\tilde{\lambda}, \tilde{\mu})$ satisfying $\tilde{\mu} > 0$, $\tilde{\lambda} \gg 1$, and let $(u_{\mathrm{lim}}, p_{\mathrm{lim}})$ be the coupled Stokes--elastic limit (Stokes in $D_\varepsilon$, elasticity in $\Omega_\varepsilon = \Omega \setminus D_\varepsilon$), normalized so that  
\[
\left(-p_{\mathrm{lim}}\, I + 2\tilde{\mu}\, D(u_{\mathrm{lim}})\right) N = \left(\lambda\, (\operatorname{div} u_{\mathrm{lim}})\, I + 2\mu\, D(u_{\mathrm{lim}})\right) N \quad \text{on } \partial D_\varepsilon.  
\]
Define the traction--jump  
\[
J_\varepsilon := \left(\sigma_{(\tilde{\lambda}, \tilde{\mu})}(u_\varepsilon)\, N\right)_- - \left(\sigma_{(\lambda, \mu)}(u_{\mathrm{lim}})\, N\right)_+  
\]
and split off the leading pressure term  
\[
J_\varepsilon = p_{\mathrm{lim}}\, N + R_\varepsilon.  
\]
Then there exists $C$ independent of $\varepsilon$ and $\tilde{\lambda}$ such that  
\[
\|R_\varepsilon\|_{H^{-1/2}(\partial D_\varepsilon)} \leq C\, \|p_{\mathrm{lim}}\|_{L^2(D_\varepsilon)}.  
\]
In particular, as $\tilde{\lambda} \to \infty$,  
\[
J_\varepsilon = p_{\mathrm{lim}}\, N + O_{H^{-1/2}}(1)\, \|p_{\mathrm{lim}}\|_{L^2}.  
\]
\end{lemma}

\textbf{correctness}: True
\begin{proof}

\begin{proof}
1. Set $w := u_\varepsilon - u_{\mathrm{lim}}$ on $D_\varepsilon$. On $\partial D_\varepsilon$ the interior compressible-elastic stress is
     \[
     \sigma_{(\tilde{\lambda},\tilde{\mu})}(u_\varepsilon)\,N = \left[ \tilde{\lambda}\,(\mathrm{div}\,u_\varepsilon)\,I + 2\tilde{\mu}\,D(u_\varepsilon) \right] N,
     \]
   while by definition of the limit solution
     \[
     \sigma_{(\lambda,\mu)}(u_{\mathrm{lim}})\,N = \left( -p_{\mathrm{lim}}\,I + 2\tilde{\mu}\,D(u_{\mathrm{lim}}) \right) N.
     \]
   Hence
     \[
     J_\varepsilon = \sigma_{(\tilde{\lambda},\tilde{\mu})}(u_\varepsilon)N - \sigma_{(\lambda,\mu)}(u_{\mathrm{lim}})N
         = \sigma_{(\tilde{\lambda},\tilde{\mu})}(w)\,N + p_{\mathrm{lim}}\,N
         =: R_\varepsilon + p_{\mathrm{lim}}\,N.
     \]

2. By the Neumann-trace continuity on the Lipschitz domain $D_\varepsilon$ (Lemma~8),
     \[
     \left\| \sigma_{(\tilde{\lambda},\tilde{\mu})}(w)\,N \right\|_{H^{-1/2}(\partial D_\varepsilon)}
     \leq C_1 \left[ \tilde{\lambda}\, \| \mathrm{div}\,w \|_{L^2(D_\varepsilon)} + \tilde{\mu}\, \| \nabla w \|_{L^2(D_\varepsilon)} \right].
     \]

3. Inside $D_\varepsilon$ the difference $w=u_\varepsilon-u_{\mathrm{lim}}$ satisfies the rate-of-convergence estimates (Lemma~7):
     \[
     \| \mathrm{div}\,w \|_{L^2(D_\varepsilon)} = \| \mathrm{div}\,u_\varepsilon \|_{L^2(D_\varepsilon)} \leq C_2\, \tilde{\lambda}^{-1}\, \| p_{\mathrm{lim}} \|_{L^2(D_\varepsilon)},
     \]
     \[
     \| \nabla w \|_{L^2(D_\varepsilon)} \leq C_3\, \tilde{\lambda}^{-1/2}\, \| p_{\mathrm{lim}} \|_{L^2(D_\varepsilon)}.
     \]

4. Plugging these into the trace-estimate,
     \[
     \| R_\varepsilon \|_{H^{-1/2}(\partial D_\varepsilon)}
     \leq C_1 \left[ \tilde{\lambda} \cdot \left( C_2\, \tilde{\lambda}^{-1} \| p_{\mathrm{lim}} \| \right) + \tilde{\mu} \cdot \left( C_3\, \tilde{\lambda}^{-1/2} \| p_{\mathrm{lim}} \| \right) \right]
     = C_1 \left[ C_2 + \tilde{\mu}\, C_3\, \tilde{\lambda}^{-1/2} \right] \| p_{\mathrm{lim}} \|_{L^2(D_\varepsilon)}
     \leq C \| p_{\mathrm{lim}} \|_{L^2(D_\varepsilon)},
     \]
   where in the last line we use $\tilde{\lambda} \geq 1$ to absorb $\tilde{\mu}\, C_3\, \tilde{\lambda}^{-1/2}$ into a constant $C$ independent of $\tilde{\lambda}, \varepsilon$.

5. This proves
     \[
     \| R_\varepsilon \|_{H^{-1/2}(\partial D_\varepsilon)} \leq C \| p_{\mathrm{lim}} \|_{L^2(D_\varepsilon)},
     \]
   and hence
     \[
     J_\varepsilon = p_{\mathrm{lim}}\,N + O_{H^{-1/2}}(1)\, \| p_{\mathrm{lim}} \|_{L^2(D_\varepsilon)},
     \]
   completing the proof.
\end{proof}
\end{proof}

\textcolor{Green}{\textbf{[Correct]} \small The agent correctly employes mathematical inequality techniques to complete the derivation and gets a reasonable conclusion.}

\textbf{comment}: None
\textbf{type}: lemma
\begin{lemma}

(Estimate of the Traction--Jump in the High-Contrast Limit)  
Let $\mathbf{u}_\epsilon$ be the solution of the full Lam\'e problem in $D_\epsilon$ with parameters $(\widetilde{\lambda},\widetilde{\mu})$ and let $(\mathbf{u}_{\lim},p_{\lim})$ be the incompressible-limit (Stokes--elastic) pair, normalized so that 
\[
(-p_{\lim}I+2\widetilde{\mu}\,D(\mathbf{u}_{\lim}))N
\;=\;
(\lambda\,\mathrm{div}\,\mathbf{u}_{\lim}\,I+2\mu\,D(\mathbf{u}_{\lim}))N
\quad\text{on }\partial D_\epsilon.
\]
Define the traction-jump
\[
J_\epsilon
:=
\bigl(\sigma_{(\widetilde{\lambda},\widetilde{\mu})}(\mathbf{u}_\epsilon)\,N\bigr)_{-}
\;-\;
\bigl(\sigma_{(\lambda,\mu)}(\mathbf{u}_{\lim})\,N\bigr)_{+}
\,=\,
p_{\lim}\,N+R_\epsilon.
\]
Then there is $C>0$, independent of $\epsilon$ and $\widetilde{\lambda}$, such that
\[
\|R_\epsilon\|_{H^{-1/2}(\partial D_\epsilon)}
\;\le\;
C\,\|p_{\lim}\|_{L^2(D_\epsilon)},
\]
and in particular
\[
\|J_\epsilon\|_{H^{-1/2}(\partial D_\epsilon)}
\;\le\;
\|p_{\lim}N\|_{H^{-1/2}(\partial D_\epsilon)}+O(1)\|p_{\lim}\|_{L^2}
\;\le\;C\,\|p_{\lim}\|_{L^2}.
\]
\end{lemma}

\textbf{correctness}: True
\begin{proof}

\begin{proof}
1. By Lemma 6 (exact energy identity) and Lemma 7 (rate inside the inclusion) one shows
   \(\|\mathrm{div}\,\mathbf{u}_\epsilon\|_{L^2(D_\epsilon)} = O(\widetilde{\lambda}^{-1})\) and
   \(\|\mathbf{u}_\epsilon - \mathbf{u}_{\lim}\|_{H^1(D_\epsilon)} = O(\widetilde{\lambda}^{-1/2})\).
2. On each connected component of \(\partial D_\epsilon\), apply the Neumann-trace continuity (Lemma 8):
   \[
     \bigl\|\sigma_{(\widetilde{\lambda},\widetilde{\mu})}(\mathbf{u}_\epsilon)\,N
           - \bigl[-p_{\lim}I + 2\widetilde{\mu} D(\mathbf{u}_{\lim})\bigr]N\bigr\|_{H^{-1/2}}
     \;\le\;
     C\bigl[\widetilde{\lambda}\,\|\mathrm{div}\,\mathbf{u}_\epsilon\|_{L^2}
             + \widetilde{\mu}\,\|\nabla(\mathbf{u}_\epsilon - \mathbf{u}_{\lim})\|_{L^2}\bigr].
   \]
3. Inserting the \(O(\widetilde{\lambda}^{-1})\) and \(O(\widetilde{\lambda}^{-1/2})\) bounds yields
   \(\|R_\epsilon\|_{H^{-1/2}(\partial D_\epsilon)} \le C\,\|p_{\lim}\|_{L^2}\).
\end{proof}
\end{proof}

\textcolor{Blue}{\textbf{[Vague]} \small The process here is lack of details of theorem derivation.}

\textbf{comment}: None
\textbf{type}: lemma
\begin{lemma}

(Energy Estimate Relating the Solution-Difference to the Traction-Jump)  
Let \(w := \mathbf{u}_\epsilon - \mathbf{u}_{\mathrm{lim}}\) on \(\Omega_\epsilon\). Then \(w\) satisfies the homogeneous Lamé system in \(\Omega_\epsilon\) with zero Neumann data on \(\partial\Omega\) and traction-jump \(J_\epsilon\) on \(\partial D_\epsilon\). There exists \(C\) (independent of \(\epsilon\), \(\widetilde{\lambda}\)) so that  
\[
\|w\|_{H^1(\Omega_\epsilon)} \leq C\,\|J_\epsilon\|_{H^{-1/2}(\partial D_\epsilon)}.
\]
\end{lemma}

\textbf{correctness}: True
\begin{proof}

\begin{proof}
1. By construction \(w\) solves
   \[
     \mathcal{L}_{\lambda,\mu}w=0\text{ in }\Omega_\epsilon,\quad
     \bigl.\partial_\nu w\bigr|_{\partial\Omega}=0,\quad
     \bigl.\partial_\nu w\bigr|_{\partial D_\epsilon}
       =J_\epsilon.
   \]
2. The variational formulation plus the uniform ellipticity (Lemma 4) and Korn’s inequality (Lemma 5) give
   \[
     a(w,w)
     =\int_{\partial D_\epsilon}J_\epsilon\cdot w
     \;\le\;\|J_\epsilon\|_{H^{-1/2}}\|w\|_{H^{1/2}(\partial D_\epsilon)}
     \;\le\;C\,\|J_\epsilon\|_{H^{-1/2}}\|w\|_{H^1(\Omega_\epsilon)}.
   \]
3. Coercivity \(a(w,w)\gtrsim\|w\|_{H^1}^2\) then yields the stated bound.
\end{proof}
\end{proof}

\textbf{comment}: None
\textbf{type}: lemma
\begin{lemma}

Let $D\subset \mathbb{R}^d$ ($d\geq 2$) be a bounded Lipschitz domain with outward unit normal $\mathbf{N}$, and let the Lam\'{e} parameters satisfy $\tilde{\mu}>0$ and $\tilde{\lambda}+2\tilde{\mu}/d>0$. For any traction $\mathbf{t}\in H^{-1/2}(\partial D;\mathbb{R}^d)$ whose $H^{-1/2}$--action annihilates all rigid motions, there is a unique (modulo rigid motions) displacement $\mathbf{u}\in H^1(D;\mathbb{R}^d)$ solving the Neumann boundary-value problem  
\[
\tilde{\mu}\,\Delta \mathbf{u} + (\tilde{\lambda}+\tilde{\mu})\,\nabla\,(\mathrm{div}\,\mathbf{u}) = 0 \quad \text{in } D,  
\]
\[
\boldsymbol{\sigma}(\mathbf{u})\,\mathbf{N} = \mathbf{t} \quad \text{on } \partial D,  
\]
where $\boldsymbol{\sigma}(\mathbf{u})=\tilde{\lambda}\,(\mathrm{div}\,\mathbf{u})\,\mathbf{I} + 2\tilde{\mu}\,\mathbf{D}(\mathbf{u})$ and $\mathbf{D}(\mathbf{u})=\big(\nabla\mathbf{u}+(\nabla\mathbf{u})^\top\big)/2$. Moreover there are constants $C_1,C_2>0$, depending only on $D$ and $d$ (but not on $\tilde{\lambda},\tilde{\mu},\mathbf{t}$), such that  
\begin{align*}
(i) & \quad \|\mathbf{u}\|_{H^1(D)} \leq C_1\,\tilde{\mu}^{-1}\,\|\mathbf{t}\|_{H^{-1/2}(\partial D)}, \\  
(ii) & \quad \|\mathrm{div}\,\mathbf{u}\|_{L^2(D)} \leq C_2\,(\tilde{\lambda}+2\tilde{\mu})^{-1/2}\,\|\mathbf{t}\|_{H^{-1/2}(\partial D)}.
\end{align*}
\end{lemma}

\textbf{correctness}: True
\begin{proof}

\begin{proof}
We divide the argument into two steps.

\textbf{Step 1.} (Existence, uniqueness, and the $H^1$-estimate.)  
Set  
\[
V = \left\{ v \in H^1(D;\mathbb{R}^d) : \int_{\partial D} v \cdot r \, dS = 0 \text{ for every rigid motion } r \right\}.
\]  
Define the bilinear form $a: V \times V \to \mathbb{R}$ and the linear functional $L: V \to \mathbb{R}$ by  
\[
a(u, v) = \int_D \left[ 2\tilde{\mu} D(u) : D(v) + \tilde{\lambda} (\operatorname{div} u)(\operatorname{div} v) \right] dx,
\]  
\[
L(v) = \langle t, v \rangle_{H^{-1/2}, H^{1/2}}.
\]  
By Korn's inequality and a Poincaré-type argument on $V$, there is $\alpha > 0$ such that  
\[
a(v, v) \geq 2\tilde{\mu} \| D(v) \|_{L^2(D)}^2 \geq \alpha \| v \|_{H^1(D)}^2.
\]  
Moreover, the trace theorem gives  
\[
|L(v)| \leq \| t \|_{H^{-1/2}(\partial D)} \| v \|_{H^{1/2}(\partial D)} \leq C_{\text{tr}} \| t \|_{H^{-1/2}(\partial D)} \| v \|_{H^1(D)}.
\]  
By Lax--Milgram, there is a unique $u \in V$ with $a(u, v) = L(v)$ for all $v \in V$, and  
\[
\| u \|_{H^1(D)} \leq \alpha^{-1} \| L \|_{V'} \leq (\alpha^{-1} C_{\text{tr}}) \tilde{\mu}^{-1} \| t \|_{H^{-1/2}(\partial D)}.
\]  
This establishes existence, uniqueness (modulo rigid motions), and the $H^1$-bound with $C_1 = \alpha^{-1} C_{\text{tr}}$.

\textbf{Step 2.} (Sharp bound for $\operatorname{div} u$ in $\tilde{\lambda}$.)  
Taking $v = u$ in the weak form $a(u, v) = L(v)$ gives the energy identity  
\[
2\tilde{\mu} \| D(u) \|_{L^2(D)}^2 + \tilde{\lambda} \| \operatorname{div} u \|_{L^2(D)}^2 = \langle t, u \rangle.
\]  
Hence in particular  
\[
\tilde{\lambda} \| \operatorname{div} u \|_{L^2(D)}^2 \leq | \langle t, u \rangle | \leq \| t \|_{H^{-1/2}} \| u \|_{H^{1/2}(\partial D)} \leq C_{\text{tr}} \| t \|_{H^{-1/2}} \| u \|_{H^1(D)}.
\]  
From the $H^1$-estimate of Step 1, we have $\| u \|_{H^1(D)} \leq C_1 \tilde{\mu}^{-1} \| t \|_{H^{-1/2}}$. Therefore,  
\[
\tilde{\lambda} \| \operatorname{div} u \|_{L^2(D)}^2 \leq C_{\text{tr}} C_1 \tilde{\mu}^{-1} \| t \|_{H^{-1/2}}^2.
\]  
Since $\tilde{\lambda} + 2\tilde{\mu} \geq \tilde{\lambda}$, we deduce  
\[
\| \operatorname{div} u \|_{L^2(D)} \leq [C_{\text{tr}} C_1]^{1/2} \tilde{\mu}^{-1/2} (\tilde{\lambda} + 2\tilde{\mu})^{-1/2} \| t \|_{H^{-1/2}(\partial D)}.
\]  
Setting $C_2 = [C_{\text{tr}} C_1]^{1/2}$ yields the claimed bound.  
\end{proof}
\end{proof}

\textcolor{Green}{\textbf{[Correct]} \small The proof logic here is clear, and the result is correct.}

\textcolor{Blue}{\textbf{[Vague]} \small The derivation process here is not
detailed enough.}

\textbf{comment}: None
\textbf{type}: lemma
\begin{lemma}

(Neumann-trace regularity for the Poisson problem: $H^{-1}(D)\to H^{-1/2}(\partial D)$)  
Let $D\subset\mathbb R^d$ be a bounded Lipschitz domain.  For every $f\in H^{-1}(D)$ let $u\in H^1_0(D)$ be the unique weak solution of  
\[
-\Delta u \;=\; f
\quad\text{in }D,\qquad
u=0\quad\text{on }\partial D.
\]
Then the normal derivative (Neumann trace) $\partial_n u$ belongs to $H^{-1/2}(\partial D)$ and there is a constant $C$, depending only on $d$ and the Lipschitz character of $D$, such that
\[
\|\partial_n u\|_{H^{-1/2}(\partial D)}
\;\le\;
C\,\|f\|_{H^{-1}(D)}.
\]

\end{lemma}

\textbf{correctness}: True
\begin{proof}

1. Existence and energy estimate.  
Define the bilinear form
\[
a(v,w) := \int_D \nabla v \cdot \nabla w \, dx,
\]
which is continuous and coercive on \(H^1_0(D)\). By the Lax--Milgram theorem there is a unique
\[
u \in H^1_0(D)
\quad\text{with}\quad
a(u,w) = \langle f,w\rangle_{H^{-1},H^1_0}
\quad\forall\,w \in H^1_0(D),
\]
and moreover
\[
\|u\|_{H^1(D)} \le C_1 \|f\|_{H^{-1}(D)}.
\]
Here and below \(C_1\) depends only on \(d\) and the Lipschitz character of \(D\).

2. Construction of the Neumann trace as a bounded functional on \(H^{1/2}(\partial D)\).  
Let 
\[
T : H^1(D) \longrightarrow H^{1/2}(\partial D)
\]
be any bounded right-inverse of the trace map (such an extension operator exists on Lipschitz domains). Given an arbitrary boundary datum \(g \in H^{1/2}(\partial D)\), pick an extension \(G \in H^1(D)\) with
\[
G|_{\partial D} = g
\quad\text{and}\quad
\|G\|_{H^1(D)} \le C_2 \|g\|_{H^{1/2}(\partial D)}.
\]
We now define a linear functional \(\tau\) on \(H^{1/2}(\partial D)\) by
\[
\langle \tau, g\rangle_{H^{-1/2},H^{1/2}} :=
\int_D \nabla u \cdot \nabla G \, dx 
-
\langle f, G\rangle_{H^{-1},H^1_0}.
\]
Because \(u \in H^1_0(D)\) and \(f \in H^{-1}(D)\), the right-hand side depends only on the trace \(g = G|_{\partial D}\), not on the particular extension. Indeed, if \(G_1\) and \(G_2\) are two extensions with the same trace then \(G_1 - G_2 \in H^1_0(D)\), and
\[
\int_D \nabla u \cdot \nabla (G_1 - G_2) \, dx
= \langle f, G_1 - G_2\rangle_{H^{-1},H^1_0},
\]
so the difference vanishes. Hence \(\tau\) is well-defined.

3. Boundedness of the Neumann trace.  
Using Cauchy--Schwarz and the two energy bounds above,
\[
|\langle \tau, g\rangle| \le
\|\nabla u\|_{L^2(D)} \|\nabla G\|_{L^2(D)}
+ \|f\|_{H^{-1}} \|G\|_{H^1(D)}
\le
(C_1 C_2 + C_2) \|f\|_{H^{-1}(D)} \|g\|_{H^{1/2}(\partial D)}.
\]
Therefore \(\tau \in H^{-1/2}(\partial D)\) and
\[
\|\tau\|_{H^{-1/2}(\partial D)}
\le
C \|f\|_{H^{-1}(D)},
\]
with \(C\) depending only on the domain constants \(C_1, C_2\).

4. Identification of \(\tau\) with the classical normal derivative.  
By standard elliptic-regularity arguments in Lipschitz domains the functional \(\tau\) just constructed coincides with the distributional normal derivative \(\partial_n u\) of the weak solution \(u\). In particular one recovers the Green identity
\[
\int_D \nabla u \cdot \nabla G \, dx 
-
\langle f, G\rangle
=
\langle \partial_n u, G|_{\partial D}\rangle_{H^{-1/2},H^{1/2}}.
\]
Hence the bound on \(\|\tau\|\) is precisely the desired estimate for \(\|\partial_n u\|\).

This completes the proof that the map
\[
f \in H^{-1}(D) \longmapsto \partial_n u \in H^{-1/2}(\partial D)
\]
is well-defined and continuous.  
\(\qquad\Box\)  
\end{proof}

\textcolor{Green}{\textbf{[Correct]} \small This constitutes a verification of a previously established conclusion.}

\textbf{comment}: None
\textbf{type}: lemma
\begin{lemma}

Let \( D \subset \mathbb{R}^d \) (\( d \geq 2 \)) be a bounded Lipschitz domain with outward unit normal \( N \). Then for every scalar \( p \in L^2(D) \) the boundary-distribution \( p \, N \) defines a continuous linear functional on \( H^{1/2}(\partial D;\mathbb{R}^d) \), i.e. \( p \, N \in H^{-1/2}(\partial D;\mathbb{R}^d) \), and there is a constant \( C \) (depending only on \( d \) and the Lipschitz character of \( D \)) such that
\[
   \| p \, N \|_{H^{-1/2}(\partial D)} \leq C \, \| p \|_{L^2(D)}.
\]
\end{lemma}

\textbf{correctness}: True
\begin{proof}

The key is the well-known continuity of the ``normal-trace'' map for $L^2$--tensor fields whose divergence lies in $H^{-1}$. We spell out the argument for the special choice of stress-field $\sigma := p\, I$, from which $p\, N$ will emerge as the boundary traction.

1. Definition of the traction functional.  
Fix any $\varphi \in H^{\frac{1}{2}}(\partial D; \mathbb{R}^d)$ and choose an $H^1$-extension $v := E\varphi \in H^1(D; \mathbb{R}^d)$ with  
\[
v|_{\partial D} = \varphi, \quad \|v\|_{H^1(D)} \leq C_{\text{ext}} \|\varphi\|_{H^{\frac{1}{2}}(\partial D)}.
\]  
Define a linear functional $T_p$ on $H^{\frac{1}{2}}(\partial D; \mathbb{R}^d)$ by the generalized Green formula
\[
\bigl\langle T_p, \varphi \bigr\rangle
:= \int_D \sigma : \nabla v \, dx + \langle \text{div}\, \sigma, v \rangle_{H^{-1}, H^1_0} \quad \text{with } \sigma := p\, I.
\]
Here $I$ denotes the $d \times d$ identity, so $\sigma \in L^2(D; \mathbb{R}^{d \times d})$ and in the sense of distributions  
\[
\text{div}\, \sigma = \nabla p \in H^{-1}(D; \mathbb{R}^d).
\]

2. Identification with the boundary integral.  
By the divergence theorem for Lipschitz domains (valid in the $H(\text{div})$--$H^1$ duality), one checks for smooth $p$ and $\varphi$ that
\[
\int_D \sigma : \nabla v + \int_D (\text{div}\, \sigma) \cdot v
= \int_{\partial D} (\sigma\, N) \cdot v
= \int_{\partial D} (p I\, N) \cdot \varphi
= \int_{\partial D} p\, N \cdot \varphi \, dS.
\]
By density this shows that $T_p$ is nothing but the boundary-pairing $\varphi \mapsto \int_{\partial D} p\, N \cdot \varphi \, dS$, i.e. $T_p$ realizes the distribution ``$p\, N$'' on $\partial D$.

3. Continuity estimate.  
From the definition of $T_p$ and Cauchy--Schwarz one obtains
\[
\bigl| \langle T_p, \varphi \rangle \bigr|
\leq
\|\sigma\|_{L^2(D)} \|\nabla v\|_{L^2(D)}
+ \|\text{div}\, \sigma\|_{H^{-1}(D)} \|v\|_{H^1(D)}
\leq
\bigl( \|p\|_{L^2(D)} + \|\nabla p\|_{H^{-1}(D)} \bigr) C_{\text{ext}} \|\varphi\|_{H^{\frac{1}{2}}(\partial D)}.
\]
Since $\nabla p \in H^{-1}(D)$ with norm $\leq \|p\|_{L^2(D)}$, we conclude
\[
\| p\, N \|_{H^{-\frac{1}{2}}(\partial D)}
= \|T_p\|_{(H^{\frac{1}{2}}(\partial D))'} 
\leq C \|p\|_{L^2(D)},
\]
for some $C$ depending only on $D$ and $d$. This completes the proof that $p\, N \in H^{-\frac{1}{2}}(\partial D; \mathbb{R}^d)$ with the desired norm bound.
\end{proof}

\textbf{comment}: None
\textbf{type}: lemma
\begin{lemma}

(Estimate of the residual traction--jump in the high-contrast limit, corrected)  

Let $D \subset \mathbb{R}^d$ be a bounded Lipschitz domain and let $\tilde{\mu} > 0$ be fixed. For each $\tilde{\lambda} > 0$ consider  
(1) the compressible-elastic displacement $u_\varepsilon$ in $D$ solving  
\[
\tilde{\mu} \, \Delta u_\varepsilon + (\tilde{\lambda} + \tilde{\mu}) \, \nabla(\mathrm{div}\, u_\varepsilon) = 0 \quad \text{in } D,  
\]  
(2) the incompressible-limit (Stokes) pair $(u_{\mathrm{lim}}, p_{\mathrm{lim}})$ in $D$ solving  
\[
\tilde{\mu} \, \Delta u_{\mathrm{lim}} + \nabla p_{\mathrm{lim}} = 0, \quad \mathrm{div}\, u_{\mathrm{lim}} = 0 \quad \text{in } D,  
\]  
both with the same Dirichlet boundary data on $\partial D$. Define the residual traction--jump  
\[
R_\varepsilon := \sigma_{(\tilde{\lambda}, \tilde{\mu})}(u_\varepsilon) \, N - \left( -p_{\mathrm{lim}} \, I + 2 \, \tilde{\mu} \, D(u_{\mathrm{lim}}) \right) \, N  
\]  
on $\partial D$. Then there is a constant $C = C(D, d, \tilde{\mu})$ independent of $\tilde{\lambda}$, such that for all $\tilde{\lambda} \gg 1$  
\[
\| R_\varepsilon \|_{H^{-1/2}(\partial D)} \leq C \, \| p_{\mathrm{lim}} \|_{L^2(D)}.  
\]  
In particular $R_\varepsilon$ remains uniformly bounded (no decay in $\tilde{\lambda}$) as $\tilde{\lambda} \to \infty$.  
\end{lemma}

\textbf{correctness}: True
\begin{proof}

\begin{proof}
We set
\[
w := u_\varepsilon - u_{\text{lim}}, \quad \varphi := \operatorname{div} u_\varepsilon.
\]
By the exact interior energy identity (Lemma\,6) one has
\[
\tilde{\mu}\int_D|\nabla w|^2 + (\tilde{\lambda}+\tilde{\mu})\int_D|\varphi|^2 = \int_D p_{\text{lim}} \varphi.
\]
Hence by Cauchy--Schwarz,
\[
(\tilde{\lambda}+\tilde{\mu})\|\varphi\|_{L^2(D)}^2 \leq \|p_{\text{lim}}\|_{L^2(D)}\|\varphi\|_{L^2(D)}
\]
\[
\Rightarrow \|\varphi\|_{L^2(D)} \leq (\tilde{\lambda}+\tilde{\mu})^{-1}\|p_{\text{lim}}\|_{L^2(D)},
\]
and then from the same identity,
\[
\tilde{\mu}\|\nabla w\|_{L^2(D)}^2 \leq \int_D|p_{\text{lim}} \varphi| \leq \|p_{\text{lim}}\|\|\varphi\|
\]
\[
\leq (\tilde{\lambda}+\tilde{\mu})^{-1}\|p_{\text{lim}}\|^2
\]
\[
\Rightarrow \|\nabla w\|_{L^2(D)} \leq C \tilde{\lambda}^{-1/2} \|p_{\text{lim}}\|_{L^2(D)},
\]
with $C$ depending only on $\tilde{\mu}$.

Next we expand the traction-jump. On $\partial D$,
\[
\sigma_{(\tilde{\lambda},\tilde{\mu})}(u_\varepsilon) = \tilde{\lambda} \varphi I + 2\tilde{\mu} D(u_\varepsilon),
\]
and
\[
(-p_{\text{lim}} I + 2\tilde{\mu} D(u_{\text{lim}})) = -p_{\text{lim}} I + 2\tilde{\mu} D(u_{\text{lim}}).
\]
Hence
\[
R_\varepsilon = \sigma(u_\varepsilon)N - (-p_{\text{lim}} I + 2\tilde{\mu} D(u_{\text{lim}}))N = (\tilde{\lambda} \varphi + p_{\text{lim}}) N + 2\tilde{\mu} D(w) N.
\]

By the Neumann-trace continuity on a Lipschitz domain (Lemma\,8),
\[
\|t\|_{H^{-1/2}(\partial D)} \leq C_1 \left[ \tilde{\lambda}\|\operatorname{div} v\|_{L^2(D)} + \tilde{\mu}\|\nabla v\|_{L^2(D)} \right]
\]
for any $v$. Applying this with $v = u_\varepsilon - u_{\text{lim}}$ for the two pieces in $R_\varepsilon$, we get
\[
\|R_\varepsilon\|_{H^{-1/2}(\partial D)} \leq C_1 \left[ \tilde{\lambda}\|\varphi\|_{L^2(D)} + \|p_{\text{lim}}\|_{L^2(D)} + \tilde{\mu}\|\nabla w\|_{L^2(D)} \right].
\]

Substitute the interior bounds:
\[
\tilde{\lambda}\|\varphi\| \leq \tilde{\lambda} \cdot (\tilde{\lambda}+\tilde{\mu})^{-1}\|p_{\text{lim}}\| \leq \|p_{\text{lim}}\|,
\]
\[
\tilde{\mu}\|\nabla w\| \leq C \tilde{\mu} \tilde{\lambda}^{-1/2} \|p_{\text{lim}}\| \leq C \|p_{\text{lim}}\| \quad (\text{since } \tilde{\lambda} \gg 1),
\]
and the middle term is $\|p_{\text{lim}}\|$. Hence
\[
\|R_\varepsilon\|_{H^{-1/2}(\partial D)} \leq C_2 \left[ \|p_{\text{lim}}\| + \|p_{\text{lim}}\| + C\|p_{\text{lim}}\| \right] = C \|p_{\text{lim}}\|_{L^2(D)},
\]
with $C$ independent of $\tilde{\lambda}$. This completes the proof.
\end{proof}
\end{proof}

\textcolor{Green}{\textbf{[Correct]} \small Building upon the aforementioned conclusions, the derivation process is rigorously extended, culminating in the corresponding result.}

\textbf{comment}: None
\textbf{type}: lemma
\begin{lemma}

(Uniform Lipschitz--Dependence of the Neumann--to--Dirichlet Map on the First Lam\'e Parameter)  
Let \(D\subset\mathbb{R}^d\) be a bounded Lipschitz domain and fix \(\mu>0\) and \(\lambda_0>0\). For each \(\lambda\geq\lambda_0\) let \(u_\lambda\in H^1(D;\mathbb{R}^d)\) (unique modulo rigid motions) solve
\[
\begin{cases}
\mu\,\Delta u_\lambda+(\lambda+\mu)\,\nabla(\mathrm{div}\,u_\lambda)=0
&\text{in }D,\\
\sigma_{\lambda,\mu}(u_\lambda)\,N = t
&\text{on }\partial D,
\end{cases}
\]
with \(\sigma_{\lambda,\mu}(u)=\lambda(\mathrm{div}\,u)\,I+2\mu\,D(u)\) and \(t\in H^{-1/2}(\partial D;\mathbb{R}^d)\) annihilating all rigid motions. Define the Neumann-to-Dirichlet map
\[
\Lambda(\lambda)\;:\;t\;\mapsto\;u_\lambda\big|_{\partial D}\,\in H^{1/2}(\partial D;\mathbb{R}^d).
\]
Then there is a constant \(C\), depending only on \(D\), \(d\), \(\mu\) and \(\lambda_0\), such that for all \(\lambda_1,\lambda_2\geq\lambda_0\) and all admissible \(t\),
\[
\bigl\|\Lambda(\lambda_1)t-\Lambda(\lambda_2)t\bigr\|_{H^{1/2}(\partial D)}
\;\leq\;
C\,|\lambda_1-\lambda_2|\;\|t\|_{H^{-1/2}(\partial D)}.
\]
\end{lemma}

\textbf{correctness}: True
\begin{proof}

We set \(u_i:=u_{\lambda_i}\) for \(i=1,2\), and put \(w=u_1-u_2\). Our goal is to estimate \(w|_{\partial D}\) in \(H^{1/2}(\partial D)\).

Step 1. Uniform well-posedness and a priori bounds.  
For each \(\lambda\ge\lambda_0\) the elasticity bilinear form
\[
a_\lambda(u,v)
=\int_D\bigl[2\mu\,D(u):D(v)+\lambda\,(\mathrm{div}\,u)(\mathrm{div}\,v)\bigr]\,dx
\]
is coercive on the subspace of \(H^1(D;\mathbb R^d)\) orthogonal to rigid motions, with coercivity constant \(\alpha=\min\{\mu,\lambda_0\}\). By Lax--Milgram and the Neumann-trace theory (Lemma 8) one obtains
\[
\|u_\lambda\|_{H^1(D)}\;\le\;C_1\,\|t\|_{H^{-1/2}(\partial D)}, 
\qquad
\|\mathrm{div}\,u_\lambda\|_{L^2(D)}
\;\le\;
C_2\,(\lambda+\mu)^{-1/2}\,\|t\|_{H^{-1/2}(\partial D)},
\]
and by the continuous trace embedding
\[
\|u_\lambda\|_{H^{1/2}(\partial D)}\;\le\;C_3\,\|u_\lambda\|_{H^1(D)}.
\]

Step 2. Equation for the difference and boundary-forcing.  
Since each \(u_i\) satisfies
\(\sigma_{\lambda_i,\mu}(u_i)N=t\), we compute on \(\partial D\)
\[
\sigma_{\lambda_1,\mu}(w)N
=\bigl[\sigma_{\lambda_1,\mu}(u_1)-\sigma_{\lambda_1,\mu}(u_2)\bigr]N
=t-\sigma_{\lambda_1,\mu}(u_2)N
=-\bigl[(\lambda_1-\lambda_2)(\mathrm{div}\,u_2)\bigr]\,N.
\]
Hence \(w\) satisfies in \(D\)
\[
\mu\,\Delta w+(\lambda_1+\mu)\,\nabla\mathrm{div}\,w
=-(\lambda_1-\lambda_2)\,\nabla(\mathrm{div}\,u_2),
\]
with non-homogeneous traction 
\[
f:=-\,(\lambda_1-\lambda_2)\,(\mathrm{div}\,u_2)\,N
\in H^{-1/2}(\partial D).
\]

Step 3. Energy identity with boundary term.  
Test the PDE for \(w\) against \(w\) itself and integrate by parts:
\[
a_{\lambda_1}(w,w)
=\underbrace{-(\lambda_1-\lambda_2)\!\int_D(\mathrm{div}\,w)\,(\mathrm{div}\,u_2)\,dx}_{I}
\;+\;\bigl\langle f\,,\,w\bigr\rangle_{\,\partial D}.
\]
Here coercivity gives
\[
a_{\lambda_1}(w,w)\;\ge\;\alpha\,\|w\|_{H^1(D)}^2,
\]
and
\[
|I|\le|\lambda_1-\lambda_2|\,\|\mathrm{div}\,w\|_{L^2(D)}\,
                        \|\mathrm{div}\,u_2\|_{L^2(D)}
        \;\le\;|\lambda_1-\lambda_2|\;\|w\|_{H^1(D)}\;\|\mathrm{div}\,u_2\|_{L^2(D)}.
\]
By the Neumann-trace continuity (Lemma 15) and the trace embedding,
\[
\|f\|_{H^{-1/2}(\partial D)}
=|\lambda_1-\lambda_2|\;\|\mathrm{div}\,u_2\|_{L^2(D)},
\qquad
|\langle f,w\rangle|\le\|f\|_{H^{-1/2}}\|w\|_{H^{1/2}(\partial D)}
\le C\,\|f\|_{H^{-1/2}}\|w\|_{H^1(D)}.
\]
Combining,
\[
\alpha\,\|w\|_{H^1(D)}^2
\;\le\;
C\,|\lambda_1-\lambda_2|\;\|\mathrm{div}\,u_2\|_{L^2(D)}\;\|w\|_{H^1(D)}.
\]
Thus
\[
\|w\|_{H^1(D)}
\;\le\;
\frac{C}{\alpha}\;|\lambda_1-\lambda_2|\;\|\mathrm{div}\,u_2\|_{L^2(D)}
\;\le\;
C'\,|\lambda_1-\lambda_2|\;\|t\|_{H^{-1/2}(\partial D)},
\]
where we used the bound on \(\|\mathrm{div}\,u_2\|\) from Step 1.

Step 4. Conclusion by trace.  
Finally
\[
\|w\|_{H^{1/2}(\partial D)}
\;\le\;
C_{\rm tr}\,\|w\|_{H^1(D)}
\;\le\;
C\,|\lambda_1-\lambda_2|\;\|t\|_{H^{-1/2}(\partial D)},
\]
which is exactly the asserted Lipschitz estimate for \(\Lambda(\lambda)\).  
\(\qed\)
\end{proof}

\textcolor{Green}{\textbf{[Correct]} \small Here, the agent correctly deprives the proof process, using accurate mathematical
techniques and proof thoughts.}

\textbf{comment}: None
\textbf{type}: lemma
\begin{lemma}

(Uniform Pressure--Estimate for the Stokes--Elastic Homogenization Limit)  
Retain all the notation and hypotheses of Lemma 9 (the coupled Stokes--elastic limit in $\Omega \setminus D_\varepsilon$ and $D_\varepsilon$ with boundary traction $g$ on $\partial\Omega$). Write its unique solution as $(u_{\text{lim}}, p_{\text{lim}})$. Then there is a constant $C$, independent of $\varepsilon$ and of the specific pattern, such that  
\[
\| p_{\text{lim}} \|_{L^2(D_\varepsilon)} \leq C \, \| g \|_{H^{-1/2}(\partial\Omega)}.
\]
\end{lemma}

\textbf{correctness}: True
\begin{proof}
  
\begin{proof}  
In Lemma 9 one proves the uniform well-posedness and the a priori bound  
\[ \|u_{\text{lim}}\|_{H^1(\Omega)} + \|p_{\text{lim}}\|_{L^2(D_e)} \leq C \|g\|_{H^{-\frac{1}{2}}(\partial\Omega)}. \]  
Since the left-hand side already controls $\|p_{\text{lim}}\|$, the desired estimate follows directly. Concretely:  

1. Lemma 9 shows coercivity of the form  
\[ a_{\text{ext}}(u,u) + a_{\text{int}}(u,u) \geq c \|u\|^2_{H^1(\Omega)}, \]  
and the inf--sup condition  
\[ \sup_{v \in W} \frac{b(v,q)}{\|v\|_{H^1(\Omega)}} \geq c' \|q\|_{L^2(D_e)}. \]  

2. By the Babuška--Brezzi theory (Lax--Milgram + inf--sup) the unique solution $(u_{\text{lim}}, p_{\text{lim}})$ satisfies  
\[ \|u_{\text{lim}}\|_{H^1(\Omega)} + \|p_{\text{lim}}\|_{L^2(D_e)} \leq \tilde{C} \sup_{v,q} \frac{L(v) + 0}{\text{coercivity} + \text{inf--sup}} = C \|g\|_{H^{-\frac{1}{2}}(\partial\Omega)}. \]  

Thus the pressure in the inclusions is uniformly bounded by the boundary data, with no dependence on $\varepsilon$.  
\end{proof}  
\end{proof}

\textcolor{Blue}{\textbf{[Vague]} \small Here, more detailed explanation is required for the application of boundary properties and theorems.}

\textbf{comment}: None
\textbf{type}: lemma
\begin{lemma}

(Global Exterior-Energy Error-Estimate for the High-Contrast Lam\'{e}--Transmission Problem, corrected)  
Let $\Omega \subset \mathbb{R}^d$ ($d \geq 2$) be a bounded connected Lipschitz domain and let $D_\varepsilon \subset \Omega$ be the $\varepsilon$--periodic union of small Lipschitz inclusions as in Assumptions\,1--2. Let $(\lambda,\mu)$ be the background Lam\'{e} parameters on $\Omega_\varepsilon := \Omega \setminus D_\varepsilon$ and $(\widetilde{\lambda},\widetilde{\mu})$ the high-contrast parameters in $D_\varepsilon$, with $\widetilde{\mu} > 0$, $\widetilde{\lambda} + 2\widetilde{\mu}/d > 0$ and $\widetilde{\lambda} \gg 1$. Let $u_\varepsilon$ solve the full compressible-elastic transmission problem in $\Omega$ with prescribed boundary traction $g \in H^{-1/2}(\partial \Omega)$ (orthogonal to rigid motions), and let $(u_{\lim},p_{\lim})$ solve the incompressible-limit (Stokes--elastic) transmission problem in $\Omega_\varepsilon \cup D_\varepsilon$ with the same $g$. Define the exterior difference  
\[
w_{\text{ext}} := u_\varepsilon - u_{\lim} \quad \text{on} \quad \Omega_\varepsilon.  
\]  
Then there is $C$ independent of $\varepsilon$ and $\widetilde{\lambda}$ such that  
\[
\|w_{\text{ext}}\|_{H^1(\Omega_\varepsilon)} \leq C \|u_\varepsilon - u_{\lim}\|_{H^{1/2}(\partial D_\varepsilon)}.  
\]  
Consequently, combining with the interior high-contrast estimate on each inclusion one obtains the global rate  
\[
\|u_\varepsilon - u_{\lim}\|_{H^1(\Omega_\varepsilon)} \leq C' \widetilde{\lambda}^{-1/2} \|g\|_{H^{-1/2}(\partial \Omega)}.
\]
\end{lemma}

\textbf{correctness}: True
\begin{proof}

\begin{proof}
We follow the same two-step outline as before, but we now track the $\tilde{\lambda}$--dependence correctly.

1. Reduction to an exterior Dirichlet problem.  
   Set $\Omega_{\mathrm{e}} := \Omega \setminus D_{\mathrm{e}}$ and define  
     $w := u_{\mathrm{e}} - u_{\mathrm{lim}}$ on $\Omega_{\mathrm{e}}$.  
   By construction both $u_{\mathrm{e}}$ and $u_{\mathrm{lim}}$ satisfy the (background) Lam\'{e} equations in $\Omega_{\mathrm{e}}$, the same Neumann traction on $\partial\Omega$, and thus $w$ satisfies  
     $\operatorname{Div} \sigma_{(\lambda,\mu)}(w) = 0$ in $\Omega_{\mathrm{e}}$,  
     $\sigma(w) N = 0$ on $\partial\Omega$,  
   with non-zero Dirichlet data on the inner boundary $\partial D_{\mathrm{e}}$,  
     $\varphi := w|_{\partial D_{\mathrm{e}}} = u_{\mathrm{e}}|_{\partial D_{\mathrm{e}}} - u_{\mathrm{lim}}|_{\partial D_{\mathrm{e}}}$.  

2. Energy-to-boundary estimate in the exterior.  
   By Lemma~12 (energy bound for the homogeneous exterior problem) and the uniform Korn coercivity (Lemma~4), the solution $w \in H^1(\Omega_{\mathrm{e}})$ of the above mixed problem satisfies
\[
   \|w\|_{H^1(\Omega_{\mathrm{e}})} 
   \;\le\; 
   C_1 \|\sigma(w) N\|_{H^{-1/2}(\partial D_{\mathrm{e}})}.
\]
   Moreover the Dirichlet-to-Neumann map on $\partial D_{\mathrm{e}}$ is continuous $H^{1/2} \to H^{-1/2}$ with norm independent of $\varepsilon$ (by standard boundary-integral or PDE arguments), so
\[
   \|\sigma(w) N\|_{H^{-1/2}(\partial D_{\mathrm{e}})}
   \;\le\;
   C_2 \|\varphi\|_{H^{1/2}(\partial D_{\mathrm{e}})}.
\]
   Hence
\[
   \|w\|_{H^1(\Omega_{\mathrm{e}})}
   \;\le\;
   C \|\varphi\|_{H^{1/2}(\partial D_{\mathrm{e}})}.
\]

3. Interior rate on each inclusion.  
   Fix one inclusion $D = D_{\varepsilon,i}$. Let $w_{\mathrm{int}} := u_{\mathrm{e}} - u_{\mathrm{lim}}$ on $D$. By Lemma~7 (interior high-contrast estimate) we have
\[
   \|w_{\mathrm{int}}\|_{H^1(D)}
   \;\le\;
   C \tilde{\lambda}^{-1/2} \|p_{\mathrm{lim}}\|_{L^2(D)}.
\]
   Tracing to $\partial D$ and summing over $i$ gives
\[
   \max_i \|u_{\mathrm{e}} - u_{\mathrm{lim}}\|_{H^{1/2}(\partial D_{\varepsilon,i})}
   \;\le\;
   C \tilde{\lambda}^{-1/2} \|p_{\mathrm{lim}}\|_{L^2(D_{\mathrm{e}})}
   \;\le\;
   C' \tilde{\lambda}^{-1/2} \|g\|_{H^{-1/2}(\partial\Omega)},
\]
   where the last step uses the uniform $L^2$--estimate on $p_{\mathrm{lim}}$ (Lemma~18).

4. Conclusion.  
   Combining the exterior bound of step~2 with the boundary-trace estimate of step~3 yields
\[
   \|u_{\mathrm{e}} - u_{\mathrm{lim}}\|_{H^1(\Omega_{\mathrm{e}})}
   \;\le\;
   C \max_i \|u_{\mathrm{e}} - u_{\mathrm{lim}}\|_{H^{1/2}(\partial D_{\varepsilon,i})}
   \;\le\;
   C' \tilde{\lambda}^{-1/2} \|g\|_{H^{-1/2}(\partial\Omega)},
\]
as claimed.
\end{proof}
\end{proof}

\textcolor{Green}{\textbf{[Correct]} \small This process is rightly reasonable.}

\textcolor{Blue}{\textbf{[Vague]} \small Here we need more detailed deprivation process.}

\textbf{comment}: None
\textbf{type}: lemma
\begin{lemma}

(Lipschitz-Dependence of the Elasticity Resolvent on the First Lam\'e Parameter)
Let $D \subset \mathbb{R}^d$ be a bounded Lipschitz domain, and fix $\mu > 0$. For each $\lambda \geq 0$ consider the bilinear form
\[
a_\lambda(u, v) := \int_D \left[ 2\mu D(u) : D(v) + \lambda (\operatorname{div} u)(\operatorname{div} v) \right] dx,
\]
on the space
\[
V := \left\{ u \in H^1(D; \mathbb{R}^d) : \int_{\partial D} u \cdot r = 0 \text{ for every rigid motion } r \right\}.
\]
By Korn's inequality and $\lambda + 2\mu/d > 0$, $a_\lambda$ is coercive and defines an isomorphism
\[
A_\lambda: V \to V^*, \quad \langle A_\lambda u, v \rangle = a_\lambda(u, v).
\]
Then for any two parameters $\lambda_1, \lambda_2 \geq 0$ one has the Lipschitz estimate
\[
\| A_{\lambda_1}^{-1} - A_{\lambda_2}^{-1} \|_{L(V^*, V)} \leq C |\lambda_1 - \lambda_2|,
\]
where $C$ depends only on $\mu$, $d$, and the Lipschitz character of $D$ (but not on $\lambda_1$ or $\lambda_2$).
\end{lemma}

\textbf{correctness}: True
\begin{proof}

\begin{proof}
1.  \textbf{Setup.} Let $t \in V^*$. Write  
\quad $u_1 = A_{\lambda_1}^{-1}t$, \quad $u_2 = A_{\lambda_2}^{-1}t$.  
\quad Then by definition  
\quad $a_{\lambda_1}(u_1, v) = \langle t, v \rangle = a_{\lambda_2}(u_2, v)$ for all $v \in V$.

2.  \textbf{Galerkin-difference equation.} Subtracting, for all $v \in V$,  
\quad $a_{\lambda_1}(u_1 - u_2, v)$  
\quad $= \left[a_{\lambda_2}(u_2, v) - a_{\lambda_1}(u_2, v)\right]$  
\quad $= (\lambda_2 - \lambda_1) \int_D (\operatorname{div} u_2)(\operatorname{div} v) \, dx$.

3.  \textbf{Coercivity and choice of test-function.} Since $a_{\lambda_1}$ is coercive on $V$, there is $\alpha > 0$ (independent of $\lambda_1 \geq 0$) so that  
\quad $a_{\lambda_1}(w, w) \geq \alpha \|w\|_{H^1(D)}^2$ for all $w \in V$.  
\quad Take $v = w := u_1 - u_2$. Then  
\quad $\alpha \|u_1 - u_2\|_{H^1(D)}^2$  
\quad $\leq a_{\lambda_1}(u_1 - u_2, u_1 - u_2)$  
\quad $= (\lambda_2 - \lambda_1) \int_D (\operatorname{div} u_2)(\operatorname{div}(u_1 - u_2)) \, dx$.

4.  \textbf{Cauchy-Schwarz.}  
\quad $\left|\int_D (\operatorname{div} u_2)(\operatorname{div}(u_1 - u_2))\right|$  
\quad $\leq \|\operatorname{div} u_2\|_{L^2(D)} \|\operatorname{div}(u_1 - u_2)\|_{L^2(D)}$  
\quad $\leq C \|u_2\|_{H^1(D)} \|u_1 - u_2\|_{H^1(D)}$,  
\quad where we used the continuous embedding $H^1 \to H(\operatorname{div})$ and Korn's inequality to bound $\|\operatorname{div} \cdot\|$ by $\|\cdot\|_{H^1}$.

5.  \textbf{Combine.} Thus  
\quad $\alpha \|u_1 - u_2\|_{H^1}^2$  
\quad $\leq |\lambda_2 - \lambda_1| \cdot C \|u_2\|_{H^1} \|u_1 - u_2\|_{H^1}$.  
Divide both sides by $\|u_1 - u_2\|_{H^1}$ (unless trivial), to get  
\quad $\|u_1 - u_2\|_{H^1(D)}$  
\quad $\leq C' |\lambda_2 - \lambda_1| \|u_2\|_{H^1(D)}$.

6.  \textbf{Bound $\|u_2\|$ in terms of $t$.} By coercivity again, $\|u_2\|_{H^1} \leq C'' \|t\|_{V^*}$. Therefore  
\quad $\|u_1 - u_2\|_{H^1(D)}$  
\quad $\leq C |\lambda_2 - \lambda_1| \|t\|_{V^*}$.

7.  \textbf{Operator-norm conclusion.} Since $t \in V^*$ was arbitrary,  
\quad $\|A_{\lambda_1}^{-1} - A_{\lambda_2}^{-1}\|_{\mathcal{L}(V^*, V)}$  
\quad $= \sup_{\|t\|_{V^*} = 1} \|u_1 - u_2\|_{H^1}$  
\quad $\leq C |\lambda_1 - \lambda_2|$,  
proving the conjecture.
\end{proof}
\end{proof}
\textcolor{Green}{\textbf{[Correct]} \small The reasoning framework demonstrates a valid and lucid proof strategy, although the intermediate derivations require additional elaboration to strengthen the rigor of the argument.}

\textbf{comment}: None
\textbf{type}: lemma
\begin{lemma}

(Sharp-rate Convergence of the Elasticity Resolvent to the Stokes Resolvent as $\lambda \to \infty$)  
Let $D \subset \mathbb{R}^d$ ($d \geq 2$) be a bounded Lipschitz domain. Fix $\mu > 0$ and $\lambda_0 > 0$. For each $\lambda \geq \lambda_0$, define the bilinear form on  
\[ V := \{u \in H^1(D; \mathbb{R}^d) : \int_{\partial D} u \cdot r = 0 \text{ for all rigid motions } r\} \]  
by  
\[ a_\lambda(u, v) = \int_D \left[2\mu \, D(u) : D(v) + \lambda \, (\operatorname{div} u)(\operatorname{div} v)\right] \, dx, \]  
and denote by $A_\lambda : V \to V^*$ its Riesz isomorphism. On the divergence-free closed subspace  
\[ V_0 = \{u \in V : \operatorname{div} u = 0\}, \]  
consider the Stokes form  
\[ b(u, v) = \int_D 2\mu \, D(u) : D(v) \, dx \]  
and its Riesz isomorphism $B : V_0 \to V_0^*$. Extend $B^{-1}$ by zero on $(V_0^*)^\perp \subset V^*$. Then there exists $C$, depending only on $D$, $d$, $\mu$, $\lambda_0$, such that for all $\lambda \geq \lambda_0$,  
\[ \|A_\lambda^{-1} - B^{-1}\|_{L(V^*, V)} \leq C \lambda^{-1/2}. \]  
Equivalently, for any traction $t \in H^{-1/2}(\partial D; \mathbb{R}^d)$ annihilating rigid motions, if $u_\lambda \in V$ solves  
\[ \mu \Delta u_\lambda + (\lambda + \mu) \nabla (\operatorname{div} u_\lambda) = 0 \text{ in } D, \quad \sigma(u_\lambda) N = t \text{ on } \partial D, \]  
and $(u_\infty, p_\infty) \in V_0 \times L^2_0(D)$ solves the incompressible-Stokes Neumann problem with the same $t$, then  
\[ \|u_\lambda - u_\infty\|_{H^1(D)} \leq C \lambda^{-1/2} \|t\|_{H^{-1/2}(\partial D)}, \]  
and hence also  
\[ \|u_\lambda - u_\infty\|_{H^{1/2}(\partial D)} \leq C \lambda^{-1/2} \|t\|_{H^{-1/2}(\partial D)}. \]  
\end{lemma}

\textbf{correctness}: True
\begin{proof}

\begin{proof}
We split the argument into steps.

1.  Existence and uniform stability.  
By standard Neumann--Stokes theory (e.g.\ Girault--Raviart) the incompressible problem  
\[
\mu\Delta u_\infty + \nabla p_\infty =0, \quad \operatorname{div} u_\infty=0 \text{ in } D, \quad \sigma_S(u_\infty,p_\infty)N=t \text{ on } \partial D,
\]
admits a unique $(u_\infty,p_\infty)\in V_0\times L^2_0(D)$ and satisfies the a priori bound  
\[
\|u_\infty\|_{H^1(D)} + \|p_\infty\|_{L^2(D)} \leq C_1 \|t\|_{H^{-1/2}(\partial D)},  
\]
with $C_1$ independent of $\lambda$. Similarly the compressible-elastic Neumann problem for $\lambda\geq\lambda_0$ is well-posed in $V$ and enjoys uniform coercivity $a_\lambda(u,u)\geq c_2\|u\|_{H^1}^2$ by Korn's inequality.

2.  Energy identity for the difference.  
Set $w = u_\lambda - u_\infty$ and $\varphi = \operatorname{div} u_\lambda$. Subtracting the two PDEs and testing against $w$ for the vector-equation and against $\varphi$ for the divergence-part yields the exact identity (Lemma 6)  
\[
\mu\int_D|\nabla w|^2\,dx + (\lambda+\mu)\int_D|\varphi|^2\,dx = \int_D p_\infty \varphi\,dx.
\]

3.  Bounding the divergence term.  
By Cauchy--Schwarz,  
\[
(\lambda+\mu)\|\varphi\|^2_{L^2} \leq \|p_\infty\|_{L^2}\|\varphi\|_{L^2}  
\]
\[
\Rightarrow \|\varphi\|_{L^2(D)} \leq (\lambda+\mu)^{-1}\|p_\infty\|_{L^2(D)}  
\leq C_2 \lambda^{-1} \|t\|_{H^{-1/2}(\partial D)}.
\]

4.  Bounding the $H^1$-difference.  
From the energy identity again,  
\[
\mu\|\nabla w\|^2_{L^2} \leq \|p_\infty\|_{L^2}\|\varphi\|_{L^2} \leq C_3 \lambda^{-1}\|t\|^2_{H^{-1/2}}.
\]
By Korn's inequality (constant independent of $\lambda$),  
\[
\|w\|_{H^1(D)} \leq C_4 \|\nabla w\|_{L^2} \leq C \lambda^{-1/2}\|t\|_{H^{-1/2}(\partial D)}.
\]

5.  Operator-norm bound.  
Since $t\mapsto u_\lambda - u_\infty$ is linear and the above estimate is uniform for all admissible $t$ and $\lambda\geq\lambda_0$, we conclude  
\[
\|A_\lambda^{-1}-B^{-1}\|_{L(V^*,V)} \leq C \lambda^{-1/2}.
\]

6.  Trace estimate.  
Finally the continuous trace $H^1(D)\to H^{1/2}(\partial D)$ gives  
\[
\|u_\lambda-u_\infty\|_{H^{1/2}(\partial D)} \leq C_T \|u_\lambda-u_\infty\|_{H^1(D)} \leq C' \lambda^{-1/2} \|t\|_{H^{-1/2}(\partial D)}.
\]

This completes the proof of the sharp $\lambda^{-1/2}$ convergence rate.
\end{proof}
\end{proof}

\textbf{comment}: None
\textbf{type}: lemma
\begin{lemma}

(Asymptotic Expansion of the Compressible-Elastic Neumann Solution in Powers of $1/\tilde{\lambda}$)  
Let $D \subset \mathbb{R}^d$ be a bounded $C^{1,1}$ domain, fix $\tilde{\mu} > 0$, and let $\tilde{\lambda} \gg 1$. For each $\tilde{\lambda}$ consider the unique (modulo rigid motions) displacement $\mathbf{u}_{\tilde{\lambda}} \in H^1(D; \mathbb{R}^d)$ solving  
\[
\tilde{\mu} \, \Delta \mathbf{u}_{\tilde{\lambda}} + (\tilde{\lambda} + \tilde{\mu}) \nabla (\mathrm{div} \, \mathbf{u}_{\tilde{\lambda}}) = 0 \quad \text{in } D,
\]  
\[
\boldsymbol{\sigma}_{(\tilde{\lambda}, \tilde{\mu})}(\mathbf{u}_{\tilde{\lambda}}) \, \mathbf{N} = \mathbf{t} \quad \text{on } \partial D,
\]  
where $\mathbf{t} \in H^{-1/2}(\partial D; \mathbb{R}^d)$ annihilates all rigid motions and $\boldsymbol{\sigma}_{(\tilde{\lambda}, \tilde{\mu})}(\mathbf{u}) = \tilde{\lambda} (\mathrm{div} \, \mathbf{u}) \mathbf{I} + 2 \tilde{\mu} \, \mathbf{D}(\mathbf{u})$. Define the ``elastic pressure''  
\[
p_{\tilde{\lambda}} := \tilde{\lambda} \, \mathrm{div} \, \mathbf{u}_{\tilde{\lambda}}.
\]  
Let $(\mathbf{u}_0, p_0) \in H^1(D; \mathbb{R}^d) \times L^2_0(D)$ be the unique solution of the incompressible-Stokes Neumann problem  
\[
\tilde{\mu} \, \Delta \mathbf{u}_0 + \nabla p_0 = 0, \quad \mathrm{div} \, \mathbf{u}_0 = 0 \quad \text{in } D,
\]  
\[
(-p_0 \, \mathbf{I} + 2 \tilde{\mu} \, \mathbf{D}(\mathbf{u}_0)) \, \mathbf{N} = \mathbf{t} \quad \text{on } \partial D.
\]  
Then for every integer $K \geq 0$ there exist coefficient fields  
\[
\mathbf{v}_j \in H^1(D; \mathbb{R}^d), \quad p_j \in L^2_0(D), \quad j = 1, \dots, K,
\]  
and remainders $\mathbf{R}_{K+1}(\tilde{\lambda}) \in H^1(D; \mathbb{R}^d)$, $Q_K(\tilde{\lambda}) \in L^2_0(D)$ such that  
\[
\mathbf{u}_{\tilde{\lambda}} = \mathbf{u}_0 + \tilde{\lambda}^{-1} \mathbf{v}_1 + \cdots + \tilde{\lambda}^{-(K+1)} \mathbf{v}_{K+1} + \mathbf{R}_{K+1}(\tilde{\lambda}),
\]  
\[
p_{\tilde{\lambda}} = p_0 + \tilde{\lambda}^{-1} p_1 + \cdots + \tilde{\lambda}^{-K} p_K + Q_K(\tilde{\lambda}),
\]  
and one has the uniform remainder estimate  
\[
\|\mathbf{R}_{K+1}(\tilde{\lambda})\|_{H^1(D)} + \|Q_K(\tilde{\lambda})\|_{L^2(D)} = O(\tilde{\lambda}^{-(K+1)})
\]  
as $\tilde{\lambda} \to \infty$. In particular the first-order corrector $(\mathbf{v}_1, p_1)$ is the unique solution of  
\[
a_\infty(\mathbf{v}_1, \mathbf{w}) - b(\mathbf{w}, p_1) = 0, \quad b(\mathbf{v}_1, q) = (p_0, q)_{L^2(D)}
\]  
for all $(\mathbf{w}, q)$, and one may write formally  
\[
\mathbf{v}_1 = - \mathbf{B}^{-1} \, \nabla p_0,
\]  
where $\mathbf{B}$ is the Stokes operator on the divergence-free subspace of $H^1(D; \mathbb{R}^d)$.  
\end{lemma}

\textbf{correctness}: True
\begin{proof}

\begin{proof}
We give a concise operator-theoretic proof based on a stabilized mixed formulation and a Neumann-series expansion.

1.  Mixed-variational reformulation.  
Define the trial spaces  
\[ V := \{ u \in H^1(D;\mathbb{R}^d) : \int_{\partial D} u \cdot r = 0 \ \forall \text{ rigid motions } r \}, \]  
\[ Q := L^2_0(D) = \{ q \in L^2(D) : \int_D q = 0 \}. \]  
Introduce the bilinear forms  
\[ a_\infty(u,v) := 2\tilde{\mu} \int_D D(u) : D(v) \, dx, \]  
\[ b(v,q) := \int_D q \, \text{div}\, v \, dx, \]  
\[ s(p,q) := \int_D p \, q \, dx, \]  
and the linear form $L(v) := \langle t, v \rangle_{\partial D}$. Then integration by parts shows that the pair $(u_{\tilde{\lambda}}, p_{\tilde{\lambda}})$ satisfies the saddle-point problem:  
\[ a_\infty(u_{\tilde{\lambda}}, v) - b(v, p_{\tilde{\lambda}}) = L(v) \quad \forall v \in V, \]  
\[ b(u_{\tilde{\lambda}}, q) - (1/\tilde{\lambda})\, s(p_{\tilde{\lambda}}, q) = 0 \quad \forall q \in Q. \]  
Indeed the second line is equivalent to $\text{div}\, u_{\tilde{\lambda}} = p_{\tilde{\lambda}} / \tilde{\lambda}$. When $\tilde{\lambda} \to \infty$ it formally reduces to $b(u_0, q) = 0$ ($\text{div}\, u_0 = 0$) and the Stokes-Neumann problem for $(u_0, p_0)$.

By the Ladyzhenskaya--Babuška--Brezzi theory the above mixed system is well-posed for all $\tilde{\lambda} > 0$ and the solution operator  
\[ A(\tilde{\lambda})^{-1}: V^* \times Q^* \to V \times Q \]  
is bounded uniformly in $\tilde{\lambda}$ (for $\tilde{\lambda} \geq \lambda_0 > 0$).  

2.  Block-operator splitting and Neumann-series.  
Write the block operator  
\[ A(\tilde{\lambda}) = \begin{bmatrix} A_{11} & A_{12} \\ A_{21} & A_{22} \end{bmatrix} \]  
with  
\[ A_{11} = a_\infty : V \to V^*, \quad A_{12} = -b^T : Q \to V^*, \]  
\[ A_{21} = b : V \to Q^*, \quad A_{22} = -(1/\tilde{\lambda})s : Q \to Q^*. \]  
Equivalently,  
\[ A(\tilde{\lambda}) = A_\infty - (1/\tilde{\lambda})\, N, \]  
where  
\[ A_\infty = \begin{bmatrix} a_\infty & -b^T \\ b & 0 \end{bmatrix} \]  
is the standard Stokes operator (an isomorphism $V \times Q \to V^* \times Q^*$), and  
\[ N = \begin{bmatrix} 0 & 0 \\ 0 & s \end{bmatrix} \]  
is bounded $Q \to Q^*$ (hence bounded $V \times Q \to V^* \times Q^*$). By continuity of $A_\infty^{-1}$ and smallness of $(1/\tilde{\lambda})N$ in operator-norm for $\tilde{\lambda}$ large, the Neumann-series converges:  
\[ A(\tilde{\lambda})^{-1} = [I - (1/\tilde{\lambda})\, A_\infty^{-1} N]^{-1}\, A_\infty^{-1} \]  
\[ = \sum_{j=0}^\infty (1/\tilde{\lambda})^j (A_\infty^{-1} N)^j\, A_\infty^{-1}. \]  
Truncation after $j=K+1$ gives  
\[ A(\tilde{\lambda})^{-1} = \sum_{j=0}^K (1/\tilde{\lambda})^j (A_\infty^{-1} N)^j\, A_\infty^{-1} + O(\tilde{\lambda}^{-(K+1)}) \]  
in operator-norm on $V^* \times Q^* \to V \times Q$. Applied to the right-hand side $(0,L)$, this yields the asserted expansions of $(u_{\tilde{\lambda}}, p_{\tilde{\lambda}})$ up to $O(\tilde{\lambda}^{-(K+1)})$ in the $H^1 \times L^2$ norm.

3.  Identification of coefficients.  
By equating powers of $1/\tilde{\lambda}$ in  
\[ (u_{\tilde{\lambda}}, p_{\tilde{\lambda}}) = \sum_{j=0}^K \tilde{\lambda}^{-j}(v_j, p_j) + (R_{K+1}, Q_K) \]  
one obtains recursively:  
\[ (v_0, p_0) = (u_0, p_0) \text{ solves } A_\infty(v_0, p_0) = (0,L), \]  
and for $j \geq 1$,  
\[ (v_j, p_j) = A_\infty^{-1}\, (0, s(p_{j-1}, \cdot)) \]  
is the unique solution of  
\[ a_\infty(v_j, w) - b(w, p_j) = 0, \quad b(v_j, q) = (p_{j-1}, q)_{L^2(D)} \]  
for all $(w,q)$. In particular $j=1$ gives the mixed Stokes problem for $(v_1, p_1)$, and in strong form  
\[ -2\tilde{\mu}\, \text{div}\, D(v_1) + \nabla p_1 = 0, \quad \text{div}\, v_1 = p_0, \]  
with homogeneous Neumann data. Its divergence-free component may be written $v_1 = -B^{-1} \nabla p_0$.

4.  Remainder estimate.  
The Neumann-series truncation at $j=K+1$ yields the error operator  
\[ E_{K+1}(\tilde{\lambda}) := A(\tilde{\lambda})^{-1} - \sum_{j=0}^K (1/\tilde{\lambda})^j (A_\infty^{-1} N)^j A_\infty^{-1} \]  
and one shows $\|E_{K+1}(\tilde{\lambda})\|_{L(V^* \times Q^*,\, V \times Q)} = O(\tilde{\lambda}^{-(K+1)})$. Since $(0,L) \in V^* \times Q^*$, applying $E_{K+1}(\tilde{\lambda})$ to $(0,L)$ gives the remainders $(R_{K+1}, Q_K)$ with  
\[ \|R_{K+1}\|_{H^1(D)} + \|Q_K\|_{L^2(D)} = O(\tilde{\lambda}^{-(K+1)}). \]  

This completes a correct, complete, and rigorous proof of the asymptotic expansion and the sharp remainder bound.  
\end{proof}
\end{proof}

\textbf{comment}: None
\textbf{type}: lemma
\begin{lemma}

(Boundary--Trace Rate--Improvement via First--Order Asymptotics---Revised Statement)  
Let $D \subset \mathbb{R}^d$ ($d \geq 2$) be a bounded $C^{1,1}$ domain and fix $\tilde{\mu} > 0$. For each $\tilde{\lambda} \gg 1$, let $u_{\tilde{\lambda}} \in H^1(D; \mathbb{R}^d)$ solve the compressible--elastic Neumann problem  
\[
\tilde{\mu} \, \Delta u_{\tilde{\lambda}} + (\tilde{\lambda} + \tilde{\mu}) \, \nabla(\mathrm{div}\, u_{\tilde{\lambda}}) = 0 \quad \text{in } D,  
\]  
\[
\sigma_{(\tilde{\lambda}, \tilde{\mu})}(u_{\tilde{\lambda}}) \, N = t \quad \text{on } \partial D,  
\]  
where $t \in H^{-\frac{1}{2}}(\partial D; \mathbb{R}^d)$ annihilates all rigid motions. Let $(u_0, p_0)$ be the incompressible--Stokes Neumann solution in $D$ with the same $t$, and let $v_1, v_2$ be the first and second correctors from the asymptotic expansion of Lemma 22. Then there is a constant $C$, depending only on $D$, $d$, $\tilde{\mu}$, and the $C^{1,1}$--character of $\partial D$, such that for all $\tilde{\lambda} \gg 1$,  
\[
\| u_{\tilde{\lambda}} - u_0 \|_{H^{\frac{1}{2}}(\partial D)} \leq C \, \tilde{\lambda}^{-1} \, \| t \|_{H^{-\frac{1}{2}}(\partial D)}.  
\]  
In other words, the boundary--trace error between the compressible--elastic solution and its Stokes--limit vanishes at rate $\mathcal{O}(1/\tilde{\lambda})$.  
\end{lemma}

\textbf{correctness}: True
\begin{proof}

\begin{proof}  
We prove the $O(\tilde{\lambda}^{-1})$ boundary-trace estimate in three steps: invoke a second-order interior expansion, bound the correctors, then apply the trace operator.  

\textbf{Step 1. Second-order asymptotic expansion.}  
By Lemma 22 (with $K=1$) there exist $v_1,v_2 \in H^1(D;\mathbb{R}^d)$, pressures $p_1 \in L^2_0(D)$, and a remainder $R_2(\tilde{\lambda}) \in H^1(D;\mathbb{R}^d)$ such that  
\[
u_{\tilde{\lambda}} = u_0 + \tilde{\lambda}^{-1}v_1 + \tilde{\lambda}^{-2}v_2 + R_2(\tilde{\lambda}),  
\]  
with the uniform remainder bound  
\[
\|R_2(\tilde{\lambda})\|_{H^1(D)} \leq C_R \tilde{\lambda}^{-3}  
\]  
for all $\tilde{\lambda} \gg 1$, where $C_R$ depends only on $D,d,\tilde{\mu}$.  

\textbf{Step 2. $H^1$-bounds for the correctors.}  
The correctors $(v_j,p_j)$ solve Stokes-type problems on $D$ with $C^{1,1}$ boundary. By coercivity of the Stokes bilinear form and standard $H^1$--$L^2$ regularity on $C^{1,1}$ domains one shows there exists $C_1$ such that for $j=1,2$  
\[
\|v_j\|_{H^1(D)} + \|p_j\|_{L^2(D)} \leq C_1 \|t\|_{H^{-1/2}(\partial D)}.  
\]  
Indeed:  
\begin{itemize}  
\item For $j=1$, $v_1$ satisfies  
\[
a_\infty(v_1,w) - b(w,p_1) = 0, \quad b(v_1,q) = (p_0,q)_{L^2(D)},  
\]  
so $\|v_1\|_{H^1} \leq C \|p_0\|_{L^2}$ and $\|p_0\|_{L^2} \leq C \|t\|_{H^{-1/2}}$ by Stokes--Neumann well-posedness and Neumann-trace regularity.  

\item For $j=2$, $(v_2,p_2)$ solves a similar Stokes system with right-hand-side given by $(p_1,\nabla \cdot v_1)$, each bounded by $C \|t\|$, hence $\|v_2\|_{H^1} \leq C \|t\|$.  
\end{itemize}  

\textbf{Step 3. Boundary-trace estimate.}  
Let $\gamma \colon H^1(D) \to H^{1/2}(\partial D)$ be the continuous trace operator, $\|\gamma(w)\|_{H^{1/2}(\partial D)} \leq C_{\text{tr}}\|w\|_{H^1(D)}$. Then  
\[
u_{\tilde{\lambda}} - u_0 = \tilde{\lambda}^{-1}v_1 + \tilde{\lambda}^{-2}v_2 + R_2(\tilde{\lambda}),  
\]  
so by $\gamma$-continuity and the triangle inequality,  
\[
\|u_{\tilde{\lambda}} - u_0\|_{H^{1/2}(\partial D)} \leq C_{\text{tr}}\left(\tilde{\lambda}^{-1}\|v_1\|_{H^1} + \tilde{\lambda}^{-2}\|v_2\|_{H^1} + \|R_2(\tilde{\lambda})\|_{H^1}\right).  
\]  
By Step 2 and the remainder estimate, for $\tilde{\lambda} \gg 1$ one obtains  
\[
\|u_{\tilde{\lambda}} - u_0\|_{H^{1/2}(\partial D)} \leq C_{\text{tr}}\left(C_1 \tilde{\lambda}^{-1} + C_1 \tilde{\lambda}^{-2} + C_R \tilde{\lambda}^{-3}\right)\|t\|_{H^{-1/2}(\partial D)} \leq C \tilde{\lambda}^{-1}\|t\|_{H^{-1/2}(\partial D)},  
\]  
where $C = 2 C_{\text{tr}} \max\{C_1,C_R\}$ depends only on $D,d,\tilde{\mu}$ and the $C^{1,1}$-character of $\partial D$. This completes the proof.  
\end{proof}
\end{proof}

\textcolor{Blue}{\textbf{[Correct]} \small The proof logic here is clear, but the process is not detailed enough.}

\textbf{comment}: None
\textbf{type}: lemma
\begin{lemma}

(Sharp Boundary-Trace Asymptotics with First-Order Corrector)  
Let $\Omega \subset \mathbb{R}^d$ be a bounded $C^{1,1}$ domain and for $\varepsilon \in (0,\varepsilon_0)$ let $D_\varepsilon \subset \Omega$ be the $\varepsilon$-periodic union of identical $C^{1,1}$ inclusions, with complement $\Omega_\varepsilon := \Omega \setminus D_\varepsilon$ connected. Fix background Lamé parameters $(\lambda,\mu)$ on $\Omega_\varepsilon$ and high-contrast parameters $(\tilde{\lambda},\tilde{\mu})$ in $D_\varepsilon$ with $\tilde{\mu} > 0$ and $\tilde{\lambda} \gg 1$. Given a boundary-traction $g \in H^{-\frac{1}{2}}(\partial \Omega)$ orthogonal to all rigid motions, let $u_\varepsilon$ solve the transmission problem (elastic in $\Omega_\varepsilon$, compressible-elastic in $D_\varepsilon$) with traction $g$ on $\partial \Omega$. Let $(u_{\text{lim}},p_{\text{lim}})$ be the incompressible-limit solution (elastic in $\Omega_\varepsilon$, Stokes in $D_\varepsilon$), and for each inclusion-cell $D$ let $v_1^D$ be the first-order corrector from the asymptotic expansion of the local Neumann-problem (cf. Lemma~28). Define the global first-order corrector  
\[
U^{(1)}(x) := \tilde{\lambda}^{-1} \sum_{\text{cells } D \subset D_\varepsilon} \chi_{D}(x) v_1^D(x),
\]
where each $v_1^D$ is the pull-back of the unit-cell corrector to $D$. Then there is a constant $C$, independent of $\varepsilon$ and of large $\tilde{\lambda}$, such that  
\[
\big\| (u_\varepsilon - u_{\text{lim}} - U^{(1)})|_{{\partial D_\varepsilon}} \big\|_{H^{\frac{1}{2}}(\partial D_\varepsilon)} \leq C \tilde{\lambda}^{-2} \| g \|_{H^{-\frac{1}{2}}(\partial \Omega)}.
\]
\end{lemma}

\textbf{correctness}: True
\begin{proof}

We replace the incomplete ``boundary-trace lemma'' in the original sketch by a resolvent-expansion argument for the Neumann-to-Dirichlet map on a $C^{1,1}$ inclusion. The proof proceeds in three main steps.

1. Traction-Jump Decomposition on each inclusion.  
On each $\varepsilon$--inclusion $D$ we denote by $N$ the outward normal and by $t := (-p_{\text{lim}} I + 2\tilde{\mu} D(u_{\text{lim}})) N$ the Stokes-limit traction. From Lemma~16 (Estimate of the Traction-Jump) we have the exact decomposition  
   
  \[
  (\sigma_{(\tilde{\lambda},\tilde{\mu})}(u_\varepsilon) N)_-
  = t + J_\varepsilon^D,
  \]
   
with the residual jump  
  \[
  \| J_\varepsilon^D \|_{H^{-1/2}(\partial D)} \leq C \| p_{\text{lim}} \|_{L^2(D)},
  \]
uniformly in $\varepsilon$, $\tilde{\lambda}$. In particular the total Neumann datum on $\partial D$ for the compressible-elastic problem is $t + J_\varepsilon^D$.

2. Local Neumann-to-Dirichlet expansion and first-order corrector.  
Fix one reference inclusion $D_0$ (say of unit size) and write $\Lambda(\tilde{\lambda})$ for its Neumann-to-Dirichlet map  
   
  \[
  \Lambda(\tilde{\lambda}): H^{-1/2}(\partial D_0) \to H^{1/2}(\partial D_0), \quad t \mapsto u_{\tilde{\lambda}}|_{\partial D_0},
  \]
   
where $u_{\tilde{\lambda}}$ solves  
\[
\tilde{\mu} \Delta u + (\tilde{\lambda} + \tilde{\mu}) \nabla (\text{div}\, u) = 0 \text{ in } D_0, \quad \sigma_{(\tilde{\lambda},\tilde{\mu})}(u)N = t \text{ on } \partial D_0,
\]
normalized to kill rigid motions. By combining the asymptotic-expansion of the resolvent (Lemma~28 and Lemma~27) with the Lipschitz-dependence estimate (Lemma~23) one shows that $\Lambda(\tilde{\lambda})$ admits a convergent expansion in powers of $1/\tilde{\lambda}$:  
   
  \[
  \Lambda(\tilde{\lambda}) = \Lambda(\infty) + \tilde{\lambda}^{-1} \Lambda_1 + R(\tilde{\lambda}), \quad \| R(\tilde{\lambda}) \|_{H^{-1/2} \to H^{1/2}} \leq C \tilde{\lambda}^{-2}.
  \]
   
Here $\Lambda(\infty)$ is the Stokes Neumann-to-Dirichlet map, and $\Lambda_1$ is the first-order corrector operator. Pulling back to any physical cell $D = \varepsilon D_0 + n$ we obtain the same expansion on $\partial D$. Applying it to the datum $t + J_\varepsilon^D$ gives  
   
  \[
  u_\varepsilon|_{\partial D} = \Lambda(\tilde{\lambda})[t + J_\varepsilon^D] = \Lambda(\infty) t + \tilde{\lambda}^{-1} \Lambda_1 t + \tilde{\lambda}^{-1} \Lambda_1 J_\varepsilon^D + R(\tilde{\lambda})[t + J_\varepsilon^D].
  \]
   
By definition the Stokes-difference $u_{\text{lim}}|_{\partial D} = \Lambda(\infty) t$, and the prescribed first-order corrector on $\partial D$ is $v_1^D|_{\partial D} = \Lambda_1 t$. Hence the boundary-trace remainder on $D$ is  
   
  \[
  R_2^D := u_\varepsilon|_{\partial D} - u_{\text{lim}}|_{\partial D} - \tilde{\lambda}^{-1} v_1^D|_{\partial D} = \tilde{\lambda}^{-1} \Lambda_1 J_\varepsilon^D + R(\tilde{\lambda})[t + J_\varepsilon^D].
  \]
   
Using the operator bounds and $\| J_\varepsilon^D \|_{H^{-1/2}} \leq C \| p_{\text{lim}} \|_{L^2(D)}$, $\| t \|_{H^{-1/2}} \leq C \| p_{\text{lim}} \|_{L^2(D)}$, we obtain for each cell $D$  
   
  \[
  \| R_2^D \|_{H^{1/2}(\partial D)} \leq \tilde{\lambda}^{-1} \| \Lambda_1 \| \| J_\varepsilon^D \|_{H^{-1/2}} + \| R(\tilde{\lambda}) \| (\| t \|_{H^{-1/2}} + \| J_\varepsilon^D \|_{H^{-1/2}}) \leq C \tilde{\lambda}^{-2} \| p_{\text{lim}} \|_{L^2(D)}.
  \]

3. Global assembly over all cells.  
By construction the $H^{1/2}$--norm on $\partial D_\varepsilon$ is the $\ell^2$--sum of the norms on each inclusion boundary:  
   
  \[
  \| (u_\varepsilon - u_{\text{lim}} - U^{(1)})|_{\partial D_\varepsilon} \|_{H^{1/2}(\partial D_\varepsilon)}^2 = \sum_{\text{cells } D} \| R_2^D \|^2_{H^{1/2}(\partial D)}.
  \]
   
Hence, using the per-cell estimate above and the uniform pressure-estimate (Lemma~24)  
\[
\| p_{\text{lim}} \|_{L^2(D_\varepsilon)} \leq C \| g \|_{H^{-1/2}(\partial \Omega)},
\]
we get  
   
  \[
  \| (u_\varepsilon - u_{\text{lim}} - U^{(1)})|_{\partial D_\varepsilon} \|_{H^{1/2}(\partial D_\varepsilon)} \leq C \tilde{\lambda}^{-2} \left( \sum_{\text{cells}} \| p_{\text{lim}} \|_{L^2(D)}^2 \right)^{1/2} = C \tilde{\lambda}^{-2} \| p_{\text{lim}} \|_{L^2(D_\varepsilon)} \leq C' \tilde{\lambda}^{-2} \| g \|_{H^{-1/2}(\partial \Omega)}.
  \]
   
This completes the proof of the claimed sharp $O(\tilde{\lambda}^{-2})$ estimate in the $H^{1/2}$--norm on $\partial D_\varepsilon$.
\end{proof}

\textbf{comment}: None
\textbf{type}: lemma
\begin{lemma}

(Global $H^1$-Error at Rate $O(\tilde{\lambda}^{-1/2})$)\\
Retain all of the standing Assumptions\,0--8 and notation. Let 
\[
u_\varepsilon \in H^1(\Omega_\varepsilon;\mathbb{R}^d)
\]
be the solution of the full high-contrast transmission problem with Lam\'e parameters $(\lambda,\mu)$ in $\Omega_\varepsilon$ and $(\tilde{\lambda},\tilde{\mu})$ in $D_\varepsilon$ ($\tilde{\mu}>0$, $\tilde{\lambda}+2\tilde{\mu}/d>0$, $\tilde{\lambda}\gg1$), subject to the prescribed boundary traction
\[
g \in H^{-1/2}(\partial\Omega;\mathbb{R}^d), \quad \int_{\partial\Omega} g \cdot r = 0 \quad \forall \text{ rigid motions } r.
\]
Let $(u_{\mathrm{lim}},p_{\mathrm{lim}})$ be the incompressible-limit (Stokes--elastic) transmission-pair. Then there is a constant $C$, independent of $\varepsilon$ and of $\tilde{\lambda}\gtrsim1$, such that
\[
\|u_\varepsilon - u_{\mathrm{lim}}\|_{H^1(\Omega_\varepsilon)} \leq C \tilde{\lambda}^{-1/2} \|g\|_{H^{-1/2}(\partial\Omega)}.
\]
\end{lemma}

\textbf{correctness}: True
\begin{proof}

\begin{proof}
We denote the exterior error by  
\quad \( w_{\text{ext}} := u_{\epsilon} - u_{\text{lim}} \) on \( \Omega_{\epsilon} = \Omega \setminus D_{\epsilon} \).  

Step 1. Exterior \( H^1 \)--energy by Dirichlet trace.  
By the corrected Lemma 25 (Global Exterior-Energy Error-Estimate) we have a uniform constant \( C_1 \) so that
\quad \( \|w_{\text{ext}}\|_{H^1(\Omega_{\epsilon})} \leq C_1 \|(u_{\epsilon} - u_{\text{lim}})\|_{H^{\frac{1}{2}}(\partial D_{\epsilon})} \).

Step 2. Boundary-trace convergence on each inclusion.  
Fix one inclusion \( D \) = one connected component of \( D_{\epsilon} \), and write \( t_{\text{ext}} = (\sigma(u_{\text{lim}}) N)_+ \in H^{-\frac{1}{2}}(\partial D) \). On \( D \) the two Neumann problems are  
\quad (1) compressible-elastic with parameter \( \tilde{\lambda} \): solution \( u_{\epsilon}|_D \),  
\quad (2) incompressible-Stokes with traction \( t_{\text{ext}} \): solution \( u_{\text{lim}}|_D \).  
By the sharp resolvent-convergence Lemma 27 (or its equivalent Neumann-to-Dirichlet rate) there is \( C_2 \) independent of \( \tilde{\lambda} \) such that on \( \partial D \)  
\quad \( \|u_{\epsilon} - u_{\text{lim}}\|_{H^{\frac{1}{2}}(\partial D)} \)  
\quad \quad \( \leq C_2 \tilde{\lambda}^{-\frac{1}{2}} \|t_{\text{ext}}\|_{H^{-\frac{1}{2}}(\partial D)} \).  
Summing over all inclusions (they are disjoint and uniformly Lipschitz) gives  
\quad \( \|u_{\epsilon} - u_{\text{lim}}\|_{H^{\frac{1}{2}}(\partial D_{\epsilon})} \)  
\quad \quad \( \leq C_2 \tilde{\lambda}^{-\frac{1}{2}} \|t_{\text{ext}}\|_{H^{-\frac{1}{2}}(\partial D_{\epsilon})} \).

Step 3. Bound \( t_{\text{ext}} \) by the given \( g \).  
Since \( (u_{\text{lim}}, p_{\text{lim}}) \) solves the well-posed Stokes--elastic transmission system in \( \Omega \) with traction \( g \) on \( \partial \Omega \), the uniform stability Lemma 15 gives  
\quad \( \|u_{\text{lim}}\|_{H^1(\Omega)} + \|p_{\text{lim}}\|_{L^2(D_{\epsilon})} \leq C_3 \|g\|_{H^{-\frac{1}{2}}(\partial \Omega)} \).  
Moreover the Neumann-trace continuity estimate (Lemma 14) on each Lipschitz inclusion shows  
\quad \( \|(\sigma(u_{\text{lim}}) N)_+\|_{H^{-\frac{1}{2}}(\partial D_{\epsilon})} \)  
\quad \quad \( \leq C_4 \left[ \lambda \|\text{div}\, u_{\text{lim}}\|_{L^2(\Omega_{\epsilon})} + \mu \|\nabla u_{\text{lim}}\|_{L^2(\Omega_{\epsilon})} \right] \)  
\quad \quad \( \leq C_5 \|u_{\text{lim}}\|_{H^1(\Omega)} \)  
\quad \quad \( \leq C_6 \|g\|_{H^{-\frac{1}{2}}(\partial \Omega)} \).  
Hence  
\quad \( \|t_{\text{ext}}\|_{H^{-\frac{1}{2}}(\partial D_{\epsilon})} \leq C_6 \|g\|_{H^{-\frac{1}{2}}(\partial \Omega)} \).

Step 4. Combine Steps 1--3.  
Putting the three estimates together,
\begin{align*}
  \|u_{\epsilon} - u_{\text{lim}}\|_{H^1(\Omega_{\epsilon})}
  &\leq C_1 \|u_{\epsilon} - u_{\text{lim}}\|_{H^{\frac{1}{2}}(\partial D_{\epsilon})} \\
  &\leq C_1 C_2 \tilde{\lambda}^{-\frac{1}{2}} \|t_{\text{ext}}\|_{H^{-\frac{1}{2}}(\partial D_{\epsilon})} \\
  &\leq C_1 C_2 C_6 \tilde{\lambda}^{-\frac{1}{2}} \|g\|_{H^{-\frac{1}{2}}(\partial \Omega)}.
\end{align*}
Setting \( C = C_1 C_2 C_6 \) completes the proof of the stated \( O(\tilde{\lambda}^{-\frac{1}{2}}) \)--rate.
\end{proof}
\end{proof}

\textcolor{Green}{\textbf{[Correct]} \small The process here is reasonable and correct.}

\textbf{comment}: None
\textbf{type}: lemma
\begin{lemma}

(Existence and $H^1$--Bound for the Global First-Order Corrector, corrected statement)  
Retain all the notation and hypotheses of Lemmas 16--17 for the high-contrast transmission problem in the perforated domain  
\[
\Omega_\varepsilon=\Omega\setminus\overline{D_\varepsilon},\quad D_\varepsilon=\bigcup_{i=1}^N D^i_\varepsilon,
\]  
and let $(u_{\mathrm{lim}},p_{\mathrm{lim}})$ be the coupled Stokes--elastic limit solution (Stokes in $D_\varepsilon$, elasticity in $\Omega_\varepsilon$) with prescribed boundary traction $g\in H^{-1/2}(\partial\Omega)$.  Define the global first-order corrector $U^{(1)}\in H^1(\Omega;\mathbb{R}^d)$, unique up to rigid motions, as the solution of the homogeneous Lam\'e transmission-Neumann problem  
\[
\begin{cases}
\displaystyle
\widetilde{\mu}\,\Delta U^{(1)}
+(\widetilde{\lambda}+\widetilde{\mu})\,\nabla(\mathrm{div}\,U^{(1)})
=0 
&\text{in }D_\varepsilon,\\[6pt]
\displaystyle
\mu\,\Delta U^{(1)}
+(\lambda+\mu)\,\nabla(\mathrm{div}\,U^{(1)})
=0 
&\text{in }\Omega_\varepsilon,\\[4pt]
\displaystyle
\sigma_{(\widetilde{\lambda},\widetilde{\mu})}(U^{(1)})\,N|_{-}
-
\sigma_{(\lambda,\mu)}(U^{(1)})\,N|_{+}
=-\,p_{\mathrm{lim}}\,N
&\text{on }\partial D_\varepsilon,\\[4pt]
\displaystyle
\sigma_{(\lambda,\mu)}(U^{(1)})\,N=0
&\text{on }\partial\Omega,
\end{cases}
\]
subject to the compatibility that the data $-p_{\mathrm{lim}}N$ annihilates all rigid motions.  Then there is $C>0$, independent of $\varepsilon$ and $\widetilde{\lambda}$, such that  
\[
\| U^{(1)}\|_{H^1(\Omega)}
\;\le\;
C\,\|p_{\mathrm{lim}}\|_{L^2(D_\varepsilon)}.
\]
Moreover, if one sets the single-corrector approximation
\[
\widetilde{u}_\varepsilon
:=
u_{\mathrm{lim}}
\;+\;
\widetilde{\lambda}^{-1}\,U^{(1)},
\]
then the resulting traction-jump on $\partial D_\varepsilon$,
\[
J_\varepsilon^{(1)}
:=
\bigl(\sigma_{(\widetilde{\lambda},\widetilde{\mu})}(\widetilde{u}_\varepsilon)\,N\bigr)_{-}
\;-\;
\bigl(\sigma_{(\lambda,\mu)}(\widetilde{u}_\varepsilon)\,N\bigr)_{+},
\]
satisfies the uniform bound
\[
\| J_\varepsilon^{(1)}\|_{H^{-1/2}(\partial D_\varepsilon)}
\;\le\;
C\,\|p_{\mathrm{lim}}\|_{L^2(D_\varepsilon)},
\]
with $C$ again independent of $\varepsilon$ and $\widetilde{\lambda}$.  

\end{lemma}

\textbf{correctness}: True
\begin{proof}

We divide the proof into two steps.

1. Existence and \(H^1\)--estimate for \(U^{(1)}\).  
Define the space  
\[
V
:=
\bigl\{\,v\in H^1(\Omega;\mathbb{R}^d)\;:\;\int_{\partial\Omega}v\cdot r\,dS
+\sum_{i=1}^N\int_{\partial D_\varepsilon^i}v\cdot r\,dS
=0\quad\forall\text{ rigid motions }r\bigr\}.
\]
On \(V\) consider the bilinear form
\[
a(v,w)
:=
\int_{\Omega_\varepsilon}\bigl[2\mu\,D(v):D(w)+\lambda\,(\operatorname{div} v)(\operatorname{div} w)\bigr]\,dx
\;+\;
\int_{D_\varepsilon}\bigl[2\widetilde{\mu}\,D(v):D(w)
+(\widetilde{\lambda}+\widetilde{\mu})\,(\operatorname{div} v)(\operatorname{div} w)\bigr]\,dx.
\]
By uniform Korn's inequality (Lemma 11) and the admissibility \(\lambda+2\mu/d>0\), \(\widetilde{\lambda}+2\widetilde{\mu}/d>0\), there is \(\alpha>0\), independent of \(\varepsilon\) and \(\widetilde{\lambda}\), such that
\[
a(v,v)\;\ge\;\alpha\,\|v\|_{H^1(\Omega)}^2
\quad\forall v\in V.
\]
The linear functional
\[
L(w)
:=
\sum_{i=1}^N\bigl\langle -\,p_{\text{lim}}N,\;w\bigr\rangle_{H^{-1/2}(\partial D_\varepsilon^i)\,,\,H^{1/2}(\partial D_\varepsilon^i)}
\]
is well defined because Lemma 21 gives
\(\|-\,p_{\text{lim}}N\|_{H^{-1/2}(\partial D_\varepsilon)}\le C\|p_{\text{lim}}\|_{L^2(D_\varepsilon)}\).  

\textcolor{Blue}{\textbf{[Vague]} \small This setting needs more verification.}

Moreover \(-p_{\text{lim}}N\) annihilates all rigid motions: for any rigid \(r\),
\[
\bigl\langle -p_{\text{lim}}N,r\bigr\rangle
=-\int_{\partial D_\varepsilon}p_{\text{lim}}\,(r\cdot N)\,dS
=-\int_{D_\varepsilon}(\operatorname{div} r)\,p_{\text{lim}}\,dx=0.
\]
By Lax--Milgram there is a unique \(U^{(1)}\in V\) solving
\[
a\bigl(U^{(1)},w\bigr)
= L(w)
\quad\forall w\in V,
\]
and coercivity yields
\[
\|U^{(1)}\|_{H^1(\Omega)}
\;\le\;
\alpha^{-1}\,\|L\|_{V^*}
\;\le\;
C\,\|p_{\text{lim}}\|_{L^2(D_\varepsilon)}.
\]

2. Traction-jump estimate for \(\widetilde u_\varepsilon\).  
Set
\[
\widetilde u_\varepsilon
:=u_{\text{lim}}+\widetilde{\lambda}^{-1}U^{(1)}.
\]
On the outer boundary \(\partial\Omega\), since \(U^{(1)}\) has zero exterior traction, the traction of \(\widetilde u_\varepsilon\) equals that of \(u_{\text{lim}}\), hence matches the prescribed \(g\).  On each inclusion boundary \(\partial D_\varepsilon\), the jump of the limit-solution traction vanishes:
\[
\bigl[\sigma(u_{\text{lim}})N\bigr]_{\partial D_\varepsilon}=0
\]
by the transmission condition in the Stokes--elastic limit.  Therefore
\[
J_\varepsilon^{(1)}
=\bigl[\sigma(\widetilde u_\varepsilon)N\bigr]
=\;\underbrace{\bigl[\sigma(u_{\text{lim}})N\bigr]}_{=0}
\;+\;\widetilde{\lambda}^{-1}\,\bigl[\sigma(U^{(1)})N\bigr]
=-\,\widetilde{\lambda}^{-1}\;p_{\text{lim}}\,N.
\]
Hence, using Lemma 21 again,
\[
\|J_\varepsilon^{(1)}\|_{H^{-1/2}(\partial D_\varepsilon)}
=\widetilde{\lambda}^{-1}\,\|p_{\text{lim}}N\|_{H^{-1/2}(\partial D_\varepsilon)}
\;\le\;
C\,\|p_{\text{lim}}\|_{L^2(D_\varepsilon)},
\]
with \(C\) independent of \(\varepsilon\) and \(\widetilde{\lambda}\).  This completes the proof.  
\end{proof}

\textbf{comment}: None
\textbf{type}: lemma
\begin{lemma}

(Global $H^1$--Error at Rate $O(\lambda^{-1})$)  
Retain all of the standing Assumptions 0--8 and notation. Let $u_e \in H^1(\Omega_e;\mathbb{R}^d)$ be the solution of the full high-contrast transmission problem in $\Omega$ with background Lam\'e parameters $(\lambda,\mu)$ on $\Omega_e$ and $(\tilde{\lambda},\tilde{\mu})$ in $D_e$, $\tilde{\mu}>0$, $\tilde{\lambda}\gg1$, subject to the prescribed boundary traction  
\quad $g\in H^{-1/2}(\partial\Omega;\mathbb{R}^d)$, \quad $\int_{\partial\Omega}g\cdot r=0$ $\forall$ rigid motions $r$.  
Let $(u_{\mathrm{lim}},p_{\mathrm{lim}})$ be the incompressible-limit (Stokes--elastic) transmission pair, and let $U^{(1)}$ be the global first-order corrector (normalized so that $U^{(1)}=O(1)$ in $H^1$ and solves the homogeneous Lam\'e system with transmission-Neumann data $-p_{\mathrm{lim}}\,N$ on each $\partial D_e$, cf. Lemma 32). Define the single-corrector approximation  
\quad $\tilde u_e := u_{\mathrm{lim}} + \tilde{\lambda}^{-1} U^{(1)}$,  
and set the remainder  
\quad $w := u_e - \tilde u_e$ on $\Omega_e$.  
Then there exists $C>0$, independent of $\varepsilon$ and $\tilde{\lambda}$, such that  
\[
\|w\|_{H^1(\Omega_e)} \;\le\; C\,\tilde{\lambda}^{-1}\,\|g\|_{H^{-1/2}(\partial\Omega)}.
\]  
In particular  
\[
\|u_e - u_{\mathrm{lim}}\|_{H^1(\Omega_e)}
\;\le\;
C\,\tilde{\lambda}^{-1}\,\|g\|_{H^{-1/2}(\partial\Omega)}.
\]
\end{lemma}

\textbf{correctness}: True
\begin{proof}

\begin{proof}
1. By construction (Lemma 32),  
\begin{itemize}
    \item $U^{(1)}$ solves the homogeneous Lam\'e-transmission problem in $D_e \cup \Omega_e$ with zero Neumann data on $\partial \Omega$ and with imposed jump  
    \[
    \sigma_{(\tilde{\lambda},\tilde{\mu})}(U^{(1)})N|_- - \sigma_{(\lambda,\mu)}(U^{(1)})N|_+ = -p_{\text{lim}} N  
    \]  
    on each $\partial D_e$,  
    \item $\|U^{(1)}\|_{H^1(\Omega)} \leq C \|p_{\text{lim}}\|_{L^2(D_e)}$.  
\end{itemize}

2. Define the single-corrector approximation  
\[
\tilde{u}_e = u_{\text{lim}} + \tilde{\lambda}^{-1} U^{(1)}.  
\]  
Its traction-jump on $\partial D_e$ is  
\[
J_e^{(1)} := \bigl(\sigma_{(\tilde{\lambda},\tilde{\mu})}(\tilde{u}_e)N\bigr)_- - \bigl(\sigma_{(\lambda,\mu)}(\tilde{u}_e)N\bigr)_+ = \tilde{\lambda}^{-1} \bigl[\sigma_{(\tilde{\lambda},\tilde{\mu})}(U^{(1)})N|_- - \sigma_{(\lambda,\mu)}(U^{(1)})N|_+\bigr].  
\]  
But by the defining jump of $U^{(1)}$,  
\[
\sigma_{(\tilde{\lambda},\tilde{\mu})}(U^{(1)})N|_- - \sigma_{(\lambda,\mu)}(U^{(1)})N|_+ = -p_{\text{lim}} N,  
\]  
so  
\[
J_e^{(1)} = -\tilde{\lambda}^{-1} p_{\text{lim}} N.  
\]  
Hence by Lemma 21 ($p N \in H^{-1/2}(\partial D_e)$) and Lemma 24 ($\|p_{\text{lim}}\|_{L^2} \leq C\|g\|$),  
\[
\|J_e^{(1)}\|_{H^{-1/2}(\partial D_e)} \leq \tilde{\lambda}^{-1} \|p_{\text{lim}} N\|_{H^{-1/2}} \leq C \tilde{\lambda}^{-1} \|g\|_{H^{-1/2}(\partial \Omega)}.  
\]

3. Set $w := u_e - \tilde{u}_e$ on $\Omega_e$. Then $w$ solves the homogeneous Lam\'e system in $\Omega_e$ with zero Neumann data on $\partial \Omega$ and with traction-jump $J_e^{(1)}$ on $\partial D_e$. By the uniform exterior-energy estimate (Lemmas 9 and 18),  
\[
\|w\|_{H^1(\Omega_e)} \leq C \|J_e^{(1)}\|_{H^{-1/2}(\partial D_e)} \leq C \tilde{\lambda}^{-1} \|g\|_{H^{-1/2}(\partial \Omega)}.  
\]

4. Finally, since  
\[
u_e - u_{\text{lim}} = (u_e - \tilde{u}_e) + \bigl(\tilde{u}_e - u_{\text{lim}}\bigr) = w + \tilde{\lambda}^{-1} U^{(1)},  
\]  
and $\|U^{(1)}\|_{H^1} = O(\|p_{\text{lim}}\|) = O(\|g\|)$, the same $O(\tilde{\lambda}^{-1})$--rate holds for $\|u_e - u_{\text{lim}}\|_{H^1}$.  

This completes the proof.  
\end{proof}
\end{proof}

\textbf{comment}: None
\textbf{type}: lemma
\begin{lemma}

(Sharp local $H^1$--error at rate $O(\tilde{\lambda}^{-1})$ for one inclusion---$C^{1,1}$-case)  
Let $D \subset \mathbb{R}^d$ be a bounded $C^{1,1}$ domain and fix $\tilde{\mu} > 0$. For each $\tilde{\lambda} \gg 1$ let $u_{\tilde{\lambda}}$ be the unique (modulo rigid motions) $H^1$--solution of the compressible-elastic Neumann problem  
\[
\tilde{\mu} \, \Delta u_{\tilde{\lambda}} + (\tilde{\lambda} + \tilde{\mu}) \nabla (\mathrm{div}\, u_{\tilde{\lambda}}) = 0 \quad \text{in } D,  
\]
\[
\sigma_{(\tilde{\lambda},\tilde{\mu})}(u_{\tilde{\lambda}}) \, N = t \quad \text{on } \partial D,  
\]  
with $t \in H^{-1/2}(\partial D;\mathbb{R}^d)$ annihilating all rigid motions. Let $(u_0,p_0)$ be the incompressible-Stokes Neumann solution in $D$ with the same traction $t$, i.e.  
\[
\tilde{\mu} \, \Delta u_0 + \nabla p_0 = 0, \quad \mathrm{div}\, u_0 = 0 \quad \text{in } D,  
\]  
\[
(-p_0 \, I + 2\tilde{\mu} \, D(u_0)) \, N = t \quad \text{on } \partial D.  
\]  
Then there is a constant $C$, depending only on $D$, $d$ and $\tilde{\mu}$ (but \textbf{not} on $\tilde{\lambda}$ or $t$), such that for all $\tilde{\lambda} \gg 1$  
\[
\| u_{\tilde{\lambda}} - u_0 \|_{H^1(D)} \leq C \, \tilde{\lambda}^{-1} \, \| t \|_{H^{-1/2}(\partial D)}.  
\]  
\end{lemma}

\textbf{correctness}: True
\begin{proof}

\begin{proof}
The proof rests on the first-order asymptotic expansion of $u_{\tilde{\lambda}}$ in powers of $1/\tilde{\lambda}$ (see Lemma~28) and on uniform a priori estimates for the Stokes Neumann problem.

1. (Asymptotic expansion.)  
By Lemma~28 (valid in the $C^{1,1}$-setting) we have for $\tilde{\lambda} \gg 1$ the decomposition  
\[
u_{\tilde{\lambda}} = u_0 + \tilde{\lambda}^{-1} v_1 + R_2(\tilde{\lambda}),
\]  
where the remainder $R_2(\tilde{\lambda})$ satisfies the sharp bound  
\[
\|R_2(\tilde{\lambda})\|_{H^1(D)} \leq C \tilde{\lambda}^{-2} \|t\|_{H^{-\frac{1}{2}}(\partial D)}.
\]  
Hence  
\[
\|u_{\tilde{\lambda}} - u_0\|_{H^1(D)} \leq \tilde{\lambda}^{-1} \|v_1\|_{H^1(D)} + C \tilde{\lambda}^{-2} \|t\|_{H^{-\frac{1}{2}}(\partial D)}.
\]

2. (Estimate of the corrector $v_1$.)  
By construction $v_1$ is the unique $H^1$--solution (modulo rigid motions) of the homogeneous Stokes system with Neumann data equal to the ``elastic pressure'' $p_0$, namely  
\[
\tilde{\mu} \Delta v_1 + \nabla p_1 = 0, \quad \text{div}\, v_1 = 0 \quad \text{in } D,
\]  
\[
(-p_1 I + 2\tilde{\mu} D(v_1)) N = p_0 N \quad \text{on } \partial D.
\]  
Standard Neumann-Stokes theory on $C^{1,1}$ domains yields the uniform stability estimate  
\[
\|v_1\|_{H^1(D)} + \|p_1\|_{L^2(D)} \leq C \|p_0 N\|_{H^{-\frac{1}{2}}(\partial D)}.
\]  
Moreover, by Lemma~21 ($p N$ is a continuous $H^{-\frac{1}{2}}$-trace) one has  
\[
\|p_0 N\|_{H^{-\frac{1}{2}}(\partial D)} \leq C \|p_0\|_{L^2(D)}.
\]  
Combining these gives  
\[
\|v_1\|_{H^1(D)} \leq C \|p_0\|_{L^2(D)}.
\]

3. (Pressure bound.)  
Likewise the incompressible-Stokes Neumann problem that defines $(u_0,p_0)$ satisfies the uniform a priori bound  
\[
\|u_0\|_{H^1(D)} + \|p_0\|_{L^2(D)} \leq C \|t\|_{H^{-\frac{1}{2}}(\partial D)}.
\]

4. (Conclusion.)  
Putting 1--3 together:  
\[
\begin{aligned}
\|u_{\tilde{\lambda}}-u_0\|_{H^1(D)}
&\;\le\;
\tilde{\lambda}^{-1} \|v_1\|_{H^1(D)} \;+\; C \tilde{\lambda}^{-2} \|t\|_{H^{-\frac{1}{2}}(\partial D)}\\
&\;\le\;
C \tilde{\lambda}^{-1} \|p_0\|_{L^2(D)} \;+\; C \tilde{\lambda}^{-2} \|t\|_{H^{-\frac{1}{2}}(\partial D)}\\
&\;\le\;
C' \tilde{\lambda}^{-1} \|t\|_{H^{-\frac{1}{2}}(\partial D)},
\end{aligned}
\]  
which is the desired estimate.  
\end{proof}
\end{proof}

\textcolor{Blue}{\textbf{[Vague]} \small 
The proof here just offers the conclusions but not the detailed process.}

\textbf{comment}: None
\textbf{type}: lemma
\begin{lemma}

(Global $H^1$-Error at Rate $O(\tilde{\lambda}^{-1})$)  
Under Assumptions 0--8, let  
\begin{itemize}
    \item $u_e$ be the solution of the high-contrast Lam\'e--transmission problem in $\Omega$ with background Lam\'e parameters $(\lambda,\mu)$ on $\Omega_e = \Omega \setminus D_e$ and high-contrast parameters $(\tilde{\lambda},\tilde{\mu})$ in $D_e$ ($\tilde{\mu} > 0$, $\tilde{\lambda} \gg 1$), driven by a boundary traction $g \in H^{-\frac{1}{2}}(\partial \Omega)$ orthogonal to all rigid motions.
    \item $(u_{\text{lim}}, p_{\text{lim}})$ be the incompressible-limit (Stokes--elastic) transmission pair for the same data $g$.
    \item $U^{(1)} \in H^1(\Omega)$ be the global first-order corrector characterized (modulo rigid motions) by  
    \begin{itemize}
        \item in $D_e$: $L_{\tilde{\lambda},\tilde{\mu}} U^{(1)} = 0$,  
        \item in $\Omega_e$: $L_{\lambda,\mu} U^{(1)} = 0$,  
    \end{itemize}
    with transmission-Neumann data  
    \[
        \sigma_{(\tilde{\lambda},\tilde{\mu})}(U^{(1)}) N|_{-} - \sigma_{(\lambda,\mu)}(U^{(1)}) N|_{+} = -p_{\text{lim}} N \text{ on } \partial D_e,  
    \]  
    and  
    \[
        \sigma_{(\lambda,\mu)}(U^{(1)}) N = 0 \text{ on } \partial \Omega,  
    \]  
    whose existence and bound $\|U^{(1)}\|_{H^1(\Omega)} \leq C \|g\|_{H^{-\frac{1}{2}}(\partial \Omega)}$ follow from Lemma 32.  
\end{itemize}
Define the single-corrector approximation  
\[
    \tilde{u}_e := u_{\text{lim}} + \tilde{\lambda}^{-1} U^{(1)}.  
\]  
Then the remainder $w := u_e - \tilde{u}_e$ satisfies the sharp global estimate  
\[
    \|w\|_{H^1(\Omega_e)} \leq C \tilde{\lambda}^{-1} \|g\|_{H^{-\frac{1}{2}}(\partial \Omega)},  
\]  
with $C$ independent of $\varepsilon$ and $\tilde{\lambda}$.
\end{lemma}

\textbf{correctness}: True
\begin{proof}

\begin{proof}
We split the argument into three steps, invoking the key lemmas from the background.

Step 1. Construction of $U^{(1)}$ and its $H^1$-bound.  
By Lemma 32 (Existence and $H^1$--Bound for the Global First-Order Corrector) there is a unique $U^{(1)} \in H^1(\Omega)$ (modulo rigid motions) solving the homogeneous Lam\'{e} system in $\Omega_e$ and in $D_e$ with  
\[
\sigma_{(\tilde{\lambda},\tilde{\mu})}(U^{(1)})N|_{-} - \sigma_{(\lambda,\mu)}(U^{(1)})N|_{+} = -p_{\mathrm{lim}} N \text{ on } \partial D_e,  
\]  
\[
\sigma_{(\lambda,\mu)}(U^{(1)})N = 0 \text{ on } \partial\Omega,  
\]  
and satisfying  
\[
\|U^{(1)}\|_{H^1(\Omega)} \leq C\|p_{\mathrm{lim}}\|_{L^2(D_e)}.  
\]  
By the uniform pressure-estimate (Lemma 24) we have $\|p_{\mathrm{lim}}\|_{L^2(D_e)} \leq C \|g\|_{H^{-\frac{1}{2}}(\partial\Omega)}$, and hence  
\[
\|U^{(1)}\|_{H^1(\Omega)} \leq C \|g\|_{H^{-\frac{1}{2}}(\partial\Omega)}.  
\]

Step 2. Boundary-trace improvement of the single-corrector approximation.  
Set $\tilde{u}_e = u_{\mathrm{lim}} + \tilde{\lambda}^{-1} U^{(1)}$. Then on each inclusion boundary $\partial D_e$ the boundary-trace error  
\[
\tau := (u_e - u_{\mathrm{lim}} - \tilde{\lambda}^{-1}U^{(1)})|_{\partial D_e}  
\]  
satisfies the sharp estimate  
\[
\|\tau\|_{H^{\frac{1}{2}}(\partial D_e)}  
= \|u_e|_{\partial D_e} - u_{\mathrm{lim}}|_{\partial D_e} - \tilde{\lambda}^{-1} U^{(1)}|_{\partial D_e}\|_{H^{\frac{1}{2}}(\partial D_e)}  
\leq C \tilde{\lambda}^{-1} \|g\|_{H^{-\frac{1}{2}}(\partial\Omega)},  
\]  
by Lemma 30 (Boundary-Trace Rate-Improvement via First-Order Asymptotics).  

Step 3. Interior $H^1$-estimate from boundary $H^{\frac{1}{2}}$-error.  
Define $w := u_e - \tilde{u}_e$ on $\Omega_e$. Then $w$ solves the homogeneous Lam\'{e} system in $\Omega_e$, with zero traction on $\partial\Omega$ and Dirichlet data $w|_{\partial D_e} = \tau$. By Lemma 25 (Global Exterior-Energy Error-Estimate), one has the a priori bound  
\[
\|w\|_{H^1(\Omega_e)} \leq C \|w\|_{H^{\frac{1}{2}}(\partial D_e)}  
\]  
and hence, combining with Step 2,  
\[
\|w\|_{H^1(\Omega_e)} \leq C' \tilde{\lambda}^{-1} \|g\|_{H^{-\frac{1}{2}}(\partial\Omega)}.  
\]  

This completes the proof of the $O(\tilde{\lambda}^{-1})$-estimate.
\end{proof}
\end{proof}

\textbf{comment}: None
\textbf{type}: lemma
\begin{lemma}

(Improved Global $H^1$--Error at Rate $O(\tilde{\lambda}^{-2})$ After Single--Corrector)  
Retain all the notation and hypotheses of Lemma 35, and in addition assume that each inclusion boundary $\partial D_i$ is $C^{1,1}$. Let $u_e$ be the high-contrast transmission solution and $(u_{\text{lim}}, p_{\text{lim}})$ the Stokes--elastic limit, and let $U^{(1)}$ be the global first-order corrector of Lemma 32, so that 
\[
\tilde{u}_e := u_{\text{lim}} + \tilde{\lambda}^{-1} U^{(1)}.
\]
Define the remainder $w := u_e - \tilde{u}_e$ on $\Omega_e$.  
Then there is $C > 0$ (independent of $\varepsilon$ and $\tilde{\lambda}$) such that
\[
\|w\|_{H^1(\Omega_e)} \leq C \tilde{\lambda}^{-2} \|g\|_{H^{-1/2}(\partial \Omega)}.
\]
\end{lemma}

\textbf{correctness}: True
\begin{proof}

We break the proof into three steps.

Step 1. Improved boundary-trace estimate.  
By Lemma 30 (Sharp-boundary-trace asymptotics with first-order corrector), under the $C^{1,1}$ assumption one has the boundary-trace improvement
  \[
  \|(u_\varepsilon - u_{\mathrm{lim}} - \tilde{\lambda}^{-1}U^{(1)})|_{\partial D_\varepsilon}\|_{H^{\frac{1}{2}}(\partial D_\varepsilon)} 
  \leq C_1\, \tilde{\lambda}^{-2}\, \|g\|_{H^{-\frac{1}{2}}(\partial \Omega)}.
  \]
Set $\delta u := (u_\varepsilon - \tilde{u}_\varepsilon)|_{\partial D_\varepsilon}$. Then
  \[
  \|\delta u\|_{H^{\frac{1}{2}}(\partial D_\varepsilon)} \leq C_1\, \tilde{\lambda}^{-2}\, \|g\|_{H^{-\frac{1}{2}}(\partial \Omega)}. \tag{1}
  \]

Step 2. From boundary-trace error to traction-jump error.  
We wish to bound the net traction-jump $J := 
  \sigma_{(\lambda,\mu)}(\tilde{u}_\varepsilon)N|_{+} - \sigma_{(\lambda,\mu)}(u_\varepsilon)N|_{-}$
in $H^{-\frac{1}{2}}(\partial D_\varepsilon)$. Since on each $\partial D_i$ the elasticity Neumann-to-Dirichlet map
  \[
  \Lambda_N: H^{\frac{1}{2}}(\partial D_i) \to H^{-\frac{1}{2}}(\partial D_i), \quad v \mapsto \sigma(v)N
  \]
is a bounded linear operator (by standard Neumann-trace continuity, Lemma 14), there is $C_2$ depending only on the $C^{1,1}$ geometry and background parameters such that
  \[
  \|J\|_{H^{-\frac{1}{2}}(\partial D_\varepsilon)} 
  = \|\Lambda_N(\delta u)\|_{H^{-\frac{1}{2}}} 
  \leq C_2\|\delta u\|_{H^{\frac{1}{2}}(\partial D_\varepsilon)}.
  \]
Combining with (1) gives
  \[
  \|J\|_{H^{-\frac{1}{2}}(\partial D_\varepsilon)} \leq C_2C_1\, \tilde{\lambda}^{-2}\, \|g\|_{H^{-\frac{1}{2}}(\partial \Omega)}. \tag{2}
  \]

\textcolor{Blue}{\textbf{[Vague]} \small This proof lacks an intermediate derivation process and only gives a conclusion.}

Step 3. Energy estimate in the exterior.  
The remainder $w$ solves in $\Omega_\varepsilon$ the homogeneous Lamé system with zero traction on $\partial \Omega$ and traction-jump $J$ on $\partial D_\varepsilon$. By the uniform exterior energy estimate (Lemma 18), there is $C_3$ (independent of $\varepsilon,\tilde{\lambda}$) such that
  \[
  \|w\|_{H^1(\Omega_\varepsilon)} \leq C_3\|J\|_{H^{-\frac{1}{2}}(\partial D_\varepsilon)}.
  \]
Substitute (2) into this bound to conclude
  \[
  \|w\|_{H^1(\Omega_\varepsilon)} 
  \leq C_3 \cdot (C_2C_1)\, \tilde{\lambda}^{-2}\, \|g\|_{H^{-\frac{1}{2}}(\partial \Omega)} 
  =: C\, \tilde{\lambda}^{-2}\, \|g\|_{H^{-\frac{1}{2}}(\partial \Omega)},
  \]
as claimed.
\end{proof}

\textbf{comment}: None
\textbf{type}: lemma
\begin{lemma}

\[
   \bigl\|\mathbf{u}_{\epsilon}-\mathbf{u}_{\lim}\bigr\|_{H^{1/2}(\partial D_\epsilon^i)}
   \;\le\;
   C\,\widetilde{\lambda}^{-1}\,\|g\|_{H^{-1/2}(\partial\Omega)}.
\]
\end{lemma}

\textbf{correctness}: True
\begin{proof}

We fix one inclusion \(D:=D_\epsilon^i\), which by Assumption 1 has a \(C^{1,1}\) boundary \(\partial D\).  Denote the traction‐jump on \(\partial D\) by  
\[
  J_\epsilon
  :=\bigl(\sigma_{(\widetilde\lambda,\widetilde\mu)}(\mathbf u_\epsilon)\,N\bigr)_{-}
   \;-\;
   \bigl(\sigma_{(\lambda,\mu)}(\mathbf u_{\lim})\,N\bigr)_{+}.
\]
By Lemma 17 (Estimate of the Traction‐Jump), one splits
\[
  J_\epsilon
  = p_{\lim}\,N + R_\epsilon,
\]
and there is \(C_1\) independent of \(\epsilon,\widetilde\lambda\) such that
\[
  \|\,R_\epsilon\|_{H^{-1/2}(\partial D)}
  \;\le\;
  C_1\,\|p_{\lim}\|_{L^2(D)}.
\]
On the other hand Lemma 24 (Uniform Pressure‐Estimate) gives
\[
  \|p_{\lim}\|_{L^2(D_\epsilon)} 
  \;\le\;
  C_2\,\|g\|_{H^{-1/2}(\partial\Omega)}.
\]
Hence
\begin{equation}\label{J-bound}
  \|J_\epsilon\|_{H^{-1/2}(\partial D)}
  \;\le\;
  \|p_{\lim}N\|_{H^{-1/2}(\partial D)}
  +\|R_\epsilon\|_{H^{-1/2}(\partial D)}
  \;\le\;
  C_3\,\|g\|_{H^{-1/2}(\partial\Omega)}.
\end{equation}

Next consider the difference
\[
  w := \mathbf u_{\epsilon}-\mathbf u_{\lim}
  \quad\text{on }D.
\]
Then \(w\in H^1(D)\) solves the compressible‐elastic Neumann problem
\[
  \widetilde\mu\,\Delta w
  +(\widetilde\lambda+\widetilde\mu)\,\nabla(\mathrm{div}\,w)
  =0\quad\text{in }D,
  \qquad
  \sigma_{(\widetilde\lambda,\widetilde\mu)}(w)\,N
  = J_\epsilon
  \quad\text{on }\partial D,
\]
with \(\int_{\partial D}J_\epsilon\cdot r=0\) for every rigid motion \(r\).  By Lemma 29 (Boundary‐Trace Rate–Improvement via First‐Order Asymptotics) on the \(C^{1,1}\) domain \(D\), there is \(C_4\) depending only on \(D,d,\widetilde\mu\) and the \(C^{1,1}\)‐character of \(\partial D\), such that
\[
  \|\,w\|_{H^{1/2}(\partial D)}
  \;\le\;
  C_4\,\widetilde\lambda^{-1}\,
  \|\,J_\epsilon\|_{H^{-1/2}(\partial D)}.
\]
Combining with \eqref{J-bound} yields
\[
  \|\mathbf u_{\epsilon}-\mathbf u_{\lim}\|_{H^{1/2}(\partial D)}
  =\|w\|_{H^{1/2}(\partial D)}
  \;\le\;
  C\,\widetilde\lambda^{-1}\,
  \|g\|_{H^{-1/2}(\partial\Omega)},
\]
with \(C=C_3\,C_4\) independent of \(\epsilon,\widetilde\lambda\).  Since there are finitely many inclusions, the same bound holds uniformly over each \(\partial D_\epsilon^i\).  This completes the proof.  
\end{proof}

\textcolor{Green}{\textbf{[Correct]} \small The conclusion presented herein is methodologically derived through an extension of previously established procedural frameworks.}

\textbf{comment}: None
\textbf{type}: lemma
\begin{lemma}

(Global Exterior-Energy Estimate in Terms of the Interior Dirichlet Trace)\\
Under Assumptions 1--2 (so that each perforated domain $\Omega_{\varepsilon} = \Omega \setminus D_{\varepsilon}$ is connected Lipschitz with constants independent of $\varepsilon$), let
\[
w_{\text{ext}} := u_{\varepsilon} - u_{\text{lim}} \quad \text{on } \Omega_{\varepsilon},
\]
where $u_{\varepsilon}$ is the full high-contrast Lam\'e solution and $u_{\text{lim}}$ the incompressible-limit (Stokes--elastic) solution, both driven by the same boundary traction $g$ on $\partial\Omega$. Then there exists $C > 0$, independent of $\varepsilon$ and of the high-contrast parameter $\tilde{\lambda}$, such that
\[
\|w_{\text{ext}}\|_{H^1(\Omega_{\varepsilon})} \leq C \|w_{\text{ext}}\|_{H^{1/2}(\partial D_{\varepsilon})}.
\]
Equivalently,
\[
\|u_{\varepsilon} - u_{\text{lim}}\|_{H^1(\Omega_{\varepsilon})} \leq C \|u_{\varepsilon} - u_{\text{lim}}\|_{H^{1/2}(\partial D_{\varepsilon})}.
\]
\end{lemma}

\textbf{correctness}: True
\begin{proof}

\begin{proof}
1. Problem for $w_{\text{ext}}$. By construction (see Lemma 18) $w_{\text{ext}}$ solves the homogeneous Lam\'e system in $\Omega_{\varepsilon}$ with zero Neumann (traction) data on the outer boundary $\partial\Omega$ and prescribed Dirichlet data  
\[ w_{\text{ext}}|_{\partial D_{\varepsilon}} =: \varphi \in H^{1/2}(\partial D_{\varepsilon};\mathbb{R}^d). \]  
We wish to show the $H^1$--norm of $w_{\text{ext}}$ in $\Omega_{\varepsilon}$ is controlled by $\|\varphi\|_{H^{1/2}(\partial D_{\varepsilon})}$ with a constant independent of $\varepsilon$.

2. Variational set-up. Define the closed subspace  
\[ V_0 := \{ v \in H^1(\Omega_{\varepsilon};\mathbb{R}^d) : v = 0 \text{ on } \partial D_{\varepsilon}, \, \sigma(v)N = 0 \text{ on } \partial\Omega \} \]  
and the bilinear form  
\[ a(v,w) := \int_{\Omega_{\varepsilon}} \big[2\mu\, D(v):D(w) + \lambda\, (\text{div}\, v)(\text{div}\, w)\big]\,dx. \]  
By uniform coercivity (Lemma 10) together with Korn’s inequality on perforated domains (Lemma 11), there exists $\alpha>0$ independent of $\varepsilon$ such that  
\[ a(v,v) \geq \alpha\, \|v\|^2_{H^1(\Omega_{\varepsilon})} \quad \forall v \in V_0. \]  

3. Right-inverse of the trace. Since $\Omega_{\varepsilon}$ is a bounded Lipschitz domain with uniform constants, the Dirichlet trace operator  
\[ T: H^1(\Omega_{\varepsilon};\mathbb{R}^d) \to H^{1/2}(\partial D_{\varepsilon};\mathbb{R}^d), \quad T(u) = u|_{\partial D_{\varepsilon}} \]  
is surjective and admits a bounded right-inverse  
\[ E: H^{1/2}(\partial D_{\varepsilon};\mathbb{R}^d) \to H^1(\Omega_{\varepsilon};\mathbb{R}^d), \]  
with  
\[ E(\varphi)|_{\partial D_{\varepsilon}} = \varphi, \quad E(\varphi)|_{\partial\Omega} = 0, \]  
and  
\[ \|E(\varphi)\|_{H^1(\Omega_{\varepsilon})} \leq C_{\text{tr}}\, \|\varphi\|_{H^{1/2}(\partial D_{\varepsilon})}, \]  
where $C_{\text{tr}}$ depends only on the uniform Lipschitz character of $\Omega_{\varepsilon}$ (hence independent of $\varepsilon$).  

\textcolor{Blue}{\textbf{[Vague]} \small The application of the theorem necessitates verification of its applicability conditions.}

\textcolor{Red}{\textbf{[Error]} \small The form of function space is not accurate.}

4. Decomposition of the solution. Set  
\[ \psi := E(\varphi), \quad z := w_{\text{ext}} - \psi. \]  
Then $z \in V_0$ (indeed $z$ vanishes on $\partial D_{\varepsilon}$ and has zero traction on $\partial\Omega$), and $w_{\text{ext}} = \psi + z$. In the weak form we have, for all $v \in V_0$,  
\[ a(w_{\text{ext}},v) = 0 \quad \text{(since $w_{\text{ext}}$ is the homogeneous-system solution)}, \]  
so  
\[ a(z,v) = -a(\psi,v). \]  

5. Application of coercivity. Taking $v=z$ in the last identity and using Cauchy--Schwarz plus coercivity,  
\[ \alpha\, \|z\|^2_{H^1(\Omega_{\varepsilon})} \leq |a(\psi,z)| \]  
\[ \leq C_a\, \|\psi\|_{H^1(\Omega_{\varepsilon})}\, \|z\|_{H^1(\Omega_{\varepsilon})}, \]  
where $C_a$ depends only on the Lam\'e parameters $(\lambda,\mu)$. Hence  
\[ \|z\|_{H^1(\Omega_{\varepsilon})} \leq (C_a/\alpha)\, \|\psi\|_{H^1(\Omega_{\varepsilon})} \]  
and therefore  
\[ \|w_{\text{ext}}\|_{H^1(\Omega_{\varepsilon})} \leq \|\psi\|_{H^1(\Omega_{\varepsilon})} + \|z\|_{H^1(\Omega_{\varepsilon})} \]  
\[ \leq (1 + C_a/\alpha)\, \|\psi\|_{H^1(\Omega_{\varepsilon})} \]  
\[ \leq C\, \|\varphi\|_{H^{1/2}(\partial D_{\varepsilon})}, \]  
with $C = (1 + C_a/\alpha)\, C_{\text{tr}}$ independent of $\varepsilon$ and $\tilde{\lambda}$.

6. Conclusion. Noting $\varphi = w_{\text{ext}}|_{\partial D_{\varepsilon}} = (u_{\varepsilon}-u_{\lim})|_{\partial D_{\varepsilon}}$, we arrive at the asserted estimate  
\[ \|u_{\varepsilon} - u_{\lim}\|_{H^1(\Omega_{\varepsilon})} \leq C\, \|u_{\varepsilon} - u_{\lim}\|_{H^{1/2}(\partial D_{\varepsilon})}. \]  

This completes the proof.  
\end{proof}
\end{proof}

\textcolor{Green}{\textbf{[Correct]} \small The proof here is correct deprivation.}

\textbf{comment}: None
\textbf{type}: theorem
\begin{theorem}

The problem construction is following. 
A real pair $(\lambda,\mu)$ is called an admissible Lamé parameter pair if it satisfies:
\[\mu>0\quad\text{and}\quad d\lambda+2\mu>0\] 
(1)
The static elasticity system (Lamé system) corresponding to the Lamé parameter pair $(\lambda,\mu)$ is:
\[\mathcal{L}_{\lambda,\mu}\mathbf{u}:=\mu\Delta\mathbf{u}+(\lambda+\mu)\nabla\mathrm{div}\,\mathbf{u}\] 
(2)
where the displacement field is denoted as $\mathbf{u}=(u^{1},\ldots,u^{d})$, and its divergence is:
\[\mathrm{div}\,\mathbf{u}=\sum_{i=1}^{d}\frac{\partial u^{i}}{\partial x_{i}}\]
The admissibility condition ensures the ellipticity of the Lamé operator, and physical laws guarantee that natural materials always satisfy this condition.

The Lamé operator can be expressed as $\nabla\cdot\sigma(\mathbf{u})$, where the stress tensor is:
\[\sigma(\mathbf{u}):=\lambda(\nabla\cdot\mathbf{u})I_{d}+2\mu\mathcal{D}(\mathbf{u})\] 
(3)
\[\mathcal{D}(\mathbf{u})=\frac{1}{2}(\nabla+\nabla^{T})\mathbf{u}=\frac{1}{2}(\partial_{i}u^{j}+\partial_{j}u^{i})_{ij}\] 
(4)
Here, $I_{d}$ is the identity matrix of order $d$, and $\mathcal{D}$ represents the symmetrized differential operator.

The normal derivative (boundary traction) on the boundary of a region $E$ is defined as:
\[\left.\frac{\partial\mathbf{u}}{\partial\nu_{\lambda,\mu}}\right|_{\partial E}:=\sigma(\mathbf{u})N=\lambda(\mathrm{div}\,\mathbf{u})N+2\mu\mathcal{D}(\mathbf{u})N\quad\text{on }\partial E\] 
(5)
where $N$ is the unit outward normal vector on the boundary.

We use the standard Sobolev space $H^{1}(\Omega)$ and its trace space $H^{\frac{1}{2}}(\partial E)$ on the boundary, with the dual space $H^{-\frac{1}{2}}(\partial E)$. Consider the space of rigid body motions in $\mathbb{R}^{d}$:
\[R:=\left\{\mathbf{r}=(r_{1},\ldots,r_{d})^{T}:\mathcal{D}(\mathbf{r})=0\,\,\text{in }\,\mathbb{R}^{d}\right\}\] 
(6)
The dimension of $R$ is $\frac{d(d+1)}{2}$, spanned by the following basis vectors:
\[\mathbf{e}_{1},\ldots,\mathbf{e}_{d},\,x_{j}\mathbf{e}_{i}-x_{i}\mathbf{e}_{j},\text{ for }1\leq i<j\leq d\]
where $\mathbf{e}_{i}$ denotes the standard basis vector. These basis vectors are denoted as $\mathbf{r}_{j},\,j=1,\ldots,\frac{d(d+1)}{2}$.

Define the space orthogonal to rigid body motions:
\[H_{R}^{-\frac{1}{2}}(\partial D_{\epsilon}):=\left\{\phi\in H^{-1}(\partial D_{\epsilon}):(\phi,\mathbf{r})_{(H^{\frac{1}{2}}(\partial D_{\epsilon i}),H^{-\frac{1}{2}}(\partial D_{\epsilon i}))}=0,\forall\mathbf{r}\in R\text{ and }1 \leq i\leq N\right\}\] 
(7)
Similarly, define $H_{R}^{\frac{1}{2}}(\partial D_{\epsilon})$ and boundary spaces.

Consider the following transmission problem:
\[
\begin{cases}
\mathcal{L}_{\lambda,\mu}\mathbf{u}_{\epsilon}=0 & \text{ in }\Omega \setminus\overline{D_{\epsilon}},\\
\mathcal{L}_{\widetilde{\lambda},\widetilde{\mu}}\mathbf{u}_{\epsilon}=0 & \text{ in }D_{\epsilon},\\
\mathbf{u}_{\epsilon}|_{-}=\mathbf{u}|_{+} \text{ and } \left.\frac{\partial\mathbf{u}_{\epsilon}}{\partial\nu_{(\lambda,\widetilde{\mu})}}\right|_{-}=\left.\frac{\partial\mathbf{u}_{\epsilon}}{\partial\nu_{(\lambda,\mu)}}\right|_{+} & \text{ on }\partial D_{\epsilon},\\
\left.\frac{\partial\mathbf{u}_{\epsilon}}{\partial\nu_{(\lambda,\mu)}}\right|_{\partial\Omega}=g\in H_{R}^{-\frac{1}{2}}(\partial\Omega) \quad\text{and}\quad \mathbf{u}_{\epsilon}|_{\partial\Omega}\in H_{R}^{\frac{1}{2}}(\partial\Omega).
\end{cases}
\] 
(8)

This is equivalent to:
\[
\begin{cases}
\mathcal{L}_{\lambda(x),\mu(x)}\mathbf{u}_{\epsilon}=\nabla\cdot[\lambda(x)(\nabla\cdot\mathbf{u}_{\epsilon})I_{d}+2\mu(x)D(\mathbf{u}_{\epsilon})]=0 & \text{in }\Omega,\\
\left.\frac{\partial\mathbf{u}_{\epsilon}}{\partial\nu_{\lambda,\mu}}\right|_{\partial\Omega}=g\in H_{R}^{-\frac{1}{2}}(\partial\Omega), \quad \mathbf{u}_{\epsilon}|_{\partial\Omega}\in H_{R}^{\frac{1}{2}}(\partial\Omega),
\end{cases}
\] 
(9)
where $\lambda(x)=\lambda I_{\Omega\setminus D_{\epsilon}}+\widetilde{\lambda}I_{D_{\epsilon}},\,\mu(x)=\mu I_{\Omega\setminus D_{\epsilon}}+\widetilde{\mu}I_{D_{\epsilon}}$. Suppose the solution of the equation (9) above is $\mathbf{u}_{\epsilon}$. The existence and uniqueness of the solution $\mathbf{u}_{\epsilon}$ are guaranteed by the ellipticity condition (1) and the Lax-Milgram theorem.

Finally, the homogenization needed to be solved is the following. When $\widetilde{\lambda}\rightarrow\infty$ and $\widetilde{\mu}$ is fixed, the coupled system in the \textbf{incompressible inclusion limit} is obtained:
\[
\begin{cases}
\mathcal{L}_{\lambda,\mu}\mathbf{u}_{\epsilon}=0 & \text{ in }\Omega \setminus\overline{D_{\epsilon}},\\
\mathcal{L}_{\widetilde{\mu}}(\mathbf{u}_{\epsilon},p_{\epsilon})=0 \text{ and } \mathrm{div}\,\mathbf{u}_{\epsilon}=0 & \text{ in }D_{\epsilon},\\
\mathbf{u}_{\epsilon}|_{-}=\mathbf{u}_{\epsilon}|_{+} \text{ and } \left.\frac{\partial(\mathbf{u}_{\epsilon},p_{\epsilon})}{\partial\nu_{(\infty,\widetilde{\mu})}}\right|_{-}=\left.\frac{\partial\mathbf{u}_{\epsilon}}{\partial\nu_{(\lambda,\mu)}}\right|_{+} & \text{ on }\partial D_{\epsilon},\\
\left.\frac{\partial\mathbf{u}_{\epsilon}}{\partial\nu_{(\lambda,\mu)}}\right|_{\partial\Omega}=g\in H_{R}^{-\frac{1}{2}}(\partial\Omega) \quad\text{and}\quad \mathbf{u}_{\epsilon}|_{\partial\Omega}\in H_{R}^{\frac{1}{2}}(\partial\Omega),
\end{cases}
\] 
(10)
where $\mathcal{L}_{\widetilde{\mu}}(\mathbf{u}_{\epsilon},p_{\epsilon})=\widetilde{\mu}\Delta\mathbf{u}_{\epsilon}+\nabla p_{\epsilon}$ represents the Stokes operator with viscosity coefficient $\widetilde{\mu}$, and $p_{\epsilon}$ is the pressure field. Its normal derivative is defined as $\left.\frac{\partial(\mathbf{u},p)}{\partial\nu_{(\infty,\mu)}}\right|_{-}:=pN+2\mu D(\mathbf{u})N$.

Suppose the solution of this limit problem (10) is $\mathbf{u}_{\text{lim}}$. Please prove for some constant $C$, there has
\[
\Vert \mathbf{u}_{\text{lim}}-\mathbf{u}_{\epsilon} \Vert \leq \frac{C}{\widetilde{\lambda}} \Vert g \Vert_{ H^{-\frac{1}{2}} (\partial \Omega)}.
\]
\end{theorem}

\textbf{correctness}: True
\begin{proof}

Combining the two conjectures above yields the desired global rate.  Indeed, from the first conjecture we have for each inclusion
\[
   \|w_{\rm ext}\|_{H^{1/2}(\partial D_\epsilon)}
   =\bigl\|\mathbf{u}_{\epsilon}-\mathbf{u}_{\lim}\bigr\|_{H^{1/2}(\partial D_\epsilon)}
   \;\le\;
   C_1\,\widetilde{\lambda}^{-1}\,\|g\|_{H^{-1/2}(\partial\Omega)}.
\]
Substituting into the second conjecture,
\[
   \|\mathbf{u}_{\epsilon}-\mathbf{u}_{\lim}\|_{H^1(\Omega_\epsilon)}
   \;=\;
   \|w_{\rm ext}\|_{H^1(\Omega_\epsilon)}
   \;\le\;
   C_2\,\|w_{\rm ext}\|_{H^{1/2}(\partial D_\epsilon)}
   \;\le\;
   \frac{C}{\widetilde{\lambda}}\,\|g\|_{H^{-1/2}(\partial\Omega)},
\]
where $C=C_1C_2$.  This is exactly the estimate required:
\[
   \|\mathbf{u}_{\epsilon}-\mathbf{u}_{\lim}\|_{H^1(\Omega_\epsilon)}
   \;\le\;
   \frac{C}{\widetilde{\lambda}}\,\|g\|_{H^{-1/2}(\partial\Omega)}.
\]
Thus the proof of the original statement is complete.
\end{proof}

\textbf{comment}: None

\newpage
\section{Proof of Homogenization Problem Generated by AIM Based on OpenAI o4-mini}

\textbf{type}: assumption
\textbf{content}: Consider a region \(D\subset\Omega\subset \mathbb{R}^{d}\,(d\geq 2)\), where \(\Omega\) represents the domain of an elastic material, and \(D\) is the inclusion region with high contrast.

\textbf{type}: assumption
\textbf{content}: 
1. \(\Omega\) is a bounded open domain with a connected Lipschitz boundary \(\partial\Omega\) \\
2. \(D\) is an open domain with a finite number of connected components, each having a Lipschitz boundary \(\partial D\). \\
3. \(\Omega\setminus D\) is connected and has a Lipschitz boundary \(\partial\Omega\cup\partial D\). \\
The connected components of \(D\) are denoted as \(D_{i}\), \(i=1,\ldots,N\), where \(N\) is finite.

\textbf{type}: assumption
\textbf{content}: 
Given \(\varepsilon\in(0,1)\), \(D=D_{\varepsilon}\) consists of small inclusions arranged in an \(\varepsilon\)-periodic pattern. The construction steps are as follows:  
1. Take the unit cell \(Y=(-\frac{1}{2},\frac{1}{2})^{d}\).  
2. \(\omega\subset Y\) is a simply connected open subset with a Lipschitz boundary and satisfies \(\text{dist}(\omega,\partial Y)>0\). Here, \(Y_{f}=Y\setminus\overline{\omega}\) represents the model environment at the unit scale.  
3. For a given \(\varepsilon>0\) and \(\mathbf{n}\in \mathbb{Z}^{d}\), denote \(\varepsilon(\mathbf{n}+Y)\) and \(\varepsilon(\mathbf{n}+\omega)\) as \(Y^{\mathbf{n}}_{\varepsilon}\) and \(\omega^{\mathbf{n}}_{\varepsilon}\), respectively. Define the lattice point set \(\Pi_{\varepsilon}\) as the index set satisfying \(\overline{Y^{\mathbf{n}}_{\varepsilon}}\subset\Omega\):  
\[\Pi_{\varepsilon}:=\left\{\mathbf{n}\in \mathbb{Z}^{d}:\overline{Y^{\mathbf{n}}_{\varepsilon}}\subset\Omega\right\}\]  
The inclusion region \(D=D_{\varepsilon}\) and the matrix region \(\Omega_{\varepsilon}\) are defined as:  
\[D_{\varepsilon}:=\bigcup_{\mathbf{n}\in\Pi_{\varepsilon}}\omega^{\mathbf{n}}_{\varepsilon},\quad\Omega_{\varepsilon}:=\Omega\setminus\overline{D_{\varepsilon}}\]  
For each fixed \(\varepsilon\), the number of connected components of \(D_{\varepsilon}\) is \(N=|\Pi_{\varepsilon}|\).  
4. Define \(Y_{\varepsilon}\) as the union of all \(\varepsilon\)-cells in \(\Omega\), and \(K_{\varepsilon}\) as the buffer region:  
\[K_{\varepsilon}=\Omega\setminus\left(\bigcup_{\mathbf{n}\in\Pi_{\varepsilon}}\overline{Y^{\mathbf{n}}_{\varepsilon}}\right),\quad Y_{\varepsilon}=\Omega\setminus\overline{K_{\varepsilon}}\]  
For \(D_{\varepsilon}\) constructed according to Assumption 2, it can be verified that all conditions of Assumption 1 are satisfied.

\textbf{type}: hint
\textbf{content}: 
You can use the two-scale expansion method to obtain the cell problem and subsequently utilize the cell problem to define or explore the homogenized equation.

\textbf{type}: hint
\textbf{content}: The estimate between the original solution and the limited solution has the form $C{\epsilon}^{\alpha}\Vert g \Vert_{H_{-\frac{1}{2}} (\partial \Omega)}$ for some $\alpha \in (0,1)$.
\textbf{type}: assumption
\textbf{content}: Consider a region \(D\subset\Omega\subset \mathbb{R}^{d}\,(d\geq 2)\), where \(\Omega\) represents the domain of an elastic material, and \(D\) is the inclusion region with high contrast.
\textbf{type}: assumption
\textbf{content}: 
1. \(\Omega\) is a bounded open domain with a connected Lipschitz boundary \(\partial\Omega\) \\
2. \(D\) is an open domain with a finite number of connected components, each having a Lipschitz boundary \(\partial D\). \\
3. \(\Omega\setminus D\) is connected and has a Lipschitz boundary \(\partial\Omega\cup\partial D\). \\
The connected components of \(D\) are denoted as \(D_{i}\), \(i=1,\ldots,N\), where \(N\) is finite.

\textbf{type}: assumption
\textbf{content}: 
Given \(\varepsilon\in(0,1)\), \(D=D_{\varepsilon}\) consists of small inclusions arranged in an \(\varepsilon\)-periodic pattern. The construction steps are as follows:  
1. Take the unit cell \(Y=(-\frac{1}{2},\frac{1}{2})^{d}\).  
2. \(\omega\subset Y\) is a simply connected open subset with a Lipschitz boundary and satisfies \(\text{dist}(\omega,\partial Y)>0\).  
Here, \(Y_{f}=Y\setminus\overline{\omega}\) represents the model environment at the unit scale.  
3. For a given \(\varepsilon>0\) and \(\mathbf{n}\in \mathbb{Z}^{d}\), denote \(\varepsilon(\mathbf{n}+Y)\) and \(\varepsilon(\mathbf{n}+\omega)\) as \(Y^{\mathbf{n}}_{\varepsilon}\) and \(\omega^{\mathbf{n}}_{\varepsilon}\), respectively. Define the lattice point set \(\Pi_{\varepsilon}\) as the index set satisfying \(\overline{Y^{\mathbf{n}}_{\varepsilon}}\subset\Omega\):  
\[\Pi_{\varepsilon}:=\left\{\mathbf{n}\in \mathbb{Z}^{d}:\overline{Y^{\mathbf{n}}_{\varepsilon}}\subset\Omega\right\}\]  
The inclusion region \(D=D_{\varepsilon}\) and the matrix region \(\Omega_{\varepsilon}\) are defined as:  
\[D_{\varepsilon}:=\bigcup_{\mathbf{n}\in\Pi_{\varepsilon}}\omega^{\mathbf{n}}_{\varepsilon},\quad\Omega_{\varepsilon}:=\Omega\setminus\overline{D_{\varepsilon}}\]  
For each fixed \(\varepsilon\), the number of connected components of \(D_{\varepsilon}\) is \(N=|\Pi_{\varepsilon}|\).  
4. Define \(Y_{\varepsilon}\) as the union of all \(\varepsilon\)-cells in \(\Omega\), and \(K_{\varepsilon}\) as the buffer region:  
\[K_{\varepsilon}=\Omega\setminus\left(\bigcup_{\mathbf{n}\in\Pi_{\varepsilon}}\overline{Y^{\mathbf{n}}_{\varepsilon}}\right),\quad Y_{\varepsilon}=\Omega\setminus\overline{K_{\varepsilon}}\]  
For \(D_{\varepsilon}\) constructed according to Assumption 2, it can be verified that all conditions of Assumption 1 are satisfied.

\textbf{type}: hint
\textbf{content}: You can use the two-scale expansion method to get the cell problem and use the cell problem to define or explore the homogenized equation.
\textbf{type}: hint
\textbf{content}: The estimate between the original solution and the limited solution has the form $C\epsilon^{\alpha}\Vert g \Vert_{H_{-\frac{1}{2}} (\partial \Omega)}$ for some $\alpha \in (0,1)$.
\textbf{type}: lemma
\begin{lemma}

\begin{conjecture}
(Uniform Korn Inequality on the periodically perforated domain)  
Let $\Omega \subset \mathbb{R}^d$ ($d \geq 2$) be a bounded Lipschitz domain. Let $Y = \left(-\frac{1}{2}, \frac{1}{2}\right)^d$ and $\omega \subset Y$ be a simply-connected open set with Lipschitz boundary, satisfying $\operatorname{dist}(\omega, \partial Y) = \delta > 0$. For $\varepsilon \in (0,1)$, form the $\varepsilon$-periodic perforation  
\[
D_\varepsilon = \bigcup_{k \in \mathbb{Z}^d, \, \varepsilon(k+Y) \subset \Omega} \varepsilon(k+\omega),
\]  
and set $\Omega_\varepsilon = \Omega \setminus D_\varepsilon$. Then there exists a constant $C$, depending only on $\Omega$, $Y$, and $\omega$ (but not on $\varepsilon$), such that for every $u \in H^1(\Omega_\varepsilon; \mathbb{R}^d)$,  
\[
\|u\|_{H^1(\Omega_\varepsilon)} \leq C \left( \|u\|_{L^2(\Omega_\varepsilon)} + \|\operatorname{sym} \nabla u\|_{L^2(\Omega_\varepsilon)} \right),
\]  
where $\operatorname{sym} \nabla u = \frac{1}{2}(\nabla u + \nabla u^T)$.  
\end{conjecture}
\end{lemma}

\textbf{correctness}: True
\begin{proof}

\begin{proof}
We shall prove the estimate by a covering--partition-of-unity argument and the fact that each perforated cell $\varepsilon Y_f$ (with $Y_f:=Y\setminus\overline{\omega}$) is a fixed Lipschitz domain up to scaling.

\textcolor{Green}{\textbf{[Correct]} \small The agent designs a proof idea.}

1. (Cellwise Korn with uniform constant.)
Since $Y_f=Y\setminus\overline{\omega}$ is a bounded Lipschitz set, the classical Korn inequality on $Y_f$ gives
\begin{equation*}
\text{for all } v\in H^1(Y_f;\mathbb{R}^d),
\quad \|v\|_{H^1(Y_f)} \leq C_0 \left(\|v\|_{L^2(Y_f)} + \|\operatorname{sym} \nabla v\|_{L^2(Y_f)}\right),
\end{equation*}
with $C_0$ depending only on $Y_f$. By the scaling $x=\varepsilon y$ one deduces that on each $\varepsilon$-cell $\varepsilon(Y_f+k)$
\begin{equation*}
\text{for all } v\in H^1(\varepsilon(Y_f+k)),
\quad \|v\|_{H^1(\varepsilon(Y_f+k))} \leq C_0 \left(\|v\|_{L^2(\varepsilon(Y_f+k))} + \|\operatorname{sym} \nabla v\|_{L^2(\varepsilon(Y_f+k))}\right).
\end{equation*}

\textcolor{Blue}{\textbf{[Vague]} \small Here the agent considers the Korn inequality but the derivation process is not detailed enough.}

2. (Partition of unity subordinate to the periodic tiling.)
Let $\{\psi_k\}_{k\in K_\varepsilon}$ be a smooth partition of unity on $\Omega$:
\begin{itemize}
\item each $\psi_k\in C_c^\infty(\varepsilon(k+Y))$,
\item $0\leq\psi_k\leq 1$, $\sum_k\psi_k\equiv 1$ on $\Omega$ except in a boundary strip of width $\leq 2\varepsilon$,
\item and $|\nabla \psi_k|\leq C/\varepsilon$.
\end{itemize}
We also choose a finite set of boundary functions $\{\psi_b\}$ supported in the $\varepsilon$-neighborhood of $\partial\Omega$ so that $\sum_k\psi_k+\sum_b\psi_b\equiv 1$ on $\Omega$. By standard Korn--Poincaré on the fixed boundary patches (of size $O(\varepsilon)$), one shows likewise
\begin{equation*}
\text{for each boundary patch with cut-off } \psi_b,
\quad \|\psi_b u\|_{H^1} \leq C \left(\|u\|_{L^2(\operatorname{supp} \psi_b)} + \|\operatorname{sym} \nabla u\|_{L^2(\operatorname{supp} \psi_b)}\right).
\end{equation*}

3. (Summation and absorbing the gradient of cut-off.)
Write $u=\sum_k u_k+\sum_b u_b$ with $u_k:=\psi_k u$, $u_b:=\psi_b u$.
On each interior cell we apply the scaled Korn of Step 1:
\begin{align*}
\|u_k\|_{H^1(\varepsilon(Y_f+k))}
&\leq C_0\left[\|\psi_k u\|_{L^2} + \|\operatorname{sym} \nabla (\psi_k u)\|_{L^2}\right] \\
&\leq C_0\left[\|u\|_{L^2(\operatorname{supp} \psi_k)} + \|\psi_k \operatorname{sym} \nabla u\|_{L^2} + \|u\otimes \nabla \psi_k\|_{L^2}\right].
\end{align*}
Since $|\nabla \psi_k|\leq C/\varepsilon$ and $\operatorname{supp} \psi_k\subset \varepsilon(k+Y)$, we have
\begin{equation*}
\|u\otimes \nabla \psi_k\|_{L^2} \leq (C/\varepsilon) \|u\|_{L^2(\operatorname{supp} \psi_k)}.
\end{equation*}
By Young's inequality and the Poincaré-type estimate $\|u\|_{L^2}/\varepsilon \leq C \left(\|u\|_{L^2}+\|\operatorname{sym} \nabla u\|_{L^2}\right)$,
one absorbs the $\nabla \psi_k$-term into the right-hand side with only a constant depending on $C_0$, $C$, and the ratio $\delta/1$.
Exactly the same argument applies to each boundary patch $\psi_b$ using the Korn--Poincaré on a fixed Lipschitz strip.

\textcolor{Blue}{\textbf{[Vague]} \small The process here should be expressed in combination with mathematical formulas. }

4. (Conclusion.)
Summing over all $k$ and $b$, since the overlap of the supports is uniformly bounded, we obtain
\begin{equation*}
\|u\|_{H^1(\Omega_\varepsilon)} \leq C \left(\|u\|_{L^2(\Omega_\varepsilon)} + \|\operatorname{sym} \nabla u\|_{L^2(\Omega_\varepsilon)}\right),
\end{equation*}
with $C$ depending only on $\Omega$, $Y$, $\omega$ (through $\delta$) and not on $\varepsilon$. This completes the proof.
\end{proof}
\end{proof}

\textbf{comment}: None
\textbf{type}: lemma
\begin{lemma}

\begin{conjecture}
(Two-Scale Compactness in Periodically Perforated Domains)  
Let $\Omega \subset \mathbb{R}^d$ ($d \geq 2$) be a bounded Lipschitz domain. Let $Y = \left(-\frac{1}{2}, \frac{1}{2}\right)^d$ and $\omega \subset Y$ a simply-connected open set with Lipschitz boundary satisfying $\operatorname{dist}(\omega, \partial Y) > 0$, and set $Y_f = Y \setminus \overline{\omega}$ (so $Y_f$ is connected with Lipschitz boundary). For $\varepsilon > 0$ define the $\varepsilon$--periodic perforation  
  \[ D_\varepsilon = \bigcup_{k \in \mathbb{Z}^d:\, \varepsilon(k + Y) \subset \Omega} \varepsilon(k + \omega), \quad \Omega_\varepsilon = \Omega \setminus \overline{D_\varepsilon}. \]  
Suppose  
  \[ \{u_\varepsilon\} \subset H^1_0(\Omega_\varepsilon; \mathbb{R}^d), \quad \sup_{\varepsilon > 0} \|u_\varepsilon\|_{H^1(\Omega_\varepsilon)} < \infty. \]  
Then there exist  
  \begin{itemize}
    \item $u_0 \in H^1_0(\Omega; \mathbb{R}^d)$,  
    \item $u_1 \in L^2(\Omega; H^1_{\mathrm{per}}(Y_f)/\mathbb{R}^d)$,  
  \end{itemize}
such that, up to a subsequence,  
  \begin{enumerate}
    \item[(i)] $u_\varepsilon \to u_0$ strongly in $L^2(\Omega; \mathbb{R}^d)$,  
    \item[(ii)] $\nabla u_\varepsilon$ two-scale converges to $\nabla_x u_0(x) + \nabla_y u_1(x, y)$ in $L^2(\Omega \times Y_f; \mathbb{R}^{d \times d})$.  
  \end{enumerate}
\end{conjecture}
\end{lemma}

\textbf{correctness}: True
\begin{proof}

\begin{proof}
We give a proof by the periodic-unfolding method, which avoids delicate global extensions and directly yields the corrector structure.

1.  Definition and basic bounds.  
   Recall the unfolding operator  
     \[ T_\varepsilon: L^2(\Omega_\varepsilon) \to L^2(\Omega \times Y_f), \]  
   which on each cell $\Omega_\varepsilon \cap \varepsilon(k + Y_f)$ is defined by  
     \[ (T_\varepsilon v)(x, y) = v(\varepsilon k + \varepsilon y), \quad x \in \varepsilon(k + Y), \quad y \in Y_f, \]  
   zero elsewhere. Standard properties (see e.g. Cioranescu--Donato) give constants $C$ independent of $\varepsilon$ such that for all $v \in H^1(\Omega_\varepsilon)$:  
     \[ \|T_\varepsilon v\|_{L^2(\Omega \times Y_f)} = \|v\|_{L^2(\Omega_\varepsilon)}, \]  
     \[ \|\nabla_x T_\varepsilon v\|_{L^2(\Omega \times Y_f)} + \|\nabla_y T_\varepsilon v\|_{L^2(\Omega \times Y_f)}  
       \leq C \|\nabla v\|_{L^2(\Omega_\varepsilon)}. \]  
   Applying this to $u_\varepsilon$ and using the uniform $H^1(\Omega_\varepsilon)$--bound, we deduce  
     \[ \{T_\varepsilon u_\varepsilon\} \text{ is bounded in } H^1(\Omega; L^2(Y_f)) \cap L^2(\Omega; H^1(Y_f)), \]  
   hence also bounded in $L^2(\Omega \times Y_f)$.

\textcolor{Green}{\textbf{[Correct]} \small Here, the agent constructs a natural extension and the continuity control of the norm.}

2.  Extraction of limits.  
   By the reflexivity of the three spaces, there exist, along a subsequence,  
     \[ U^*(x, y) \in L^2(\Omega; H^1(Y_f)), \]  
     \[ W(x, y) \in L^2(\Omega \times Y_f; \mathbb{R}^{d \times d}) \]  
   such that as $\varepsilon \to 0$:  
     \[ T_\varepsilon u_\varepsilon \rightharpoonup U^* \text{ weakly in } L^2(\Omega; H^1(Y_f)), \]  
     \[ \nabla_x T_\varepsilon u_\varepsilon \rightharpoonup \nabla_x U^*, \quad  
     \nabla_y T_\varepsilon u_\varepsilon \rightharpoonup W \text{ weakly in } L^2(\Omega \times Y_f). \]  

\textcolor{Green}{\textbf{[Correct]} \small The accurate convergence of the function spaces.}

3.  Independence of $U^*$ on $y$ and strong $L^2$-convergence.  
   Because $\|\nabla_y T_\varepsilon u_\varepsilon\|_{L^2(\Omega \times Y_f)}$ remains bounded but multiplied by $\varepsilon$ vanishes in the original variables, one shows by the unfolding-to-two-scale characterization that $\partial_{y_i} U^* = 0$ in the distributional sense. Hence $U^*(x, y) =: u_0(x)$ is independent of $y$, and $u_0 \in H^1(\Omega)$. Moreover, the standard compact embedding $H^1 \to L^2$ in $x$ implies  
     \[ T_\varepsilon u_\varepsilon \to u_0 \text{ strongly in } L^2(\Omega \times Y_f), \]  
   which by unfolding-to-two-scale equivalence yields  
     \[ u_\varepsilon \to u_0 \text{ strongly in } L^2(\Omega). \]  

\textcolor{Blue}{\textbf{[Vague]} \small The process of the two-scale expansion here is not detailed enough. The proof of the convergence of the function lacks details.}

4.  Identification of the oscillating part as a $y$-gradient.  
   We set  
     \[ W(x, y) = \lim_{\varepsilon \to 0} \nabla_y T_\varepsilon u_\varepsilon \text{ in } L^2(\Omega \times Y_f). \]  
   On one hand, by commutation of partial derivatives and the smoothness of the unfolding map, one shows in distributions on $Y_f$ that  
     \[ \partial_{y_j} W_i = \partial_{y_i} W_j. \]  
   Since $Y_f$ is simply-connected with Lipschitz boundary, the Poincaré (or de~Rham) lemma on $Y_f$ implies the existence of $u_1(x, \cdot) \in H^1_{\mathrm{per}}(Y_f)/\mathbb{R}^d$ such that  
     \[ W(x, y) = \nabla_y u_1(x, y), \]  
   and we normalize by $\langle u_1(x, \cdot) \rangle_{Y_f} = 0$.
   
\textcolor{Red}{\textbf{[Error]} \small Here the process is lack of deprivation details. And the condition of Poincaré or de~Rham lemma should be verified.}

5.  Passage back to two-scale convergence.  
   By the equivalence between periodic unfolding and two-scale convergence, one concludes that, up to the same subsequence,  
     \[ \nabla u_\varepsilon \text{ two-scale converges to } \nabla_x u_0(x) + \nabla_y u_1(x, y) \]  
   in $L^2(\Omega \times Y_f; \mathbb{R}^{d \times d})$.  
   All claimed properties follow.
\end{proof}
\end{proof}

\textbf{comment}: None
\textbf{type}: lemma
\begin{lemma}

\begin{conjecture}
(Identification of the Homogenized Elastic--Stokes Limit, rigorous form)  
Let $\Omega\subset\mathbb{R}^d$ ($d\geq2$) be a bounded Lipschitz domain and for each $\varepsilon>0$ let  
\quad $D_\varepsilon\subset\Omega$  
be an $\varepsilon$--periodic array of simply-connected, incompressible inclusions with Lipschitz boundary, satisfying the usual separation-of-scales assumptions. Set $\Omega_\varepsilon=\Omega\setminus\overline{D_\varepsilon}$. Fix an elastic stiffness tensor $C\colon\mathbb{R}^{d\times d}_{\mathrm{sym}}\to\mathbb{R}^{d\times d}_{\mathrm{sym}}$ (uniformly positive definite) on the matrix and a viscosity $\mu>0$ in the inclusions. Consider the transmission problem: find  
\quad $(u_\varepsilon,p_\varepsilon) \in H^1(\Omega_\varepsilon;\mathbb{R}^d) \times L^2_0(D_\varepsilon)$  
such that  
\begin{enumerate}
    \item[(i)] In the elastic matrix $\Omega_\varepsilon$:  
           $-\mathrm{div}\,[C\,D(u_\varepsilon)] = 0$,  
    \item[(ii)] In the Stokes inclusions $D_\varepsilon$:  
           $-\mu\,\Delta u_\varepsilon + \nabla p_\varepsilon = 0$, \quad $\mathrm{div}\,u_\varepsilon=0$,  
    \item[(iii)] On each $\partial D_\varepsilon$: continuity of $u_\varepsilon$ and of normal stress,  
    \item[(iv)] On $\partial\Omega$: $C\,D(u_\varepsilon)\cdot n = g \in H_R^{-1/2}(\partial\Omega)$, plus a rigid-motion normalization.  
\end{enumerate}

Then as $\varepsilon\to0$ one has, up to a subsequence,  
\quad $u_\varepsilon \rightharpoonup u_0$ in $H^1(\Omega)$, \quad $u_\varepsilon\to u_0$ in $L^2(\Omega)$,  
\quad $p_\varepsilon \stackrel{\text{two-scale}}{\to} p_1(x,y)$ in $\Omega\times Y_i$,  
where $u_0\in H^1_R(\Omega;\mathbb{R}^d)$ is the unique solution of the homogenized Lam\'e system  
\quad $-\mathrm{div}_x[C^{\mathrm{hom}}\,D_x(u_0)] = 0$ in $\Omega$, \quad $C^{\mathrm{hom}}\,D_x(u_0)\cdot n = g$ on $\partial\Omega$,  
and the effective fourth-order tensor $C^{\mathrm{hom}}$ is given by the following periodic cell transmission problem: for each fixed symmetric $E\in\mathbb{R}^{d\times d}_{\mathrm{sym}}$ find  
\quad $(\chi_e^E,\chi_i^E,\pi^E)\in H^1_{\mathrm{per}}(Y_e;\mathbb{R}^d)\times H^1_{\mathrm{per}}(Y_i;\mathbb{R}^d)\times L^2_{\mathrm{per}}(Y_i)/\mathbb{R}$  
solving  
\begin{enumerate}
    \item[(1)] In the elastic cell $Y_e$:  
           $-\mathrm{div}_y[C\,(E + D_y\,\chi_e^E)] = 0$, \quad $\mathrm{div}_y(Ey+\chi_e^E)=0$,  
    \item[(2)] In the fluid cell $Y_i$:  
           $-\mu\,\Delta_y\,\chi_i^E + \nabla_y\,\pi^E = 0$, \quad $\mathrm{div}_y(Ey+\chi_i^E)=0$,  
    \item[(3)] On the material interface $\Gamma=\partial Y_i$:  
           $\chi_e^E = \chi_i^E$,  
           $[C\,(E+D_y\,\chi_e^E)\cdot n] = [2\mu\,D_y\,\chi_i^E - \pi^E\,I]\cdot n$,  
    \item[(4)] $(Ey+\chi^E)$ is $Y$--periodic, with zero-mean normalization on $\chi_e^E$ and $\chi_i^E$.  
\end{enumerate}

Then  
\quad $C^{\mathrm{hom}}\,E : E$  
\quad $= \int_{Y_e}(E+D_y\,\chi_e^E):C(E+D_y\,\chi_e^E)\,\mathrm{d}y$  
\quad $\quad + \int_{Y_i}2\mu\,D_y\,\chi_i^E : D_y\,\chi_i^E\,\mathrm{d}y$.  

\end{conjecture}
\end{lemma}

\textcolor{Red}{\textbf{[Error]} \small Here the cell equation may not be accurate.}

\textbf{correctness}: True
\begin{proof}

\begin{proof}
We outline a fully rigorous justification in six steps. Precise references to standard lemmas are given in parentheses.

1. Variational formulation and uniform a-priori bounds  
Define  
\[ V_\varepsilon = \{v \in H^1(\Omega_\varepsilon; \mathbb{R}^d): \text{$v$ satisfies the rigid--motion normalization on } \partial \Omega\}, \]  
\[ Q_\varepsilon = \{q \in L^2(D_\varepsilon): \textstyle \int_{D_\varepsilon} q = 0\}. \]  

\textcolor{Red}{\textbf{[Error]} \small The test function space may not be accurate since the divergence of the function is not concluded in this conclusion.}

Set the bilinear form  
\[ a_\varepsilon((u,p),(v,q))  
= \int_{\Omega_\varepsilon} C D(u):D(v)  
+ \int_{D_\varepsilon} 2\mu D(u):D(v) - \int_{D_\varepsilon} p \, \mathrm{div}\, v - \int_{D_\varepsilon} q \, \mathrm{div}\, u \]  
and the linear form $\ell(v) = \langle g, v \rangle_{H^{-1/2},H^{1/2}}$.  
By the uniform Korn--Poincar\'e inequality on $\Omega_\varepsilon$ (Lemma A.1) and the Stokes inf--sup estimate on $D_\varepsilon$ (Lemma A.2), $a_\varepsilon$ is coercive--inf--sup uniformly in $\varepsilon$. Hence there is a unique solution $(u_\varepsilon, p_\varepsilon) \in V_\varepsilon \times Q_\varepsilon$ of  
\[ a_\varepsilon((u_\varepsilon, p_\varepsilon), (v,q)) = \ell(v) \quad \forall (v,q) \in V_\varepsilon \times Q_\varepsilon, \]  
and one obtains the uniform bound  
\[ \|u_\varepsilon\|_{H^1(\Omega_\varepsilon)} + \|p_\varepsilon\|_{L^2(D_\varepsilon)} \leq C \|g\|_{H_R^{-1/2}(\partial \Omega)}. \]  

\textcolor{Blue}{\textbf{[Vague]} \small This conclusion needs more detailed process.}

2. Two-scale compactness  
Extend $u_\varepsilon$ by zero into $D_\varepsilon$ and $p_\varepsilon$ by zero into $\Omega_\varepsilon$; still denote the extensions by $\hat{u}_\varepsilon \in H^1(\Omega)$ and $\dot{p}_\varepsilon \in L^2(\Omega)$. From the uniform bound and the periodic unfolding or classical two-scale compactness theorems (Lemma A.3, A.4) we extract a subsequence and limits  
\[ u_0 \in H^1(\Omega; \mathbb{R}^d), \quad u_1 \in L^2(\Omega; H^1_{\mathrm{per}}(Y)^d), \quad p_1 \in L^2(\Omega \times Y_i) \]  
such that  
\[ \hat{u}_\varepsilon \rightharpoonup u_0 \text{ in } H^1(\Omega), \]  
\[ \hat{u}_\varepsilon \to u_0 \text{ in } L^2(\Omega), \]  
\[ \nabla \hat{u}_\varepsilon \overset{\text{two-scale}}{\to} \nabla_x u_0(x) + \nabla_y u_1(x,y), \]  
\[ \dot{p}_\varepsilon \overset{\text{two-scale}}{\to} p_1(x,y) \text{ in } \Omega \times Y_i. \]  
One further shows $u_1(x,\cdot) = 0$ in $Y_e$ and $p_1 = 0$ in $Y_e$ by uniqueness of two-scale limits on each subphase.

\textcolor{Blue}{\textbf{[Vague]} \small The convergence especially the two-scale convergence needs more process to be proved.}

3. Identification of the periodic cell problem  
Take test-functions of the form  
\[ v_\varepsilon(x) = \varphi(x) + \varepsilon \psi(x, x/\varepsilon), \quad q_\varepsilon(x) = \varepsilon \theta(x, x/\varepsilon), \]  
with $\varphi \in C^\infty_c(\Omega; \mathbb{R}^d)$, $\psi \in C^\infty_c(\Omega; C^\infty_{\mathrm{per}}(Y)^d)$, $\theta \in C^\infty_c(\Omega; C^\infty_{\mathrm{per}}(Y_i))$, satisfying $\mathrm{div}_y \psi = 0$ in $Y_i$ and $\psi = 0$ in $Y_e$. Inserting $(v_\varepsilon, q_\varepsilon)$ into the variational equation and passing to the limit by two-scale convergence (unfolding arguments in Lemma A.4) yields for a.e. $x \in \Omega$ the local transmission problem in $Y$ as stated in the conjecture. This uniquely characterizes $(\chi_e^E, \chi_i^E, \pi^E)$ for $E = D_x u_0(x)$.

4. Homogenized tensor and macroscopic equation  
By choosing in the cell system the test-function $\theta = \chi^E$, one derives the energy representation  
\[ C^{\mathrm{hom}} E:E  
= \int_{Y_e} (E + D_y \chi_e^E):C(E + D_y \chi_e^E) \, dy  
+ \int_{Y_i} 2\mu D_y \chi_i^E : D_y \chi_i^E \, dy. \]  
On the other hand, testing the original problem against $\varphi(x)$ alone, passing to the limit shows  
\[ \int_\Omega C^{\mathrm{hom}} D_x u_0 : D_x \varphi = \langle g, \varphi \rangle \quad \forall \varphi \in H^1_R(\Omega; \mathbb{R}^d), \]  
hence $u_0$ solves  
\[ -\mathrm{div}_x [C^{\mathrm{hom}} D_x u_0] = 0 \text{ in } \Omega, \]  
\[ C^{\mathrm{hom}} D_x u_0 \cdot n = g \text{ on } \partial \Omega. \]

\textcolor{Blue}{\textbf{[Vague]} \small Here the agent gets the homogenization equation. But whether this conclusion is correct or not, more verification processes are needed.}

5. Boundary-layer estimates  
By constructing $\varepsilon$--tubular cut-off functions near $\partial D_\varepsilon$ and $\partial \Omega$ (supported in $O(\varepsilon)$--neighborhoods) and using the uniform $H^1$--bound together with trace and Poincar\'e inequalities, one shows that all interface and boundary-layer remainders vanish as $o(1)$ when $\varepsilon \to 0$ (see Lemma A.5 for details).

\textcolor{Blue}{\textbf{[Vague]} \small The conditions for using the theorem need to be verified.}

6. Error estimate and strong convergence  
Define the two-scale corrector  
\[ u_\varepsilon^{\mathrm{app}}(x) = u_0(x) + \varepsilon \chi^{D_x u_0(x)}(x/\varepsilon) \text{ in } \Omega_\varepsilon, \]  
extended in each $D_\varepsilon$ by $Ey + \chi_i^E(y)$. A C\'ea-lemma--type argument in the perforated domain (again relying on uniform Korn and Stokes inf--sup, Lemma A.1--A.2) yields for some $\alpha \in (0,1)$  
\[ \|u_\varepsilon - u_\varepsilon^{\mathrm{app}}\|_{H^1(\Omega_\varepsilon)} \leq C \varepsilon^\alpha \|g\|_{H_R^{-1/2}(\partial \Omega)}. \]  
From this one deduces $u_\varepsilon \to u_0$ strongly in $L^2(\Omega)$ (and even in $H^1(\Omega_\varepsilon)$ up to the corrector).  

This completes a fully rigorous derivation of the homogenized elastic limit and the explicit formula for $C^{\mathrm{hom}}$.  

\textcolor{Red}{\textbf{[Error]} \small The conditions for using the theorem need to be verified for satisfaction.}

\end{proof}
\end{proof}

\textbf{comment}: None
\textbf{type}: lemma
\begin{lemma}

\begin{conjecture}
(\text{$\varepsilon$--Rate of Convergence in $H^1$--Norm, precise version}) \\
Let $\Omega \subset \mathbb{R}^d$ ($d \geq 2$) be a bounded $C^{1,1}$ domain. Let $D_\varepsilon \subset \Omega$ be the standard $\varepsilon$--periodic perforation generated by a reference inclusion $\omega \subset Y = (-\frac{1}{2}, \frac{1}{2})^d$ with $\operatorname{dist}(\omega, \partial Y) > 0$, and set $\Omega_\varepsilon = \Omega \setminus \overline{D_\varepsilon}$. In $\Omega_\varepsilon$ solve the transmission problem
\begin{align*}
-\operatorname{div} [C D(u_\varepsilon)] &= 0 \quad \text{in } \Omega_\varepsilon, \\
-\mu \Delta u_\varepsilon + \nabla p_\varepsilon &= 0, \quad \operatorname{div} u_\varepsilon = 0 \quad \text{in } D_\varepsilon,
\end{align*}
with continuity of displacement and normal stress across $\partial D_\varepsilon$, and Neumann data $C D(u_\varepsilon) \cdot n = g \in H_R^{-\frac{1}{2}}(\partial \Omega)$ on $\partial \Omega$, plus the rigid-motion normalization $\int_{\Omega_\varepsilon} u_\varepsilon \, dx = 0$. Let $u_0 \in H^1_R(\Omega)$ be the unique solution of the homogenized Lam\'e system
\begin{align*}
-\operatorname{div} [C^{\text{hom}} D(u_0)] &= 0 \quad \text{in } \Omega, \quad C^{\text{hom}} D(u_0) \cdot n = g \quad \text{on } \partial \Omega,
\end{align*}
and assume in addition that $u_0 \in H^2(\Omega)$ with the a priori bound
\[
\|u_0\|_{H^2(\Omega)} \leq C \|g\|_{H_R^{-\frac{1}{2}}(\partial \Omega)}.
\]
For each symmetric $E \in \mathbb{R}^{d \times d}$ let $(\chi_e^E, \chi_i^E, \pi^E)$ be the unique periodic cell-transmission corrector from Lemma~3, and assume the map $E \mapsto \chi^E$ enjoys the uniform $C^1$--dependence property
\[
\sup_{|E| \leq M} \|\partial_E \chi^E\|_{H^1_{\text{per}}(Y_e \cup Y_i)} \leq C(M).
\]
Define the two-scale approximation on $\Omega_\varepsilon$ by
\[
U_\varepsilon(x) = u_0(x) + \varepsilon \chi^{D_x u_0(x)}\left(\frac{x}{\varepsilon}\right),
\]
extended by zero in each inclusion. Then there exist $\alpha \in (0,1)$ and $C > 0$, independent of $\varepsilon$ and $g$, such that for all sufficiently small $\varepsilon > 0$,
\[
\|u_\varepsilon - U_\varepsilon\|_{H^1(\Omega_\varepsilon)} \leq C \varepsilon^\alpha \|g\|_{H_R^{-\frac{1}{2}}(\partial \Omega)}.
\]
\end{conjecture}
\end{lemma}

\textbf{correctness}: True
\begin{proof}

We split the argument into four parts: (I) preliminaries and parametric regularity; (II) definition of the two-scale ansatz and residual decomposition; (III) estimate of the bulk and boundary-layer consistency errors; (IV) energy estimate and conclusion.

\textcolor{Green}{\textbf{[Correct]} \small Actually, this decomposition of the problem here is reasonable and feasible. }

I. Preliminaries and parametric regularity  
1. By standard Neumann-problem regularity on a $C^{1,1}$ domain (cf.\,Grisvard), the homogenized solution $u_0$ satisfies  
\quad $\|u_0\|_{H^2(\Omega)} \leq C \|g\|_{H_R^{-\frac{1}{2}}(\partial\Omega)}$.  

\textcolor{Red}{\textbf{[Error]} \small Here, the regularity cannot be directly applied to this problem, and more proof details are required.}

2. From Lemma\,3 (the periodic cell-transmission problem) and the uniform positive-definiteness of $C$ and $\mu$, one proves by the implicit-function theorem (or by differentiating the variational formulation in the matrix $E$) that the mapping  
\quad $E \in \mathrm{Sym}(\mathbb{R}^{d \times d}) \mapsto \chi^E \in H^1_{\mathrm{per}}(Y_e \cup Y_i;\mathbb{R}^d)$  
is $C^1$ on bounded sets, with a uniform bound on $\partial_E \chi^E$. In particular, for any smooth strain field $E(x) = D_x u_0(x)$, one has by the chain rule  
\quad $\nabla_x \chi^{E(x)}(y) = (\partial_E \chi^E)(y)[D_x^2 u_0(x)]$  
and hence  
\quad $\|\nabla_x \chi^{D_x u_0(\cdot)}(\cdot/\varepsilon)\|_{L^2(\Omega_\varepsilon)} \leq C \|u_0\|_{H^2(\Omega)}$.  

II. Two-scale ansatz and residual decomposition  
3. Define the global approximation  
\quad $U_\varepsilon(x) = u_0(x) + \varepsilon \chi^{D_x u_0(x)}(x/\varepsilon)$ \quad for $x \in \Omega_\varepsilon$,  
and extend $U_\varepsilon$ by zero inside each $D_\varepsilon$. By the $H^1$-trace lemma and the fact $\chi^E$ has zero average in the matrix cell $Y_e$, one checks $U_\varepsilon \in H^1(\Omega_\varepsilon;\mathbb{R}^d)$.  

\textcolor{Blue}{\textbf{[Vague]} \small More verification processes are needed. And the results above are not entirely correct, but the approach is right.}

4. Denote the error $w_\varepsilon = u_\varepsilon - U_\varepsilon$. Testing the variational formulation of $u_\varepsilon$ against any $v \in H^1(\Omega_\varepsilon)$ and subtracting the corresponding form for $U_\varepsilon$ yields the residual functional  
\quad $R_\varepsilon(v) = \langle g, v \rangle_{\partial\Omega} - a_\varepsilon(U_\varepsilon, v)$,  
so that  
\quad $a_\varepsilon(w_\varepsilon, v) = R_\varepsilon(v)$ \quad for all $v \in H^1(\Omega_\varepsilon)$.  

III. Consistency error estimates  
5. Decompose $R_\varepsilon = R_\varepsilon^{\mathrm{bulk}} + R_\varepsilon^{\mathrm{bl}}$.  

5.1 Bulk term. A classical two-scale expansion and the cell-problem identities show that all $O(1)$ terms cancel, leaving  
\quad $|R_\varepsilon^{\mathrm{bulk}}(v)| \leq C \varepsilon \|u_0\|_{H^2(\Omega)} \|v\|_{H^1(\Omega_\varepsilon)}$.  
The key ingredients are: the chain-rule bound on $\nabla_x \chi$ and the parametric-regularity control of $\partial_E \chi$.  

5.2 Boundary-layer term. Near each perforation $\partial D_\varepsilon$ one constructs a local boundary-layer corrector $\psi_{\varepsilon,k}$ supported in a tubular neighborhood of width $\ell(\varepsilon) = \varepsilon^\tau$ $(0 < \tau < 1)$ around $\partial D_\varepsilon$ by solving the linear transmission problem in an $\varepsilon$--rescaled half-space; classical layer-potential and Agmon-type estimates (see e.g.\,Allaire--Amar '99, Section\,4) yield exponential decay of $\psi_{\varepsilon,k}$ away from $\partial D_\varepsilon$. Gluing these cut-off boundary layers produces a global corrector whose contribution can be summarily estimated as  
\quad $|R_\varepsilon^{\mathrm{bl}}(v)| \leq C e^{-c \varepsilon^{-\tau}} \varepsilon^{-d/2} \|u_0\|_{H^2(\Omega)} \|v\|_{H^1(\Omega_\varepsilon)}$  
and hence, by choosing $\tau$ sufficiently small, there exists $\alpha \in (0,1)$ so that for all small $\varepsilon$  
\quad $|R_\varepsilon^{\mathrm{bl}}(v)| \leq C \varepsilon^\alpha \|u_0\|_{H^2(\Omega)} \|v\|_{H^1(\Omega_\varepsilon)}$.  

6. Combining these yields the $H^{-1}$--bound  
\quad $\|R_\varepsilon\|_{H^{-1}(\Omega_\varepsilon)}$  
\quad $= \sup_{0 \neq v \in H^1} |R_\varepsilon(v)| / \|v\|_{H^1}$  
\quad $\leq C \varepsilon^\alpha \|u_0\|_{H^2(\Omega)}$  
\quad $\leq C' \varepsilon^\alpha \|g\|_{H_R^{-\frac{1}{2}}(\partial\Omega)}$.

\textcolor{Blue}{\textbf{[Vague]} \small The derivation process here is in urgent need of proof details.}

IV. Energy estimate and conclusion  
7. By Korn's inequality on perforated domains (Lemma\,1) and the uniform ellipticity of $a_\varepsilon$, one has for all $v \in H^1(\Omega_\varepsilon)$,  
\quad $a_\varepsilon(v, v) \geq c \|v\|_{H^1(\Omega_\varepsilon)}^2$.  
Taking $v = w_\varepsilon$ in the variational identity gives  
\quad $c \|w_\varepsilon\|_{H^1(\Omega_\varepsilon)}^2 \leq a_\varepsilon(w_\varepsilon, w_\varepsilon) = R_\varepsilon(w_\varepsilon) \leq \|R_\varepsilon\|_{H^{-1}} \|w_\varepsilon\|_{H^1}$,  
whence  
\quad $\|w_\varepsilon\|_{H^1(\Omega_\varepsilon)} \leq C \|R_\varepsilon\|_{H^{-1}} \leq C \varepsilon^\alpha \|g\|_{H_R^{-\frac{1}{2}}(\partial\Omega)}$.  

\textcolor{Blue}{\textbf{[Vague]} \small The derivation process here is not detailed enough, and the above conclusions used are also incorrect.}

This completes the proof of the $\varepsilon^\alpha$--rate in $H^1$--norm. \qed
\end{proof}

\textbf{comment}: None
\textbf{type}: lemma
\begin{lemma}

\begin{conjecture}
(Well--posedness of the periodic transmission cell--problem and strict ellipticity of the homogenized tensor)  
Let $Y=(-\frac{1}{2},\frac{1}{2})^d$ be the reference cell, decomposed into two disjoint Lipschitz--subdomains $Y_e$ (elastic) and $Y_i$ (incompressible fluid), with common interface $\Gamma$. Let $\mathbf{C}(y):\mathbb{R}^{d\times d}_{\text{sym}}\to\mathbb{R}^{d\times d}_{\text{sym}}$ be a uniformly positive--definite fourth--order elasticity tensor on $Y_e$ and $\mu>0$ a constant viscosity in $Y_i$. For each given $\mathbf{E}\in\mathbb{R}^{d\times d}_{\text{sym}}$ we seek  
\quad $\boldsymbol{\chi}_e\in H^1_{\text{per}}(Y_e;\mathbb{R}^d)$, \quad $\boldsymbol{\chi}_i\in H^1_{\text{per}}(Y_i;\mathbb{R}^d)$, \quad $p_e\in L^2_{\text{per}}(Y_e)/\mathbb{R}$, \quad $p_i\in L^2_{\text{per}}(Y_i)/\mathbb{R}$  
such that the following mixed--variational problem holds:  
\begin{enumerate}
\item (Bulk equilibrium and divergence--constraints)  
   $\forall(\mathbf{v}_e,\mathbf{v}_i)\in H^1_{\text{per}}(Y_e;\mathbb{R}^d)\times H^1_{\text{per}}(Y_i;\mathbb{R}^d)$,  
   \begin{align*}
     \int_{Y_e} \mathbf{C}(y)\bigl(\mathbf{E} + D_y\boldsymbol{\chi}_e\bigr):D_y \mathbf{v}_e\,dy  
     &+ \int_{Y_i} 2\mu\, D_y\boldsymbol{\chi}_i : D_y \mathbf{v}_i\,dy \\
     &- \int_{Y_e} p_e\,\text{div}_y \mathbf{v}_e\,dy - \int_{Y_i} p_i\,\text{div}_y \mathbf{v}_i\,dy \\
     &= -\int_{Y_e} \mathbf{C}(y)\mathbf{E} : D_y \mathbf{v}_e\,dy - \int_{Y_i}2\mu\, \mathbf{E} : D_y \mathbf{v}_i\,dy.
   \end{align*}
\item (Incompressibility constraints)  
   $\forall(q_e,q_i)\in L^2_{\text{per}}(Y_e)/\mathbb{R} \times L^2_{\text{per}}(Y_i)/\mathbb{R}$,  
   \begin{align*}
     \int_{Y_e} q_e\,\text{div}_y\bigl(\mathbf{E}\,y + \boldsymbol{\chi}_e\bigr)\,dy + \int_{Y_i} q_i\,\text{div}_y\bigl(\mathbf{E}\,y + \boldsymbol{\chi}_i\bigr)\,dy = 0.
   \end{align*}
\item (Transmission and periodicity)  
   $\boldsymbol{\chi}_e$ and $\boldsymbol{\chi}_i$ are $Y$--periodic, coincide trace--wise on $\Gamma$, and satisfy the zero--mean normalizations  
   \begin{align*}
     \int_{Y_e}\boldsymbol{\chi}_e\,dy = 0, \quad \int_{Y_i}\boldsymbol{\chi}_i\,dy = 0.
   \end{align*}
\end{enumerate}

Define the homogenized tensor $\mathbf{C}^{\text{hom}}$ by  
   $\forall\,\mathbf{E}\in\mathbb{R}^{d\times d}_{\text{sym}}$,  
   \begin{align*}
     \mathbf{C}^{\text{hom}}\,\mathbf{E} : \mathbf{E}
     &= \int_{Y_e} (\mathbf{E} + D_y\boldsymbol{\chi}_e):\mathbf{C}(y)(\mathbf{E} + D_y\boldsymbol{\chi}_e)\,dy \\
     &\quad + 2\mu\,\int_{Y_i} D_y\boldsymbol{\chi}_i : D_y\boldsymbol{\chi}_i\,dy.
   \end{align*}

Then:  

For every $\mathbf{E}\in\mathbb{R}^{d\times d}_{\text{sym}}$ there exists a unique quadruple $(\boldsymbol{\chi}_e,\boldsymbol{\chi}_i,p_e,p_i)$ solving the above mixed problem.  
    \item The mapping $\mathbf{E}\mapsto\mathbf{C}^{\text{hom}}\,\mathbf{E}$ is linear and the tensor $\mathbf{C}^{\text{hom}}$ is symmetric.  
    \item There is $\alpha>0$, depending only on the coercivity constants of $\mathbf{C}$ and on $\mu$ (and on the geometry of $Y_e$, $Y_i$), such that for all nonzero $\mathbf{E}$.
   \begin{align*}
     \mathbf{C}^{\text{hom}}\,\mathbf{E} : \mathbf{E} \geq \alpha\,|\mathbf{E}|^2.
   \end{align*}

\end{conjecture}
\end{lemma}

\textbf{correctness}: True
\begin{proof}

\begin{proof}
We recast the cell-problem as a mixed variational problem on the broken domain $Y_e \cup Y_i$, with two pressure--fields enforcing incompressibility in each subdomain simultaneously.

1. Function spaces and bilinear forms.  
Define  
  $V_e = H^1_{\mathrm{per}}(Y_e;\mathbb{R}^d), \quad V_i = H^1_{\mathrm{per}}(Y_i;\mathbb{R}^d),$  
  $Q_e = L^2_{\mathrm{per}}(Y_e)/\mathbb{R}, \quad Q_i = L^2_{\mathrm{per}}(Y_i)/\mathbb{R}.$  
Let $V = \{ (v_e,v_i) \in V_e \times V_i : \mathrm{trace}\,v_e = \mathrm{trace}\,v_i \text{ on } \Gamma, \int_{Y_e}v_e\,dy + \int_{Y_i}v_i\,dy = 0 \}$.  
On $V \times (Q_e \times Q_i)$ define the bilinear forms  
  \begin{align*}
  a((u_e,u_i),(v_e,v_i)) 
    &= \int_{Y_e} C(y)\bigl(Du_e:Dv_e\bigr)\,dy 
      + 2\mu \int_{Y_i} (Du_i: Dv_i)\,dy, \\
  b((v_e,v_i);q_e,q_i) 
    &= -\int_{Y_e} q_e\,\mathrm{div}\,v_e\,dy - \int_{Y_i} q_i\,\mathrm{div}\,v_i\,dy,
  \end{align*}
and the linear functional  
  \[
  \ell((v_e,v_i)) 
    = - \int_{Y_e} C(y)E : Dv_e\,dy - 2\mu\int_{Y_i} E : Dv_i\,dy.
  \]

The mixed problem is: find $(\chi_e,\chi_i) \in V$ and $(p_e,p_i) \in Q_e \times Q_i$ such that  
  \begin{align*}
  a((\chi_e,\chi_i),(v_e,v_i)) + b((v_e,v_i);p_e,p_i) &= \ell((v_e,v_i)) 
    \quad \forall(v_e,v_i) \in V, \\
  b((\chi_e,\chi_i);q_e,q_i) + \int_{Y_e} q_e\,\mathrm{div}(E\,y)\,dy + \int_{Y_i} q_i\,\mathrm{div}(E\,y)\,dy &= 0 
    \quad \forall(q_e,q_i) \in Q_e \times Q_i.
  \end{align*}

\textcolor{Green}{\textbf{[Correct]} \small These bilinear forms is correctly constructed.}

2. Verification of Brezzi’s conditions.  
(i) Coercivity on the kernel of $b$. Let $K = \{ (u_e,u_i) \in V : \mathrm{div}\,u_e = \mathrm{div}\,u_i \equiv 0 \text{ a.e.} \}$. Then by Korn’s inequality on each connected Lipschitz subdomain and uniform ellipticity of $C$, there is $c>0$ so that  
  \[
  a((u_e,u_i),(u_e,u_i)) 
    \geq c (\|u_e\|^2_{H^1(Y_e)} + \|u_i\|^2_{H^1(Y_i)}) 
    \quad \forall (u_e,u_i) \in K.
  \]

\textcolor{Red}{\textbf{[Error]} \small The result ``$\mathrm{div}\,u_e \equiv 0$'' is not correct.}

(ii) Inf--sup condition. We must show that for all $(q_e,q_i) \neq (0,0)$ there exists $(v_e,v_i) \in V$ with  
  \[
  b((v_e,v_i);q_e,q_i) \geq \beta (\|v_e\|_{H^1(Y_e)} + \|v_i\|_{H^1(Y_i)}) (\|q_e\| + \|q_i\|).
  \]
Since $Y_e$ and $Y_i$ are connected Lipschitz domains, the classical Stokes inf--sup estimate on each subdomain guarantees $\beta_e,\beta_i>0$ and divergence--test functions $v_e,q_e$ and $v_i,q_i$. Gluing these two vector--fields into a single $(v_e,v_i) \in V$ (by matching traces on $\Gamma$, which is possible via a smooth partition of unity in a neighborhood of $\Gamma$) yields a uniform $\beta>0$ for the combined problem.

By Brezzi’s theory the mixed problem has a unique solution $(\chi_e,\chi_i,p_e,p_i)$. Standard interior and boundary--regularity arguments on each Lipschitz subdomain upgrade this to the strong form and enforce the continuity of normal tractions and of displacements on $\Gamma$.

3. Linearity and symmetry.  
Linearity of $(\chi_e,\chi_i)$ and of $(p_e,p_i)$ in the data $E$ is immediate from the linearity of $a$, $b$, $\ell$. The quadratic form $E \mapsto C^{\mathrm{hom}}\,E:E$ is obtained by inserting $(\chi_e,\chi_i)$ into the energy identity  
  \[
  C^{\mathrm{hom}}\,E:E = a((E\,y+\chi_e,E\,y+\chi_i),(E\,y+\chi_e,E\,y+\chi_i)),
  \]
which is manifestly symmetric and gives $C^{\mathrm{hom}} \in \mathbb{R}^{d \times d}_{\mathrm{sym}} \otimes \mathbb{R}^{d \times d}_{\mathrm{sym}}$.

\textcolor{Blue}{\textbf{[Vague]} \small The derivation process here is not detailed enough, and the above conclusions used are also incorrect.}

4. Strict ellipticity.  
Coercivity on $K$ plus the zero--mean and periodicity constraints imply that if $C^{\mathrm{hom}}\,E:E = 0$ then $\chi_e=\chi_i=0$ in $H^1$ and hence $E\,y$ must itself be a rigid motion on each connected cell. Periodicity forbids any nonzero skew--part, forcing $E=0$. Moreover, the coercivity constant gives the uniform bound  
  \[
  C^{\mathrm{hom}}\,E : E \geq \alpha\,|E|^2,
  \]
with $\alpha$ depending only on the minimum eigenvalue of $C$, on $\mu$, and on the geometric constants of the domains.

Thus assertions (a)--(c) hold and the proof is complete.
\end{proof}
\end{proof}

\textbf{comment}: None
\textbf{type}: lemma
\begin{lemma}

\begin{conjecture}
(Variational minimization principle for the periodic cell--correctors)  
With the notation of Conjecture 1, define the closed affine space of admissible correctors  
\[
\mathcal{A}_E = \Big\{ (u_e,u_i) \in H^1_{\text{per}}(Y_e;\mathbb{R}^d) \times H^1_{\text{per}}(Y_i;\mathbb{R}^d) : 
\int_{Y_e} u_e = 0, \int_{Y_i} u_i = 0,
\]
\[
\operatorname{div}_y(E y + u_e) = 0 \text{ in } Y_e,
\operatorname{div}_y(E y + u_i) = 0 \text{ in } Y_i,
\operatorname{trace}(u_e) = \operatorname{trace}(u_i) \text{ on } \Gamma \Big\}.
\]
Then the unique solution $(\chi_e,\chi_i)$ of the cell-problem is the unique minimizer of the strictly convex functional  
\[
J_E(u_e,u_i) = \frac{1}{2} \int_{Y_e} [E + D_y u_e] : C [E + D_y u_e] \, dy
+ \mu \int_{Y_i} D_y u_i : D_y u_i \, dy
\]
over $\mathcal{A}_E$. Moreover, the homogenized energy is the minimum value  
\[
C^{\text{hom}} E : E = \min_{(u_e,u_i) \in \mathcal{A}_E} J_E(u_e,u_i).
\]
\end{conjecture}
\end{lemma}

\textbf{correctness}: True
\begin{proof}

\begin{proof}
1. Convexity and coercivity. The integrand in $J_E$ is strictly convex in the pair $(D_y u_e, D_y u_i)$ by positivity of $C$ and $\mu$, and by Korn’s inequality on each subdomain it controls the $H^1$--norm of $(u_e, u_i)$ on $\mathcal{A}_E$.

\textcolor{Red}{\textbf{[Error]} \small The  above conclusions "$\int_{Y_e} u_e = 0;\operatorname{div}_y(E y + u_e) = 0 \text{ in } Y_e,$" are incorrect.}

2. Existence of a minimizer. Since $\mathcal{A}_E$ is a closed affine subspace of $H^1_{\mathrm{per}}(Y_e) \oplus H^1_{\mathrm{per}}(Y_i)$, the direct method applies: any minimizing sequence is bounded in $H^1$, hence converges weakly to some $(\chi_e, \chi_i) \in \mathcal{A}_E$, and by weak lower-semicontinuity of $J_E$ one obtains a minimizer.

3. Euler--Lagrange (stationarity) $\Rightarrow$ cell-equations. One computes the first variation of $J_E$ under admissible perturbations $(v_e, v_i)$ preserving divergence and trace continuity. Integration by parts produces precisely the weak form of the transmission PDEs (1)--(3) with an accompanying Lagrange multiplier $p \in L^2_{\mathrm{per}}(Y_i)/\mathbb{R}$ enforcing $\mathrm{div}_y (E y + \chi_i) = 0$ in $Y_i$. This shows that any minimizer solves the strong cell-problem.

4. Uniqueness. Strict convexity of $J_E$ on the linear space parallel to $\mathcal{A}_E$ implies the minimizer is unique.

\textcolor{Red}{\textbf{[Error]} \small The conclusions applied are not entirely correct. And the process is also lack of details.}

5. Identification of $C^{\mathrm{hom}}$. By definition the minimum value is $J_E(\chi_e, \chi_i)$, which coincides with the representation of $C^{\mathrm{hom}} E:E$ in Conjecture 1.

Hence all statements are proved.
\end{proof}
\end{proof}

\textbf{comment}: None
\textbf{type}: theorem
\begin{theorem}

The problem construction is following. 
A real pair $(\lambda,\mu)$ is called an admissible Lamé parameter pair if it satisfies:
\[\mu>0\quad\text{and}\quad d\lambda+2\mu>0\] 
(1)
The static elasticity system (Lamé system) corresponding to the Lamé parameter pair $(\lambda,\mu)$ is:
\[\mathcal{L}_{\lambda,\mu}\mathbf{u}:=\mu\Delta\mathbf{u}+(\lambda+\mu)\nabla\mathrm{div}\,\mathbf{u}\] 
(2)
where the displacement field is denoted as $\mathbf{u}=(u^{1},\ldots,u^{d})$, and its divergence is:
\[\mathrm{div}\,\mathbf{u}=\sum_{i=1}^{d}\frac{\partial u^{i}}{\partial x_{i}}\]
The admissibility condition ensures the ellipticity of the Lamé operator, and physical laws guarantee that natural materials always satisfy this condition. 
The Lamé operator can be expressed as $\nabla\cdot\sigma(\mathbf{u})$, where the stress tensor is:
\[\sigma(\mathbf{u}):=\lambda(\nabla\cdot\mathbf{u})I_{d}+2\mu\mathcal{D}(\mathbf{u})\] 
(3)
\[\mathcal{D}(\mathbf{u})=\frac{1}{2}(\nabla+\nabla^{T})\mathbf{u}=\frac{1}{2}(\partial_{i}u^{j}+\partial_{j}u^{i})_{ij}\] 
(4)
Here, $I_{d}$ is the identity matrix of order $d$, and $\mathcal{D}$ represents the symmetrized differential operator. 
The normal derivative (boundary traction) on the boundary of a region $E$ is defined as:
\[\left.\frac{\partial\mathbf{u}}{\partial\nu_{\lambda,\mu}}\right|_{\partial E}:=\sigma(\mathbf{u})N=\lambda(\mathrm{div}\,\mathbf{u})N+2\mu\mathcal{D}(\mathbf{u})N\quad\text{on }\partial E\] 
(5)
where $N$ is the unit outward normal vector on the boundary. 
We use the standard Sobolev space $H^{1}(\Omega)$ and its trace space $H^{\frac{1}{2}}(\partial E)$ on the boundary, with the dual space $H^{-\frac{1}{2}}(\partial E)$. Consider the space of rigid body motions in $\mathbb{R}^{d}$:
\[R:=\left\{\mathbf{r}=(r_{1},\ldots,r_{d})^{T}:\mathcal{D}(\mathbf{r})=0\,\,\text{in }\,\mathbb{R}^{d}\right\}\] 
(6)
The dimension of $R$ is $\frac{d(d+1)}{2}$, spanned by the following basis vectors:
\[\mathbf{e}_{1},\ldots,\mathbf{e}_{d},\,x_{j}\mathbf{e}_{i}-x_{i}\mathbf{e}_{j},\text{ for }1\leq i<j\leq d\]
where $\mathbf{e}_{i}$ denotes the standard basis vector. These basis vectors are denoted as $\mathbf{r}_{j},\,j=1,\ldots,\frac{d(d+1)}{2}$. 
Define the space orthogonal to rigid body motions:
\[H_{R}^{-\frac{1}{2}}(\partial D_{\epsilon}):=\left\{\phi\in H^{-1}(\partial D_{\epsilon}):(\phi,\mathbf{r})_{(H^{\frac{1}{2}}(\partial D_{\epsilon i}),H^{-\frac{1}{2}}(\partial D_{\epsilon i}))}=0,\forall\mathbf{r}\in R\text{ and }1 \leq i\leq N\right\}\] 
(7)
Similarly, define $H_{R}^{\frac{1}{2}}(\partial D_{\epsilon})$ and boundary spaces.  
Consider the following transmission problem:
\[\begin{cases}
\mathcal{L}_{\lambda,\mu}\mathbf{u}_{\epsilon}=0 & \text{ in }\Omega \setminus\overline{D_{\epsilon}},\\
\mathcal{L}_{\widetilde{\lambda},\widetilde{\mu}}\mathbf{u}_{\epsilon}=0 & \text{ in }D_{\epsilon},\\
\mathbf{u}_{\epsilon}|_{-}=\mathbf{u}|_{+} \text{ and } \left.\frac{\partial\mathbf{u}_{\epsilon}}{\partial\nu_{(\lambda,\widetilde{\mu})}}\right|_{-}=\left.\frac{\partial\mathbf{u}_{\epsilon}}{\partial\nu_{(\lambda,\mu)}}\right|_{+} & \text{ on }\partial D_{\epsilon},\\
\left.\frac{\partial\mathbf{u}_{\epsilon}}{\partial\nu_{(\lambda,\mu)}}\right|_{\partial\Omega}=g\in H_{R}^{-\frac{1}{2}}(\partial\Omega) \quad\text{and}\quad \mathbf{u}_{\epsilon}|_{\partial\Omega}\in H_{R}^{\frac{1}{2}}(\partial\Omega).
\end{cases}\] 
(8)
This is equivalent to:
\[\begin{cases}
\mathcal{L}_{\lambda(x),\mu(x)}\mathbf{u}_{\epsilon}=\nabla\cdot[\lambda(x)(\nabla\cdot\mathbf{u}_{\epsilon})I_{d}+2\mu(x)D(\mathbf{u}_{\epsilon})]=0 & \text{in }\Omega,\\
\left.\frac{\partial\mathbf{u}_{\epsilon}}{\partial\nu_{\lambda,\mu}}\right|_{\partial\Omega}=g\in H_{R}^{-\frac{1}{2}}(\partial\Omega), \quad \mathbf{u}_{\epsilon}|_{\partial\Omega}\in H_{R}^{\frac{1}{2}}(\partial\Omega),
\end{cases}\] 
(9)
where $\lambda(x)=\lambda I_{\Omega\setminus D_{\epsilon}}+\widetilde{\lambda}I_{D_{\epsilon}},\,\mu(x)=\mu I_{\Omega\setminus D_{\epsilon}}+\widetilde{\mu}I_{D_{\epsilon}}$. Suppose the solution of the equation (9) above is $\mathbf{u}_{\epsilon}$. The existence and uniqueness of the solution $\mathbf{u}_{\epsilon}$ are guaranteed by the ellipticity condition (1) and the Lax-Milgram theorem. Finally, the homogenization needed to be solved is the following. When $\widetilde{\lambda}\rightarrow\infty$ and $\widetilde{\mu}$ is fixed, the coupled system in the \textbf{incompressible inclusion limit} is obtained:
\[\begin{cases}
\mathcal{L}_{\lambda,\mu}\mathbf{u}_{\epsilon}=0 & \text{ in }\Omega \setminus\overline{D_{\epsilon}},\\
\mathcal{L}_{\widetilde{\mu}}(\mathbf{u}_{\epsilon},p_{\epsilon})=0 \text{ and } \mathrm{div}\,\mathbf{u}_{\epsilon}=0 & \text{ in }D_{\epsilon},\\
\mathbf{u}_{\epsilon}|_{-}=\mathbf{u}_{\epsilon}|_{+} \text{ and } \left.\frac{\partial(\mathbf{u}_{\epsilon},p_{\epsilon})}{\partial\nu_{(\infty,\widetilde{\mu})}}\right|_{-}=\left.\frac{\partial\mathbf{u}_{\epsilon}}{\partial\nu_{(\lambda,\mu)}}\right|_{+} & \text{ on }\partial D_{\epsilon},\\
\left.\frac{\partial\mathbf{u}_{\epsilon}}{\partial\nu_{(\lambda,\mu)}}\right|_{\partial\Omega}=g\in H_{R}^{-\frac{1}{2}}(\partial\Omega) \quad\text{and}\quad \mathbf{u}_{\epsilon}|_{\partial\Omega}\in H_{R}^{\frac{1}{2}}(\partial\Omega),
\end{cases}\] 
(10)
where $\mathcal{L}_{\widetilde{\mu}}(\mathbf{u}_{\epsilon},p_{\epsilon})=\widetilde{\mu}\Delta\mathbf{u}_{\epsilon}+\nabla p_{\epsilon}$ represents the Stokes operator with viscosity coefficient $\widetilde{\mu}$, and $p_{\epsilon}$ is the pressure field. Its normal derivative is defined as $\left.\frac{\partial(\mathbf{u},p)}{\partial\nu_{(\infty,\mu)}}\right|_{-}:=pN+2\mu D(\mathbf{u})N$. 
Now let $\epsilon \to 0$, we are interested in the limited equation and the limited solution $\mathbf{u}_{\text{lim}}$. Please analysis and explore when $\epsilon \to 0$, what equation does $\mathbf{u}_{\epsilon}$ satisfy and the estimate between the original solution $\mathbf{u}_{\epsilon}$ and the limited solution $\mathbf{u}_{\text{lim}}$ corresponding $\epsilon$.
\end{theorem}

\textbf{correctness}: True
\begin{proof}

\begin{theorem}[Homogenized limit and $\varepsilon$--rate of convergence]
Let $\Omega \subset \mathbb{R}^d$ ($d \geq 2$) be a bounded $C^{1,1}$ domain, and for each $\varepsilon > 0$ let $D_\varepsilon \subset \Omega$ be the $\varepsilon$--periodic array of incompressible inclusions as in Assumptions 1--2 above. Denote $\Omega_\varepsilon = \Omega \setminus \overline{D_\varepsilon}$ and let $(u_\varepsilon, p_\varepsilon)$ solve the transmission problem (10). Then:

\begin{enumerate}
    \item (Homogenized limit.) As $\varepsilon \to 0$, $u_\varepsilon \rightharpoonup u_0$ weakly in $H^1(\Omega;\mathbb{R}^d)$ and $u_\varepsilon \to u_0$ strongly in $L^2(\Omega;\mathbb{R}^d)$, where $u_0 \in H^1(\Omega;\mathbb{R}^d)$ is the unique solution of the homogeneous Lam\'{e} system
    \begin{align*}
        -\operatorname{div} [C^{\mathrm{hom}} D(u_0)] &= 0 \quad \text{in } \Omega, \\
        C^{\mathrm{hom}} D(u_0) \cdot n &= g \quad \text{on } \partial \Omega,
    \end{align*}
    with the effective fourth--order tensor $C^{\mathrm{hom}}$ given by the periodic cell--transmission problem of Lemma~3.

    \item (Convergence--rate in $H^1$--norm.) Moreover, if in addition $u_0 \in H^2(\Omega;\mathbb{R}^d)$ with
    \[
        \|u_0\|_{H^2(\Omega)} \leq C \|g\|_{H_R^{-1/2}(\partial \Omega)},
    \]
    then there exist $\alpha \in (0,1)$ and $C > 0$ independent of $\varepsilon$ and $g$ such that for all sufficiently small $\varepsilon > 0$
    \[
        \|u_\varepsilon - U_\varepsilon\|_{H^1(\Omega_\varepsilon)} \leq C \varepsilon^\alpha \|g\|_{H_R^{-1/2}(\partial \Omega)},
    \]
    where
    \[
        U_\varepsilon(x) := u_0(x) + \varepsilon \chi^{D_x u_0(x)}(x/\varepsilon)
    \]
    is the two--scale approximation built from the cell--correctors $\chi^E$ of Lemma~3 (extended by zero inside the inclusions).
\end{enumerate}
\end{theorem}

\begin{proof}
\textbf{Step 1: Uniform a priori bounds.}
From the transmission system and the coercivity of the elastic tensor $C$ on $\Omega_\varepsilon$ plus the Korn inequality of Lemma~1, one obtains a uniform $H^1$--bound
\[
    \|u_\varepsilon\|_{H^1(\Omega_\varepsilon)} \leq C \|g\|_{H_R^{-1/2}(\partial \Omega)}.
\]
By standard extension--by--zero and the two--scale compactness of Lemma~2, up to a subsequence
\begin{align*}
    u_\varepsilon &\rightharpoonup u_0 \quad \text{in } H^1(\Omega), \\
    \nabla u_\varepsilon &\to \nabla_x u_0 + \nabla_y u_1 \quad \text{two--scale in } \Omega \times Y_f,
\end{align*}
for some $u_0 \in H^1(\Omega;\mathbb{R}^d)$ and corrector $u_1(x,y)$. A matching argument in the inclusions shows that $u_0$ satisfies the homogenized elasticity system with flux $C^{\mathrm{hom}} D(u_0)$ and traction $g$ on $\partial \Omega$. Uniqueness of that limit determines the full sequence $u_\varepsilon \to u_0$ in $L^2(\Omega)$.

\textbf{Step 2: Construction of the two--scale approximation.}
Let $\chi^E(y) = (\chi_e^E, \chi_i^E)(y)$ be the periodic cell--correctors of Lemma~3 for each macroscopic strain $E$. Define the two--scale ansatz
\[
    U_\varepsilon(x) = u_0(x) + \varepsilon \chi^{D_x u_0(x)}(x/\varepsilon)
\]
in $\Omega_\varepsilon$, extended by zero on $D_\varepsilon$. A standard calculation (see Lemma~4) shows that $U_\varepsilon$ nearly solves the $\varepsilon$--problem up to an $O(\varepsilon^\alpha)$ remainder in $H^{-1}$--norm.

\textbf{Step 3: Energy--error estimate.}
Set $w_\varepsilon = u_\varepsilon - U_\varepsilon$. Subtracting the weak forms of the $\varepsilon$--problem for $u_\varepsilon$ and the one nearly satisfied by $U_\varepsilon$ and testing with $w_\varepsilon$ itself yields, via coercivity and the uniform Korn bound,
\[
    \|w_\varepsilon\|_{H^1(\Omega_\varepsilon)}^2 \leq C \varepsilon^\alpha \|g\|_{H_R^{-1/2}(\partial \Omega)} \|w_\varepsilon\|_{H^1(\Omega_\varepsilon)}.
\]
Divide through by $\|w_\varepsilon\|_{H^1(\Omega_\varepsilon)}$ to conclude
\[
    \|u_\varepsilon - U_\varepsilon\|_{H^1(\Omega_\varepsilon)} \leq C \varepsilon^\alpha \|g\|_{H_R^{-1/2}(\partial \Omega)}.
\]

\textbf{Step 4: Passage to the limit.}
Since $U_\varepsilon \to u_0$ strongly in $H^1(\Omega_\varepsilon)$ (by periodicity and $u_0 \in H^2$), the triangle inequality immediately gives $u_\varepsilon \to u_0$ with the same rate.

This completes the proof of both the homogenized limit and the $O(\varepsilon^\alpha)$--rate.

\textcolor{Green}{\textbf{[Correct]} \small The derivation ideas and approaches here are reasonable and feasible, and the conclusions are also in line with expectations and cognition.}

\textcolor{Blue}{\textbf{[Vague]} \small 
The derivation process is seriously lacking in details.}

\textcolor{Red}{\textbf{[Error]} \small  Some of the conclusions used are incorrect, which also leads to the proof here being imprecise and incomplete.}
\end{proof}
\end{proof}

\textbf{comment}: None

\newpage

%%%%%%%%%%%%%%%%%%%%%%%%%%%%%%%%%%%%%%%%%%%%%%%%%%%%%%%%%%%%
\end{document}